\documentclass[pdflatex,sn-apa]{sn-jnl}

\usepackage{graphicx}%
\usepackage{multirow}%
\usepackage{amsmath,amssymb,amsfonts}%
\usepackage{amsthm}%
\usepackage{mathrsfs}%
\usepackage[title]{appendix}%
\usepackage{xcolor}%
\usepackage{textcomp}%
\usepackage{manyfoot}%
\usepackage{booktabs}%
\usepackage{algorithm}%
\usepackage{algorithmicx}%
\usepackage{algpseudocode}%
\usepackage{listings}%

\usepackage{hyperref}       
\usepackage{url}            
\usepackage{graphicx}
\usepackage{subcaption}
\usepackage{tikz,pgfplots}
\usepackage{multirow}
\usepackage{longtable}
\usepackage{stmaryrd}
\usepackage{pdflscape}
\usepackage{bm}
\usepackage{adjustbox}
\renewcommand{\vec}[1]{\boldsymbol{#1}}

\theoremstyle{thmstyleone}%
\theoremstyle{thmstyletwo}%

\theoremstyle{thmstylethree}%

\begin{document}

\title[Article Title]{Aleatoric and Epistemic Uncertainty Measures for Ordinal Classification through Binary Reduction}


\author*[1,2]{\fnm{Stefan} \sur{Haas}}\email{stefan.haas@campus.lmu.de}\email{stefan.sh.haas@bmwgroup.com}

\author[1,3,4]{\fnm{Eyke} \sur{Hüllermeier}}\email{eyke@lmu.de}

\affil[1]{\orgdiv{Institute of Informatics}, \orgname{LMU Munich}, \orgaddress{\country{Germany}}}

\affil[2]{ \orgname{BMW Group}, \orgaddress{\city{Munich},\country{Germany}}}

\affil[3]{ \orgname{Munich Center for Machine Learning}, \orgaddress{\country{Germany}}}

\affil[4]{ \orgname{German Centre for Artificial Intelligence (DFKI)}, \orgaddress{\city{Kaiserslautern},\country{Germany}}}


\abstract{ Ordinal classification problems, where labels exhibit a natural order, are prevalent in high-stakes fields such as medicine and finance. Accurate uncertainty quantification, including the decomposition into aleatoric (inherent variability) and epistemic (lack of knowledge) components, is crucial for reliable decision-making. However, existing research has primarily focused on nominal classification and regression. In this paper, we introduce a novel class of measures of aleatoric and epistemic uncertainty in ordinal classification, which is based on a suitable reduction to (entropy- and variance-based) measures for the binary case. These measures effectively capture the trade-off in ordinal classification between exact hit-rate and minimial error distances. We demonstrate the effectiveness of our approach on various tabular ordinal benchmark datasets using ensembles of gradient-boosted trees and multi-layer perceptrons for approximate Bayesian inference. Our method significantly outperforms standard and label-wise entropy and variance-based measures in error detection, as indicated by misclassification rates and mean absolute error. Additionally, the ordinal measures show competitive performance in out-of-distribution (OOD) detection. Our findings highlight the importance of considering the ordinal nature of classification problems when assessing uncertainty.}




\maketitle

\section{Introduction}

Supervised machine learning models are increasingly used for high-stakes decision-making in domains such as medicine and finance. Consider predicting treatment effects \citep{RAFIQUE20214003} or automating loan approvals \citep{UDDIN2023327}. Likewise, quantifying the \emph{predictive uncertainty} associated with a query $\vec{x}_q$
becomes more and more important for reliable and safe decision-making. Information about the predictive uncertainty could, for instance, be used in a downstream selective classification \citep{DBLP:conf/nips/GeifmanE17,DBLP:journals/ml/HendrickxPPMD24} approach in which only certain enough queries are processed automatically while uncertain ones are delegated to human experts. This, in turn, reduces the risk of wrong predictions and increases the overall accuracy of the predictor \citep{haas2024conformalized}.

A common distinction in uncertainty quantification is drawn between so-called \emph{aleatoric} and \emph{epistemic} uncertainty \citep{DBLP:journals/isci/SengeBDHHDH14,DBLP:journals/ml/HullermeierW21}. The latter refers to the uncertainty that arises due to a lack of knowledge or information. It can be reduced with additional training data or by selecting a better predictor or model for the specific task. In contrast, aleatoric uncertainty is irreducible and arises from the inherent randomness in the data. 
Identifying and measuring these uncertainties allows for more nuanced detection of issues related to the learning and prediction process. For example, information about aleatoric uncertainty empowers decision-makers to assess whether a dependable decision can be reached at all. Likewise, understanding epistemic uncertainty provides valuable insights into whether a given predictor is sufficiently informed to make reliable decisions or requires additional training data or a different type of model or model class.

In general, the focus of uncertainty quantification in machine learning has primarily been on nominal classification and regression, with the Bayesian approach prevailing in the literature, where epistemic uncertainty is represented by a second-order probability distribution representing the posterior over the hypothesis space: a probability distribution of probability distributions \citep{DBLP:journals/ml/HullermeierW21}. In practice, such second-order distributions are commonly approximated through Monte Carlo sampling using ensembles \citep{DBLP:conf/iclr/MalininPU21,DBLP:conf/ida/ShakerH20,DBLP:conf/uai/WimmerSHBH23,labelwise} or other variational techniques such as dropout \citep{DBLP:conf/icml/GalG16} or drop connect \citep{DBLP:journals/corr/abs-1906-04569} in deep learning.

\begin{figure}[!htbp]
  \centering
   \begin{subfigure}[t]{0.3\linewidth}
           \centering
           \includegraphics[width=\linewidth]{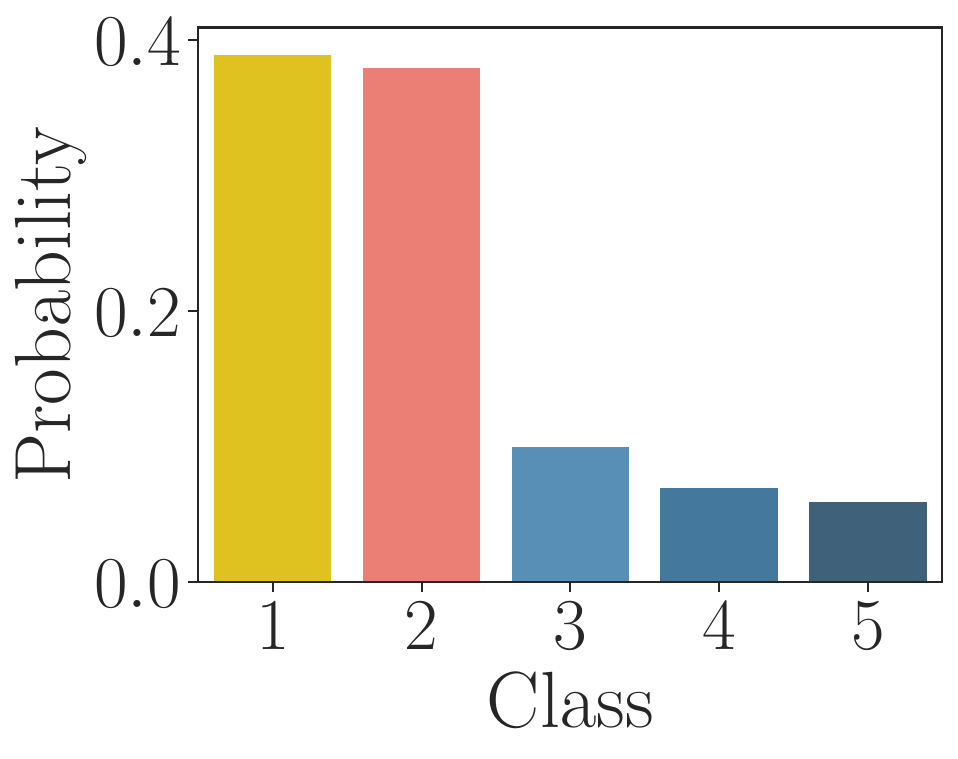}
       \end{subfigure} 
         \begin{subfigure}[t]{0.3\linewidth}
           \centering
           \includegraphics[width=\linewidth]{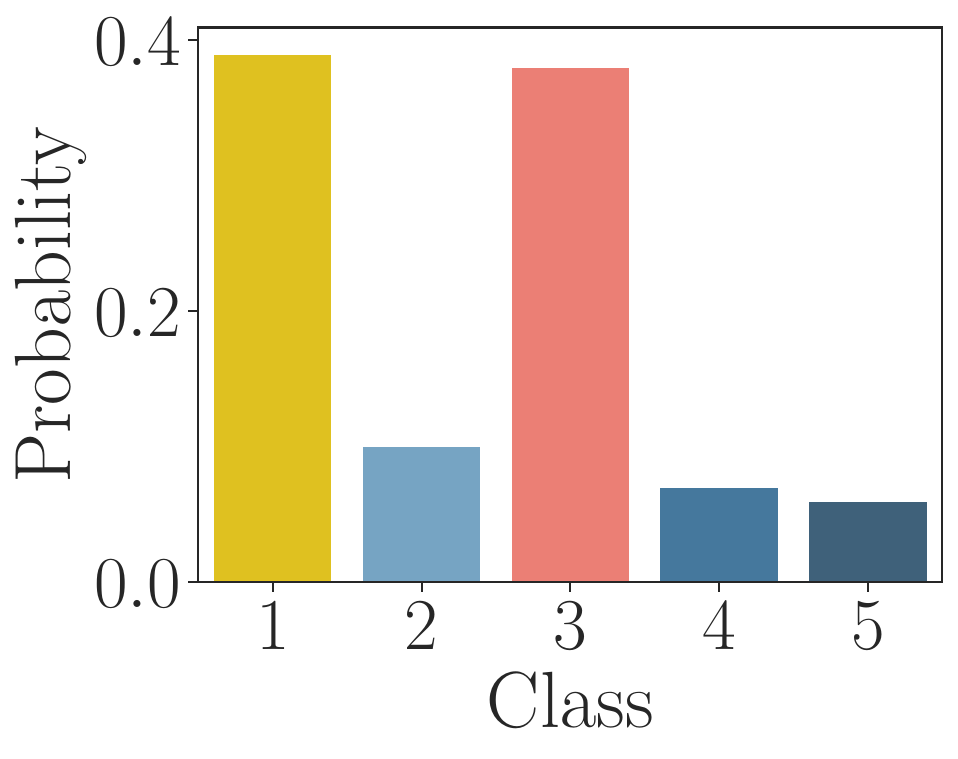}
       \end{subfigure} 
       \begin{subfigure}[t]{0.3\linewidth}
        \centering
        \includegraphics[width=\linewidth]{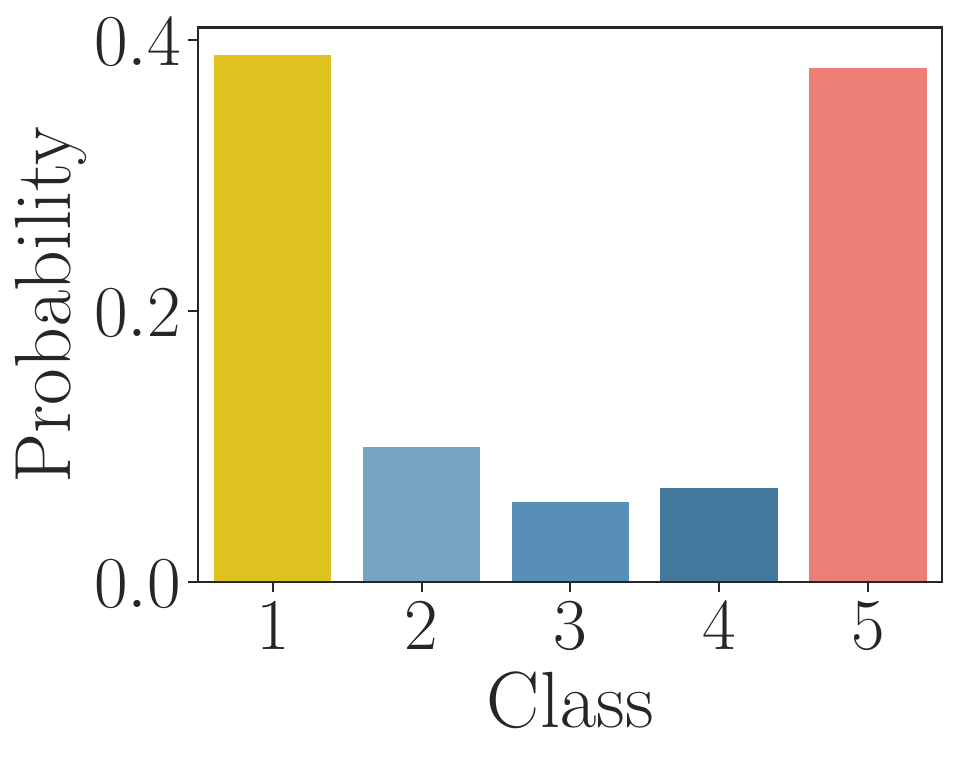}
    \end{subfigure}      
        \caption{Several probability distributions which lead to the same (Shannon) entropy ($\mathbb{H}$ = 1.9 with base 2). In ordinal classification, where the minimization of error distance is an important factor in addition to the exact hit-rate, this behavior has been shown to be problematic \citep{haas4965714uncertainty}. Obviously, the different distributions should be associated with different degrees of uncertainty, arguably increasing from left to right in this example. }
          \label{fig:entropy_example}
\end{figure}

A principled approach to measuring and separating aleatoric and epistemic uncertainty on the basis of classical information-theoretic measures of (Shannon) entropy \citep{shannon1948mathematical} was proposed by \cite{pmlr-v80-depeweg18a}. The approach has widely been adopted for nominal classification \citep{DBLP:conf/iclr/MalininPU21,DBLP:conf/ida/ShakerH20,DBLP:journals/corr/abs-1906-04569,shaker2021ensemble,DBLP:conf/aime/LohrIH24,HosseinIce2024}.
However, particularly in high-stakes use-cases, the class labels $y \in \mathcal{Y}$ often exhibit a natural order relation, with $y_{1} \prec y_{2} \prec \cdots \prec y_{K}$. Think of credit scoring with $\mathcal{Y}=\{$poor, fair, good, very good, excellent$\}$ or any other rating application, such as disease severity in medicine or employee performance evaluation in human resources.
Since ordinal classification lies somewhat between classification and regression, commonly used entropy and variance-based approaches are also applicable in the ordinal case \citep{DBLP:conf/iclr/MalininPU21}. However, these approaches may not ideally reflect the inherent trade-off within ordinal classification between exact hit-rate and minimized error distances, as has been shown by \cite{haas4965714uncertainty} for entropy. Entropy is not a good choice for uncertainty quantification in ordinal classification, as it does not take into account the dispersion of a probability distribution and is invariant to the redistribution of probability mass (cf.\ Figure \ref{fig:entropy_example} for an illustration).
To the best of our knowledge, no efforts have been devoted to disentangling aleatoric and epistemic uncertainty in the context of ordinal classification yet, which is crucial for reliable decision-making in many high-stakes use cases.

In this paper, we make the following key contributions:
\begin{itemize}
  \item We explore methods to disentangle aleatoric and epistemic uncertainty in ordinal classification based on commonly used entropy and variance-based decomposition approaches.
  \item We propose a novel binary decomposition method that builds upon commonly used entropy and total variance-based decompositions for quantifying aleatoric and epistemic uncertainty, taking the ordinal structure into account.
  \item We compare our novel method with standard entropy- and total variance-based approaches, as well as with label-wise entropy and variance-based binary decompositions, in an extensive study on twenty-three common tabular ordinal benchmark datasets using ensembles of gradient-boosted trees and multi-layer perceptrons for approximate Bayesian inference, specifically evaluating performance in error detection and out-of-distribution detection.
  \item Moreover, we demonstrate that the cross-entropy (CE) loss, as a proper scoring rule, is advantageous over ordinal losses for uncertainty quantification in ordinal classification in general. While ordinal losses often induce compressed unimodal predictive probability distributions---which can serve as a beneficial inductive bias for predictions, particularly for discretized continuous ordinal targets---they negatively impact uncertainty quantification.
\end{itemize}

\section{Learning Probabilistic Predictors}

We consider the setting of probabilistic supervised machine learning, in which a learner is given access to a set of training data
$$
\mathcal{D} = \{(\vec{x}_1,y_{1}),\ldots,(\vec{x}_{n},y_{n})\} \subset \mathcal{X} \times \mathcal{Y} \, ,
$$ 
with $\vec{x}_i \in \mathcal{X} \subseteq \mathbb{R}^m$ a feature vector from an instance space $\mathcal{X}$, and  $y_i \in \mathcal{Y}$ the corresponding class label or outcome from a set of outcomes $\mathcal{Y}$ that can be associated with an instance. 
In particular, we focus on the ordinal classification scenario, where $\mathcal{Y}=\{y_{1},\ldots,y_{K}\}$ consist of a finite set of class labels equipped with a natural (linear) order relation: 
$$y_{1} \prec y_{2} \prec \cdots \prec y_{K}$$ 

Suppose a model or hypothesis space $\mathcal{H}$ to be given,
where a hypothesis $h \in \mathcal{H}$ is a predictive model in the form of a mapping $\mathcal{X} \rightarrow \mathbb{P}(\mathcal{Y})$ from instances to probability distributions on outcomes. Assuming that training data as well as future (test) data is independently distributed according to an underlying (unknown) joint probability $P$ on $\mathcal{X} \times \mathcal{Y}$, the goal in probabilistic supervised learning is to induce a 
hypothesis $h^* \in \mathcal{H}$ with low risk (expected loss)
$$
R(h) := \mathbb{E}_{(\vec{x},y) \sim P} l(h(\vec{x}),y) = \int_{\mathcal{X} \times \mathcal{Y}} l(h(\vec{x}),y)\,dP(\vec{x},y) \, ,
$$
where $l: \, \mathbb{P}(\mathcal{Y}) \times  \mathcal{Y}  \rightarrow \mathbb{R}_+$ is
a loss (error) function. Training probabilistic predictors is typically accomplished by minimizing the (perhaps regularized) empirical risk 
\begin{equation}\label{eq:remp}
R_{emp}( h) := \frac{1}{n} \sum_{i=1}^{n} l(h(\bm{x}_i), y_i)
\end{equation}
as an estimate of the true risk (generalization performance). Then, the empirical risk minimizer 
$$
h :=  \underset{h' \in \mathcal{H}}{\arg\min} \ \mathcal{R}_{emp}(h')
$$ 
serves as an approximation of the true risk minimizing hypothesis $h^{*}$. Given a query instance $\vec{x}_q \in \mathcal{X}$ as input, it produces a probabilistic prediction 
\begin{equation}\label{eq:propre}
\vec{p} =  h(\vec{x}_q) = (p(y_1), \ldots , p(y_K)) = (p_{1},\ldots,p_{K}) \in \mathbb{P}(\mathcal{Y}) 
\end{equation}
as output, where $p_{k}$ is the predicted probability for the $k^{th}$ class $y_{k}$.

So-called (strictly) proper scoring rules \citep{gneiting2007strictly} are commonly used as loss functions $l$ in (\ref{eq:remp}). These have the nice theoretical property of being minimized (in expectation) by the true conditional probabilities, hence incentivizing the learner to produce well-calibrated probability estimates (\ref{eq:propre}). An important example of such loss functions is the log-loss or cross-entropy loss (CE) 
\begin{equation}\label{eq:ce}
  l_{CE}( \vec{p} , y) = -\sum_{k=1}^K \llbracket y = y_k \rrbracket \, \log(p_k) \, ,
  \end{equation}
where $\llbracket \cdot \rrbracket$ denotes the indicator function.

In the context of ordinal classification, one might be tempted to prefer dedicated ordinal losses, such as the quadratic weighted kappa (QWK) \citep{DBLP:journals/prl/TorrePV18} or the squared Earth Mover's Distance (EMD) loss \citep{DBLP:journals/corr/HouYS16}, which are designed to produce accurate predictions while accounting for the ordinal structure of $\mathcal{Y}$. One should note, however, that accurate prediction is not the same as faithful uncertainty representation, and indeed, from a probability estimation point of view, ordinal losses of that kind provide the wrong incentive. Imagine, for example, that the probability $p( \cdot \, | \,  \vec{x}_q)$ is uniform over the three classes $y_1, y_2, y_3$. Then, CE is minimized (in expectation) by predicting exactly this distribution, whereas the $L_1$-loss, which takes the order of the classes into account, is minimized by the distribution that assigns probability 1 to $y_2$ (and hence suggests complete certainty).  

This problem is confirmed by \cite{DBLP:journals/prl/TorrePV18} and \cite{DBLP:journals/ijon/LiuFKDXLY20}, who demonstrate that ordinal losses inherently bias the predictive probability distributions towards unimodality by penalizing more distant errors at the loss level. While this approach can be effective for loss minimization, it does not necessarily promote truthful uncertainty representation and, depending on the application and type of data, may introduce an undesirable inductive bias. The literature on ordinal classification places a strong emphasis on what \cite{anderson1984regression} refers to as ``grouped continuous'' ordered categorical variables, where an inherently continuous variable, such as age, is discretized into groups \citep{DBLP:conf/cvpr/NiuZWGH16,DBLP:journals/prl/CaoMR20,DBLP:conf/cvpr/LiWYL0YWP22,DBLP:conf/caepia/YunGGBGH24}. For such kind of variables, the assumption of unimodality may still appear to be reasonable. However, Anderson also identifies a second type, termed ``assessed'' ordered categorical variables, where an \emph{assessor} provides a \emph{judgment} \citep{DBLP:conf/pkdd/HaasH22}. For this second type, the inductive bias of unimodality appears rather arbitrary. Anderson further notes that errors for assessed variables are likely to be greater. We refer to Appendix \ref{appendix:sec:comparison_prrs} for experiments demonstrating the superiority of proper scores (in particular the CE loss) for uncertainty quantification in ordinal classification compared to dedicated ordinal losses.

\section{Aleatoric and Epistemic Uncertainty}

\begin{figure}[!htbp]
  \centering
   \begin{subfigure}[t]{0.3\linewidth}
           \centering
           \includegraphics[width=\linewidth]{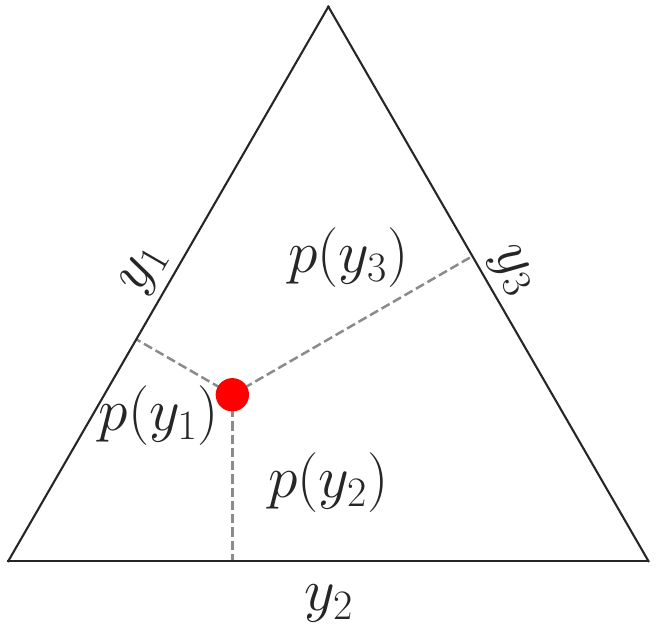}
       \end{subfigure} 
         \begin{subfigure}[t]{0.3\linewidth}
           \centering
           \includegraphics[width=\linewidth]{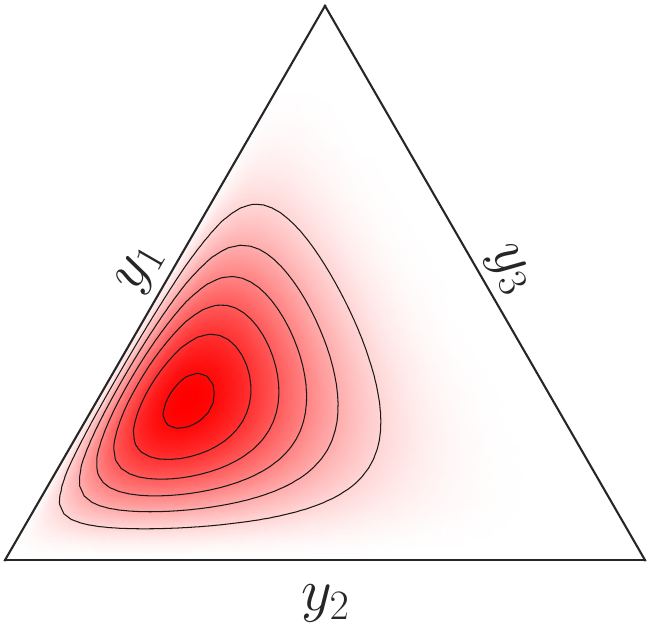}
       \end{subfigure} 
       \begin{subfigure}[t]{0.3\linewidth}
        \centering
        \includegraphics[width=\linewidth]{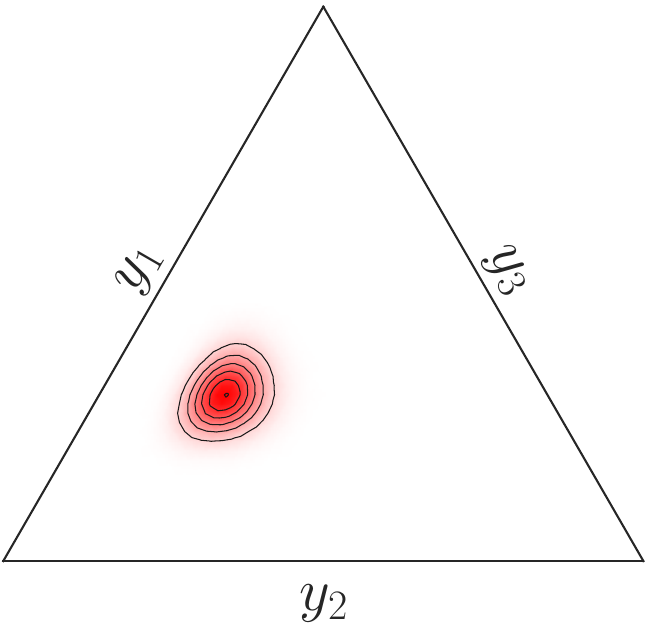}
    \end{subfigure}      
        \caption{Uncertainty awareness, illustrated on the probability simplex for $\mathcal{Y}=\{y_1,y_2,y_3\}$. From left to right: Aleatoric uncertainty without any epistemic uncerainty awareness, Bayesian representation of epistemic uncertainty in the form of a probability distribution over probability distributions, a more concentrated or ``peaked'' second-order distribution compared to the previous one.}
          \label{fig:aleatoric_epistemic_vis}
\end{figure}

A probabilistic predictor's uncertainty is purely aleatoric, as it fully commits to a single hypothesis $h$, which in turn fully commits to a single probability distribution (\ref{eq:propre}) when making a prediction. Hence, it does not represent any epistemic uncertainty about the hypothesis itself, nor about the probability (\ref{eq:propre}). 
A popular framework that caters for the representation of epistemic uncertainty on top of aleatoric uncertainty is Bayesian inference. Here, instead of committing to a single hypothesis, a prior $p(h)$ is placed over the candidates $h \in \mathcal{H}$  \citep{DBLP:conf/icml/GalG16,DBLP:conf/nips/KendallG17}.
Learning essentially consists of updating the prior distribution $p(h)$  to the posterior distribution $p(h \, | \, \mathcal{D})$ in light of the training data $\mathcal{D}$:
$$
p(h \, | \, \mathcal{D}) = \frac{p(h) \cdot p(\mathcal{D} \, | \, h ) }{p(\mathcal{D})} \propto p(h) \cdot p(\mathcal{D} \, | \, h ) \, ,
$$
where $p(\mathcal{D} \, | \, h)$ is the likelihood of the hypothesis $h$ (i.e., the probability of the data given $h$) and $p(\mathcal{D})$ is the marginal probability of the data, which serves as a normalization factor.
Intuitively, $p(h \, | \, \mathcal{D})$ captures the state of knowledge of the learner and hence its epistemic uncertainty. The more concentrated (or ``peaked'') the probability mass in a small region of $\mathcal{H}$, the less uncertain the learner is.
Since every $h \in \mathcal{H}$ produces a probabilistic prediction, the belief about the outcome $y$ is represented by a second-order probability: a probability distribution of probability distributions \citep{shaker2021ensemble} (cf.\ Figure \ref{fig:aleatoric_epistemic_vis} for an illustration).

More concretely, the predictive posterior distribution specifies the posterior probability of each outcome $y \in \mathcal{Y}$. It is defined in terms of the \emph{expected} probability $p(y \, | \, \vec{x}, h)$, where the expectation is taken with respect to the posterior distribution $p(h \, | \, \mathcal{D})$:
\begin{equation}
  \label{eq:pred_post}
  p(y \, | \, \vec{x}) = \mathbb{E}_{p(h \, | \, \mathcal{D})}[ \, p(y \, | \, \vec{x}, h) \,] =  \int_{\mathcal{H}} p(y \, | \, \vec{x}, h)\, d\, p(h \, | \, \mathcal{D})   
  \end{equation}
In practice, as computing the exact expectation is intractable, (\ref{eq:pred_post}) is commonly approximated using Monte Carlo integration techniques. In the simplest case, a finite ensemble $H = \{h_1,\ldots,h_M\}$ consisting of a set of $M$ 
 models is trained, and integration is replaced by the arithmetic average \citep{DBLP:conf/iclr/MalininPU21,DBLP:conf/ida/ShakerH20,shaker2021ensemble}:
\begin{equation}
  \label{eq:ensemble}
  p(y \, | \, \vec{x}) \approx \frac{1}{M} \sum_{m=1}^M p(y \, | \, \vec{x}, h_m)
  \end{equation}

\subsection{Entropy}
\label{sec:ent}

A principled and popular approach to measuring and separating aleatoric and epistemic uncertainty in (Bayesian) machine learning, though criticized for not satisfying certain theoretical axioms \citep{DBLP:conf/uai/WimmerSHBH23}, is based on the classical information-theoretic measure of (Shannon) entropy \citep{pmlr-v80-depeweg18a}.
Total uncertainty (TU) is hereby measured in terms of the entropy of the posterior predictive distribution:
\begin{align}\label{eq:eu_entropy}
  \text{TU}(\vec{x}) = \mathbb{H} \big[ p(y \, | \, \vec{x}) \big] = \mathbb{H} \Bigl[\mathbb{E}_{p(h \, | \, \mathcal{D})}[ \, p(y \, | \, \vec{x}, h) \,]\Bigr] \, ,
\end{align}
where the Shannon entropy of a (discrete) probability distribution $\vec{p}=(p_1, \ldots , p_K)$ is given by
$$
\mathbb{H}(\vec{p}) = - \sum_{k=1}^K p_k \cdot \log(p_k) \, .
$$
By fixing a hypothesis $h \in \mathcal{H}$, the epistemic uncertainty is essentially removed, and only aleatoric uncertainty remains. Therefore, a natural measure of aleatoric uncertainty (AU) is the conditional entropy 
(i.e., the expected entropy of $p(y \, | \, \vec{x}, h)$, with the expectation taken with regard to the posterior $p(h \, | \, \mathcal{D})$):
\begin{align}\label{eq:au_ent}
  \text{AU}(\vec{x}) = \mathbb{E}_{p(h \, | \, \mathcal{D})}\mathbb{H} \big[ \, p(y \, | \, \vec{x}, h) \, \big] =  \int_{\mathcal{H}} \, p(h \, | \, \mathcal{D}) \, \mathbb{H} \big[ \, p(y \, | \, \vec{x}, h) \, \big] \; d\,h  
  \end{align}
  Eventually, the epistemic uncertainty (EU) is measured in terms of the \emph{mutual information} between hypotheses $h$ and outcomes $y$, which is obtained as the difference between total and aleatoric uncertainty:
\begin{equation}
  \label{eq:epistemic_entropy}
\underbrace{\mathbb{I} \big[ y,h \, | \, \vec{x},\mathcal{D} \big]}_{\text{EU}(\vec{x})} 
\, = \, \underbrace{\mathbb{H} \big[p(y \, | \, \vec{x}) \big]}_{\text{TU}(\vec{x})} 
\, - \, \underbrace{\mathbb{E}_{p(h \, | \, \mathcal{D})}\mathbb{H} \big[ \, p(y \, | \, \vec{x}, h) \, \big]}_{\text{AU}(\vec{x})}
\end{equation}
In practice, approximations of (\ref{eq:eu_entropy}) and (\ref{eq:au_ent}), and hence of (\ref{eq:epistemic_entropy}), are again obtained by replacing integration over $\mathcal{H}$ with averaging over a finite ensemble:
\begin{align}
  \text{TU}(\vec{x}) &\approx \mathbb{H}\Biggl[ \frac{1}{M} \sum_{m=1}^M p(y \, | \, \vec{x},h_m) \Biggr] \\
  \text{AU}(\vec{x})  &\approx \frac{1}{M} \sum_{m=1}^M \mathbb{H} \big[ \, p(y \, | \, \vec{x}, h_m) \, \big]  
  \end{align}

Since labels in ordinal classification are of categorical nature, the entropy-based approach is also applicable to probabilistic ordinal classification. Note, however, that it does not take the ordinal structure into account and is invariant to permutations of the probability degrees \citep{DBLP:journals/ml/HullermeierW21,haas4965714uncertainty}.

\subsection{Variance}
\label{sec:variance}
Another principled approach to separating aleatoric and epistemic uncertainty, also originally proposed by \cite{pmlr-v80-depeweg18a}, is based on the \emph{law of total variance}. This measure is conceptually similar to the entropy measure defined in (\ref{eq:epistemic_entropy}), making use of the variance $\mathbb{V}$ as the base measure.
The total uncertainty can hereby be decomposed into its aleatoric and epistemic parts as follows:
\begin{equation}
  \label{eq:law_total_variance}
  \underbrace{\mathbb{V}_{p(y \, | \, \vec{x}, \mathcal{D})}[y  \, | \, \vec{x}]}_{\text{TU}(\vec{x})} = \ \underbrace{\mathbb{V}_{p(h \, | \, \mathcal{D})}\bigl[\mathbb{E}_{p( y \, | \, \vec{x}, h )}[y  \, | \, \vec{x}]\bigr]}_{\text{EU}(\vec{x})} +  \underbrace{\mathbb{E}_{p(h \, | \, \mathcal{D})} \bigl[ \mathbb{V}_{p(y \, | \, \vec{x}, h)}[y  \, | \, \vec{x}] \bigr]}_{\text{AU}(\vec{x})}
\end{equation}
However, the above decomposition, is primarily applicable to numerical targets $y$, for which variance is well-defined, but not for categorical or ordinal targets.
Therefore, this alternative measure has so far been primarily used for quantifying uncertainty in regression tasks, where the target variable is continuous or integer-valued \citep{DBLP:conf/iclr/MalininPU21}. 

In practice, ordinal targets $y_1, \ldots , y_K$ are often encoded in terms of numbers $1, \ldots, K$, and thereby embedded in the metric space $(\mathbb{R}, |\cdot| )$. Obviously, this embedding is debatable, as it postulates equal distances between all neighbored ordinal categories\,---\,an assumption that, even if acceptable as an approximation in some cases, is disputable in general. Nevertheless, if one is willing to accept this assumption, the measures in (\ref{eq:law_total_variance}) can be computed. Their ensemble-based approximations are given as follows:
\begin{align}
  \text{AU}(\vec{x}) & \approx  \frac{1}{M} \sum_{m=1}^M   \sum_{k=1}^K p( k \, | \, \vec{x}, h_m ) \cdot (k - \mu_m)^2  \, , \\
   \text{EU}(\vec{x}) & \approx \frac{1}{M} \sum_{m=1}^M \big [ \mu - \mu_m \big ]^2 \, , \\
   \text{TU}(\vec{x}) & \approx \sum_{k=1}^K  p( k \, | \, \vec{x} ) \cdot (k - \mu)^2  \, ,
\end{align}
with
$$
\mu = \frac{1}{M} \sum_{m=1}^M \mu_m \, , \quad \mu_m = \sum_{k=1}^K p( k \, | \, \vec{x}, h_m ) \cdot k \quad (m=1, \ldots , M) \, .
$$

As a dispersion measure, variance takes the distance or dispersion of probability mass into account. Thus, compared to entropy, it has the advantage of not being invariant to permutation of probability degrees. However, as already mentioned, the assumption of equal distances in ordinal classification remains disputable.

\section{Uncertainty Measures through Binary Reduction}
\label{sec:bin_decomp}

So far, we considered probabilistic multinomial classifiers $h: \mathcal{X} \rightarrow \mathbb{P}(\mathcal{Y})$ with $\mathcal{Y} = \{ y_1, \ldots , y_K \}$. Such classifiers produce predictions 
$$
p \left( \cdot \, | \, \vec{x} , h \right) = (p(y_1 \, | \, \vec{x}, h), \ldots , p( y_K \, | \, \vec{x}, h)) \, , 
$$
where $p(y_k \, | \, \vec{x},h)$ denotes the probability of the class $y_k$ predicted by $h$ for the query instance $\vec{x}$. For any classifier $h$, let $h^{(k)}: \mathcal{X} \rightarrow \mathbb{P}(\{0,1\})$ denote the (derived) binary classifier that predicts the Bernoulli distribution 
$$
p\left(\cdot \, | \, \vec{x} , h^{(k)} \right) = \left(p \big( 0 \, | \, \vec{x}, h^{(k)} \big), p \big(1 \, | \, \vec{x}, h^{(k)} \big) \right) \, ,
$$ 
with $p(1 \, | \, \vec{x}, h^{(k)}) = p(y_k \, | \, \vec{x}, h)$ and $p( 0 \, | \, \vec{x}, h^{(k)}) = \sum_{1 \leq i \neq k \leq K} p(y_i \, | \, \vec{x}, h)$. In other words, $h^{(k)}$ predicts whether class $y_k$ will occur as an outcome or not; it treats this class as positive and all other classes as negative, and adopts the probabilities for these two cases from the probabilities predicted by $h$.

Obviously, all uncertainty measures introduced in the previous section can also be computed for the binary case ($K=2$). If $U$ is any such measure, then we denote by $U^{(k)}$ the measure that is obtained by replacing the multinomial distributions $p(\cdot \, | \, \vec{x}, h)$ with the binary distributions $p(\cdot \, | \, \vec{x} , h^{(k)})$. 
In particular, 
\begin{align*} 
  \text{TU}^{(k)}_{\mathbb{U}}(\vec{x}) &  =  \; \mathbb{U} \Bigl[\mathbb{E}_{p(h \, | \, \mathcal{D})} \left[ \, p( \cdot  \, | \, \vec{x}, h^{(k)}) \, \right]\Bigr]  &  \approx \; \mathbb{U} \left[ \frac{1}{M} \sum_{m=1}^M p( \cdot  \, | \, \vec{x}, h^{(k)}_m) \right] \, ,\\
   \text{AU}^{(k)}_{\mathbb{U}}(\vec{x}) &  = \; \mathbb{E}_{p(h \, | \, \mathcal{D})}\mathbb{U} \big[ \, p( \cdot \, | \, \vec{x}, h^{(k)}) \, \big] &  \approx  \; \frac{1}{M} \sum_{m=1}^M \mathbb{U}  \left[ p( \cdot  \, | \, \vec{x}, h^{(k)}_m) \right] \, ,
\end{align*}
where $\mathbb{U} = \mathbb{H}$ (entropy-based) or $\mathbb{U} = \mathbb{V}$ (variance-based).

Given label-wise measures of that kind, \cite{labelwise} propose to define overall measures of uncertainty for the original multinomial case as follows:
\begin{equation}\label{eq:ucat}
U^{cat}_{\mathbb{U}}(\vec{x}) = \sum_{k=1}^K U^{(k)}_{\mathbb{U}}(\vec{x}) 
\end{equation}
This approach is inspired by binary decomposition techniques for reducing a multinomial classification problem to several binary problems \citep{allw_rm01}, in particular the one-versus-rest decomposition \citep{DBLP:journals/jmlr/RifkinK03}. Here, the same idea of reduction is applied to uncertainty measures. Intuitively, the uncertainty in the answer to the question ``Which class will occur?'' is defined as an aggregation of the uncertainties in the questions ``Will class $y_k$ occur, yes or no?'', $k=1, \ldots, K$. One obvious advantage of this approach is that measures like variance can be used in a theoretically sound manner, because variance is well defined for the binary case\,---\,as opposed to the multinomial case. Another advantage is that uncertainty quantification becomes somewhat more nuanced, for example because the overall uncertainty can be attributed to specific class labels \citep{labelwise}.  

\begin{figure}[!htbp]
  \centering
   \begin{subfigure}[t]{0.24\linewidth}
           \centering
           \includegraphics[width=\linewidth]{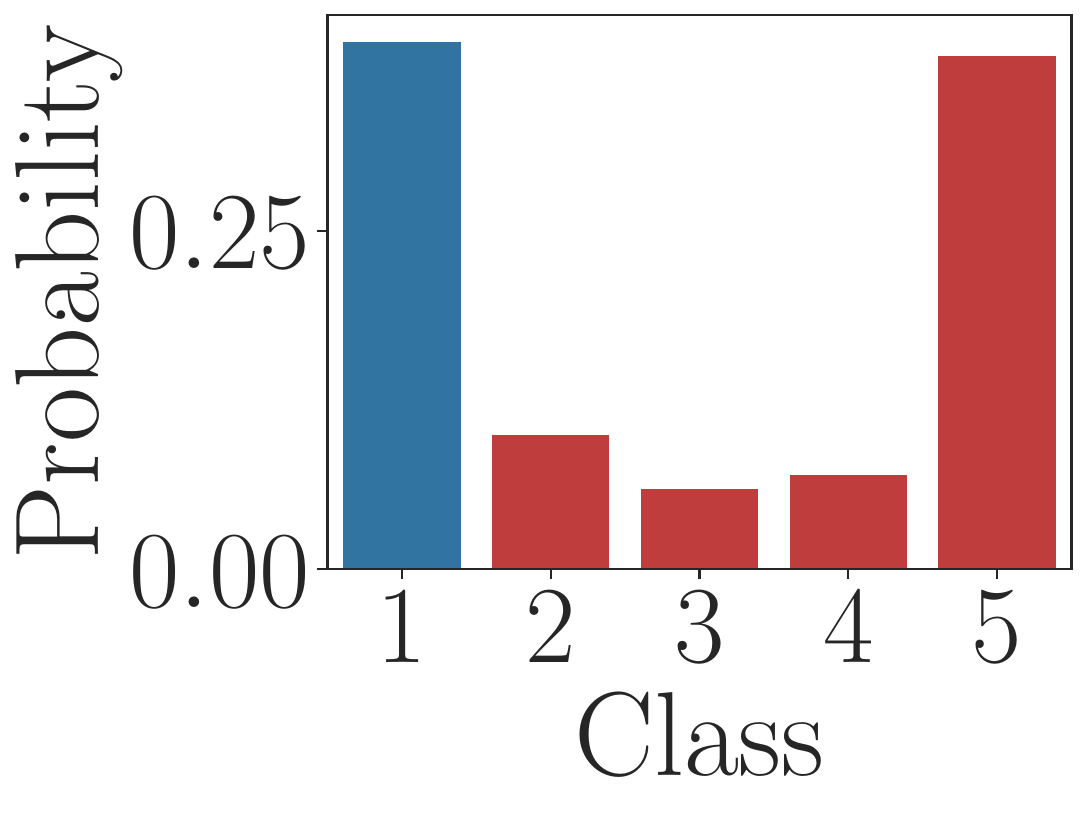}           \captionsetup{justification=centering}
           \caption{Class 1 vs.\ classes 2 to 5.}
       \end{subfigure} 
         \begin{subfigure}[t]{0.24\linewidth}
           \centering
           \includegraphics[width=\linewidth]{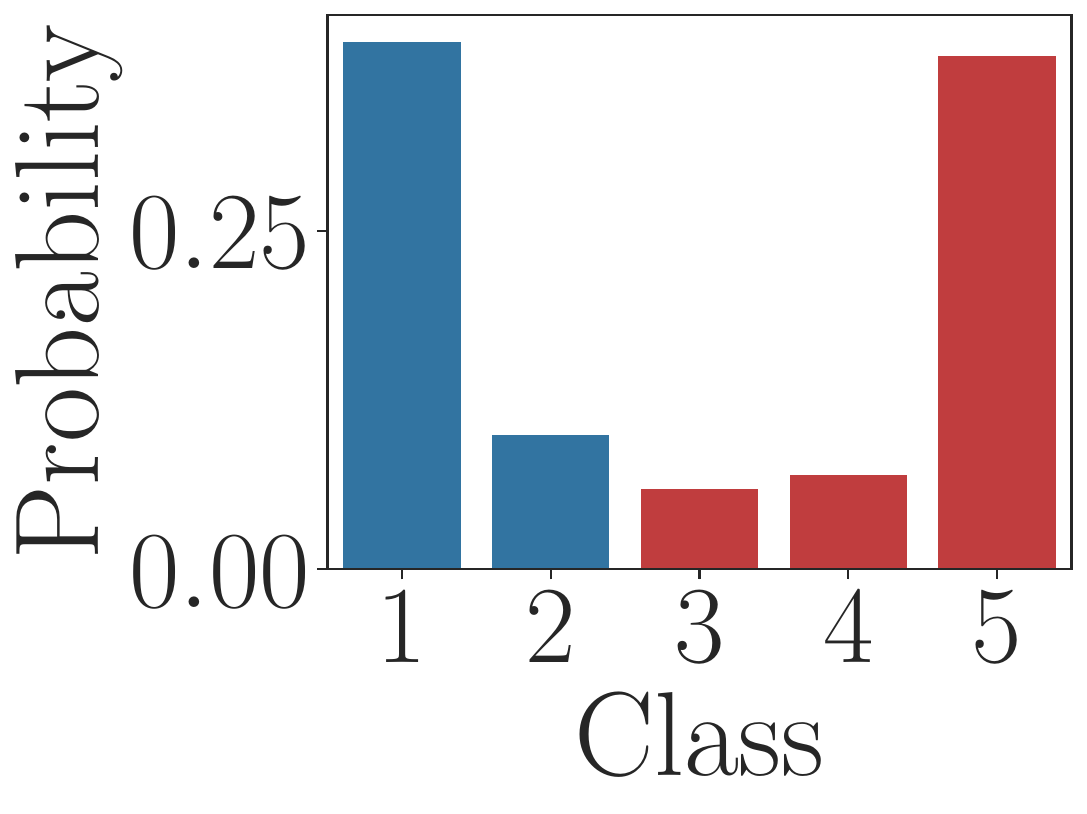}
           \captionsetup{justification=centering}
           \caption{Classes 1 and 2 vs.\ classes 3 to 5.}
       \end{subfigure}
         \begin{subfigure}[t]{0.24\linewidth}
           \centering
           \includegraphics[width=\linewidth]{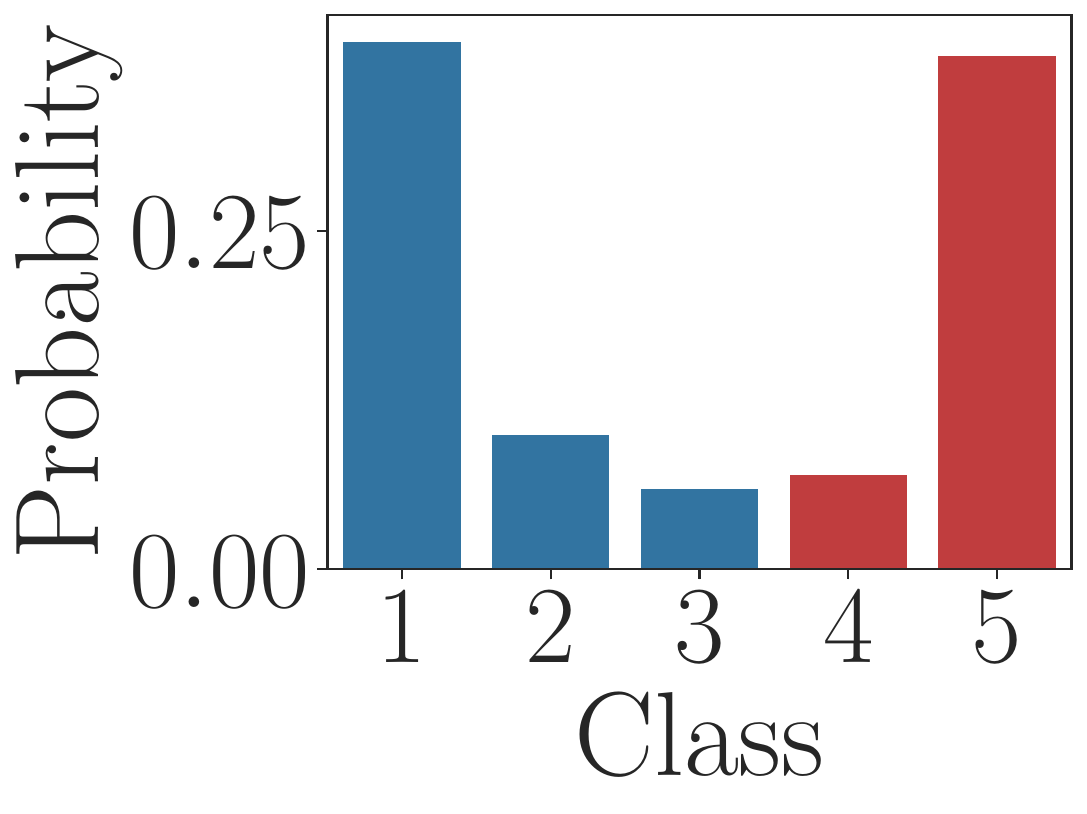}
            \captionsetup{justification=centering}
           \caption{Classes 1 to 3 vs.\ classes 4 and 5.}
  \end{subfigure}  
   \begin{subfigure}[t]{0.24\linewidth}
           \centering
           \includegraphics[width=\linewidth]{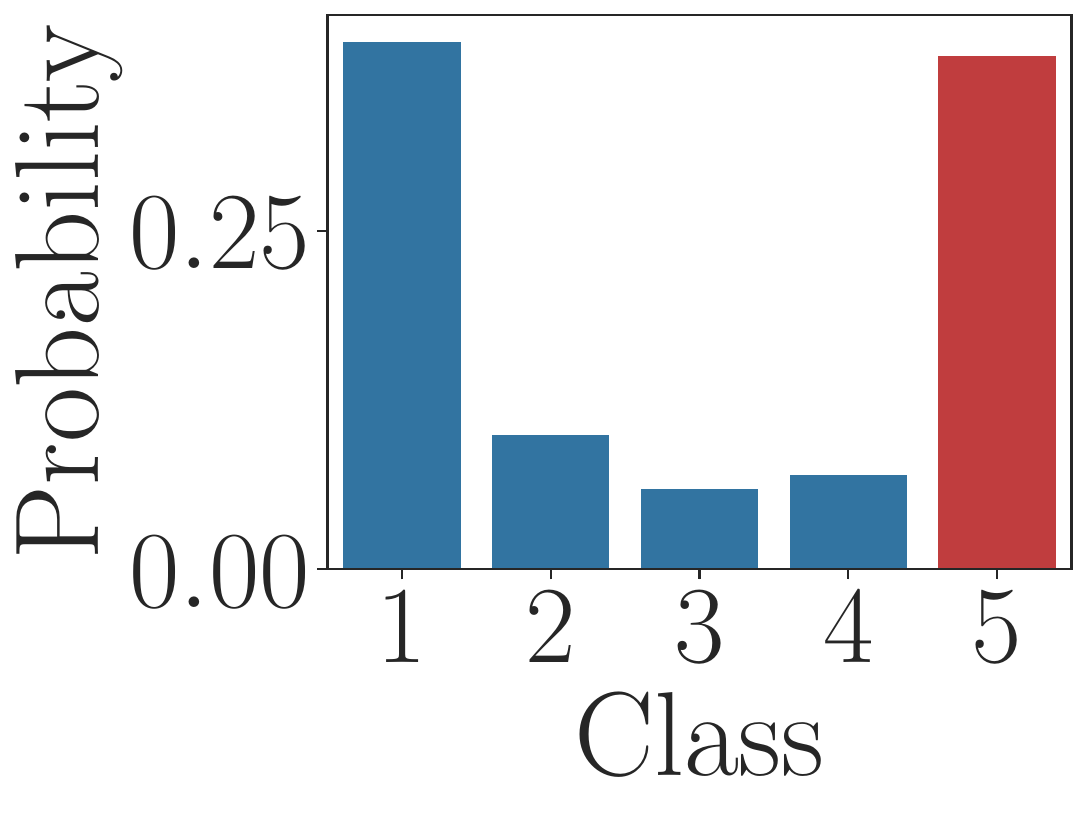}
            \captionsetup{justification=centering}
           \caption{Classes 1 to 4 vs.\ class 5.}
       \end{subfigure}       
          \caption{Example of an OCS decomposition with five classes.}
           \label{fig:bin_splits}
\end{figure}

A one-versus-rest decomposition is still invariant toward the permutation of class labels and does not take the ordinal structure of the problem into account. In particular, with variance as a base measure, the overall measure of uncertainty does not capture any notion of dispersion or distance.
Thus, one-versus-rest decomposition seems to be appropriate for the multinomial but not for the ordinal case.
Indeed, the arguably most natural reduction in the case of ordinal classification is not achieved by means of a one-versus-rest decomposition, but rather through binary splits of the ordinal scale, separating a lower part $\{ y_1, \ldots , y_k \}$ of the scale from an upper part $\{y_{k+1},\ldots,y_K\}$~\citep{DBLP:conf/ecml/FrankH01,DBLP:conf/nips/LiL06}. In the following, we will refer to this approach as the \emph{order-consistent split} (OCS) reduction. Again, this reduction allows for solving the original classification task\,---\,the correct ordinal category can be recovered from consistent answers to $K-1$ queries of the form ``Does the class label exceed level $k$?'', $k=1, \ldots , K-1$ (see Figure \ref{fig:bin_splits} for an illustration). Moreover, it does take the ordinal structure of the problem into account~\citep{DBLP:journals/ijdmmm/HuhnH08}.

Correspondingly, for any classifier $h$, let $h^{(\leq k)}: \mathcal{X} \rightarrow \mathbb{P}(\{0,1\})$ now denote the (derived) binary classifier that predicts 
$$
p \left( \cdot \, | \, \vec{x}, h^{(\leq k)} \right) = \left(p \big( 0 \, | \, \vec{x}, h^{(\leq k)} \big), p \big(1 \, | \, \vec{x}, h^{(\leq k)} \big) \right)
$$ 
with 
\begin{align*}
p\left(  0 \, | \, \vec{x}, h^{(\leq k)} \right) & = \sum_{1 \leq i \leq k} p(y_i \, | \, \vec{x}, h) \\ 
p \left(1 \, | \, \vec{x}, h^{(\leq k)} \right) & = \sum_{k+1 \leq i \leq K}  p(y_i \, | \, \vec{x}, h) \, .
\end{align*}
Thus, $h^{(\leq k)}$ treats the classes $y_1, \ldots, y_k$ as negative and the classes $y_{k+1}, \ldots , y_K$ as positive, and again adopts the probabilities for these two cases from the probabilities predicted by $h$.
Similar to above, we obtain 
\begin{align*} 
  \text{TU}^{(\leq k)}_{\mathbb{U}}(\vec{x}) &  =  \; \mathbb{U} \Bigl[\mathbb{E}_{p(h \, | \, \mathcal{D})} \left[ \, p( \cdot  \, | \, \vec{x}, h^{(\leq k)}) \, \right]\Bigr]  & \approx \;  \mathbb{U} \left[ \frac{1}{M} \sum_{m=1}^M p( \cdot  \, | \, \vec{x}, h^{(\leq k)}_m) \right] \, ,\\
   \text{AU}^{(\leq k)}_{\mathbb{U}}(\vec{x}) & = \; \mathbb{E}_{p(h \, | \, \mathcal{D})}\mathbb{U} \big[ \, p( \cdot \, | \, \vec{x}, h^{(\leq k)}) \, \big] & \approx  \; \frac{1}{M} \sum_{m=1}^M \mathbb{U}  \left[ p( \cdot  \, | \, \vec{x}, h^{(\leq k)}_m) \right] \, ,
\end{align*}
where $\mathbb{U} = \mathbb{H}$ (entropy-based) or $\mathbb{U} = \mathbb{V}$ (variance-based).
Moreover, the overall measures of uncertainty for the original ordinal problem is given by
\begin{equation}\label{eq:uord}
U^{ord}_{\mathbb{U}}(\vec{x}) = \sum_{k=1}^{K-1} U^{(\leq k)}_{\mathbb{U}}(\vec{x}) \, . 
\end{equation}

For the special case of total uncertainty, the above construction has already been considered by \cite{haas4965714uncertainty}, who argue for its suitability in the context of ordinal classification. They show, for example, that both $\text{TU}^{ord}_{\mathbb{H}}$ and $\text{TU}^{ord}_{\mathbb{V}}$ are maximized by the bimodal distribution that assigns probability $1/2$ to the extreme classes $1$ and $K$, respectively, and not by the uniform distribution.

\section{Comparative Analysis of the Measures}

\begin{figure}[!htbp]
  \centering
    \begin{subfigure}[t]{0.3\linewidth}
      \centering
      \includegraphics[width=\linewidth]{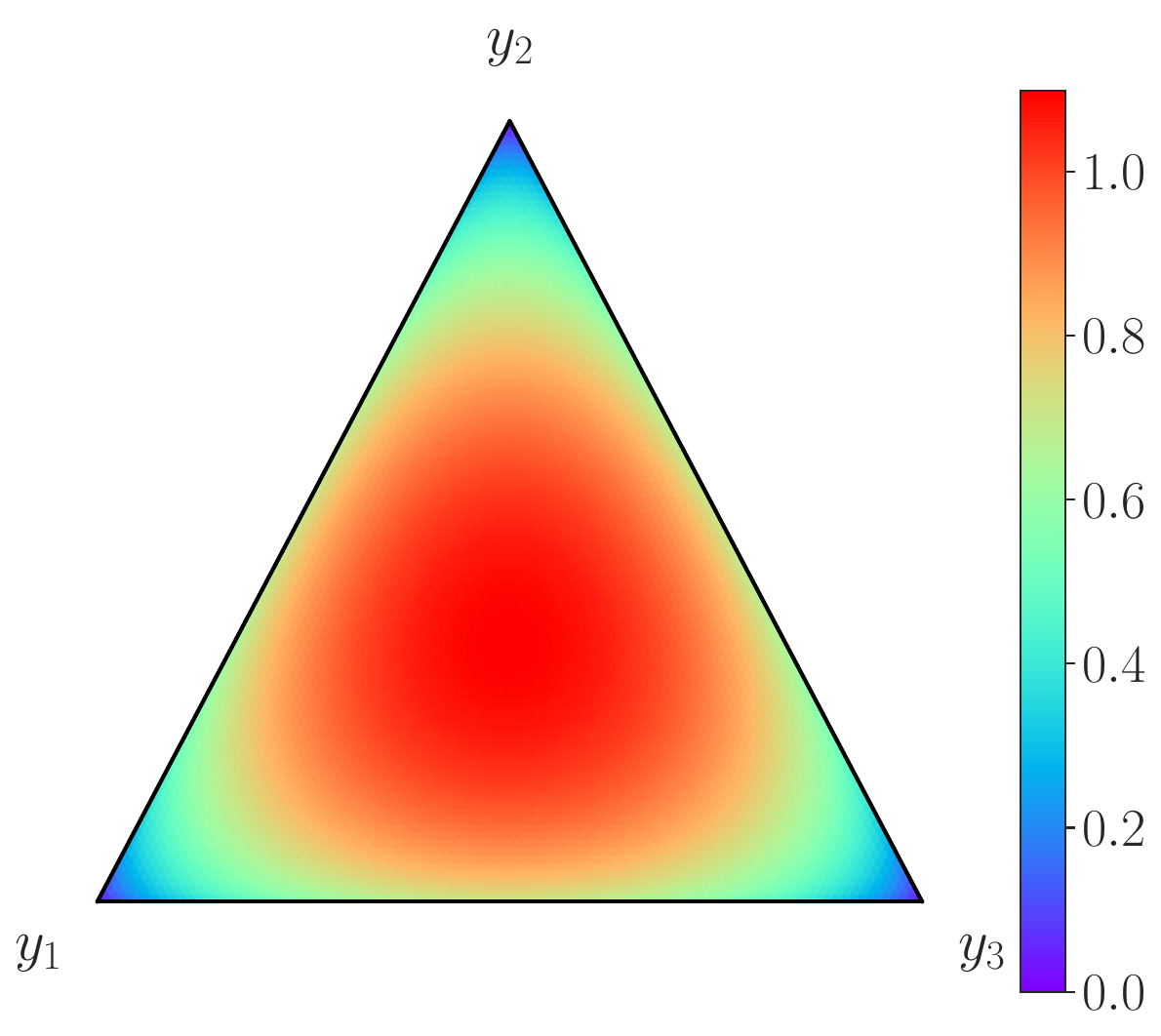}
      \subcaption{Entropy}
  \end{subfigure} 
    \begin{subfigure}[t]{0.3\linewidth}
      \centering
      \includegraphics[width=\linewidth]{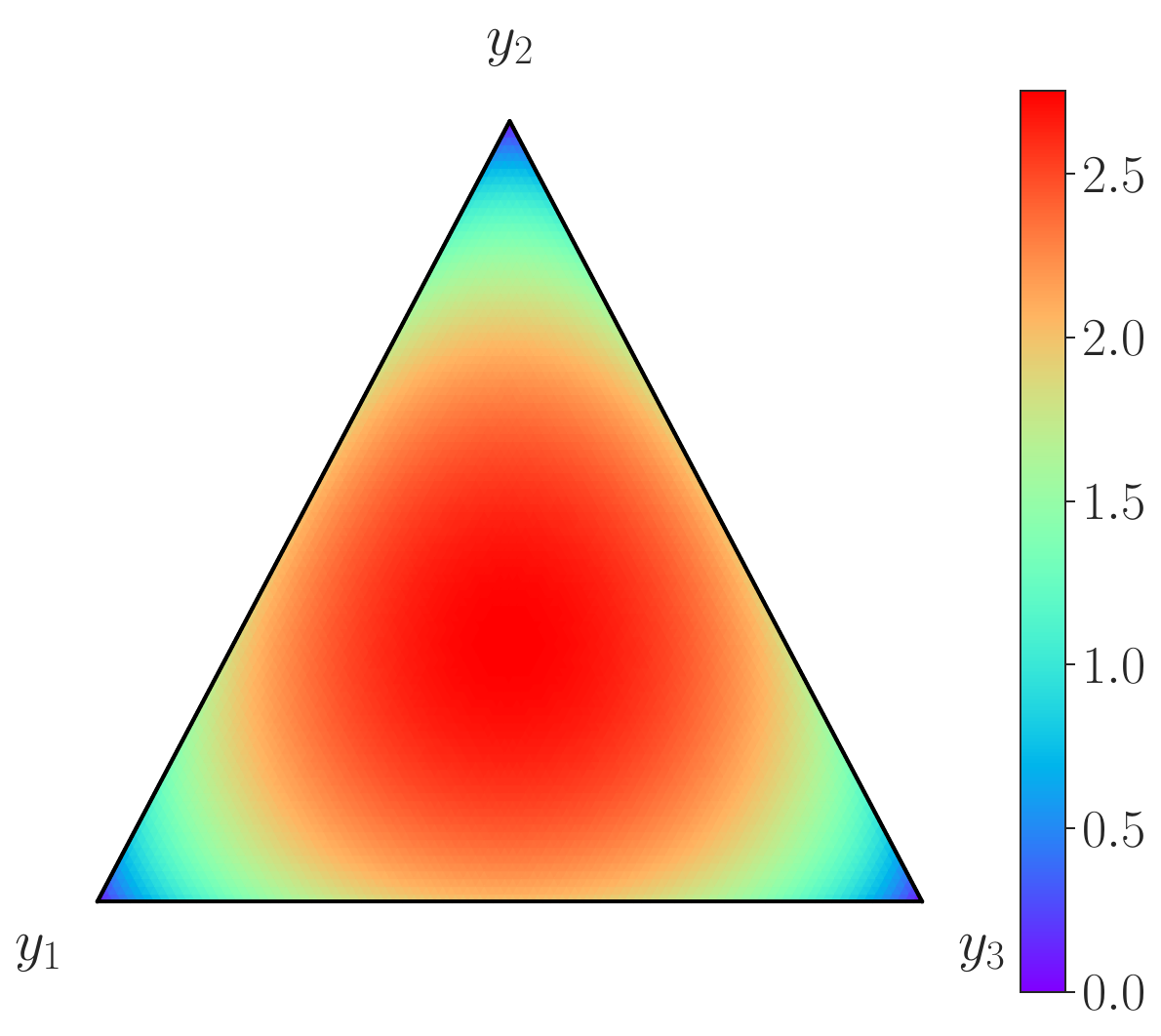}
      \subcaption{Label-wise binary entropy (one vs. rest)}
  \end{subfigure}  
  \begin{subfigure}[t]{0.3\linewidth}
  \centering
  \includegraphics[width=\linewidth]{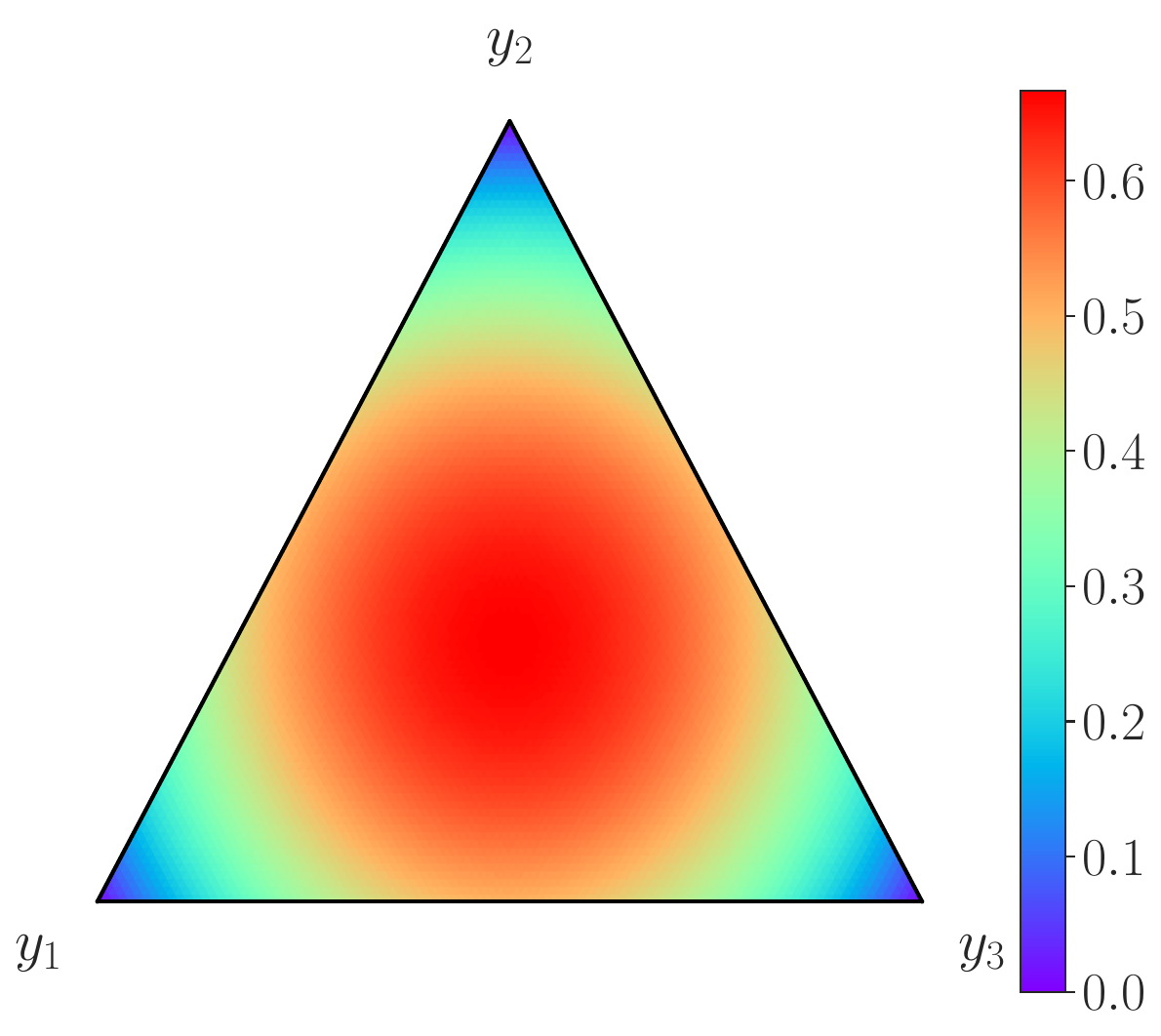}
  \subcaption{Label-wise  binary variance (one vs. rest)}
  \end{subfigure}  
  \begin{subfigure}[t]{0.3\linewidth}
    \centering
    \includegraphics[width=\linewidth]{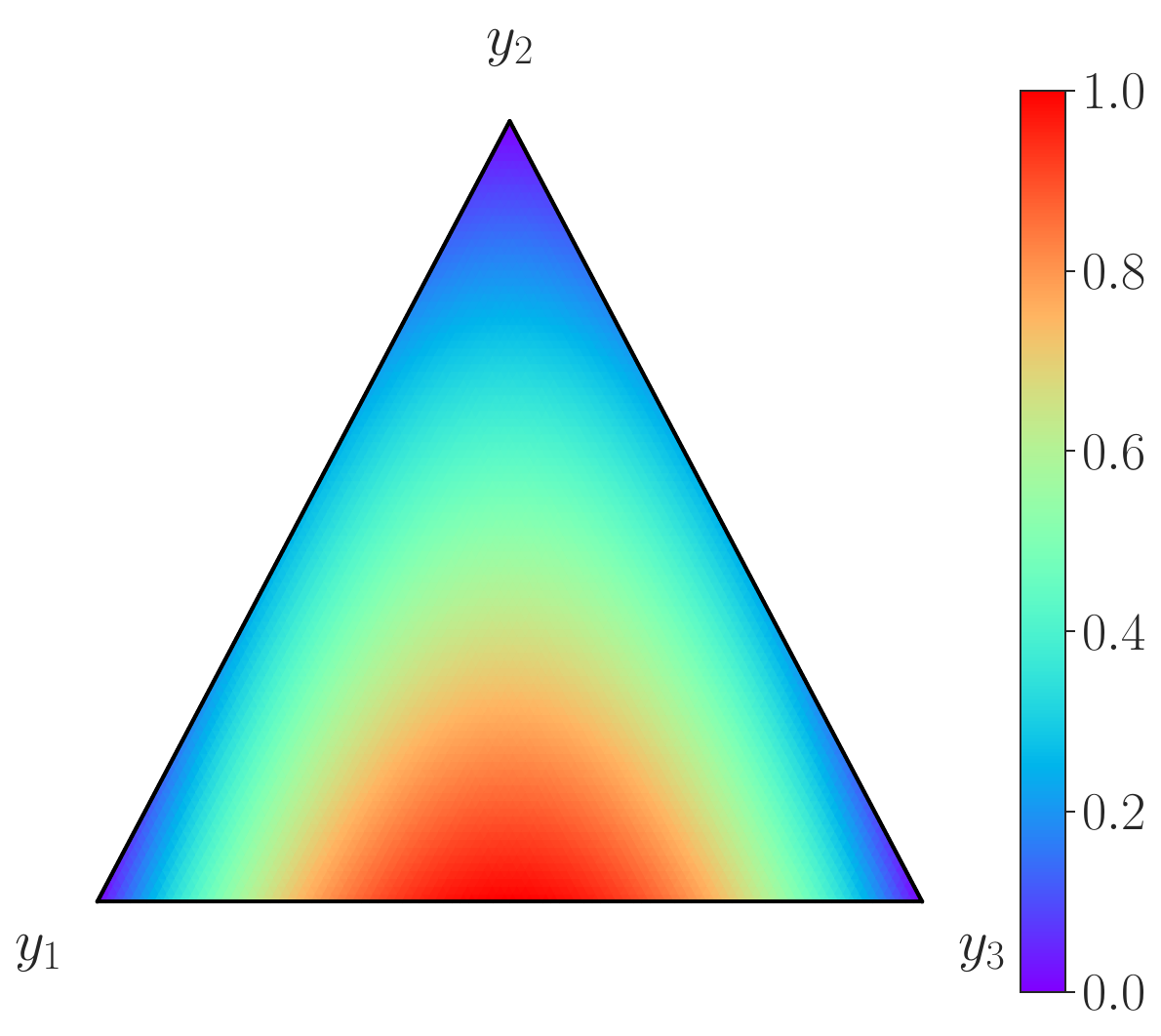}
    \subcaption{Variance}
    \label{subfig:var}
\end{subfigure} 
  \begin{subfigure}[t]{0.3\linewidth}
    \centering
    \includegraphics[width=\linewidth]{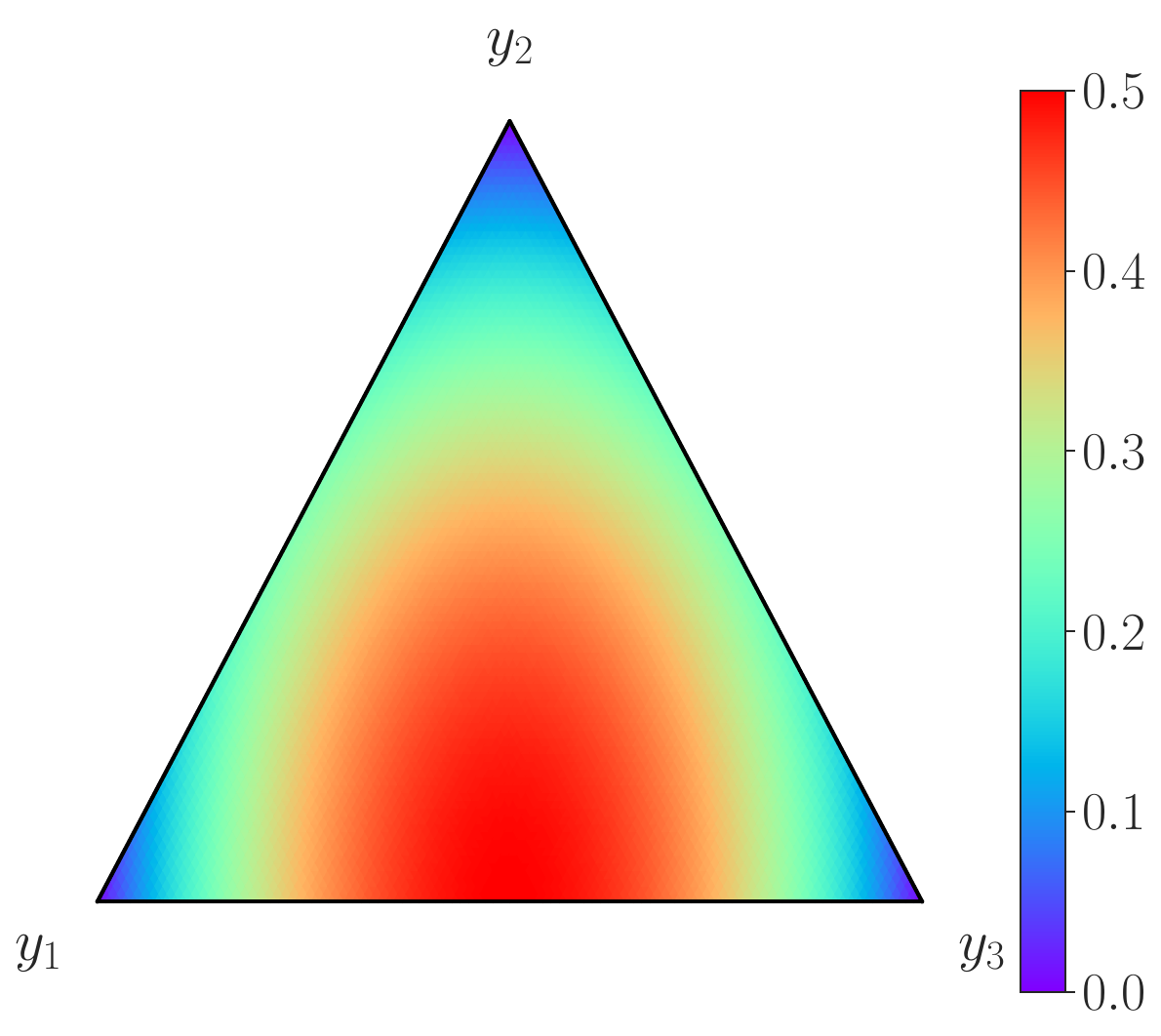}
    \subcaption{Ordinal binary variance}
    \label{subfig:ord_bin_var}
\end{subfigure}  
\begin{subfigure}[t]{0.3\linewidth}
\centering
\includegraphics[width=\linewidth]{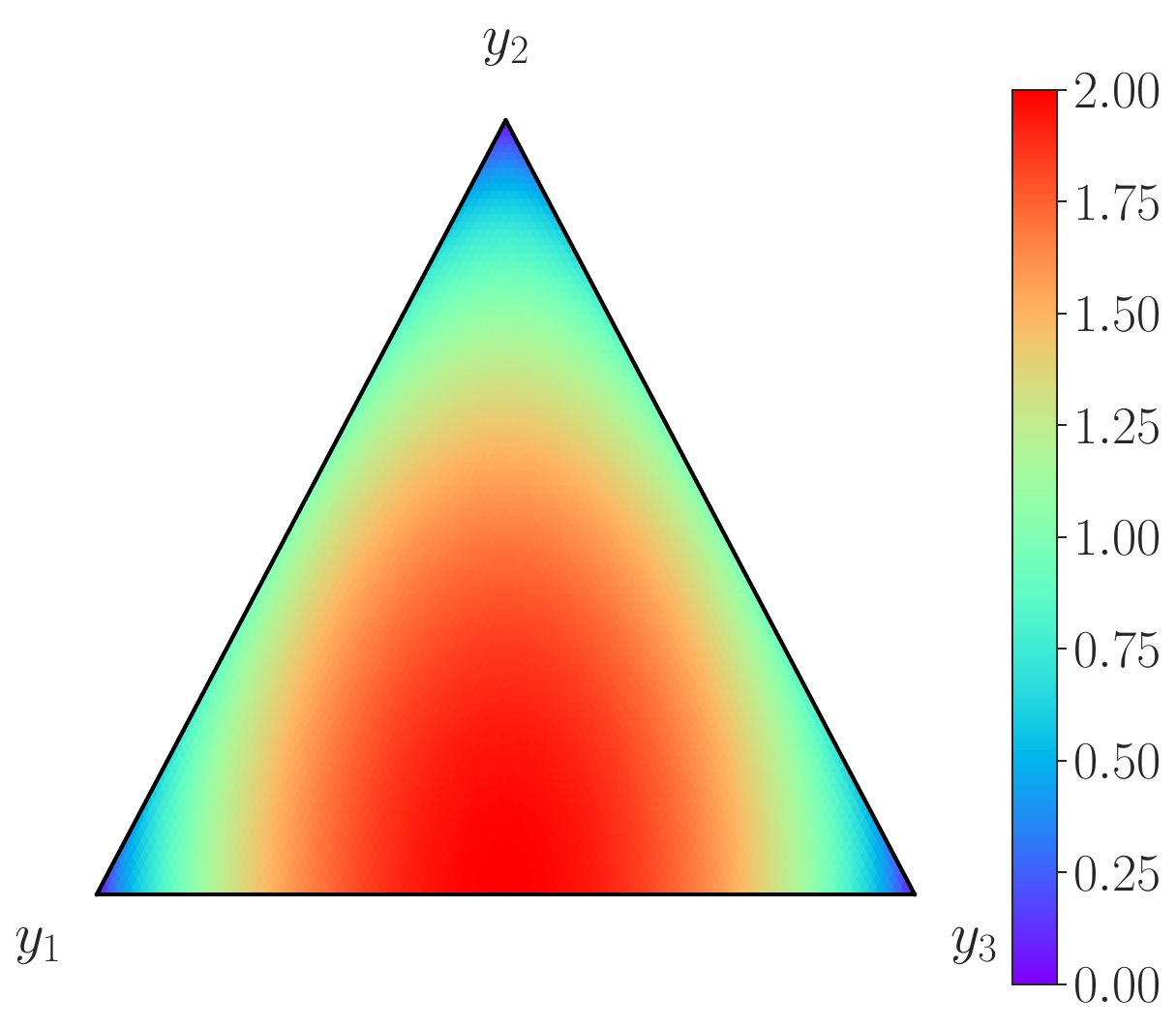}
\subcaption{Ordinal binary entropy}
\label{subfig:ord_bin_ent}
\end{subfigure}  
\caption{(Total) uncertainty heatmaps for the different uncertainty measures using the probability simplex over $\mathcal{Y}=\{y_1,y_2,y_3\}$.}
\label{fig:prob_simplex}
\end{figure}

In this section, we conduct a brief comparative analysis of the different measures. Figure \ref{fig:prob_simplex} illustrates the measured (total) uncertainty of the above-presented uncertainty measures via heatmaps using the probability simplex over $\mathcal{Y}=\{y_1,y_2,y_3\}$. As one can clearly see, the first row of measures is maximized by the uniform distribution, where all probability mass is equally distributed among all three classes and radiates from the center of the simplex. In contrast, the measures in the second row are maximized by the extreme bimodal distribution, with all probability mass equally concentrated at the extreme classes (in this case at $y_1$ and $y_3$). In this case, the uncertainty radiates from the center between $y_1$ and $y_3$.
While the heatmaps for entropy, label-wise binary entropy, and variance look quite similar, there is a bigger difference between variance (cf.\ Figure \ref{subfig:var}) and the OCS decompositions for variance and entropy (cf.\ Figures \ref{subfig:ord_bin_var} and \ref{subfig:ord_bin_ent}). Variance appears to have a very strong focus on the extreme bimodal distribution when it comes to uncertainty quantification. For probability mass moving towards the uniform distribution, it significantly measures less uncertainty than the OCS decompositions. Thus, one could conclude that the OCS reductions might strike a better balance and can be seen as standing between uncertainty measures maximized by the uniform distribution and strict dispersion measures like variance.

When comparing the uncertainties of the OCS decompositions, the decomposition with variance as the base measure (cf.\ Figure \ref{subfig:ord_bin_var}) appears slightly less extended toward the uniform distribution than the decomposition with entropy as the base measure (cf.\ Figure \ref{subfig:ord_bin_ent}).
Since ordinal classification lies between regression and nominal classification, it can be hypothesized that OCS-decomposition-based uncertainty measures are well-suited for uncertainty quantification in this context. These measures could effectively address two key aspects of uncertainty quantification: they may enhance the exact hit rate by indicating uncertainty for uniform distributions, and they may reduce error distances by indicating uncertainty in extreme bimodal cases. In contrast, other measures tend to focus primarily on one of these aspects.

\section{Experiments on Ordinal Benchmark Datasets}
\label{sec:experiments}

In the following sections, we evaluate the approaches described above for disentangling aleatoric and epistemic uncertainty using common tabular ordinal benchmark datasets. Our focus is on how effectively the different measures enhance predictive performance and decision-making, taking into account standard ordinal classification metrics. Specifically, we assess the measures' effectiveness in error detection through rejection-based experiments and evaluate their performance in out-of-distribution detection.\footnote{The source code for the experiments is made avaliable at \url{https://github.com/stefanahaas41/ordinal-aleatoric-epistemic-uncertainty}}

\subsection{Experimental Setup}

To approximate Bayesian inference, we create ensembles consisting of 10 independent gradient boosted trees (GBT) \citep{friedman2001greedy} (refer to Appendix \ref{asec:prrs_mlp} for additional experimental results using ensembles of Multi-Layer Perceptrons (MLPs) instead). We use GBTs rather than deep neural networks as they provide state-of-the-art performance on tabular datasets \citep{DBLP:conf/nips/GrinsztajnOV22,DBLP:journals/inffus/Shwartz-ZivA22}, and tabular datasets are prevalent in high-stakes settings like finance or medicine. Hence, GBTs are also highly relevant for practitioners. Furthermore, unlike Random Forests, they also enable flexible usage of loss functions, including proper scoring rules such as cross-entropy loss \citep{gneiting2007strictly}.

According to \cite{DBLP:conf/iclr/MalininPU21}, we set the \emph{subsample} rate to 0.5 to induce stochasticity in the sequential training process and eventually in the resulting trees.
In the context of gradient boosting, subsampling refers to using a subset of the training data to train each individual tree in the ensemble. All other parameters are left with the default values. Concretely, we use LightGBM \citep{DBLP:conf/nips/KeMFWCMYL17} as a fast and popular gradient boosting library for our ensemble implementations with the cross-entropy (CE) loss for multi-class classification (refer to Appendix \ref{asec:add_exp_gbts} for additional experimental results using other popular GBT libraries).
This approach enables us to obtain conditional probability distributions $p(y \, | \, \vec{x_q})$, which serve as the foundation for evaluating various uncertainty measures.

Table \ref{tab:ord_benchmark} presents some attributes of the twenty-three ordinal benchmark datasets utilized for our evaluation \citep{kelly2023uci,vanschoren2014openml}. These datasets are widely recognized within the realm of ordinal classification research and are characterized by variability in size, number of features, and classes, offering a robust foundation for a thorough assessment of the presented uncertainty measures.

\begin{table}[!htbp]
  \caption{Twenty-three common ordinal benchmark datasets used for evaluating the different uncertainty measures.}
  \label{tab:ord_benchmark}
   \begin{tabular}{l|cccc}
   \toprule
  Dataset  & \# instances & \# features & \# classes\\
  \midrule
  Grub Damage & 155 & 8 & 4\\
  Obesity & 2,111 &  16 & 7  \\
  CMC & 1,473 & 9 & 3\\
  New Thyroid  & 215 & 5 & 3 \\
  Balance Scale  & 625 & 4 & 3\\
  Automobile  & 205 & 25 & 7\\
  Eucalyptus  & 736 & 19 & 5 \\
  TAE  & 151 & 5 & 3 \\
  Heart (CLE)  & 303 & 13 &  5 \\
  SWD   & 1,000 & 10 &  4 \\
  ERA   & 1,000 & 4 & 9\\
  ESL   & 488 & 4 & 9\\
  LEV  & 1,000 & 4 & 5 \\
  Red Wine  & 1,599 & 11 & 6 \\
  White Wine   & 4,898 & 11 & 7 \\
  Triazines & 186 & 60 & 5 \\
  Machine CPU & 209 & 6 & 10 \\
  Auto MPG & 392 & 7 & 10 \\
  Boston Housing & 506 & 13 & 5\\
  Pyrimidines & 74 & 27 & 10 \\
  Abalone & 4,177 & 8 & 10 \\
  Wisconsin Breast Cancer & 194 & 32 & 5 \\
  Stocks Domain & 950 & 9 & 5 \\
  \bottomrule
  \end{tabular}
\end{table}
In terms of evaluating the predictive performance in relation to the quantified uncertainties, we rely on the two most popular metrics in the realm of ordinal classification: Accuracy (ACC) (or its inverse misclassification rate (MCR) or mean zero-one error (MZE)) and mean absolute error (MAE) \citep{DBLP:journals/tkde/GutierrezPSFH16,DBLP:conf/ai/GaudetteJ09}.
Another very popular performance measure for ordinal classification is the quadratic weighted kappa (QWK) \citep{DBLP:journals/prl/TorrePV18,cohen1960coefficient}. However, QWK poses some challenges in rejection-based evaluation in uncertainty quantification when the required confusion matrix becomes sparse, which is why we exclude it here. The mean squared error (MSE) is also commonly used when evaluating ordinal classification, emphasizing larger error distances \citep{DBLP:conf/ai/GaudetteJ09,DBLP:conf/isda/BaccianellaES09}.
There are also some dedicated performance measures for imbalanced ordinal classification, such as average MAE (AMAE) \citep{DBLP:conf/isda/BaccianellaES09} and maximum MAE (MMAE) \citep{DBLP:journals/ijon/Cruz-RamirezHSG14}. However, since our focus is on general uncertainty quantification and not specifically on uncertainty quantification for imbalanced data, we do not consider them here.
Consequently, we believe that ACC and MAE best capture the fundamental trade-off in ordinal classification between exact hit rate and error distance minimization.

In general, we use 10-fold cross-validation for all our experiments to ensure robust and fair comparison of all uncertainty measures. 
In terms of preprocessing the datasets for the experimental evaluation, all categorical features were one-hot encoded and the ordinal labels $y_1, \ldots, y_K$ were integer encoded from $1, \ldots, K$.

\subsection{Accuracy-Rejection Curves}
\label{subsec:rejection_curves}
A common approach for evaluating the quality of uncertainty quantification methods are \emph{accuracy-rejection curves}, which depict the accuracy of a predictor as a function of the percentage of rejections \citep{DBLP:journals/jmlr/NadeemZH10, DBLP:journals/tfs/HuhnH09}. A predictor that is allowed to abstain from predicting a certain percentage $p$ of queries will only predict the $(1-p)\%$ of queries that it feels most certain about. Ideally, the accuracy should increase (for performance measures that are supposed to be maximized like ACC) or the error metric should decrease (for performance measures that are supposed to be minimized like MAE) with increasing $p$, leading to a monotonically increasing or decreasing curve, unlike a flat random accuracy-rejection curve.

\begin{figure}[!htbp]
  \centering
     \begin{subfigure}[t]{0.3\linewidth}
           \centering
           \includegraphics[width=\linewidth]{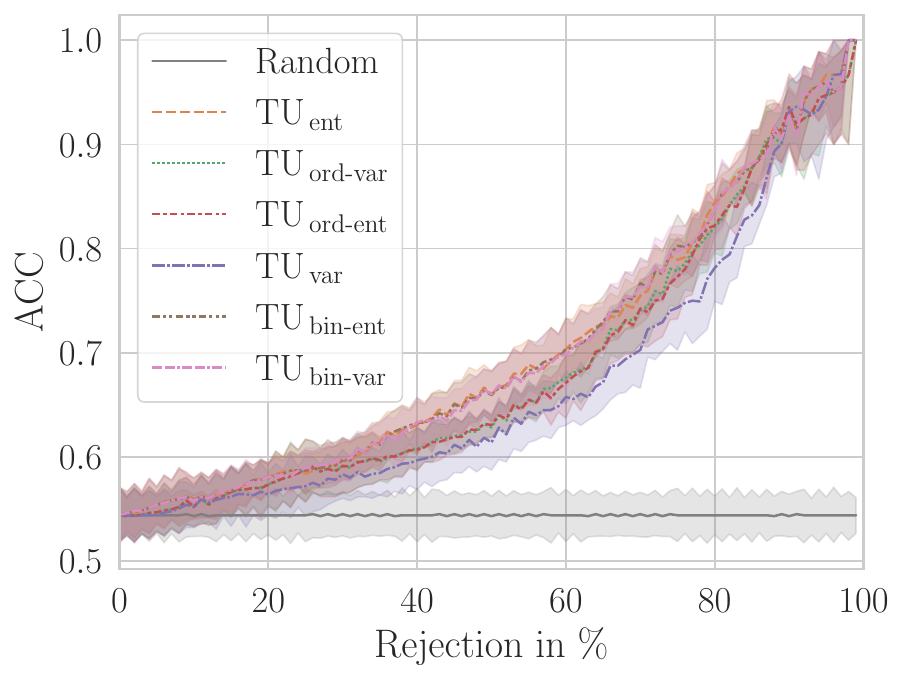}
           \subcaption{CMC TU}
       \end{subfigure} 
         \begin{subfigure}[t]{0.3\linewidth}
           \centering
           \includegraphics[width=\linewidth]{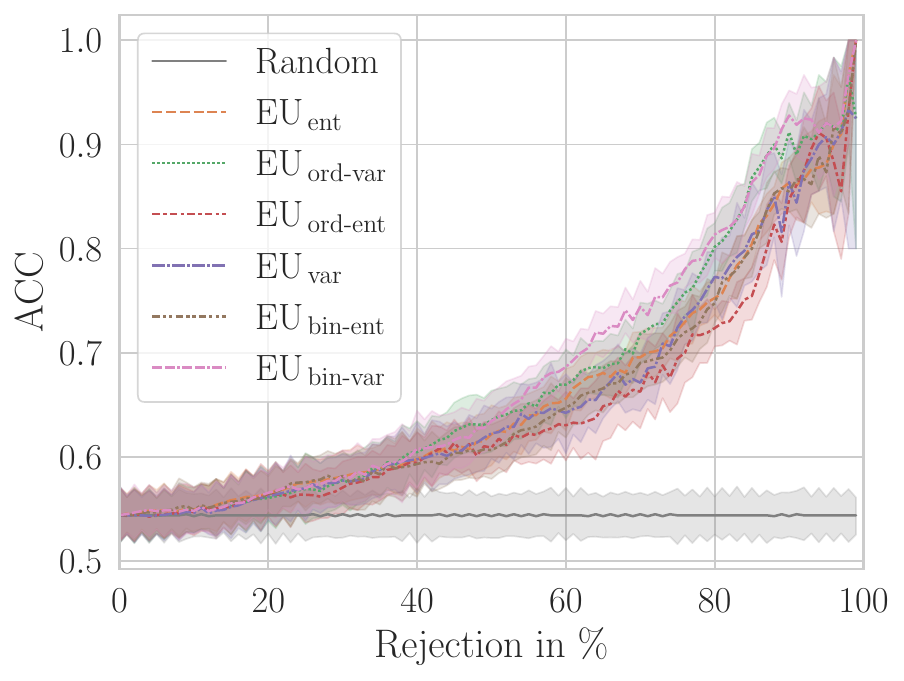}
           \subcaption{CMC EU}
       \end{subfigure}  
       \begin{subfigure}[t]{0.3\linewidth}
        \centering
        \includegraphics[width=\linewidth]{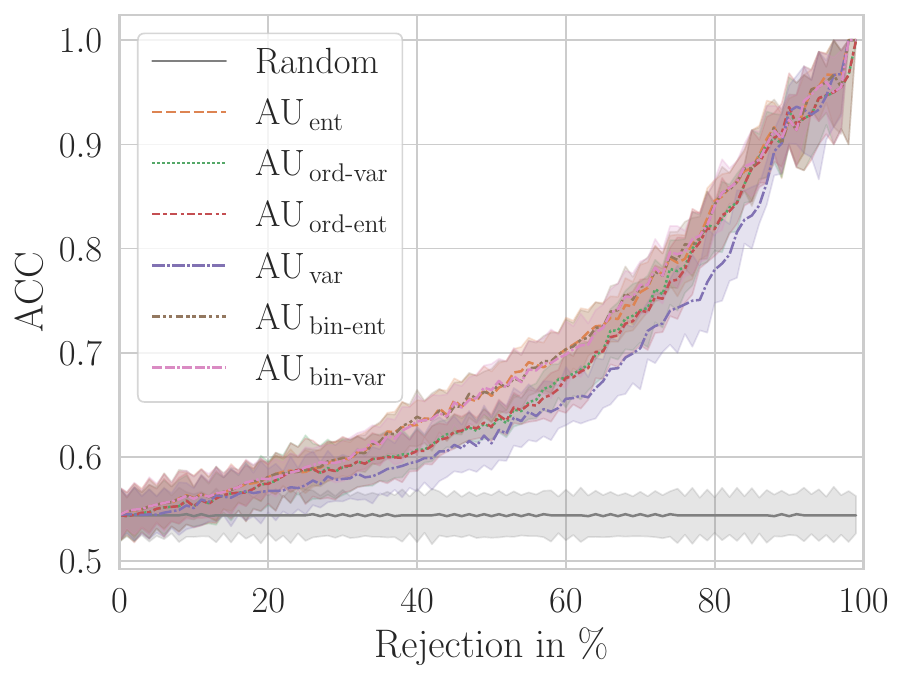}
        \subcaption{CMC AU}
    \end{subfigure}  
    \begin{subfigure}[t]{0.3\linewidth}
      \centering
      \includegraphics[width=\linewidth]{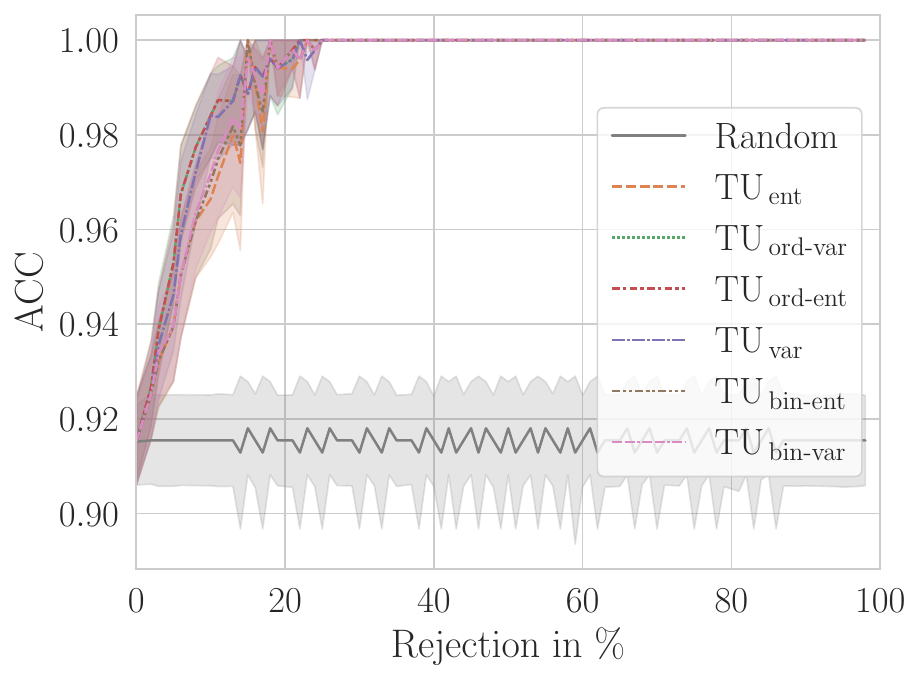}
      \subcaption{Balance Scale TU }
  \end{subfigure} 
    \begin{subfigure}[t]{0.3\linewidth}
      \centering
      \includegraphics[width=\linewidth]{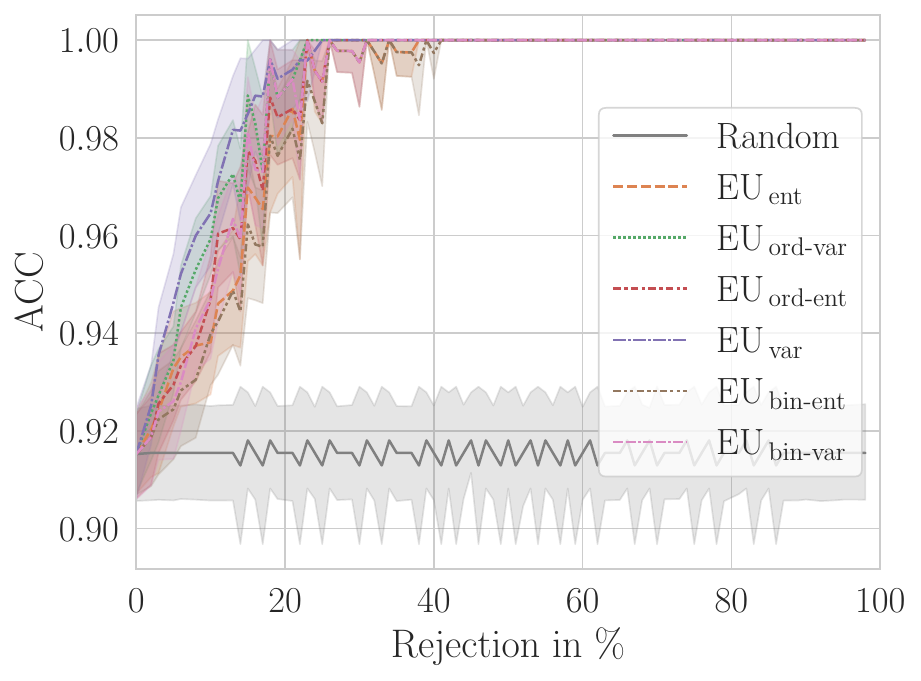}
      \subcaption{Balance Scale EU}
  \end{subfigure}  
  \begin{subfigure}[t]{0.3\linewidth}
   \centering
   \includegraphics[width=\linewidth]{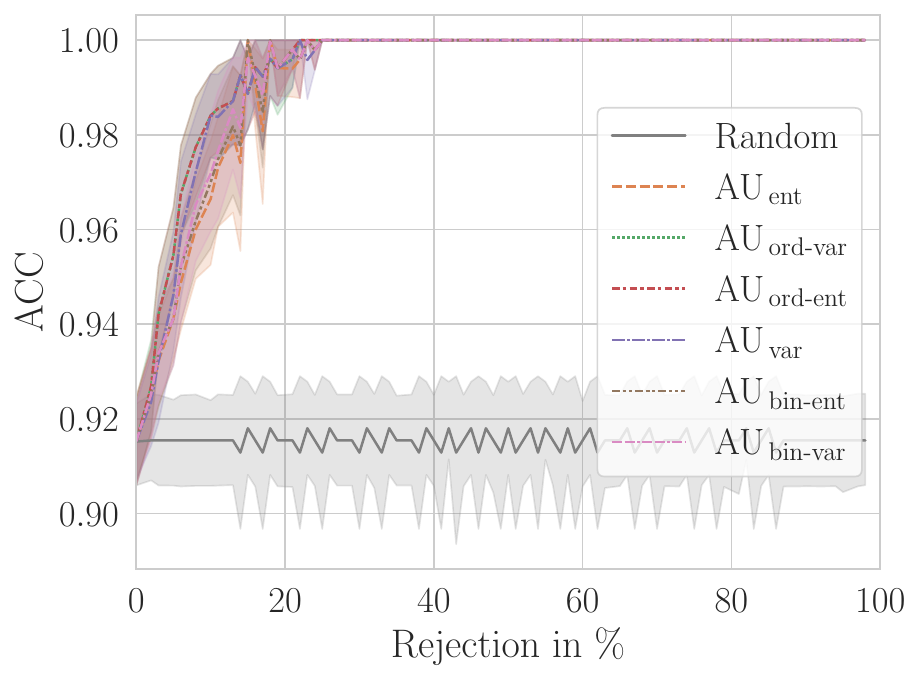}
   \subcaption{Balance Scale AU}
\end{subfigure}    
\begin{subfigure}[t]{0.3\linewidth}
  \centering
  \includegraphics[width=\linewidth]{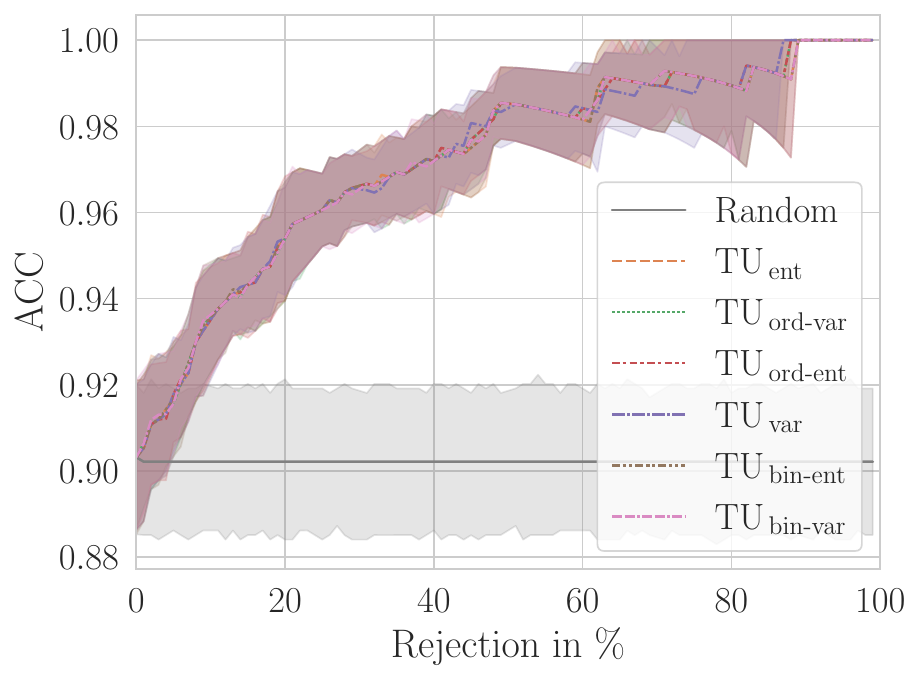}
  \subcaption{Stocks Domain TU }
\end{subfigure} 
\begin{subfigure}[t]{0.3\linewidth}
  \centering
  \includegraphics[width=\linewidth]{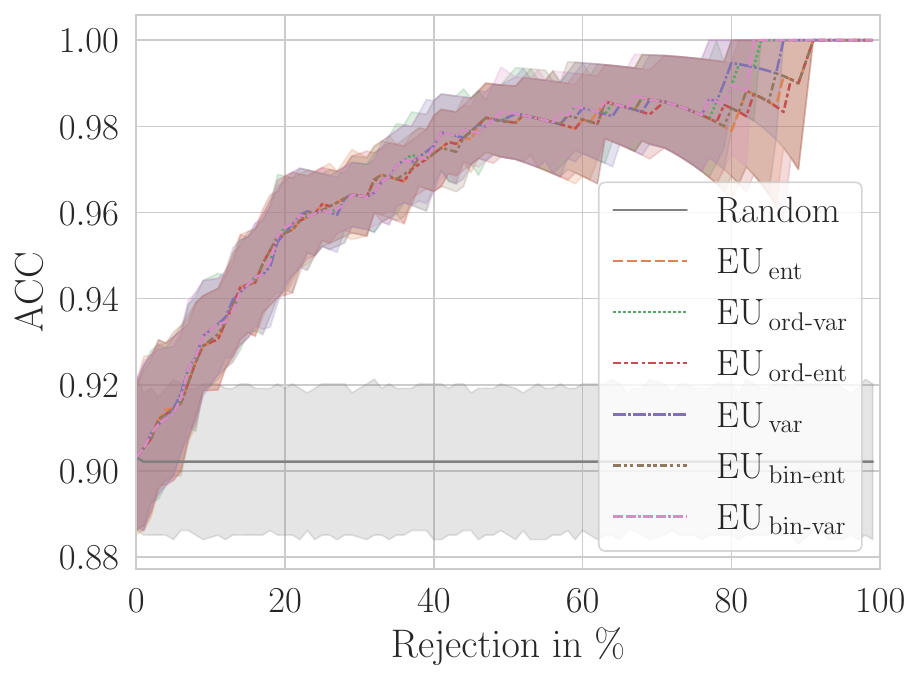}
  \subcaption{Stocks Domain EU}
\end{subfigure}  
\begin{subfigure}[t]{0.3\linewidth}
\centering
\includegraphics[width=\linewidth]{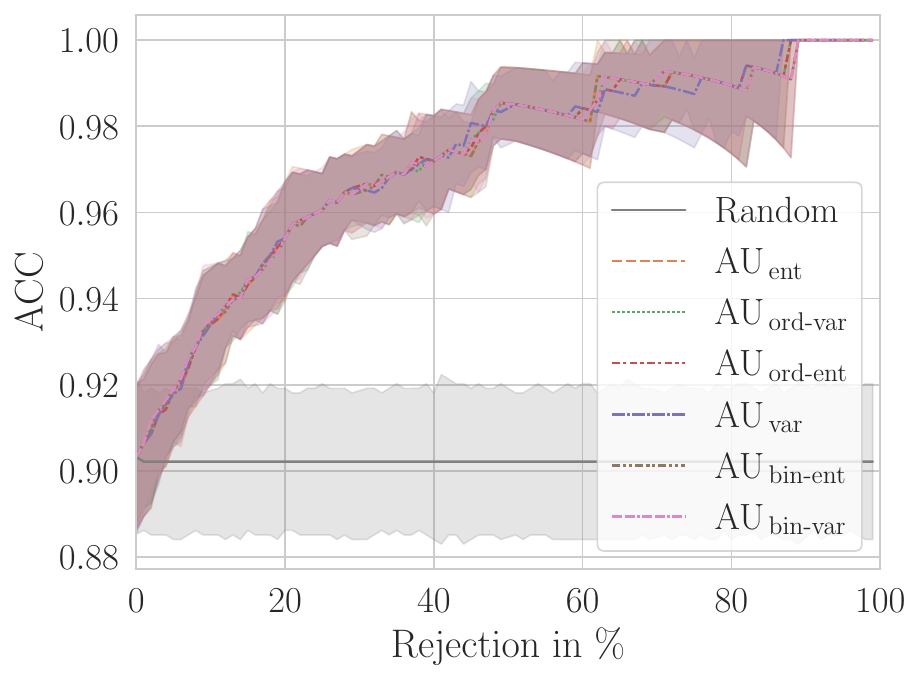}
\subcaption{Stocks Domain AU}
\end{subfigure}     
\begin{subfigure}[t]{0.3\linewidth}
  \centering
  \includegraphics[width=\linewidth]{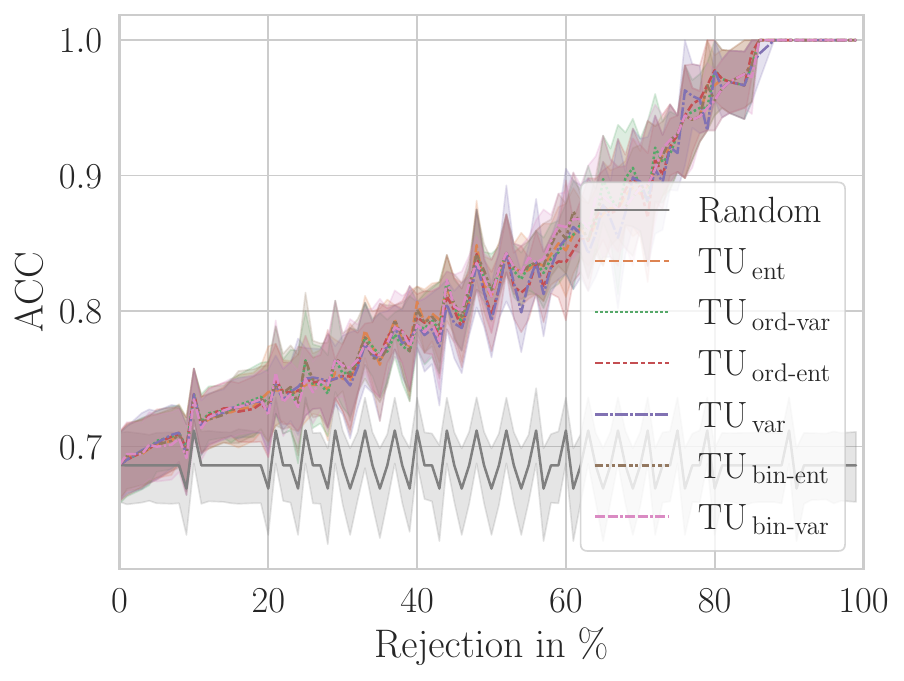}
  \subcaption{Eucalyptus TU}
\end{subfigure} 
\begin{subfigure}[t]{0.3\linewidth}
  \centering
  \includegraphics[width=\linewidth]{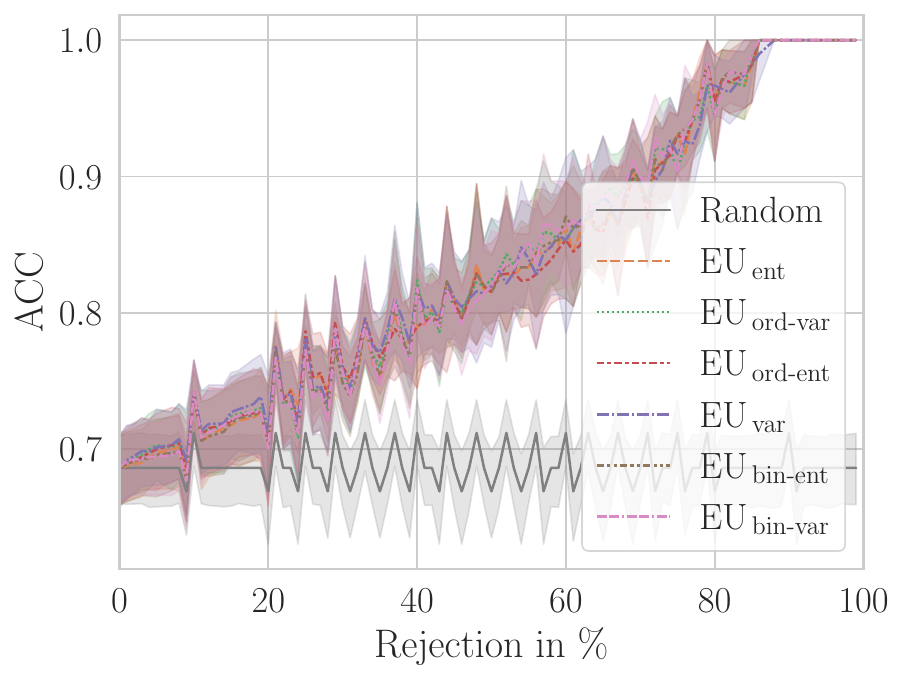}
  \subcaption{Eucalyptus EU}
\end{subfigure}  
\begin{subfigure}[t]{0.3\linewidth}
\centering
\includegraphics[width=\linewidth]{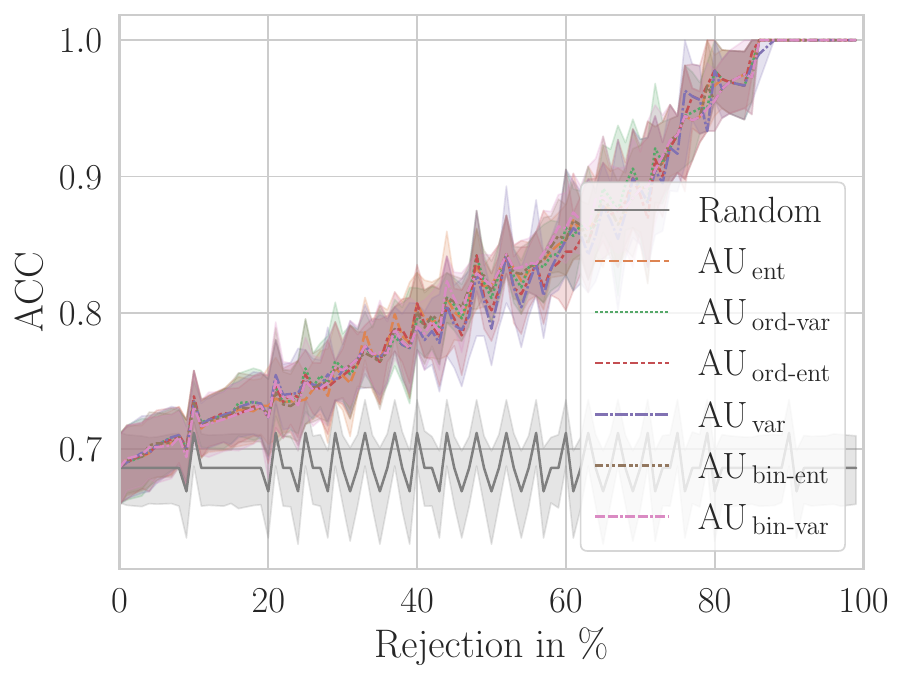}
\subcaption{Eucalyptus AU}
\end{subfigure}  
\begin{subfigure}[t]{0.3\linewidth}
  \centering
  \includegraphics[width=\linewidth]{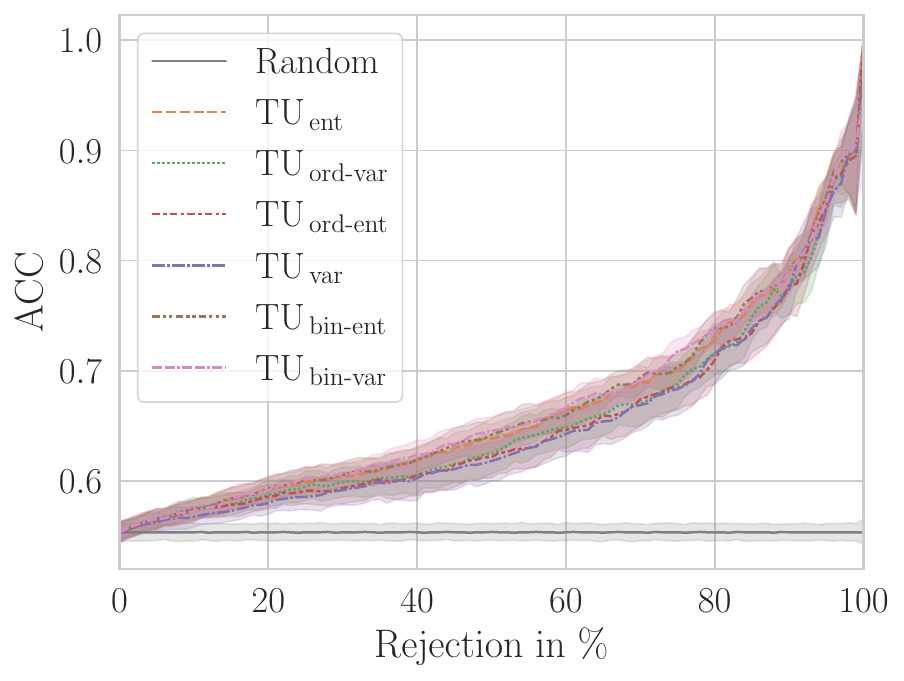}
  \subcaption{Abalone TU}
\end{subfigure} 
\begin{subfigure}[t]{0.3\linewidth}
  \centering
  \includegraphics[width=\linewidth]{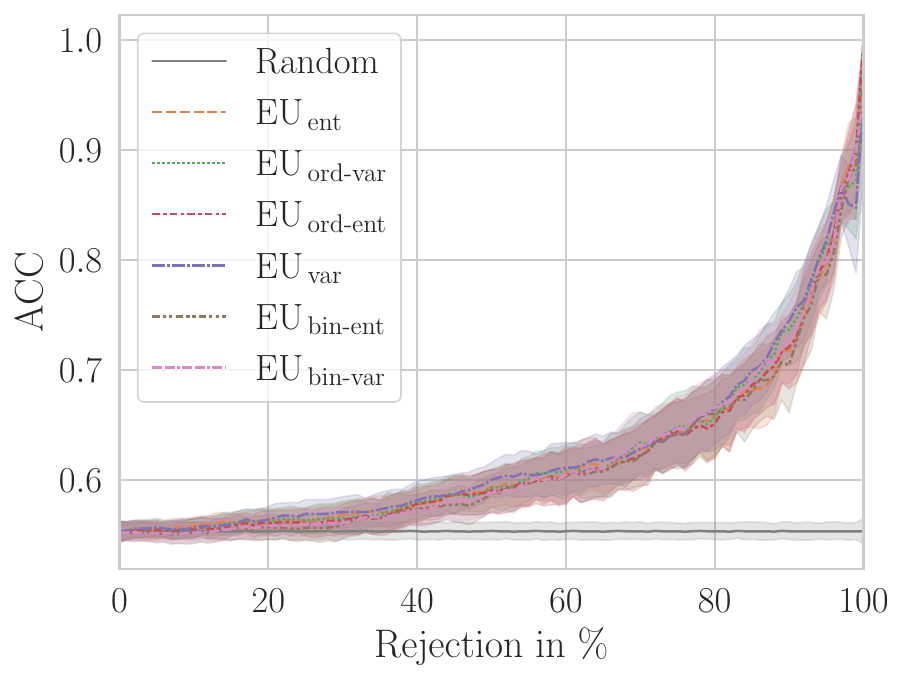}
  \subcaption{Abalone EU}
\end{subfigure}  
\begin{subfigure}[t]{0.3\linewidth}
\centering
\includegraphics[width=\linewidth]{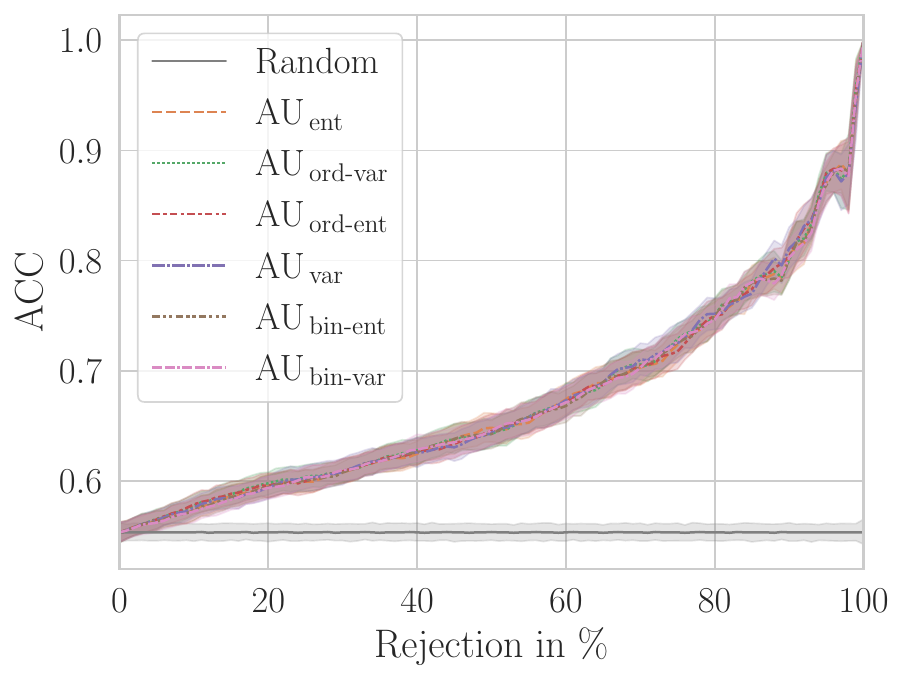}
\subcaption{Abalone AU}
\end{subfigure}  
        \caption{Accuracy rejection curves for different datasets, uncertainty types (TU, EU, AU) and measures using an ensemble of GBTs (LightGBM \citep{DBLP:conf/nips/KeMFWCMYL17}).}
        \label{fig:acc_reject}
\end{figure}

\begin{figure}[!htbp]
  \centering
     \begin{subfigure}[t]{0.3\linewidth}
           \centering
           \includegraphics[width=\linewidth]{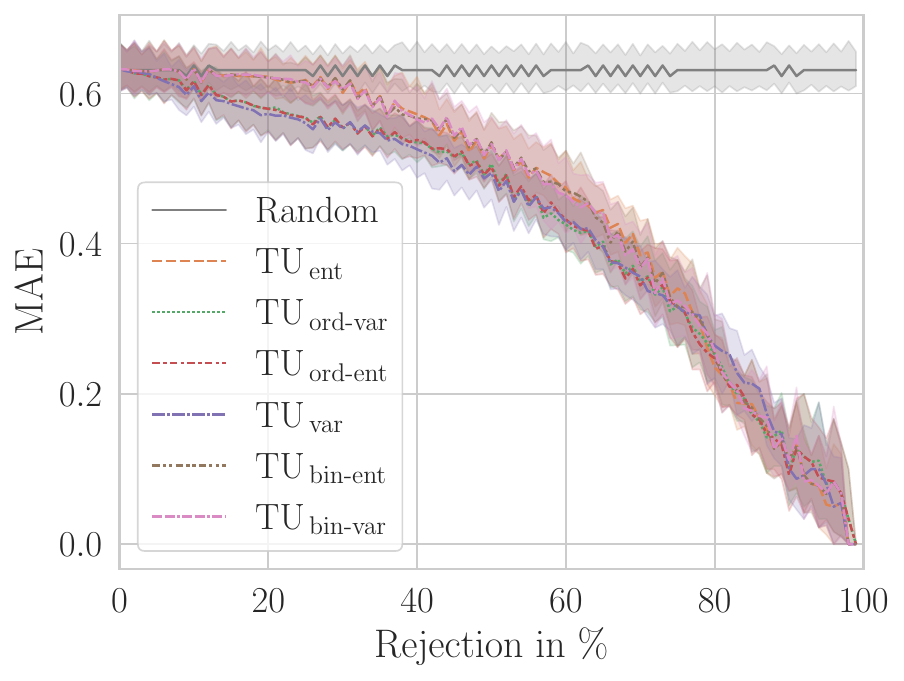}
           \subcaption{CMC TU }
       \end{subfigure} 
         \begin{subfigure}[t]{0.3\linewidth}
           \centering
           \includegraphics[width=\linewidth]{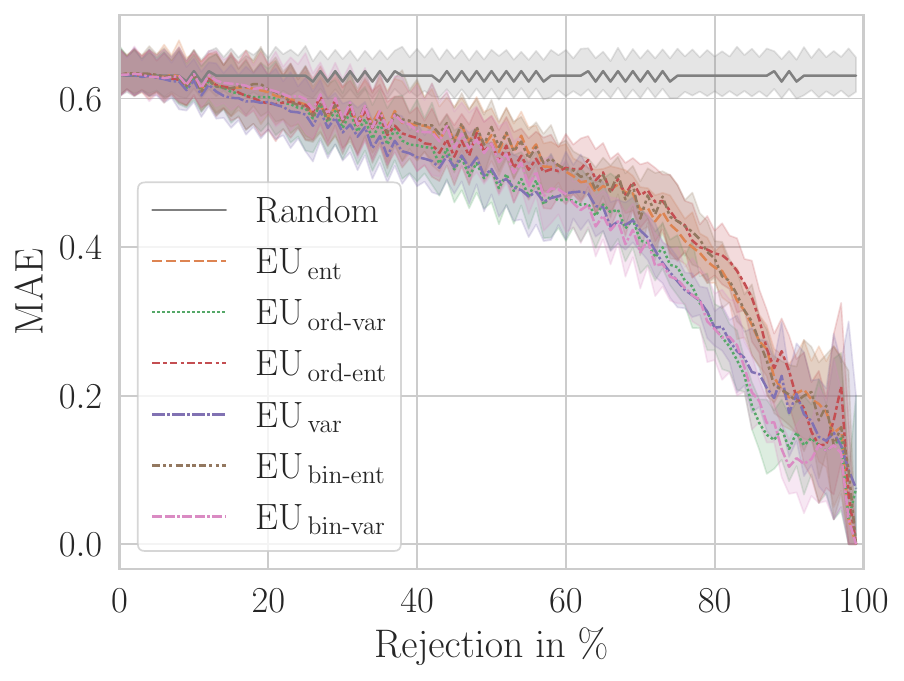}
           \subcaption{CMC EU}
       \end{subfigure}  
       \begin{subfigure}[t]{0.3\linewidth}
        \centering
        \includegraphics[width=\linewidth]{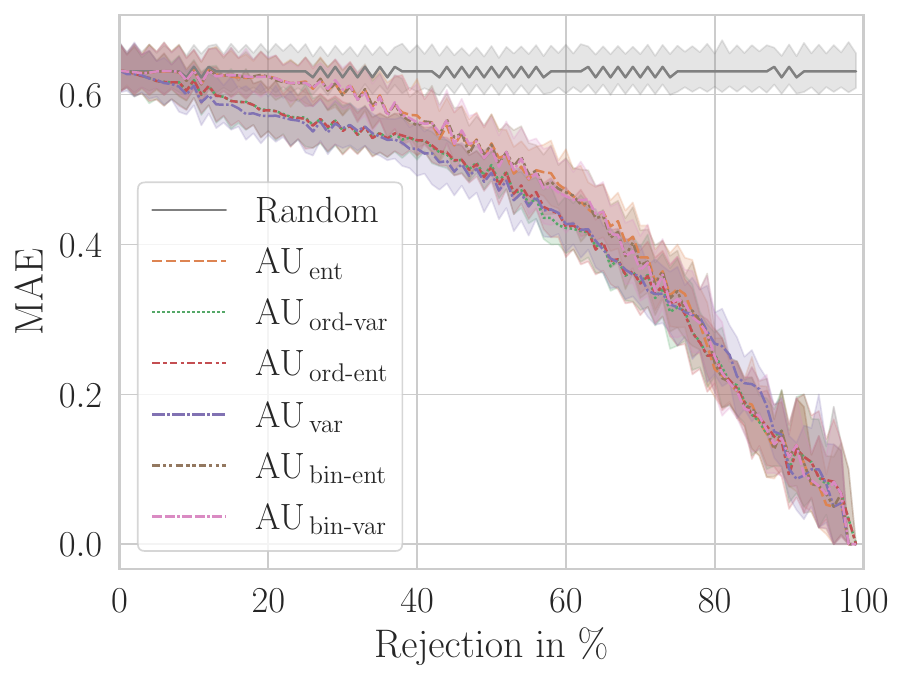}
        \subcaption{CMC AU}
    \end{subfigure}    
    \begin{subfigure}[t]{0.3\linewidth}
      \centering
      \includegraphics[width=\linewidth]{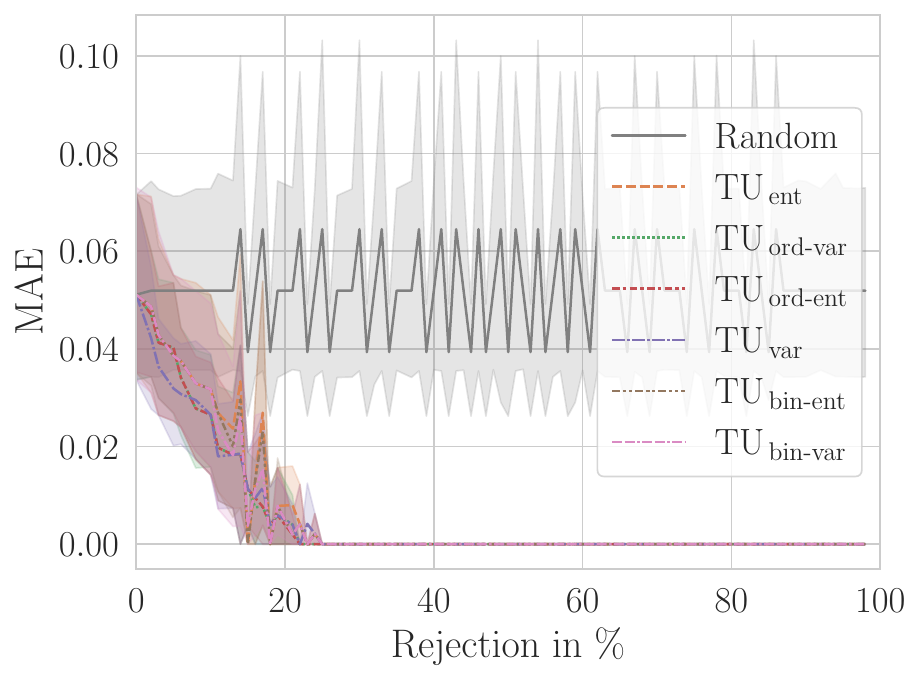}
      \subcaption{Balance Scale TU }
  \end{subfigure} 
    \begin{subfigure}[t]{0.3\linewidth}
      \centering
      \includegraphics[width=\linewidth]{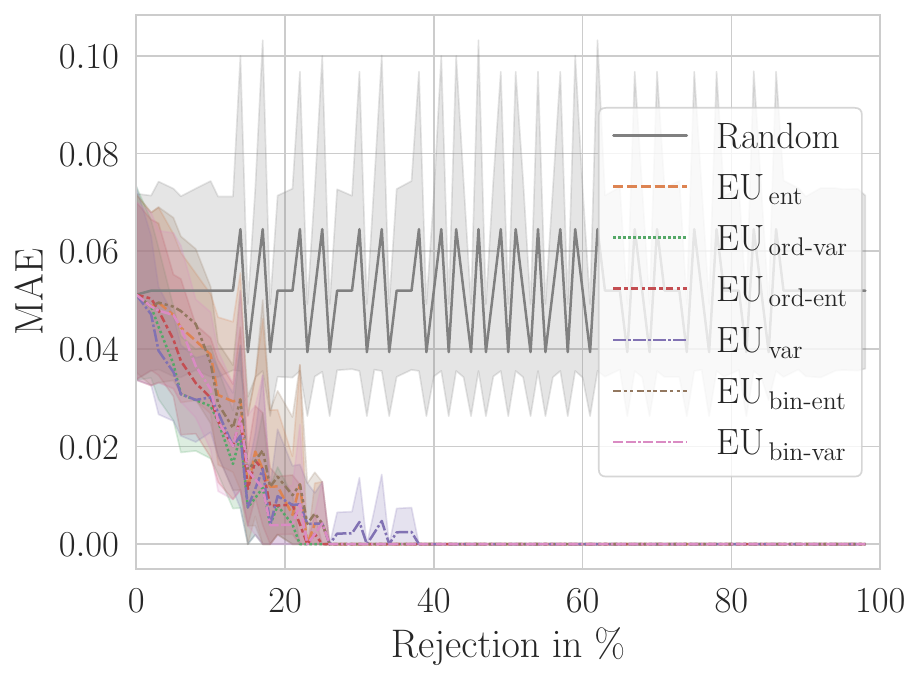}
      \subcaption{Balance Scale EU}
  \end{subfigure}  
  \begin{subfigure}[t]{0.3\linewidth}
   \centering
   \includegraphics[width=\linewidth]{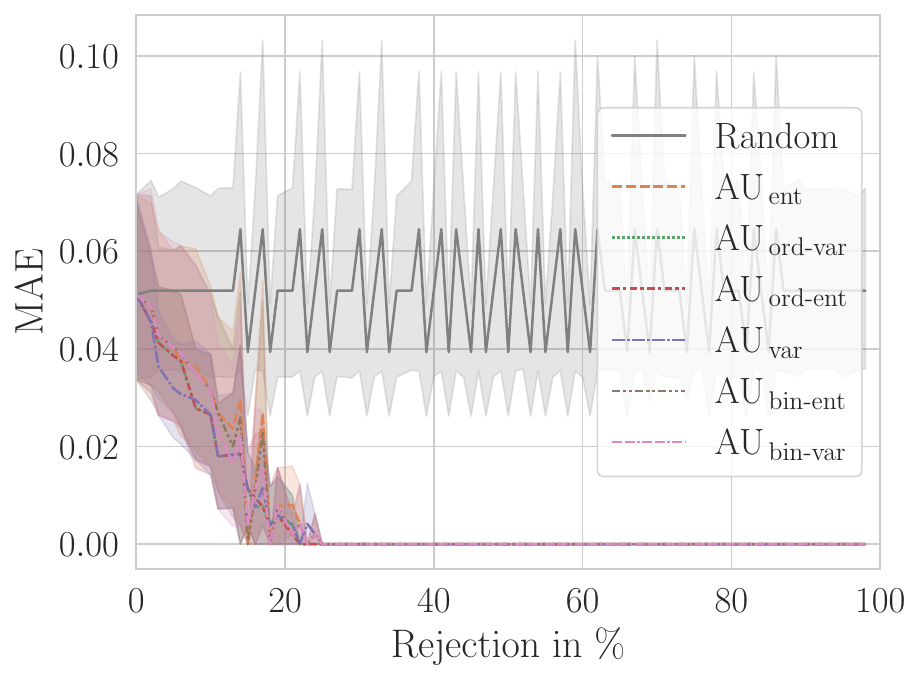}
   \subcaption{Balance Scale AU}
\end{subfigure}    
\begin{subfigure}[t]{0.3\linewidth}
  \centering
  \includegraphics[width=\linewidth]{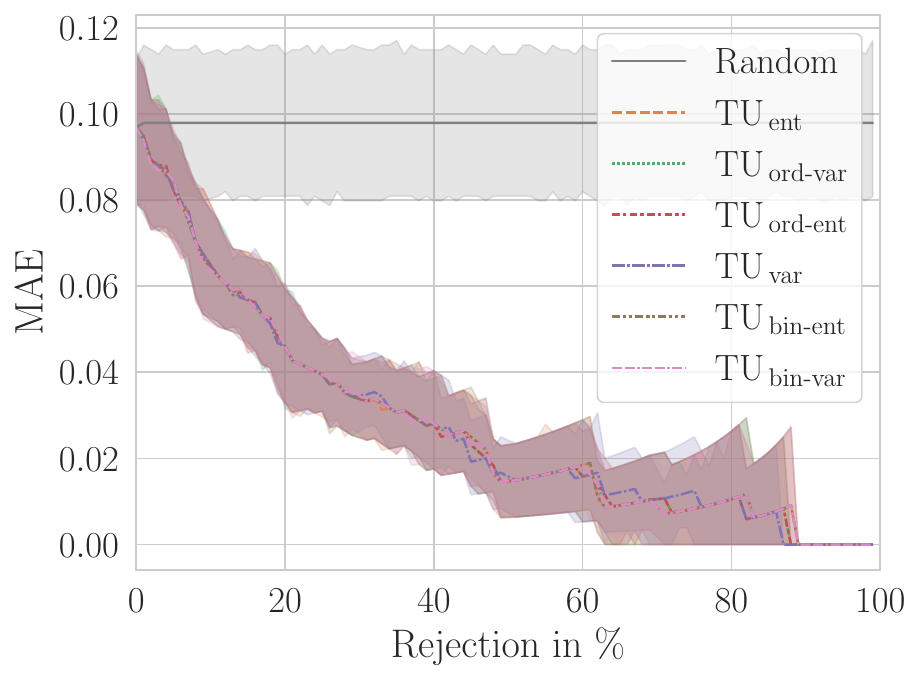}
  \subcaption{Stocks Domain TU }
\end{subfigure} 
\begin{subfigure}[t]{0.3\linewidth}
  \centering
  \includegraphics[width=\linewidth]{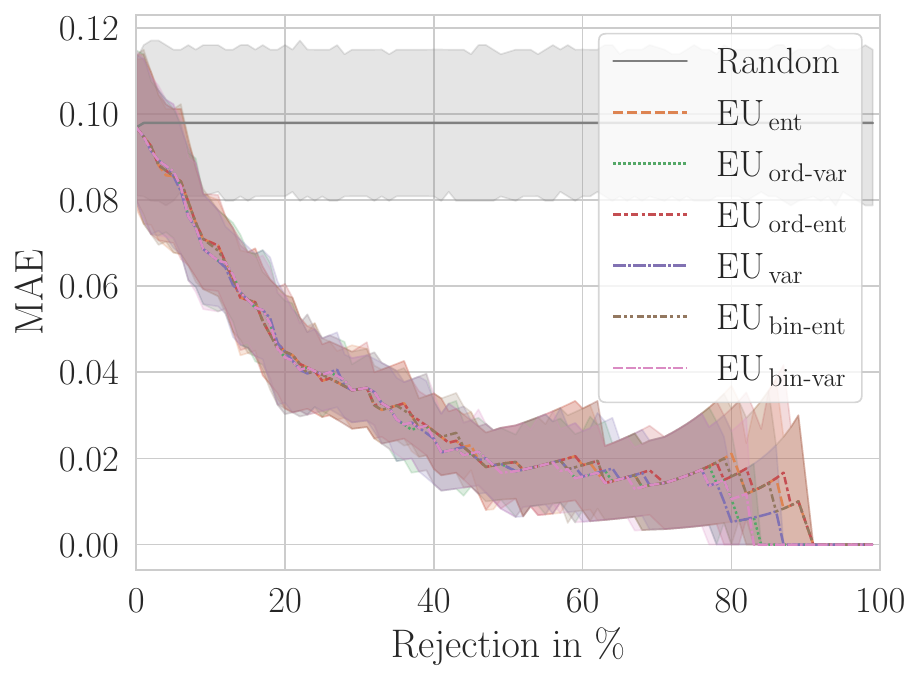}
  \subcaption{Stocks Domain EU}
\end{subfigure}  
\begin{subfigure}[t]{0.3\linewidth}
\centering
\includegraphics[width=\linewidth]{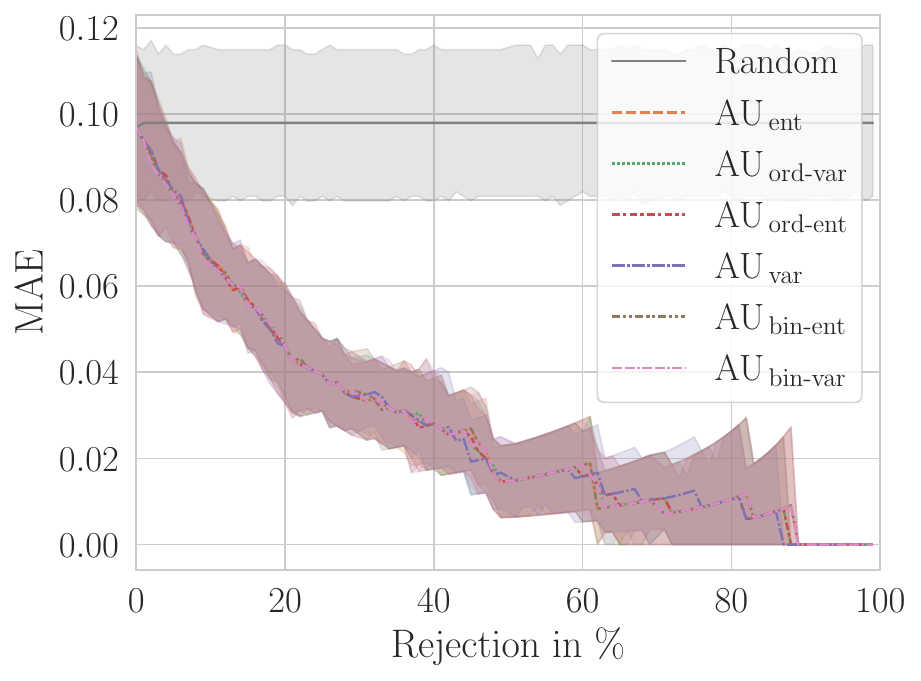}
\subcaption{Stocks Domain AU}
\end{subfigure}     
\begin{subfigure}[t]{0.3\linewidth}
  \centering
  \includegraphics[width=\linewidth]{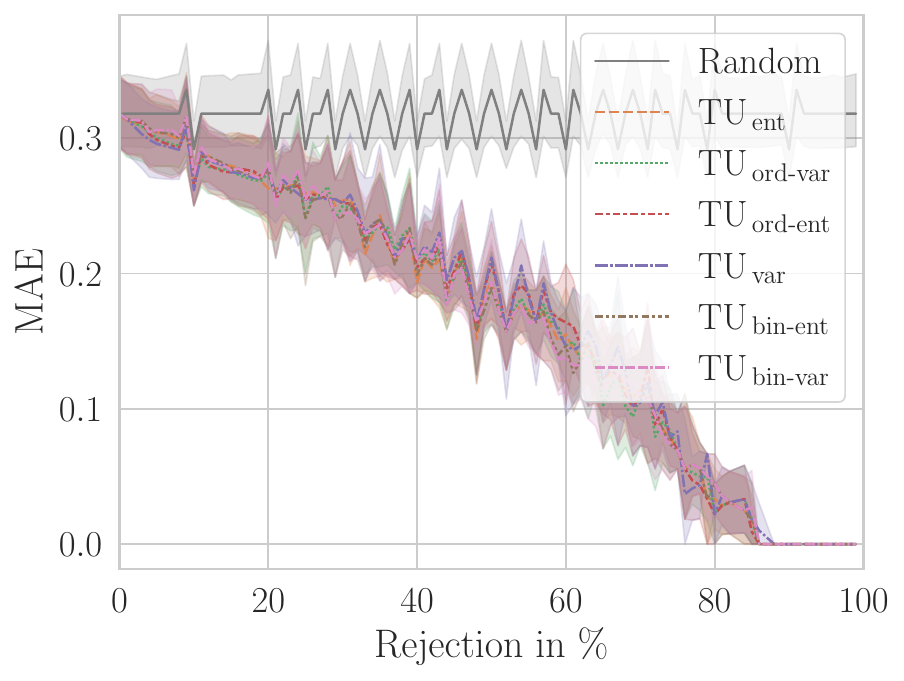}
  \subcaption{Eucalyptus TU }
\end{subfigure} 
\begin{subfigure}[t]{0.3\linewidth}
  \centering
  \includegraphics[width=\linewidth]{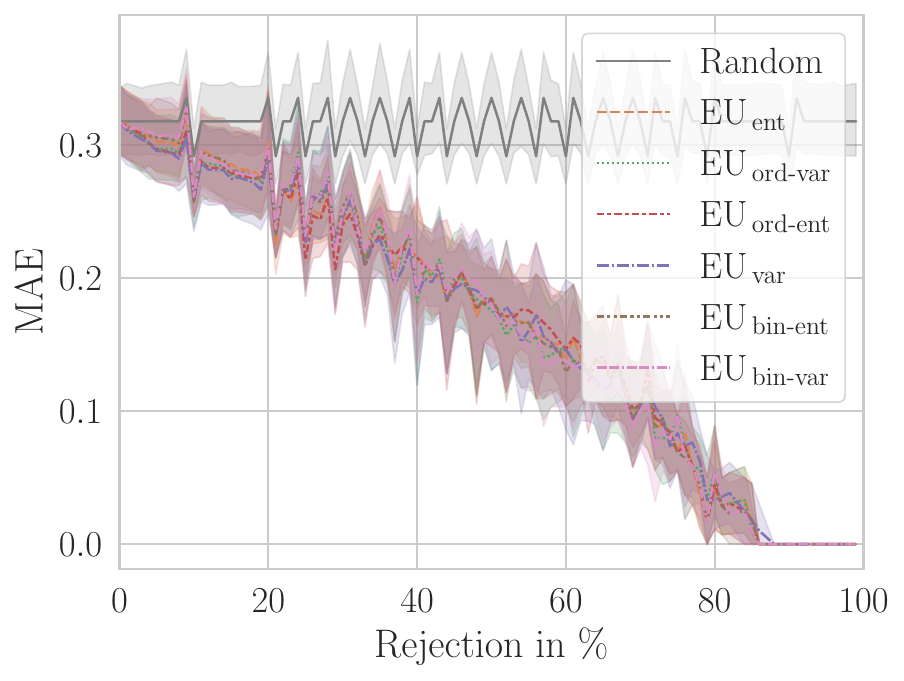}
  \subcaption{Eucalyptus EU}
\end{subfigure}  
\begin{subfigure}[t]{0.3\linewidth}
\centering
\includegraphics[width=\linewidth]{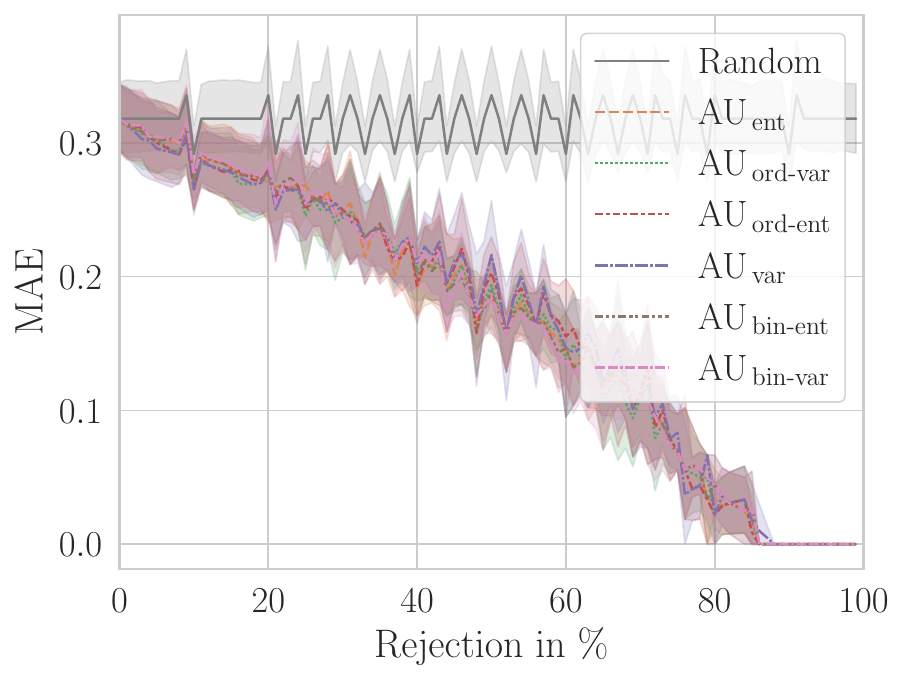}
\subcaption{Eucalyptus AU}
\end{subfigure}   
\begin{subfigure}[t]{0.3\linewidth}
  \centering
  \includegraphics[width=\linewidth]{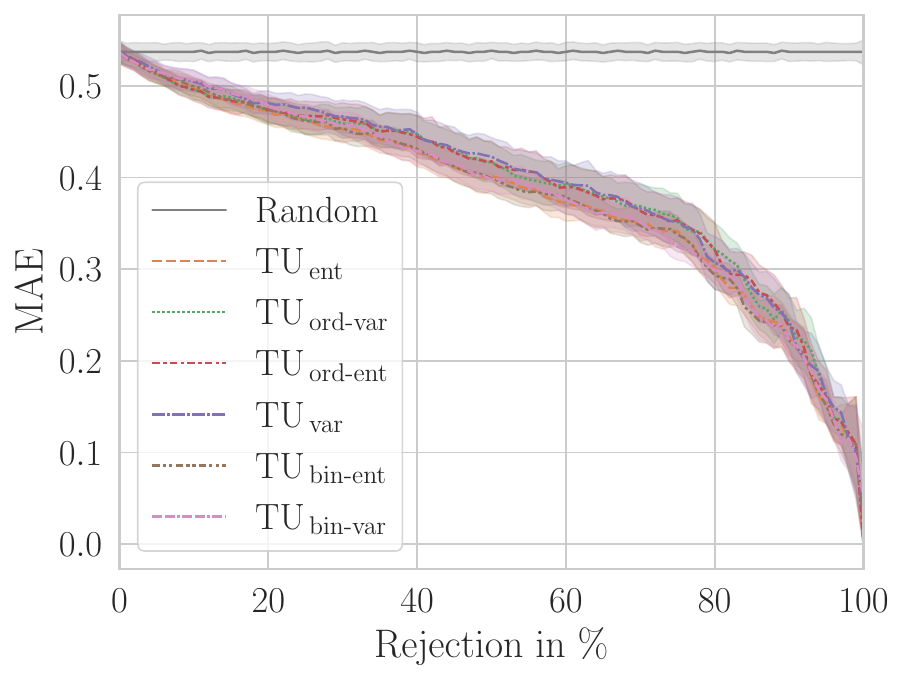}
  \subcaption{Abalone TU }
\end{subfigure} 
\begin{subfigure}[t]{0.3\linewidth}
  \centering
  \includegraphics[width=\linewidth]{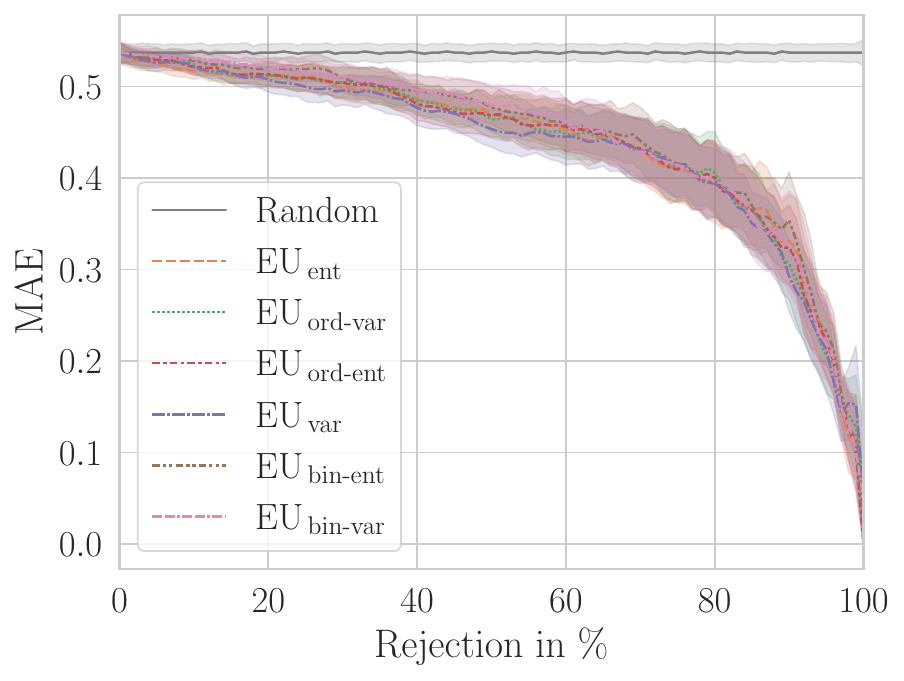}
  \subcaption{Abalone EU}
\end{subfigure}  
\begin{subfigure}[t]{0.3\linewidth}
\centering
\includegraphics[width=\linewidth]{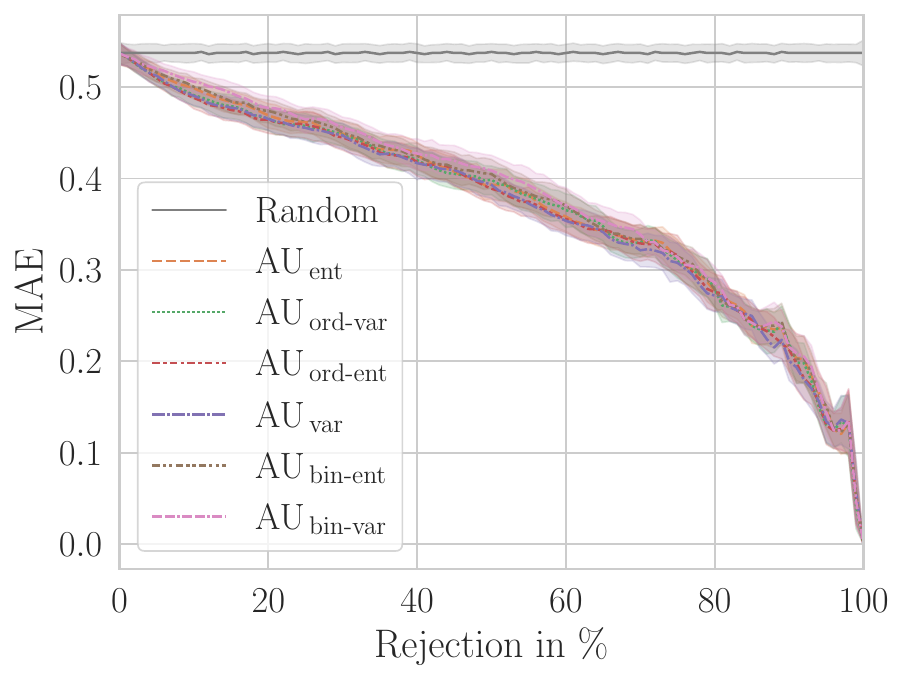}
\subcaption{Abalone AU}
\end{subfigure}   
        \caption{Mean absolute error rejection curves for different datasets, uncertainty types (TU, EU, AU) and measures using an ensemble of GBTs (LightGBM \citep{DBLP:conf/nips/KeMFWCMYL17}).}
        \label{fig:mae_reject}
\end{figure}

Figures \ref{fig:acc_reject} and \ref{fig:mae_reject} display accuracy-rejection curves for some selected ordinal benchmark datasets for ACC and MAE based on the different uncertainty types (AU, EU, and TU) and uncertainty measures:
$\mathbb{H}$ (ent), $\mathbb{V}$ (var), 
$U^{cat}_{\mathbb{H}}$ (bin-ent), $U^{cat}_{\mathbb{V}}$ (bin-var),  $U^{ord}_{\mathbb{H}}$ (ord-ent), and $U^{ord}_{\mathbb{V}}$ (ord-var). As one can clearly see, all measures are capable of quantifying the different uncertainty types properly, as all accuracy-rejection curves either increase or decrease monotonically for ACC and MAE, respectively. Hence, all the above-presented measures appear to be viable solutions for uncertainty quantification in ordinal classification. In spite of capturing different types of uncertainty, AU, EU, and TU lead to highly correlated curves, which is in line with observations in \citep{DBLP:journals/corr/abs-2402-19460}. However, accuracy-rejection curves only provide a coarse visual way to assess the quality of uncertainty quantification methods and do not allow for rigorous statistical comparisons.

\subsection{Prediction-Rejection-Ratios (PRRs)}
\label{subsec:prrs}

\begin{figure}[!htbp]
  \centering
   \begin{subfigure}[t]{0.4\linewidth}
           \centering
           \includegraphics[width=\linewidth]{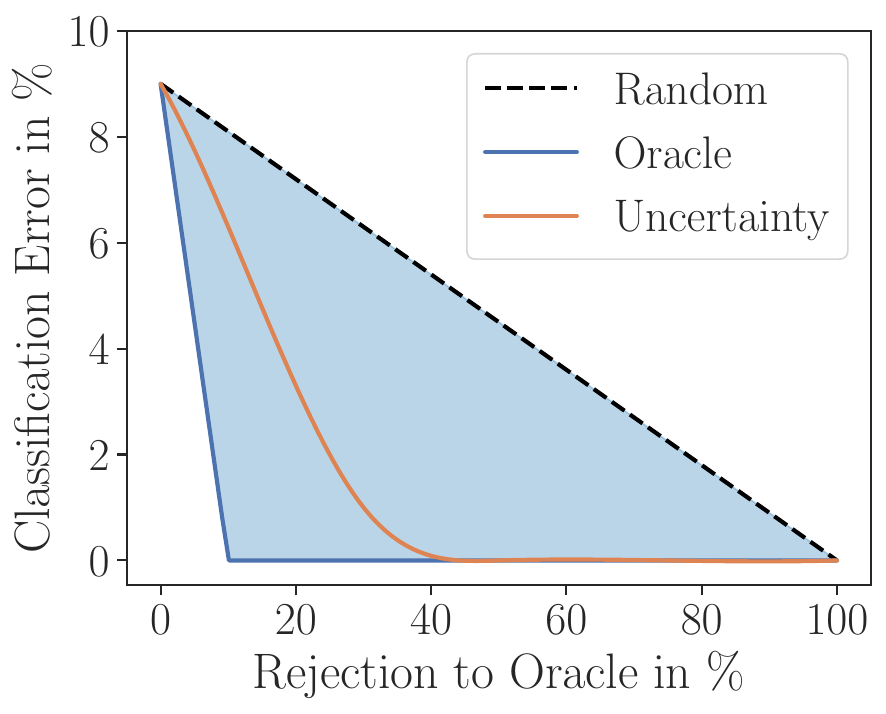}
           \subcaption{Shaded area is $AR_{orc}$. }
       \end{subfigure} 
         \begin{subfigure}[t]{0.4\linewidth}
           \centering
           \includegraphics[width=\linewidth]{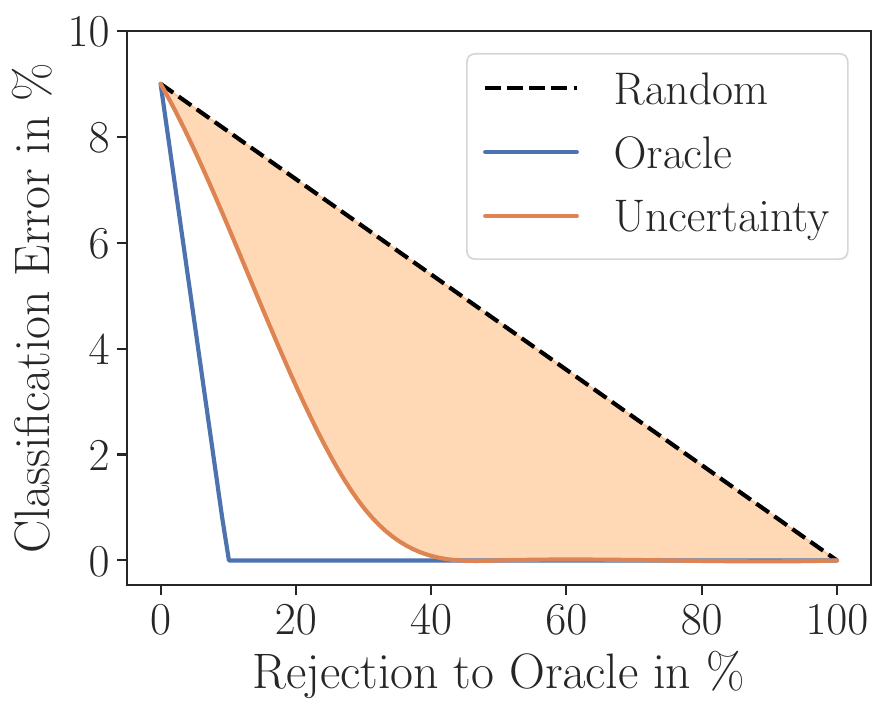}
           \subcaption{Shaded area is $AR_{unc}$. }
       \end{subfigure}    
        \caption{Example Prediction Rejection Curves \citep{malinin2019uncertainty}.}
          \label{fig:prr}
\end{figure}

To compare the different uncertainty quantification methods on a more fine-grained numerical level, \emph{prediction-rejection ratios} (PRRs), as introduced by \cite{malinin2019uncertainty}, provide a good solution. Essentially, PRRs summarize rejection curves in a single numerical value, where the area between the curve obtained for the uncertainty method and the curve obtained for random rejection, $AR_{unc}$, is compared to the corresponding area produced by a perfect oracle-based rejection, $AR_{orc}$ (cf.\ Figure \ref{fig:prr}):
$$
PRR = \frac{AR_{unc}}{AR_{orc}} \, .
$$
Unlike in accuracy-rejection curves, random rejection will not produce a flat line but a line that decreases linearly in expectation. This is because rejected queries are supposedly delegated to an oracle that will answer queries correctly.

Another property of PRRs is that they assess uncertainty quantification quality independently of classification performance. A PRR value of 1 indicates perfect rejection, and hence perfect uncertainty uqantification, whereas a value of 0 indicates random rejection. The PRR can also become negative, which indicates worse than random uncertainty quantification. To calculate PRR values for uncertainty quantification evaluation, one also needs to select a performance measure, just as for rejection curves.
To make all rejection curves go in the same direction, we measure the MCR instead of ACC, as is commonly done \citep{DBLP:conf/iclr/MalininPU21,DBLP:journals/ml/LahotiGW23}, and again MAE.

To compare the PRR values calculated for the different uncertainty methods and datasets, we conduct a Friedman test followed by a Holm-adjusted Wilcoxon signed-rank test \citep{demsar2006statistical,benavoli2016should}. 
The non-parametric Friedman test will first determine whether there is a significant difference in the performance of the uncertainty measures across the datasets overall (with a significance level $p=0.05$). If it indicates a significant difference, the Wilcoxon signed-rank test will then conduct pairwise comparisons between the uncertainty measures to determine which specific differences are statistically significant (again at a significance level $p=0.05$).
Furthermore, we depict the results by uncertainty type (AU, EU, TU, or All) as well as performance measure (MCR, MAE, or both) and visualize these using critical difference diagrams (CD).
The critical difference diagrams show the average ranks of the different measures. If two measures do not differ significantly, they are connected by a horizontal bar or line.
The full results of our comparison can be found in the tables presented in Appendix \ref{asec:prrs}.

Figure \ref{fig:cd_total} displays the CD diagrams for total uncertainty. Though there is no statistically significant difference among the different measures for total uncertainty, binary methods outperform standard entropy and variance-based measures, with the variance-based OCS decomposition (ord-var) leading the field when considering MCR and MAE.

\begin{figure}[!htbp]
  \centering 
  \caption{CD diagrams for Aleatoric Uncertainty (AU) using an ensemble of GBTs (LightGBM \citep{DBLP:conf/nips/KeMFWCMYL17}).}
  \label{fig:cd_aleatoric}
  \begin{subfigure}[t]{\linewidth}
    \centering 
\begin{tikzpicture}[
  treatment line/.style={rounded corners=1.5pt, line cap=round, shorten >=1pt},
  treatment label/.style={font=\small},
  group line/.style={ultra thick},
]

\begin{axis}[
  clip={false},
  axis x line={center},
  axis y line={none},
  axis line style={-},
  xmin={1},
  ymax={0},
  scale only axis={true},
  width={\axisdefaultwidth},
  ticklabel style={anchor=south, yshift=1.3*\pgfkeysvalueof{/pgfplots/major tick length}, font=\small},
  every tick/.style={draw=black},
  major tick style={yshift=.5*\pgfkeysvalueof{/pgfplots/major tick length}},
  minor tick style={yshift=.5*\pgfkeysvalueof{/pgfplots/minor tick length}},
  title style={yshift=\baselineskip},
  xmax={6},
  ymin={-4.5},
  height={5\baselineskip},
  xtick={1,2,3,4,5,6},
  minor x tick num={3},
  title={AU - PRR (MCR), PRR (MAE)},
]

\draw[treatment line] ([yshift=-2pt] axis cs:3.131521739130435, 0) |- (axis cs:2.631521739130435, -2.0)
  node[treatment label, anchor=east] {$\text{AU}_{\,\text{ord-var}}$};
\draw[treatment line] ([yshift=-2pt] axis cs:3.348913043478261, 0) |- (axis cs:2.631521739130435, -3.0)
  node[treatment label, anchor=east] {$\text{AU}_{\,\text{ord-ent}}$};
\draw[treatment line] ([yshift=-2pt] axis cs:3.5478260869565217, 0) |- (axis cs:2.631521739130435, -4.0)
  node[treatment label, anchor=east] {$\text{AU}_{\,\text{bin-var}}$};
\draw[treatment line] ([yshift=-2pt] axis cs:3.635869565217391, 0) |- (axis cs:4.168478260869565, -4.0)
  node[treatment label, anchor=west] {$\text{AU}_{\,\text{ent}}$};
\draw[treatment line] ([yshift=-2pt] axis cs:3.667391304347826, 0) |- (axis cs:4.168478260869565, -3.0)
  node[treatment label, anchor=west] {$\text{AU}_{\,\text{var}}$};
\draw[treatment line] ([yshift=-2pt] axis cs:3.6684782608695654, 0) |- (axis cs:4.168478260869565, -2.0)
  node[treatment label, anchor=west] {$\text{AU}_{\,\text{bin-ent}}$};
\draw[group line] (axis cs:3.348913043478261, -1.3333333333333333) -- (axis cs:3.6684782608695654, -1.3333333333333333);

\end{axis}
\end{tikzpicture}
\captionsetup{justification=centering}
\subcaption{Result for MCR and MAE.}
\label{fig:cd_aleatoric_mcr_mae}
\end{subfigure}
\begin{subfigure}[t]{0.48\linewidth}
  \centering
\begin{tikzpicture}[scale=0.65,
  treatment line/.style={rounded corners=1.5pt, line cap=round, shorten >=1pt},
  treatment label/.style={font=\small},
  group line/.style={ultra thick},
]

\begin{axis}[
  clip={false},
  axis x line={center},
  axis y line={none},
  axis line style={-},
  xmin={1},
  ymax={0},
  scale only axis={true},
  width={\axisdefaultwidth},
  ticklabel style={anchor=south, yshift=1.3*\pgfkeysvalueof{/pgfplots/major tick length}, font=\small},
  every tick/.style={draw=black},
  major tick style={yshift=.5*\pgfkeysvalueof{/pgfplots/major tick length}},
  minor tick style={yshift=.5*\pgfkeysvalueof{/pgfplots/minor tick length}},
  title style={yshift=\baselineskip},
  xmax={6},
  ymin={-4.5},
  height={5\baselineskip},
  xtick={1,2,3,4,5,6},
  minor x tick num={3},
  title={AU - PRR (MCR)},
]

\draw[treatment line] ([yshift=-2pt] axis cs:3.139130434782609, 0) |- (axis cs:2.639130434782609, -2.0)
  node[treatment label, anchor=east] {$\text{AU}_{\,\text{ord-var}}$};
\draw[treatment line] ([yshift=-2pt] axis cs:3.4413043478260867, 0) |- (axis cs:2.639130434782609, -3.0)
  node[treatment label, anchor=east] {$\text{AU}_{\,\text{bin-var}}$};
\draw[treatment line] ([yshift=-2pt] axis cs:3.4521739130434783, 0) |- (axis cs:2.639130434782609, -4.0)
  node[treatment label, anchor=east] {$\text{AU}_{\,\text{ord-ent}}$};
\draw[treatment line] ([yshift=-2pt] axis cs:3.5217391304347827, 0) |- (axis cs:4.356521739130435, -4.0)
  node[treatment label, anchor=west] {$\text{AU}_{\,\text{bin-ent}}$};
\draw[treatment line] ([yshift=-2pt] axis cs:3.5891304347826085, 0) |- (axis cs:4.356521739130435, -3.0)
  node[treatment label, anchor=west] {$\text{AU}_{\,\text{ent}}$};
\draw[treatment line] ([yshift=-2pt] axis cs:3.856521739130435, 0) |- (axis cs:4.356521739130435, -2.0)
  node[treatment label, anchor=west] {$\text{AU}_{\,\text{var}}$};
\draw[group line] (axis cs:3.139130434782609, -1.3333333333333333) -- (axis cs:3.5891304347826085, -1.3333333333333333);
\draw[group line] (axis cs:3.4413043478260867, -1.5333333333333332) -- (axis cs:3.856521739130435, -1.5333333333333332);

\end{axis}
\end{tikzpicture}
\captionsetup{justification=centering}
\subcaption{Result only for MCR.}
\label{fig:cd_aleatoric_mcr}
\end{subfigure}
\begin{subfigure}[t]{0.48\linewidth}
  \centering
\begin{tikzpicture}[scale=0.65,
  treatment line/.style={rounded corners=1.5pt, line cap=round, shorten >=1pt},
  treatment label/.style={font=\small},
  group line/.style={ultra thick},
]

\begin{axis}[
  clip={false},
  axis x line={center},
  axis y line={none},
  axis line style={-},
  xmin={1},
  ymax={0},
  scale only axis={true},
  width={\axisdefaultwidth},
  ticklabel style={anchor=south, yshift=1.3*\pgfkeysvalueof{/pgfplots/major tick length}, font=\small},
  every tick/.style={draw=black},
  major tick style={yshift=.5*\pgfkeysvalueof{/pgfplots/major tick length}},
  minor tick style={yshift=.5*\pgfkeysvalueof{/pgfplots/minor tick length}},
  title style={yshift=\baselineskip},
  xmax={6},
  ymin={-4.5},
  height={5\baselineskip},
  xtick={1,2,3,4,5,6},
  minor x tick num={3},
  title={AU - PRR (MAE)},
]

\draw[treatment line] ([yshift=-2pt] axis cs:3.123913043478261, 0) |- (axis cs:2.623913043478261, -2.0)
  node[treatment label, anchor=east] {$\text{AU}_{\,\text{ord-var}}$};
\draw[treatment line] ([yshift=-2pt] axis cs:3.2456521739130433, 0) |- (axis cs:2.623913043478261, -3.0)
  node[treatment label, anchor=east] {$\text{AU}_{\,\text{ord-ent}}$};
\draw[treatment line] ([yshift=-2pt] axis cs:3.4782608695652173, 0) |- (axis cs:2.623913043478261, -4.0)
  node[treatment label, anchor=east] {$\text{AU}_{\,\text{var}}$};
\draw[treatment line] ([yshift=-2pt] axis cs:3.6543478260869566, 0) |- (axis cs:4.315217391304348, -4.0)
  node[treatment label, anchor=west] {$\text{AU}_{\,\text{bin-var}}$};
\draw[treatment line] ([yshift=-2pt] axis cs:3.6826086956521737, 0) |- (axis cs:4.315217391304348, -3.0)
  node[treatment label, anchor=west] {$\text{AU}_{\,\text{ent}}$};
\draw[treatment line] ([yshift=-2pt] axis cs:3.8152173913043477, 0) |- (axis cs:4.315217391304348, -2.0)
  node[treatment label, anchor=west] {$\text{AU}_{\,\text{bin-ent}}$};
\draw[group line] (axis cs:3.2456521739130433, -2.0) -- (axis cs:3.6826086956521737, -2.0);
\draw[group line] (axis cs:3.123913043478261, -1.3333333333333333) -- (axis cs:3.4782608695652173, -1.3333333333333333);
\draw[group line] (axis cs:3.4782608695652173, -1.5333333333333332) -- (axis cs:3.8152173913043477, -1.5333333333333332);

\end{axis}
\end{tikzpicture}

\captionsetup{justification=centering}
\subcaption{Result only for MAE.}
\label{fig:cd_aleatoric_mae}
         \end{subfigure}                  
\end{figure}
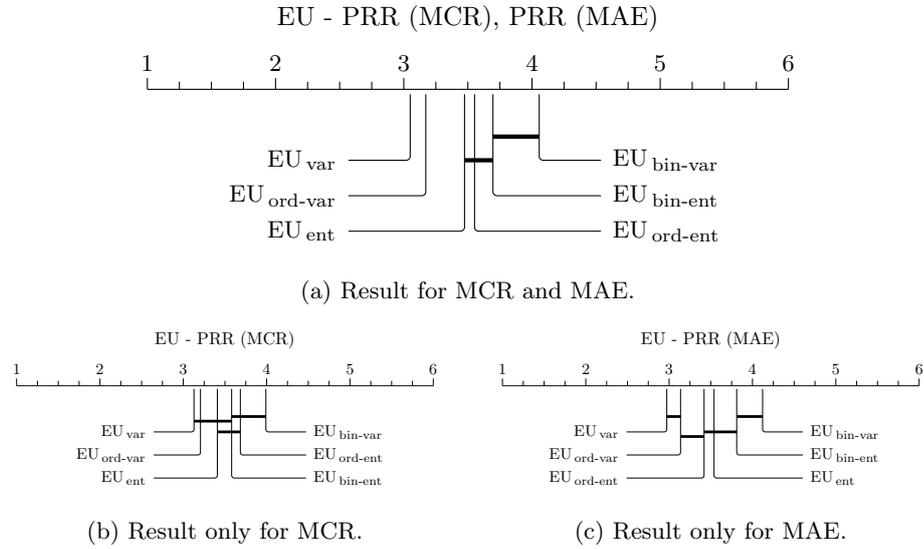

When looking at the results for quantifying epistemic uncertainty in Figure \ref{fig:cd_epistemic_mcr_mae}, var and ord-var significantly outperform the other measures when simultaneously considering MCR and MAE. With regards to ord-var, the same applies for aleatoric uncertainty (cf.\ Figure \ref{fig:cd_aleatoric_mcr_mae}).

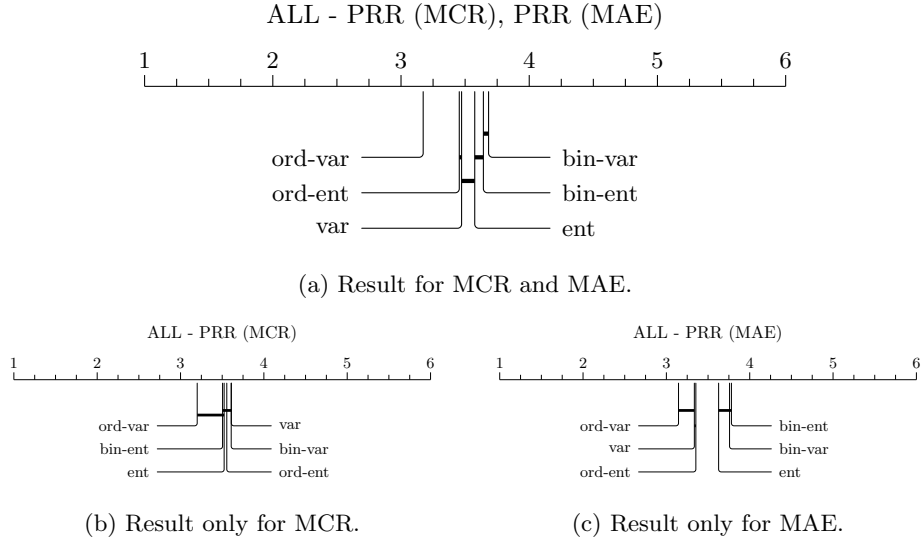
\begin{figure}[!htbp]
  \centering 
  \caption{CD diagrams for all uncertainty types (AU, EU and TU) using an ensemble of GBTs (LightGBM \citep{DBLP:conf/nips/KeMFWCMYL17}).}
  \label{fig:cd_all}
  \begin{subfigure}[t]{\linewidth}
    \centering 
\begin{tikzpicture}[
  treatment line/.style={rounded corners=1.5pt, line cap=round, shorten >=1pt},
  treatment label/.style={font=\small},
  group line/.style={ultra thick},
]

\begin{axis}[
  clip={false},
  axis x line={center},
  axis y line={none},
  axis line style={-},
  xmin={1},
  ymax={0},
  scale only axis={true},
  width={\axisdefaultwidth},
  ticklabel style={anchor=south, yshift=1.3*\pgfkeysvalueof{/pgfplots/major tick length}, font=\small},
  every tick/.style={draw=black},
  major tick style={yshift=.5*\pgfkeysvalueof{/pgfplots/major tick length}},
  minor tick style={yshift=.5*\pgfkeysvalueof{/pgfplots/minor tick length}},
  title style={yshift=\baselineskip},
  xmax={6},
  ymin={-4.5},
  height={5\baselineskip},
  xtick={1,2,3,4,5,6},
  minor x tick num={3},
  title={ALL - PRR (MCR), PRR (MAE)},
]

\draw[treatment line] ([yshift=-2pt] axis cs:3.1728260869565217, 0) |- (axis cs:2.6728260869565217, -2.0)
  node[treatment label, anchor=east] {ord-var};
\draw[treatment line] ([yshift=-2pt] axis cs:3.4543478260869565, 0) |- (axis cs:2.6728260869565217, -3.0)
  node[treatment label, anchor=east] {ord-ent};
\draw[treatment line] ([yshift=-2pt] axis cs:3.472463768115942, 0) |- (axis cs:2.6728260869565217, -4.0)
  node[treatment label, anchor=east] {var};
\draw[treatment line] ([yshift=-2pt] axis cs:3.575, 0) |- (axis cs:4.182971014492754, -4.0)
  node[treatment label, anchor=west] {ent};
\draw[treatment line] ([yshift=-2pt] axis cs:3.642391304347826, 0) |- (axis cs:4.182971014492754, -3.0)
  node[treatment label, anchor=west] {bin-ent};
\draw[treatment line] ([yshift=-2pt] axis cs:3.6829710144927534, 0) |- (axis cs:4.182971014492754, -2.0)
  node[treatment label, anchor=west] {bin-var};
\draw[group line] (axis cs:3.642391304347826, -1.3333333333333333) -- (axis cs:3.6829710144927534, -1.3333333333333333);
\draw[group line] (axis cs:3.575, -2.0) -- (axis cs:3.642391304347826, -2.0);
\draw[group line] (axis cs:3.472463768115942, -2.6666666666666665) -- (axis cs:3.575, -2.6666666666666665);
\draw[group line] (axis cs:3.4543478260869565, -2.0) -- (axis cs:3.472463768115942, -2.0);

\end{axis}
\end{tikzpicture}
\captionsetup{justification=centering}
\subcaption{Result for MCR and MAE.}
\label{fig:cd_all_mcr_mae}
\end{subfigure}
\begin{subfigure}[t]{0.48\linewidth}
  \centering
\begin{tikzpicture}[scale=0.65,
  treatment line/.style={rounded corners=1.5pt, line cap=round, shorten >=1pt},
  treatment label/.style={font=\small},
  group line/.style={ultra thick},
]

\begin{axis}[
  clip={false},
  axis x line={center},
  axis y line={none},
  axis line style={-},
  xmin={1},
  ymax={0},
  scale only axis={true},
  width={\axisdefaultwidth},
  ticklabel style={anchor=south, yshift=1.3*\pgfkeysvalueof{/pgfplots/major tick length}, font=\small},
  every tick/.style={draw=black},
  major tick style={yshift=.5*\pgfkeysvalueof{/pgfplots/major tick length}},
  minor tick style={yshift=.5*\pgfkeysvalueof{/pgfplots/minor tick length}},
  title style={yshift=\baselineskip},
  xmax={6},
  ymin={-4.5},
  height={5\baselineskip},
  xtick={1,2,3,4,5,6},
  minor x tick num={3},
  title={ALL - PRR (MCR)},
]

\draw[treatment line] ([yshift=-2pt] axis cs:3.2, 0) |- (axis cs:2.7, -2.0)
  node[treatment label, anchor=east] {ord-var};
\draw[treatment line] ([yshift=-2pt] axis cs:3.5043478260869567, 0) |- (axis cs:2.7, -3.0)
  node[treatment label, anchor=east] {bin-ent};
\draw[treatment line] ([yshift=-2pt] axis cs:3.5231884057971015, 0) |- (axis cs:2.7, -4.0)
  node[treatment label, anchor=east] {ent};
\draw[treatment line] ([yshift=-2pt] axis cs:3.555072463768116, 0) |- (axis cs:4.1094202898550725, -4.0)
  node[treatment label, anchor=west] {ord-ent};
\draw[treatment line] ([yshift=-2pt] axis cs:3.6079710144927537, 0) |- (axis cs:4.1094202898550725, -3.0)
  node[treatment label, anchor=west] {bin-var};
\draw[treatment line] ([yshift=-2pt] axis cs:3.6094202898550725, 0) |- (axis cs:4.1094202898550725, -2.0)
  node[treatment label, anchor=west] {var};
\draw[group line] (axis cs:3.5043478260869567, -1.3333333333333333) -- (axis cs:3.6094202898550725, -1.3333333333333333);
\draw[group line] (axis cs:3.2, -1.5333333333333332) -- (axis cs:3.5231884057971015, -1.5333333333333332);

\end{axis}
\end{tikzpicture}
\captionsetup{justification=centering}
\subcaption{Result only for MCR.}
\label{fig:cd_all_mcr}
\end{subfigure}
\begin{subfigure}[t]{0.48\linewidth}
  \centering
\begin{tikzpicture}[scale=0.65,
  treatment line/.style={rounded corners=1.5pt, line cap=round, shorten >=1pt},
  treatment label/.style={font=\small},
  group line/.style={ultra thick},
]

\begin{axis}[
  clip={false},
  axis x line={center},
  axis y line={none},
  axis line style={-},
  xmin={1},
  ymax={0},
  scale only axis={true},
  width={\axisdefaultwidth},
  ticklabel style={anchor=south, yshift=1.3*\pgfkeysvalueof{/pgfplots/major tick length}, font=\small},
  every tick/.style={draw=black},
  major tick style={yshift=.5*\pgfkeysvalueof{/pgfplots/major tick length}},
  minor tick style={yshift=.5*\pgfkeysvalueof{/pgfplots/minor tick length}},
  title style={yshift=\baselineskip},
  xmax={6},
  ymin={-4.5},
  height={5\baselineskip},
  xtick={1,2,3,4,5,6},
  minor x tick num={3},
  title={ALL - PRR (MAE)},
]

\draw[treatment line] ([yshift=-2pt] axis cs:3.1456521739130436, 0) |- (axis cs:2.6456521739130436, -2.0)
  node[treatment label, anchor=east] {ord-var};
\draw[treatment line] ([yshift=-2pt] axis cs:3.3355072463768116, 0) |- (axis cs:2.6456521739130436, -3.0)
  node[treatment label, anchor=east] {var};
\draw[treatment line] ([yshift=-2pt] axis cs:3.353623188405797, 0) |- (axis cs:2.6456521739130436, -4.0)
  node[treatment label, anchor=east] {ord-ent};
\draw[treatment line] ([yshift=-2pt] axis cs:3.6268115942028984, 0) |- (axis cs:4.280434782608696, -4.0)
  node[treatment label, anchor=west] {ent};
\draw[treatment line] ([yshift=-2pt] axis cs:3.7579710144927536, 0) |- (axis cs:4.280434782608696, -3.0)
  node[treatment label, anchor=west] {bin-var};
\draw[treatment line] ([yshift=-2pt] axis cs:3.7804347826086957, 0) |- (axis cs:4.280434782608696, -2.0)
  node[treatment label, anchor=west] {bin-ent};
\draw[group line] (axis cs:3.6268115942028984, -1.3333333333333333) -- (axis cs:3.7804347826086957, -1.3333333333333333);
\draw[group line] (axis cs:3.1456521739130436, -1.3333333333333333) -- (axis cs:3.3355072463768116, -1.3333333333333333);
\draw[group line] (axis cs:3.3355072463768116, -2.0) -- (axis cs:3.353623188405797, -2.0);

\end{axis}
\end{tikzpicture}

\captionsetup{justification=centering}
\subcaption{Result only for MAE.}
\label{fig:cd_all_mae}
         \end{subfigure}              
\end{figure}

Eventually, when looking at the results over all uncertainty types (AU, EU, and TU) in Figure \ref{fig:cd_all}, the results become very distinct. Again, ord-var significantly outperforms the other measures (cf.\ Figure \ref{fig:cd_all_mcr_mae}), but also the distinction between measures taking distance into account becomes clear, with ord-var, var, and ord-ent significantly outperforming the other measures on MAE (cf.\ Figure \ref{fig:cd_all_mae}). Interestingly, for MCR there is no significant difference between the measures (cf.\ Figure \ref{fig:cd_all_mcr}).

In the end, taking distance into account is an important property a measure needs to fulfill in uncertainty quantification for ordinal classification.
This becomes also obvious when looking at the overall ranking of measures in Figure \ref{fig:cd_all_mcr_mae} for MCR and MAE, with ord-var, ord-ent, and var outperforming ent, bin-ent, and bin-var.

Referring back to Figure \ref{fig:prob_simplex}, we may conclude that ord-var best represents the inherent trade-off in ordinal classification between exact hit-rate and minimized error distances. This conclusion is based on its qualitative position between variance and ord-ent, as it balances the focus on the extreme bimodal distribution and indicates uncertainty for the uniform distribution (see Figures \ref{subfig:var}, \ref{subfig:ord_bin_var}, and \ref{subfig:ord_bin_ent}). The focus of var on the extreme bimodal distribution appears too extreme and results in worse performance on MCR. Conversely, the focus of ord-ent on the extreme bimodal distribution is not strong enough, leading to worse performance on MAE. Overall, ord-var seems to best capture this trade-off.


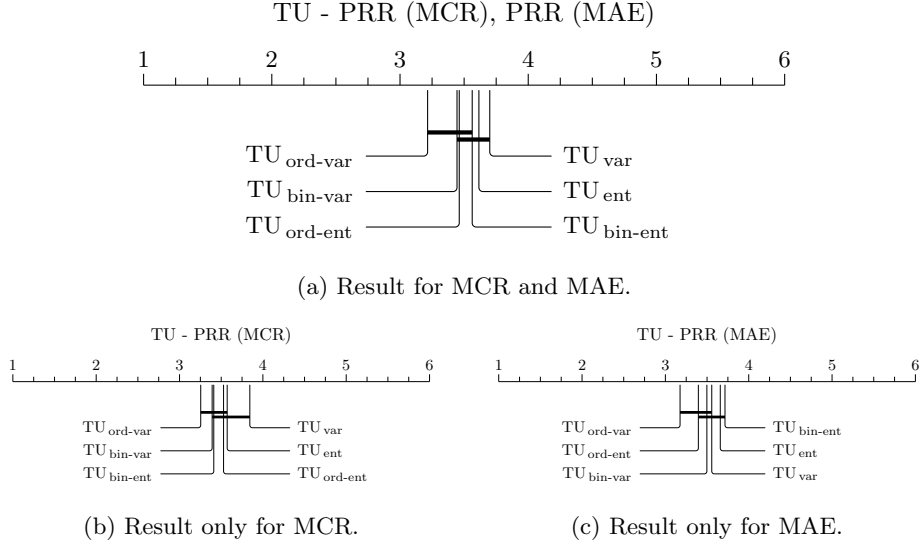
\begin{figure}[!htbp]
  \centering 
  \caption{Critical difference (CD) diagrams\protect\footnotemark  for the evaluated Total Uncertainty (TU) measures and performance metrics based on a Friedman test followed by a post-hoc Holm-adjusted Wilcoxon signed-rank test \citep{demsar2006statistical,benavoli2016should} using an ensemble of GBTs (LightGBM \citep{DBLP:conf/nips/KeMFWCMYL17}).
  Groups of uncertainty measures that are
  not significantly different (at p = 0.05) are connected.}
  \label{fig:cd_total}
  \begin{subfigure}[t]{\linewidth}
    \centering 
\begin{tikzpicture}[
  treatment line/.style={rounded corners=1.5pt, line cap=round, shorten >=1pt},
  treatment label/.style={font=\small},
  group line/.style={ultra thick},
]

\begin{axis}[
  clip={false},
  axis x line={center},
  axis y line={none},
  axis line style={-},
  xmin={1},
  ymax={0},
  scale only axis={true},
  width={\axisdefaultwidth},
  ticklabel style={anchor=south, yshift=1.3*\pgfkeysvalueof{/pgfplots/major tick length}, font=\small},
  every tick/.style={draw=black},
  major tick style={yshift=.5*\pgfkeysvalueof{/pgfplots/major tick length}},
  minor tick style={yshift=.5*\pgfkeysvalueof{/pgfplots/minor tick length}},
  title style={yshift=\baselineskip},
  xmax={6},
  ymin={-4.5},
  height={5\baselineskip},
  xtick={1,2,3,4,5,6},
  minor x tick num={3},
  title={TU - PRR (MCR), PRR (MAE)},
]

\draw[treatment line] ([yshift=-2pt] axis cs:3.215217391304348, 0) |- (axis cs:2.715217391304348, -2.0)
  node[treatment label, anchor=east] {$\text{TU}_{\,\text{ord-var}}$};
\draw[treatment line] ([yshift=-2pt] axis cs:3.4445652173913044, 0) |- (axis cs:2.715217391304348, -3.0)
  node[treatment label, anchor=east] {$\text{TU}_{\,\text{bin-var}}$};
\draw[treatment line] ([yshift=-2pt] axis cs:3.4619565217391304, 0) |- (axis cs:2.715217391304348, -4.0)
  node[treatment label, anchor=east] {$\text{TU}_{\,\text{ord-ent}}$};
\draw[treatment line] ([yshift=-2pt] axis cs:3.5630434782608695, 0) |- (axis cs:4.2, -4.0)
  node[treatment label, anchor=west] {$\text{TU}_{\,\text{bin-ent}}$};
\draw[treatment line] ([yshift=-2pt] axis cs:3.615217391304348, 0) |- (axis cs:4.2, -3.0)
  node[treatment label, anchor=west] {$\text{TU}_{\,\text{ent}}$};
\draw[treatment line] ([yshift=-2pt] axis cs:3.7, 0) |- (axis cs:4.2, -2.0)
  node[treatment label, anchor=west] {$\text{TU}_{\,\text{var}}$};
\draw[group line] (axis cs:3.215217391304348, -1.3333333333333333) -- (axis cs:3.5630434782608695, -1.3333333333333333);
\draw[group line] (axis cs:3.4445652173913044, -1.5333333333333332) -- (axis cs:3.7, -1.5333333333333332);

\end{axis}
\end{tikzpicture}
\captionsetup{justification=centering}
\subcaption{Result for MCR and MAE.}
\label{fig:cd_total_mcr_mae}
\end{subfigure}
\begin{subfigure}[t]{0.48\linewidth}
  \centering
\begin{tikzpicture}[scale=0.65,
  treatment line/.style={rounded corners=1.5pt, line cap=round, shorten >=1pt},
  treatment label/.style={font=\small},
  group line/.style={ultra thick},
]

\begin{axis}[
  clip={false},
  axis x line={center},
  axis y line={none},
  axis line style={-},
  xmin={1},
  ymax={0},
  scale only axis={true},
  width={\axisdefaultwidth},
  ticklabel style={anchor=south, yshift=1.3*\pgfkeysvalueof{/pgfplots/major tick length}, font=\small},
  every tick/.style={draw=black},
  major tick style={yshift=.5*\pgfkeysvalueof{/pgfplots/major tick length}},
  minor tick style={yshift=.5*\pgfkeysvalueof{/pgfplots/minor tick length}},
  title style={yshift=\baselineskip},
  xmax={6},
  ymin={-4.5},
  height={5\baselineskip},
  xtick={1,2,3,4,5,6},
  minor x tick num={3},
  title={TU - PRR (MCR)},
]

\draw[treatment line] ([yshift=-2pt] axis cs:3.2543478260869567, 0) |- (axis cs:2.7543478260869567, -2.0)
  node[treatment label, anchor=east] {$\text{TU}_{\,\text{ord-var}}$};
\draw[treatment line] ([yshift=-2pt] axis cs:3.391304347826087, 0) |- (axis cs:2.7543478260869567, -3.0)
  node[treatment label, anchor=east] {$\text{TU}_{\,\text{bin-var}}$};
\draw[treatment line] ([yshift=-2pt] axis cs:3.4108695652173915, 0) |- (axis cs:2.7543478260869567, -4.0)
  node[treatment label, anchor=east] {$\text{TU}_{\,\text{bin-ent}}$};
\draw[treatment line] ([yshift=-2pt] axis cs:3.5282608695652176, 0) |- (axis cs:4.343478260869565, -4.0)
  node[treatment label, anchor=west] {$\text{TU}_{\,\text{ord-ent}}$};
\draw[treatment line] ([yshift=-2pt] axis cs:3.5717391304347825, 0) |- (axis cs:4.343478260869565, -3.0)
  node[treatment label, anchor=west] {$\text{TU}_{\,\text{ent}}$};
\draw[treatment line] ([yshift=-2pt] axis cs:3.8434782608695652, 0) |- (axis cs:4.343478260869565, -2.0)
  node[treatment label, anchor=west] {$\text{TU}_{\,\text{var}}$};
\draw[group line] (axis cs:3.2543478260869567, -1.3333333333333333) -- (axis cs:3.5717391304347825, -1.3333333333333333);
\draw[group line] (axis cs:3.391304347826087, -1.5333333333333332) -- (axis cs:3.8434782608695652, -1.5333333333333332);

\end{axis}
\end{tikzpicture}
\captionsetup{justification=centering}
\subcaption{Result only for MCR.}
\label{fig:cd_total_mcr}
\end{subfigure}
\begin{subfigure}[t]{0.48\linewidth}
\centering
\begin{tikzpicture}[scale=0.65,
  treatment line/.style={rounded corners=1.5pt, line cap=round, shorten >=1pt},
  treatment label/.style={font=\small},
  group line/.style={ultra thick},
]

\begin{axis}[
  clip={false},
  axis x line={center},
  axis y line={none},
  axis line style={-},
  xmin={1},
  ymax={0},
  scale only axis={true},
  width={\axisdefaultwidth},
  ticklabel style={anchor=south, yshift=1.3*\pgfkeysvalueof{/pgfplots/major tick length}, font=\small},
  every tick/.style={draw=black},
  major tick style={yshift=.5*\pgfkeysvalueof{/pgfplots/major tick length}},
  minor tick style={yshift=.5*\pgfkeysvalueof{/pgfplots/minor tick length}},
  title style={yshift=\baselineskip},
  xmax={6},
  ymin={-4.5},
  height={5\baselineskip},
  xtick={1,2,3,4,5,6},
  minor x tick num={3},
  title={TU - PRR (MAE)},
]

\draw[treatment line] ([yshift=-2pt] axis cs:3.1760869565217393, 0) |- (axis cs:2.6760869565217393, -2.0)
  node[treatment label, anchor=east] {$\text{TU}_{\,\text{ord-var}}$};
\draw[treatment line] ([yshift=-2pt] axis cs:3.3956521739130436, 0) |- (axis cs:2.6760869565217393, -3.0)
  node[treatment label, anchor=east] {$\text{TU}_{\,\text{ord-ent}}$};
\draw[treatment line] ([yshift=-2pt] axis cs:3.497826086956522, 0) |- (axis cs:2.6760869565217393, -4.0)
  node[treatment label, anchor=east] {$\text{TU}_{\,\text{bin-var}}$};
\draw[treatment line] ([yshift=-2pt] axis cs:3.5565217391304347, 0) |- (axis cs:4.215217391304348, -4.0)
  node[treatment label, anchor=west] {$\text{TU}_{\,\text{var}}$};
\draw[treatment line] ([yshift=-2pt] axis cs:3.658695652173913, 0) |- (axis cs:4.215217391304348, -3.0)
  node[treatment label, anchor=west] {$\text{TU}_{\,\text{ent}}$};
\draw[treatment line] ([yshift=-2pt] axis cs:3.715217391304348, 0) |- (axis cs:4.215217391304348, -2.0)
  node[treatment label, anchor=west] {$\text{TU}_{\,\text{bin-ent}}$};
\draw[group line] (axis cs:3.1760869565217393, -1.3333333333333333) -- (axis cs:3.5565217391304347, -1.3333333333333333);
\draw[group line] (axis cs:3.3956521739130436, -1.5333333333333332) -- (axis cs:3.715217391304348, -1.5333333333333332);

\end{axis}
\end{tikzpicture}
\captionsetup{justification=centering}
             \subcaption{Result only for MAE.}
             \label{fig:cd_total_mae}
         \end{subfigure}     

\end{figure} 

\footnotetext{\url{https://github.com/mirkobunse/critdd}}


\begin{figure}[!htbp]
  \centering 
  \caption{CD diagrams for Epistemic Uncertainty (EU) using an ensemble of GBTs (LightGBM \citep{DBLP:conf/nips/KeMFWCMYL17}).}
  \label{fig:cd_epistemic}
  \begin{subfigure}[t]{\linewidth}
    \centering 
\begin{tikzpicture}[
  treatment line/.style={rounded corners=1.5pt, line cap=round, shorten >=1pt},
  treatment label/.style={font=\small},
  group line/.style={ultra thick},
]

\begin{axis}[
  clip={false},
  axis x line={center},
  axis y line={none},
  axis line style={-},
  xmin={1},
  ymax={0},
  scale only axis={true},
  width={\axisdefaultwidth},
  ticklabel style={anchor=south, yshift=1.3*\pgfkeysvalueof{/pgfplots/major tick length}, font=\small},
  every tick/.style={draw=black},
  major tick style={yshift=.5*\pgfkeysvalueof{/pgfplots/major tick length}},
  minor tick style={yshift=.5*\pgfkeysvalueof{/pgfplots/minor tick length}},
  title style={yshift=\baselineskip},
  xmax={6},
  ymin={-4.5},
  height={5\baselineskip},
  xtick={1,2,3,4,5,6},
  minor x tick num={3},
  title={EU - PRR (MCR), PRR (MAE)},
]

\draw[treatment line] ([yshift=-2pt] axis cs:3.05, 0) |- (axis cs:2.55, -2.0)
  node[treatment label, anchor=east] {$\text{EU}_{\,\text{var}}$};
\draw[treatment line] ([yshift=-2pt] axis cs:3.1717391304347826, 0) |- (axis cs:2.55, -3.0)
  node[treatment label, anchor=east] {$\text{EU}_{\,\text{ord-var}}$};
\draw[treatment line] ([yshift=-2pt] axis cs:3.473913043478261, 0) |- (axis cs:2.55, -4.0)
  node[treatment label, anchor=east] {$\text{EU}_{\,\text{ent}}$};
\draw[treatment line] ([yshift=-2pt] axis cs:3.5521739130434784, 0) |- (axis cs:4.556521739130435, -4.0)
  node[treatment label, anchor=west] {$\text{EU}_{\,\text{ord-ent}}$};
\draw[treatment line] ([yshift=-2pt] axis cs:3.6956521739130435, 0) |- (axis cs:4.556521739130435, -3.0)
  node[treatment label, anchor=west] {$\text{EU}_{\,\text{bin-ent}}$};
\draw[treatment line] ([yshift=-2pt] axis cs:4.056521739130435, 0) |- (axis cs:4.556521739130435, -2.0)
  node[treatment label, anchor=west] {$\text{EU}_{\,\text{bin-var}}$};
\draw[group line] (axis cs:3.6956521739130435, -1.3333333333333333) -- (axis cs:4.056521739130435, -1.3333333333333333);
\draw[group line] (axis cs:3.473913043478261, -2.0) -- (axis cs:3.6956521739130435, -2.0);

\end{axis}
\end{tikzpicture}
\captionsetup{justification=centering}
\subcaption{Result for MCR and MAE.}
\label{fig:cd_epistemic_mcr_mae}
\end{subfigure}
\begin{subfigure}[t]{0.48\linewidth}
  \centering
\begin{tikzpicture}[scale=0.65,
  treatment line/.style={rounded corners=1.5pt, line cap=round, shorten >=1pt},
  treatment label/.style={font=\small},
  group line/.style={ultra thick},
]

\begin{axis}[
  clip={false},
  axis x line={center},
  axis y line={none},
  axis line style={-},
  xmin={1},
  ymax={0},
  scale only axis={true},
  width={\axisdefaultwidth},
  ticklabel style={anchor=south, yshift=1.3*\pgfkeysvalueof{/pgfplots/major tick length}, font=\small},
  every tick/.style={draw=black},
  major tick style={yshift=.5*\pgfkeysvalueof{/pgfplots/major tick length}},
  minor tick style={yshift=.5*\pgfkeysvalueof{/pgfplots/minor tick length}},
  title style={yshift=\baselineskip},
  xmax={6},
  ymin={-4.5},
  height={5\baselineskip},
  xtick={1,2,3,4,5,6},
  minor x tick num={3},
  title={EU - PRR (MCR)},
]

\draw[treatment line] ([yshift=-2pt] axis cs:3.128260869565217, 0) |- (axis cs:2.628260869565217, -2.0)
  node[treatment label, anchor=east] {$\text{EU}_{\,\text{var}}$};
\draw[treatment line] ([yshift=-2pt] axis cs:3.2065217391304346, 0) |- (axis cs:2.628260869565217, -3.0)
  node[treatment label, anchor=east] {$\text{EU}_{\,\text{ord-var}}$};
\draw[treatment line] ([yshift=-2pt] axis cs:3.408695652173913, 0) |- (axis cs:2.628260869565217, -4.0)
  node[treatment label, anchor=east] {$\text{EU}_{\,\text{ent}}$};
\draw[treatment line] ([yshift=-2pt] axis cs:3.5804347826086955, 0) |- (axis cs:4.4913043478260875, -4.0)
  node[treatment label, anchor=west] {$\text{EU}_{\,\text{bin-ent}}$};
\draw[treatment line] ([yshift=-2pt] axis cs:3.6847826086956523, 0) |- (axis cs:4.4913043478260875, -3.0)
  node[treatment label, anchor=west] {$\text{EU}_{\,\text{ord-ent}}$};
\draw[treatment line] ([yshift=-2pt] axis cs:3.991304347826087, 0) |- (axis cs:4.4913043478260875, -2.0)
  node[treatment label, anchor=west] {$\text{EU}_{\,\text{bin-var}}$};
\draw[group line] (axis cs:3.5804347826086955, -1.3333333333333333) -- (axis cs:3.991304347826087, -1.3333333333333333);
\draw[group line] (axis cs:3.408695652173913, -2.0) -- (axis cs:3.6847826086956523, -2.0);
\draw[group line] (axis cs:3.128260869565217, -1.5333333333333332) -- (axis cs:3.5804347826086955, -1.5333333333333332);

\end{axis}
\end{tikzpicture}
\captionsetup{justification=centering}
\subcaption{Result only for MCR.}
\label{fig:cd_epistemic_mcr}
\end{subfigure}
\begin{subfigure}[t]{0.48\linewidth}
  \centering
\begin{tikzpicture}[scale=0.65,
  treatment line/.style={rounded corners=1.5pt, line cap=round, shorten >=1pt},
  treatment label/.style={font=\small},
  group line/.style={ultra thick},
]

\begin{axis}[
  clip={false},
  axis x line={center},
  axis y line={none},
  axis line style={-},
  xmin={1},
  ymax={0},
  scale only axis={true},
  width={\axisdefaultwidth},
  ticklabel style={anchor=south, yshift=1.3*\pgfkeysvalueof{/pgfplots/major tick length}, font=\small},
  every tick/.style={draw=black},
  major tick style={yshift=.5*\pgfkeysvalueof{/pgfplots/major tick length}},
  minor tick style={yshift=.5*\pgfkeysvalueof{/pgfplots/minor tick length}},
  title style={yshift=\baselineskip},
  xmax={6},
  ymin={-4.5},
  height={5\baselineskip},
  xtick={1,2,3,4,5,6},
  minor x tick num={3},
  title={EU - PRR (MAE)},
]

\draw[treatment line] ([yshift=-2pt] axis cs:2.9717391304347824, 0) |- (axis cs:2.4717391304347824, -2.0)
  node[treatment label, anchor=east] {$\text{EU}_{\,\text{var}}$};
\draw[treatment line] ([yshift=-2pt] axis cs:3.1369565217391306, 0) |- (axis cs:2.4717391304347824, -3.0)
  node[treatment label, anchor=east] {$\text{EU}_{\,\text{ord-var}}$};
\draw[treatment line] ([yshift=-2pt] axis cs:3.4195652173913045, 0) |- (axis cs:2.4717391304347824, -4.0)
  node[treatment label, anchor=east] {$\text{EU}_{\,\text{ord-ent}}$};
\draw[treatment line] ([yshift=-2pt] axis cs:3.5391304347826087, 0) |- (axis cs:4.621739130434783, -4.0)
  node[treatment label, anchor=west] {$\text{EU}_{\,\text{ent}}$};
\draw[treatment line] ([yshift=-2pt] axis cs:3.8108695652173914, 0) |- (axis cs:4.621739130434783, -3.0)
  node[treatment label, anchor=west] {$\text{EU}_{\,\text{bin-ent}}$};
\draw[treatment line] ([yshift=-2pt] axis cs:4.121739130434783, 0) |- (axis cs:4.621739130434783, -2.0)
  node[treatment label, anchor=west] {$\text{EU}_{\,\text{bin-var}}$};
\draw[group line] (axis cs:3.8108695652173914, -1.3333333333333333) -- (axis cs:4.121739130434783, -1.3333333333333333);
\draw[group line] (axis cs:2.9717391304347824, -1.3333333333333333) -- (axis cs:3.1369565217391306, -1.3333333333333333);
\draw[group line] (axis cs:3.4195652173913045, -2.0) -- (axis cs:3.8108695652173914, -2.0);
\draw[group line] (axis cs:3.1369565217391306, -2.2) -- (axis cs:3.4195652173913045, -2.2);

\end{axis}
\end{tikzpicture}
\captionsetup{justification=centering}
\subcaption{Result only for MAE.}
\label{fig:cd_epistemic_mae}
         \end{subfigure}       
\end{figure}


\subsection{Out-Of-Distribution (OOD) Detection}
\label{subsec:ood_gbt}

A very critical and practically highly relevant challenge for machine learning models is the detection of out-of-distribution (OOD) data, which is data the learner has not seen during training, and which is sampled from a distribution that differs from the distribution of the training data. Think of malicious loan approval requests or unknown clinical conditions.
In such cases, the model should ideally be aware of its own incompetency and trigger appropriate fallback scenarios.
it is commonly assumed that OOD samples lead to high epistemic uncertainty.

To test this assumption in an OOD evaluation experiment, one typically first trains a model on an in-distribution (ID) dataset and computes uncertainty values on its corresponding test set.
Subsequently, the model is exposed to OOD data sampled from out-of-domain datasets. The model, which has not seen this data before, is then expected to exhibit increased epistemic uncertainty. Note that OOD detection for the ordinal case does not inherently differ from the nominal case, as the detection primarily operates in the input space $\mathcal{X}$. The purpose of this experiment is to ensure that our proposed OCS reduction is also competitive when it comes to OOD detection and not only on error detection. Concretely, one can evaluate the quality of OOD detection using the computed area under the receiver operating characteristic curve (AUC-ROC), in which OOD and ID data are given binary labels (0 for ID and 1 for OOD). The determined epistemic uncertainty values represent target scores \citep{DBLP:conf/iclr/HendrycksG17}.

For our used tabular ordinal datasets, we use the same approach as done by \cite{DBLP:conf/iclr/MalininPU21}. For each dataset, we take its test set as ID data. The OOD data of the same size is sampled from the Year MSD dataset \citep{kelly2023uci}. All numerical features are normalized by the per-column mean and variance obtained on the ID training data, and categorical features are uniformly sampled at random from the set of all categories of the particular feature.

\begin{figure}[!htbp]
  \centering 
  \caption{CD diagram for OOD detection on the basis of epistemic uncertainty quantified by the different measures using an ensemble of GBTs (LightGBM \citep{DBLP:conf/nips/KeMFWCMYL17}).}
  \label{fig:ood_lgbm}

  \begin{subfigure}[t]{\linewidth}
    \centering

\begin{tikzpicture}[
  treatment line/.style={rounded corners=1.5pt, line cap=round, shorten >=1pt},
  treatment label/.style={font=\small},
  group line/.style={ultra thick},
]

\begin{axis}[
  clip={false},
  axis x line={center},
  axis y line={none},
  axis line style={-},
  xmin={1},
  ymax={0},
  scale only axis={true},
  width={\axisdefaultwidth},
  ticklabel style={anchor=south, yshift=1.3*\pgfkeysvalueof{/pgfplots/major tick length}, font=\small},
  every tick/.style={draw=black},
  major tick style={yshift=.5*\pgfkeysvalueof{/pgfplots/major tick length}},
  minor tick style={yshift=.5*\pgfkeysvalueof{/pgfplots/minor tick length}},
  title style={yshift=\baselineskip},
  xmax={6},
  ymin={-4.5},
  height={5\baselineskip},
  xtick={1,2,3,4,5,6},
  minor x tick num={3},
  title={EU - OOD},
]

\draw[treatment line] ([yshift=-2pt] axis cs:2.378260869565217, 0) |- (axis cs:1.8782608695652172, -2.0)
  node[treatment label, anchor=east] {ord-ent};
\draw[treatment line] ([yshift=-2pt] axis cs:2.7934782608695654, 0) |- (axis cs:1.8782608695652172, -3.0)
  node[treatment label, anchor=east] {ent};
\draw[treatment line] ([yshift=-2pt] axis cs:3.152173913043478, 0) |- (axis cs:1.8782608695652172, -4.0)
  node[treatment label, anchor=east] {bin-ent};
\draw[treatment line] ([yshift=-2pt] axis cs:3.6565217391304348, 0) |- (axis cs:5.469565217391304, -4.0)
  node[treatment label, anchor=west] {var};
\draw[treatment line] ([yshift=-2pt] axis cs:4.05, 0) |- (axis cs:5.469565217391304, -3.0)
  node[treatment label, anchor=west] {ord-var};
\draw[treatment line] ([yshift=-2pt] axis cs:4.969565217391304, 0) |- (axis cs:5.469565217391304, -2.0)
  node[treatment label, anchor=west] {bin-var};
\draw[group line] (axis cs:3.152173913043478, -2.6666666666666665) -- (axis cs:3.6565217391304348, -2.6666666666666665);

\end{axis}
\end{tikzpicture}
\end{subfigure}

\end{figure}
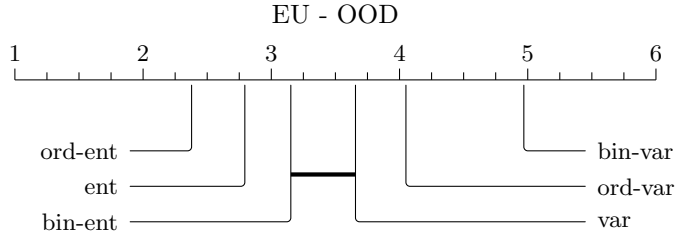

In the CD diagram in Figure \ref{fig:ood_lgbm}, we report the overall result of our OOD experiment based on epistemic uncertainty using AUC-ROC as the OOD performance metric and 10-fold cross-validation over all datasets (see Table \ref{tab:ood_gbt_detailed} for detailed results).
It appears as if entropy-based measures have a clear edge over variance-based measures when it comes to OOD detection. This is also underpinned when looking at the results obtained for an ensemble of MLPs (cf.\ Figure \ref{fig:ood_mlp}).
For detailed experimental OOD results, we refer to Appendix \ref{appendix:ood}. In general, we can conclude that our proposed OCS decomposition method appears competitive with standard and label-wise uncertainty approaches. However, unlike in error detection, entropy seems more appropriate for OOD detection than variance.


\section{Conclusion and Future Work}

In this paper, we have presented and experimentally evaluated several methods for quantifying aleatoric and epistemic uncertainty for probabilistic ordinal classification. Through visualizing accuracy-rejection curves and calculating prediction rejection ratios, we have demonstrated how all explored methods increase predictive performance and improve decision-making, with our OCS decomposition method using variance as base measure (ord-var) overall leading the field.
Furthermore, we have demonstrated the competitiveness of our ordinal approach compared to existing methods for OOD detection. Interestingly, for OOD detection, entropy emerges as the superior choice, not only as a base measure for the binary decomposition but also in general.

All in all, we were able to experimentally prove our hypothesis that measures disregarding the ordinal structure, such as entropy and the labelwise approaches, are not ideal candidates for quantifying uncertainty and disentangling aleatoric and epistemic uncertainty in ordinal classification.
To this end, one should rather consider our binary decomposition method or variance instead, as minimizing error distances is a crucial factor in ordinal classification. In particular, our binary decomposition method with variance as base measure (ord-var) strikes the best balance between exact hit-rate and reduced error distances, which is crucial in uncertainty quantification for ordinal classification. Despite the assumption of equal distances between classes, variance performs surprisingly well, suggesting that this assumption may not be a significant limitation in practice, or at least is justified for many ordinal datasets.

As a general recommendation, we suggest using variance if the predictor tends to commit large distance errors, as variance has a strong focus on the extreme bimodal distribution with all mass equally allocated to the extreme classes. Furthermore, it is also quite effective in detecting in-distribution instances with high epistemic uncertainty. However, if errors are not too widespread, our proposed OCS reduction achieves a favorable balance between exact hit-rate and error distance minimization.

Moreover, in ordinal classification, there exists a significant gap between predictive performance on ordinal metrics, such as QWK or RPS, and uncertainty quantification. While ordinal losses achieve strong results on these metrics, they negatively impact uncertainty quantification due to the strong inductive bias toward unimodal predictive probability distributions. This bias reduces variance, which, as demonstrated, is essential for reliable uncertainty quantification. Addressing this gap should be a key focus for future research. 
If reliable uncertainty quantification in ordinal classification is crucial, the CE loss appears to be the most suitable choice, as it leads to unbiased predictive probability distributions.

Additionally, future work could investigate uncertainty quantification for ordinal classification from a more theoretical point of view and try to assess or justify specific uncertainty measures axiomatically \citep{DBLP:conf/uai/WimmerSHBH23,bulte2025axiomatic}. Furthermore, from an experimental point of view, it might be interesting to explore additional non-tabular ordinal datasets (e.g., image datasets).

\begin{appendices}

\section{Prediction Rejection Ratios (PRR) - Detailed Results}
\label{asec:prrs}

In this section, we display the detailed prediction rejection ratio results for the different tabular ordinal benchmark datasets, uncertainty types and measures on the basis of an ensemble of GBTs (LightGBM \citep{DBLP:conf/nips/KeMFWCMYL17}) (cf.\ Subsection \ref{subsec:prrs}).
Table \ref{tab:prr_acc} shows results for missclassification rate (MCR) and Table \ref{tab:prr_mae} for mean absolute error (MAE).

\tiny 

\begin{longtable}{llrrrrrr}
  \caption{PRRs (MCR) using an ensemble of GBTs (LightGBM \citep{DBLP:conf/nips/KeMFWCMYL17}).} \label{tab:prr_acc} \\
  \toprule
   &  & \multicolumn{6}{c}{\textbf{PRR (MCR)} ($\uparrow$) } \\
   \midrule
   & \textbf{Measure} & bin-ent ($\uparrow$) & bin-var ($\uparrow$) & ent ($\uparrow$) & ord-ent ($\uparrow$) & ord-var ($\uparrow$) & var ($\uparrow$) \\
   \midrule
   \textbf{Dataset} & \textbf{Type} & & &  &  &  &  \\
  \endfirsthead
  \toprule
  &  & \multicolumn{6}{c}{\textbf{PRR (MCR)} ($\uparrow$)} \\
  \midrule
  & \textbf{Measure} & bin-ent ($\uparrow$) & bin-var ($\uparrow$) & ent ($\uparrow$) & ord-ent ($\uparrow$) & ord-var ($\uparrow$) & var ($\uparrow$) \\
  \midrule
  \textbf{Dataset} & \textbf{Type} &  &  &  & & & \\
  \midrule
  \endhead
  \midrule
  \multicolumn{6}{r}{Continued on next page} \\
  \midrule
  \endfoot
  \bottomrule
  \endlastfoot
  \midrule
  \multirow[t]{3}{*}{Abalone} & AU & 0.282\textpm0.06 & 0.286\textpm0.06 & 0.289\textpm0.06 & 0.293\textpm0.06 & \textbf{0.295}\textpm0.06 & \textbf{0.295}\textpm0.06 \\
 & EU & 0.1\textpm0.17 & 0.096\textpm0.16 & 0.118\textpm0.15 & 0.113\textpm0.15 & 0.121\textpm0.15 & \textbf{0.128}\textpm0.15 \\
 & TU & \textbf{0.282}\textpm0.05 & 0.278\textpm0.05 & 0.274\textpm0.04 & 0.251\textpm0.04 & 0.259\textpm0.04 & 0.24\textpm0.05 \\
\midrule
\multirow[t]{3}{*}{Auto MPG} & AU & 0.362\textpm0.17 & 0.383\textpm0.16 & 0.394\textpm0.15 & 0.429\textpm0.15 & 0.436\textpm0.14 & \textbf{0.446}\textpm0.15 \\
 & EU & 0.338\textpm0.17 & 0.349\textpm0.14 & 0.36\textpm0.16 & 0.394\textpm0.16 & 0.396\textpm0.17 & \textbf{0.424}\textpm0.18 \\
 & TU & 0.363\textpm0.18 & 0.383\textpm0.16 & 0.389\textpm0.16 & 0.435\textpm0.14 & 0.435\textpm0.14 & \textbf{0.451}\textpm0.16 \\
\midrule
\multirow[t]{3}{*}{Automobile} & AU & 0.637\textpm0.47 & 0.635\textpm0.47 & 0.625\textpm0.48 & 0.623\textpm0.47 & \textbf{0.644}\textpm0.48 & 0.581\textpm0.51 \\
 & EU & 0.597\textpm0.47 & 0.587\textpm0.47 & 0.609\textpm0.48 & 0.607\textpm0.47 & \textbf{0.622}\textpm0.47 & 0.617\textpm0.56 \\
 & TU & 0.637\textpm0.47 & 0.638\textpm0.48 & 0.634\textpm0.48 & 0.623\textpm0.47 & \textbf{0.645}\textpm0.48 & 0.581\textpm0.51 \\
\midrule
\multirow[t]{3}{*}{Balance Scale} & AU & 0.929\textpm0.06 & 0.924\textpm0.06 & 0.913\textpm0.07 & \textbf{0.96}\textpm0.04 & \textbf{0.96}\textpm0.04 & 0.945\textpm0.05 \\
 & EU & 0.842\textpm0.07 & 0.798\textpm0.08 & 0.818\textpm0.08 & 0.836\textpm0.06 & 0.879\textpm0.04 & \textbf{0.911}\textpm0.06 \\
 & TU & 0.927\textpm0.06 & 0.921\textpm0.06 & 0.912\textpm0.07 & \textbf{0.959}\textpm0.04 & \textbf{0.959}\textpm0.04 & 0.946\textpm0.05 \\
\midrule
\multirow[c]{3}{*}{\shortstack{Wisconsin \\Breast Cancer}} & AU & 0.275\textpm0.27 & 0.29\textpm0.27 & \textbf{0.294}\textpm0.27 & 0.255\textpm0.27 & 0.224\textpm0.29 & 0.175\textpm0.23 \\
 & EU & 0.131\textpm0.43 & \textbf{0.177}\textpm0.41 & \textbf{0.177}\textpm0.38 & 0.123\textpm0.41 & 0.104\textpm0.45 & 0.1\textpm0.44 \\
 & TU & 0.27\textpm0.3 & 0.285\textpm0.28 & \textbf{0.29}\textpm0.29 & 0.248\textpm0.28 & 0.224\textpm0.31 & 0.181\textpm0.25 \\
\midrule
\multirow[t]{3}{*}{ERA} & AU & 0.151\textpm0.11 & 0.153\textpm0.12 & \textbf{0.155}\textpm0.12 & 0.08\textpm0.08 & 0.098\textpm0.08 & 0.043\textpm0.07 \\
 & EU & \textbf{0.047}\textpm0.1 & 0.022\textpm0.11 & 0.025\textpm0.12 & -0.013\textpm0.12 & 0.012\textpm0.13 & -0.022\textpm0.13 \\
 & TU & 0.148\textpm0.11 & 0.153\textpm0.12 & \textbf{0.154}\textpm0.12 & 0.079\textpm0.08 & 0.099\textpm0.08 & 0.042\textpm0.06 \\
\midrule
\multirow[t]{3}{*}{ESL} & AU & 0.247\textpm0.13 & 0.241\textpm0.12 & 0.235\textpm0.11 & 0.262\textpm0.1 & 0.27\textpm0.12 & \textbf{0.278}\textpm0.11 \\
 & EU & 0.237\textpm0.13 & 0.229\textpm0.15 & 0.223\textpm0.14 & 0.226\textpm0.14 & 0.263\textpm0.11 & \textbf{0.27}\textpm0.11 \\
 & TU & 0.25\textpm0.13 & 0.241\textpm0.12 & 0.241\textpm0.11 & 0.263\textpm0.09 & 0.268\textpm0.11 & \textbf{0.277}\textpm0.11 \\
\midrule
\multirow[t]{3}{*}{Eucalyptus} & AU & \textbf{0.437}\textpm0.1 & 0.431\textpm0.11 & 0.424\textpm0.11 & 0.418\textpm0.12 & 0.428\textpm0.11 & 0.414\textpm0.12 \\
 & EU & 0.44\textpm0.08 & 0.434\textpm0.08 & 0.439\textpm0.08 & 0.437\textpm0.07 & \textbf{0.449}\textpm0.06 & 0.447\textpm0.06 \\
 & TU & \textbf{0.44}\textpm0.1 & 0.435\textpm0.1 & 0.428\textpm0.11 & 0.424\textpm0.11 & 0.433\textpm0.1 & 0.416\textpm0.12 \\
\midrule
\multirow[t]{3}{*}{Heart (CLE)} & AU & 0.618\textpm0.16 & 0.621\textpm0.16 & \textbf{0.629}\textpm0.16 & 0.616\textpm0.14 & 0.602\textpm0.14 & 0.582\textpm0.14 \\
 & EU & 0.518\textpm0.19 & 0.52\textpm0.17 & \textbf{0.559}\textpm0.17 & 0.549\textpm0.14 & 0.548\textpm0.15 & 0.541\textpm0.15 \\
 & TU & 0.615\textpm0.16 & 0.62\textpm0.16 & \textbf{0.623}\textpm0.16 & 0.616\textpm0.14 & 0.6\textpm0.14 & 0.582\textpm0.13 \\
\midrule
\multirow[t]{3}{*}{Boston Housing} & AU & \textbf{0.406}\textpm0.15 & 0.397\textpm0.15 & 0.393\textpm0.15 & 0.39\textpm0.15 & 0.398\textpm0.15 & 0.396\textpm0.14 \\
 & EU & 0.415\textpm0.15 & \textbf{0.416}\textpm0.15 & 0.41\textpm0.15 & 0.41\textpm0.14 & 0.414\textpm0.15 & 0.41\textpm0.14 \\
 & TU & \textbf{0.412}\textpm0.15 & 0.404\textpm0.15 & 0.4\textpm0.16 & 0.399\textpm0.15 & 0.408\textpm0.15 & 0.405\textpm0.14 \\
\midrule
\multirow[t]{3}{*}{LEV} & AU & 0.181\textpm0.1 & \textbf{0.182}\textpm0.1 & 0.174\textpm0.1 & 0.167\textpm0.11 & 0.178\textpm0.1 & 0.167\textpm0.1 \\
 & EU & 0.165\textpm0.12 & 0.161\textpm0.13 & 0.175\textpm0.13 & 0.185\textpm0.14 & 0.174\textpm0.11 & \textbf{0.196}\textpm0.11 \\
 & TU & \textbf{0.181}\textpm0.1 & \textbf{0.181}\textpm0.1 & 0.176\textpm0.1 & 0.166\textpm0.11 & 0.177\textpm0.1 & 0.168\textpm0.1 \\
\midrule
\multirow[t]{3}{*}{Machine CPU} & AU & 0.664\textpm0.11 & 0.692\textpm0.11 & 0.692\textpm0.1 & 0.727\textpm0.11 & 0.734\textpm0.1 & \textbf{0.742}\textpm0.11 \\
 & EU & 0.582\textpm0.16 & 0.619\textpm0.14 & 0.65\textpm0.13 & 0.675\textpm0.15 & 0.649\textpm0.15 & \textbf{0.702}\textpm0.15 \\
 & TU & 0.668\textpm0.1 & 0.684\textpm0.11 & 0.696\textpm0.1 & 0.727\textpm0.11 & 0.732\textpm0.11 & \textbf{0.742}\textpm0.11 \\
\midrule
\multirow[t]{3}{*}{New Thyroid} & AU & 0.54\textpm0.47 & 0.54\textpm0.47 & 0.54\textpm0.47 & \textbf{0.565}\textpm0.49 & \textbf{0.565}\textpm0.49 & 0.56\textpm0.48 \\
 & EU & 0.545\textpm0.48 & 0.545\textpm0.48 & 0.545\textpm0.48 & \textbf{0.57}\textpm0.49 & 0.56\textpm0.49 & 0.565\textpm0.49 \\
 & TU & 0.54\textpm0.47 & 0.535\textpm0.47 & 0.535\textpm0.47 & \textbf{0.565}\textpm0.49 & \textbf{0.565}\textpm0.49 & 0.56\textpm0.48 \\
\midrule
\multirow[t]{3}{*}{Pyrimidines} & AU & -0.094\textpm0.4 & -0.062\textpm0.35 & -0.062\textpm0.35 & 0.038\textpm0.56 & \textbf{0.099}\textpm0.31 & -0.01\textpm0.64 \\
 & EU & -0.058\textpm0.49 & -0.059\textpm0.37 & -0.041\textpm0.39 & -0.096\textpm0.52 & -0.039\textpm0.5 & \textbf{0.054}\textpm0.47 \\
 & TU & -0.094\textpm0.4 & -0.077\textpm0.43 & -0.077\textpm0.43 & 0.056\textpm0.58 & \textbf{0.099}\textpm0.31 & -0.01\textpm0.64 \\
\midrule
\multirow[t]{3}{*}{Red Wine} & AU & 0.432\textpm0.12 & 0.436\textpm0.12 & \textbf{0.438}\textpm0.12 & 0.433\textpm0.12 & 0.434\textpm0.12 & 0.429\textpm0.13 \\
 & EU & 0.438\textpm0.09 & 0.446\textpm0.08 & \textbf{0.456}\textpm0.09 & \textbf{0.456}\textpm0.09 & 0.443\textpm0.1 & 0.439\textpm0.11 \\
 & TU & 0.436\textpm0.11 & 0.442\textpm0.12 & \textbf{0.444}\textpm0.12 & 0.44\textpm0.12 & 0.437\textpm0.12 & 0.432\textpm0.13 \\
\midrule
\multirow[t]{3}{*}{SWD} & AU & 0.189\textpm0.09 & 0.188\textpm0.08 & 0.182\textpm0.08 & 0.189\textpm0.08 & \textbf{0.194}\textpm0.09 & 0.191\textpm0.09 \\
 & EU & 0.142\textpm0.11 & 0.115\textpm0.08 & 0.136\textpm0.09 & 0.142\textpm0.09 & 0.163\textpm0.1 & \textbf{0.185}\textpm0.11 \\
 & TU & 0.193\textpm0.09 & 0.187\textpm0.08 & 0.182\textpm0.08 & 0.189\textpm0.08 & \textbf{0.195}\textpm0.09 & 0.192\textpm0.09 \\
\midrule
\multirow[t]{3}{*}{Stocks Domain} & AU & \textbf{0.668}\textpm0.06 & \textbf{0.668}\textpm0.06 & \textbf{0.668}\textpm0.06 & 0.666\textpm0.07 & 0.666\textpm0.06 & 0.665\textpm0.07 \\
 & EU & 0.643\textpm0.07 & 0.627\textpm0.08 & 0.628\textpm0.08 & 0.627\textpm0.08 & \textbf{0.644}\textpm0.07 & 0.643\textpm0.07 \\
 & TU & 0.668\textpm0.06 & \textbf{0.669}\textpm0.06 & 0.666\textpm0.06 & 0.664\textpm0.06 & 0.668\textpm0.06 & 0.666\textpm0.07 \\
\midrule
\multirow[t]{3}{*}{TAE} & AU & \textbf{0.229}\textpm0.32 & 0.218\textpm0.34 & 0.213\textpm0.34 & 0.164\textpm0.31 & 0.17\textpm0.31 & 0.16\textpm0.26 \\
 & EU & 0.037\textpm0.27 & 0.001\textpm0.23 & 0.022\textpm0.27 & 0.013\textpm0.23 & 0.094\textpm0.24 & \textbf{0.108}\textpm0.16 \\
 & TU & \textbf{0.233}\textpm0.32 & 0.224\textpm0.33 & 0.197\textpm0.36 & 0.157\textpm0.31 & 0.166\textpm0.3 & 0.16\textpm0.26 \\
\midrule
\multirow[t]{3}{*}{Triazines} & AU & \textbf{0.308}\textpm0.2 & 0.284\textpm0.18 & 0.262\textpm0.18 & 0.255\textpm0.22 & 0.259\textpm0.22 & 0.232\textpm0.24 \\
 & EU & 0.241\textpm0.25 & 0.218\textpm0.23 & 0.236\textpm0.2 & 0.267\textpm0.23 & \textbf{0.276}\textpm0.24 & 0.272\textpm0.23 \\
 & TU & \textbf{0.315}\textpm0.19 & 0.287\textpm0.18 & 0.265\textpm0.18 & 0.255\textpm0.22 & 0.262\textpm0.22 & 0.234\textpm0.24 \\
\midrule
\multirow[t]{3}{*}{White Wine} & AU & \textbf{0.378}\textpm0.05 & 0.375\textpm0.05 & 0.368\textpm0.05 & 0.355\textpm0.05 & 0.366\textpm0.05 & 0.345\textpm0.05 \\
 & EU & 0.436\textpm0.05 & 0.438\textpm0.05 & \textbf{0.444}\textpm0.05 & 0.439\textpm0.04 & 0.433\textpm0.05 & 0.419\textpm0.05 \\
 & TU & \textbf{0.394}\textpm0.05 & 0.391\textpm0.05 & 0.383\textpm0.05 & 0.367\textpm0.05 & 0.379\textpm0.05 & 0.355\textpm0.05 \\
\midrule
\multirow[t]{3}{*}{CMC} & AU & 0.359\textpm0.1 & \textbf{0.36}\textpm0.1 & \textbf{0.36}\textpm0.1 & 0.302\textpm0.05 & 0.305\textpm0.05 & 0.252\textpm0.05 \\
 & EU & \textbf{0.285}\textpm0.09 & 0.236\textpm0.09 & 0.25\textpm0.09 & 0.214\textpm0.08 & 0.262\textpm0.08 & 0.235\textpm0.07 \\
 & TU & 0.358\textpm0.1 & \textbf{0.359}\textpm0.1 & \textbf{0.359}\textpm0.1 & 0.3\textpm0.05 & 0.303\textpm0.05 & 0.253\textpm0.05 \\
\midrule
\multirow[t]{3}{*}{Grub Damage} & AU & 0.222\textpm0.38 & 0.249\textpm0.39 & 0.248\textpm0.4 & \textbf{0.278}\textpm0.34 & 0.27\textpm0.33 & 0.221\textpm0.33 \\
 & EU & 0.142\textpm0.31 & 0.174\textpm0.3 & \textbf{0.223}\textpm0.37 & 0.174\textpm0.35 & 0.155\textpm0.37 & 0.135\textpm0.36 \\
 & TU & 0.222\textpm0.38 & 0.242\textpm0.39 & 0.251\textpm0.4 & \textbf{0.277}\textpm0.34 & \textbf{0.277}\textpm0.33 & 0.229\textpm0.34 \\
\midrule
\multirow[t]{3}{*}{Obesity} & AU & 0.891\textpm0.08 & 0.892\textpm0.08 & 0.893\textpm0.08 & 0.895\textpm0.08 & 0.895\textpm0.08 & \textbf{0.898}\textpm0.09 \\
 & EU & 0.892\textpm0.08 & 0.892\textpm0.09 & 0.893\textpm0.09 & 0.896\textpm0.09 & 0.895\textpm0.08 & \textbf{0.899}\textpm0.09 \\
 & TU & 0.891\textpm0.08 & 0.892\textpm0.08 & 0.893\textpm0.09 & 0.896\textpm0.09 & 0.895\textpm0.09 & \textbf{0.899}\textpm0.09 \\
    \end{longtable}

\begin{longtable}{llrrrrrr}
  \caption{PRRs (MAE) using an ensemble of GBTs (LightGBM \citep{DBLP:conf/nips/KeMFWCMYL17}).} \label{tab:prr_mae} \\
  \toprule
   &  & \multicolumn{6}{c}{\textbf{PRR (MAE)} ($\uparrow$) } \\
   \midrule
   & \textbf{Measure} & bin-ent ($\uparrow$) & bin-var ($\uparrow$) & ent ($\uparrow$) & ord-ent ($\uparrow$) & ord-var ($\uparrow$) & var ($\uparrow$) \\
   \midrule
   \textbf{Dataset} & \textbf{Type} & & &  &  &  &  \\
  \endfirsthead
  \toprule
  &  & \multicolumn{6}{c}{\textbf{PRR (MAE)} ($\uparrow$)} \\
  \midrule
  & \textbf{Measure} & bin-ent ($\uparrow$) & bin-var ($\uparrow$) & ent ($\uparrow$) & ord-ent ($\uparrow$) & ord-var ($\uparrow$) & var ($\uparrow$) \\
  \midrule
  \textbf{Dataset} & \textbf{Type} &  &  &  & & & \\
  \midrule
  \endhead
  \midrule
  \multicolumn{6}{r}{Continued on next page} \\
  \midrule
  \endfoot
  \bottomrule
  \endlastfoot
  \midrule
  \multirow[t]{3}{*}{Abalone} & AU & 0.299\textpm0.05 & 0.315\textpm0.05 & 0.323\textpm0.05 & 0.334\textpm0.05 & 0.328\textpm0.05 & \textbf{0.335}\textpm0.04 \\
 & EU & 0.113\textpm0.18 & 0.115\textpm0.18 & 0.141\textpm0.16 & 0.136\textpm0.16 & 0.138\textpm0.16 & \textbf{0.15}\textpm0.15 \\
 & TU & 0.304\textpm0.03 & 0.312\textpm0.03 & \textbf{0.313}\textpm0.02 & 0.295\textpm0.02 & 0.295\textpm0.02 & 0.282\textpm0.03 \\
\midrule
\multirow[t]{3}{*}{Auto MPG} & AU & 0.332\textpm0.17 & 0.361\textpm0.16 & 0.38\textpm0.14 & 0.47\textpm0.12 & 0.47\textpm0.12 & \textbf{0.495}\textpm0.12 \\
 & EU & 0.332\textpm0.15 & 0.376\textpm0.12 & 0.39\textpm0.13 & 0.437\textpm0.15 & 0.434\textpm0.14 & \textbf{0.479}\textpm0.14 \\
 & TU & 0.328\textpm0.19 & 0.362\textpm0.16 & 0.381\textpm0.14 & 0.475\textpm0.12 & 0.476\textpm0.11 & \textbf{0.501}\textpm0.12 \\
\midrule
\multirow[t]{3}{*}{Automobile} & AU & 0.557\textpm0.46 & 0.556\textpm0.47 & 0.549\textpm0.47 & 0.562\textpm0.47 & \textbf{0.581}\textpm0.47 & 0.516\textpm0.5 \\
 & EU & 0.534\textpm0.46 & 0.528\textpm0.47 & 0.542\textpm0.49 & \textbf{0.556}\textpm0.47 & 0.555\textpm0.45 & 0.545\textpm0.54 \\
 & TU & 0.554\textpm0.46 & 0.561\textpm0.47 & 0.557\textpm0.47 & 0.563\textpm0.47 & \textbf{0.582}\textpm0.47 & 0.517\textpm0.5 \\
\midrule
\multirow[t]{3}{*}{Balance Scale} & AU & \textbf{0.87}\textpm0.06 & 0.864\textpm0.06 & 0.856\textpm0.07 & 0.848\textpm0.09 & 0.85\textpm0.08 & 0.854\textpm0.09 \\
 & EU & 0.804\textpm0.09 & 0.766\textpm0.1 & 0.78\textpm0.09 & 0.791\textpm0.11 & \textbf{0.821}\textpm0.11 & 0.813\textpm0.1 \\
 & TU & \textbf{0.863}\textpm0.06 & 0.861\textpm0.06 & 0.856\textpm0.07 & 0.847\textpm0.09 & 0.848\textpm0.09 & 0.857\textpm0.1 \\
\midrule
\multirow[c]{3}{*}{\shortstack{Wisconsin \\Breast Cancer}} & AU & 0.155\textpm0.33 & 0.179\textpm0.3 & \textbf{0.191}\textpm0.28 & 0.143\textpm0.27 & 0.123\textpm0.29 & 0.105\textpm0.27 \\
 & EU & -0.004\textpm0.21 & 0.031\textpm0.21 & \textbf{0.05}\textpm0.19 & 0.004\textpm0.26 & -0.009\textpm0.24 & 0.009\textpm0.26 \\
 & TU & 0.148\textpm0.33 & 0.181\textpm0.3 & \textbf{0.188}\textpm0.28 & 0.13\textpm0.27 & 0.123\textpm0.29 & 0.111\textpm0.26 \\
\midrule
\multirow[t]{3}{*}{ERA} & AU & 0.024\textpm0.08 & 0.013\textpm0.07 & 0.014\textpm0.07 & 0.02\textpm0.1 & \textbf{0.034}\textpm0.09 & 0.017\textpm0.11 \\
 & EU & -0.024\textpm0.13 & -0.019\textpm0.12 & -0.021\textpm0.12 & -0.033\textpm0.12 & -0.024\textpm0.13 & \textbf{-0.013}\textpm0.13 \\
 & TU & 0.022\textpm0.08 & 0.011\textpm0.07 & 0.01\textpm0.07 & 0.019\textpm0.1 & \textbf{0.035}\textpm0.09 & 0.016\textpm0.11 \\
\midrule
\multirow[t]{3}{*}{ESL} & AU & 0.257\textpm0.12 & 0.255\textpm0.12 & 0.254\textpm0.11 & 0.278\textpm0.11 & 0.285\textpm0.12 & \textbf{0.295}\textpm0.12 \\
 & EU & 0.262\textpm0.12 & 0.255\textpm0.16 & 0.249\textpm0.16 & 0.252\textpm0.16 & 0.288\textpm0.11 & \textbf{0.293}\textpm0.12 \\
 & TU & 0.262\textpm0.11 & 0.257\textpm0.11 & 0.259\textpm0.11 & 0.28\textpm0.11 & 0.284\textpm0.11 & \textbf{0.294}\textpm0.12 \\
\midrule
\multirow[t]{3}{*}{Eucalyptus} & AU & \textbf{0.42}\textpm0.1 & 0.417\textpm0.1 & 0.412\textpm0.11 & 0.41\textpm0.12 & \textbf{0.42}\textpm0.11 & 0.407\textpm0.12 \\
 & EU & 0.43\textpm0.08 & 0.429\textpm0.08 & 0.434\textpm0.08 & 0.439\textpm0.08 & \textbf{0.444}\textpm0.07 & \textbf{0.444}\textpm0.08 \\
 & TU & \textbf{0.425}\textpm0.1 & 0.421\textpm0.1 & 0.416\textpm0.11 & 0.417\textpm0.11 & \textbf{0.425}\textpm0.11 & 0.41\textpm0.12 \\
\midrule
\multirow[t]{3}{*}{Heart (CLE)} & AU & 0.56\textpm0.14 & 0.565\textpm0.14 & 0.577\textpm0.13 & \textbf{0.581}\textpm0.1 & 0.571\textpm0.1 & 0.556\textpm0.1 \\
 & EU & 0.487\textpm0.19 & 0.491\textpm0.17 & 0.519\textpm0.17 & 0.533\textpm0.13 & 0.534\textpm0.12 & \textbf{0.535}\textpm0.1 \\
 & TU & 0.553\textpm0.14 & 0.566\textpm0.14 & 0.568\textpm0.14 & \textbf{0.583}\textpm0.1 & 0.569\textpm0.09 & 0.559\textpm0.09 \\
\midrule
\multirow[t]{3}{*}{Boston Housing} & AU & 0.402\textpm0.2 & 0.394\textpm0.19 & 0.391\textpm0.2 & 0.391\textpm0.19 & 0.398\textpm0.19 & \textbf{0.403}\textpm0.17 \\
 & EU & 0.413\textpm0.19 & 0.42\textpm0.18 & 0.416\textpm0.18 & \textbf{0.421}\textpm0.18 & 0.413\textpm0.19 & 0.413\textpm0.18 \\
 & TU & 0.409\textpm0.2 & 0.401\textpm0.2 & 0.397\textpm0.2 & 0.4\textpm0.19 & 0.407\textpm0.19 & \textbf{0.412}\textpm0.18 \\
\midrule
\multirow[t]{3}{*}{LEV} & AU & 0.176\textpm0.1 & \textbf{0.178}\textpm0.1 & 0.173\textpm0.1 & 0.168\textpm0.1 & 0.175\textpm0.09 & 0.166\textpm0.1 \\
 & EU & 0.164\textpm0.12 & 0.167\textpm0.12 & 0.181\textpm0.13 & 0.191\textpm0.13 & 0.174\textpm0.1 & \textbf{0.193}\textpm0.1 \\
 & TU & \textbf{0.177}\textpm0.1 & \textbf{0.177}\textpm0.1 & 0.175\textpm0.1 & 0.167\textpm0.1 & 0.175\textpm0.1 & 0.166\textpm0.1 \\
\midrule
\multirow[t]{3}{*}{Machine CPU} & AU & 0.643\textpm0.16 & 0.684\textpm0.16 & 0.69\textpm0.15 & 0.733\textpm0.1 & 0.734\textpm0.09 & \textbf{0.747}\textpm0.08 \\
 & EU & 0.571\textpm0.18 & 0.623\textpm0.16 & 0.651\textpm0.16 & 0.695\textpm0.14 & 0.653\textpm0.16 & \textbf{0.712}\textpm0.13 \\
 & TU & 0.648\textpm0.16 & 0.677\textpm0.16 & 0.691\textpm0.15 & 0.731\textpm0.1 & 0.73\textpm0.1 & \textbf{0.747}\textpm0.08 \\
\midrule
\multirow[t]{3}{*}{New Thyroid} & AU & 0.54\textpm0.47 & 0.54\textpm0.47 & 0.54\textpm0.47 & \textbf{0.565}\textpm0.49 & \textbf{0.565}\textpm0.49 & 0.56\textpm0.48 \\
 & EU & 0.545\textpm0.48 & 0.545\textpm0.48 & 0.545\textpm0.48 & \textbf{0.57}\textpm0.49 & 0.56\textpm0.49 & 0.565\textpm0.49 \\
 & TU & 0.54\textpm0.47 & 0.535\textpm0.47 & 0.535\textpm0.47 & \textbf{0.565}\textpm0.49 & \textbf{0.565}\textpm0.49 & 0.56\textpm0.48 \\
\midrule
\multirow[t]{3}{*}{Pyrimidines} & AU & 0.231\textpm0.45 & 0.225\textpm0.4 & 0.192\textpm0.42 & 0.257\textpm0.34 & 0.255\textpm0.33 & \textbf{0.327}\textpm0.43 \\
 & EU & \textbf{0.219}\textpm0.34 & 0.04\textpm0.35 & 0.024\textpm0.34 & -0.013\textpm0.51 & -0.011\textpm0.51 & -0.071\textpm0.53 \\
 & TU & 0.237\textpm0.42 & 0.214\textpm0.39 & 0.216\textpm0.39 & 0.268\textpm0.36 & 0.248\textpm0.33 & \textbf{0.29}\textpm0.49 \\
\midrule
\multirow[t]{3}{*}{Red Wine} & AU & 0.429\textpm0.1 & 0.435\textpm0.11 & \textbf{0.44}\textpm0.11 & 0.436\textpm0.11 & 0.433\textpm0.11 & 0.432\textpm0.12 \\
 & EU & 0.431\textpm0.08 & 0.437\textpm0.08 & \textbf{0.446}\textpm0.09 & \textbf{0.446}\textpm0.09 & 0.436\textpm0.09 & 0.431\textpm0.1 \\
 & TU & 0.434\textpm0.1 & 0.442\textpm0.11 & \textbf{0.447}\textpm0.11 & 0.443\textpm0.11 & 0.436\textpm0.11 & 0.435\textpm0.11 \\
\midrule
\multirow[t]{3}{*}{SWD} & AU & 0.136\textpm0.06 & 0.135\textpm0.07 & 0.13\textpm0.07 & 0.14\textpm0.07 & 0.144\textpm0.07 & \textbf{0.149}\textpm0.08 \\
 & EU & 0.129\textpm0.09 & 0.11\textpm0.07 & 0.123\textpm0.08 & 0.122\textpm0.07 & 0.138\textpm0.07 & \textbf{0.154}\textpm0.07 \\
 & TU & 0.138\textpm0.06 & 0.134\textpm0.07 & 0.13\textpm0.07 & 0.141\textpm0.07 & 0.146\textpm0.07 & \textbf{0.15}\textpm0.08 \\
\midrule
\multirow[t]{3}{*}{Stocks Domain} & AU & \textbf{0.668}\textpm0.06 & \textbf{0.668}\textpm0.06 & \textbf{0.668}\textpm0.06 & 0.666\textpm0.07 & 0.666\textpm0.06 & 0.665\textpm0.07 \\
 & EU & 0.643\textpm0.07 & 0.627\textpm0.08 & 0.628\textpm0.08 & 0.627\textpm0.08 & \textbf{0.644}\textpm0.07 & 0.643\textpm0.07 \\
 & TU & 0.668\textpm0.06 & \textbf{0.669}\textpm0.06 & 0.666\textpm0.06 & 0.664\textpm0.06 & 0.668\textpm0.06 & 0.666\textpm0.07 \\
\midrule
\multirow[t]{3}{*}{TAE} & AU & 0.059\textpm0.3 & 0.048\textpm0.33 & 0.044\textpm0.33 & 0.235\textpm0.22 & 0.243\textpm0.21 & \textbf{0.256}\textpm0.2 \\
 & EU & 0.094\textpm0.13 & 0.08\textpm0.11 & 0.091\textpm0.15 & 0.116\textpm0.16 & 0.213\textpm0.17 & \textbf{0.224}\textpm0.22 \\
 & TU & 0.063\textpm0.3 & 0.059\textpm0.31 & 0.041\textpm0.34 & 0.229\textpm0.21 & 0.239\textpm0.21 & \textbf{0.265}\textpm0.2 \\
\midrule
\multirow[t]{3}{*}{Triazines} & AU & \textbf{0.348}\textpm0.21 & 0.332\textpm0.17 & 0.314\textpm0.16 & 0.326\textpm0.21 & 0.337\textpm0.22 & 0.299\textpm0.22 \\
 & EU & 0.304\textpm0.22 & 0.32\textpm0.16 & 0.333\textpm0.13 & 0.362\textpm0.17 & 0.364\textpm0.18 & \textbf{0.365}\textpm0.19 \\
 & TU & \textbf{0.352}\textpm0.2 & 0.336\textpm0.18 & 0.316\textpm0.16 & 0.326\textpm0.21 & 0.335\textpm0.22 & 0.301\textpm0.22 \\
\midrule
\multirow[t]{3}{*}{White Wine} & AU & 0.374\textpm0.05 & \textbf{0.376}\textpm0.04 & 0.375\textpm0.04 & 0.365\textpm0.04 & 0.37\textpm0.04 & 0.354\textpm0.04 \\
 & EU & 0.419\textpm0.06 & 0.422\textpm0.06 & \textbf{0.433}\textpm0.06 & 0.427\textpm0.05 & 0.42\textpm0.06 & 0.41\textpm0.06 \\
 & TU & 0.39\textpm0.05 & \textbf{0.393}\textpm0.04 & 0.388\textpm0.04 & 0.376\textpm0.04 & 0.383\textpm0.04 & 0.364\textpm0.04 \\
\midrule
\multirow[t]{3}{*}{CMC} & AU & 0.221\textpm0.06 & 0.219\textpm0.06 & 0.218\textpm0.06 & 0.288\textpm0.06 & 0.289\textpm0.05 & \textbf{0.294}\textpm0.07 \\
 & EU & 0.23\textpm0.07 & 0.203\textpm0.07 & 0.207\textpm0.07 & 0.215\textpm0.08 & 0.253\textpm0.08 & \textbf{0.267}\textpm0.08 \\
 & TU & 0.22\textpm0.06 & 0.219\textpm0.06 & 0.217\textpm0.06 & 0.288\textpm0.06 & 0.289\textpm0.05 & \textbf{0.295}\textpm0.07 \\
\midrule
\multirow[t]{3}{*}{Grub Damage} & AU & 0.231\textpm0.28 & 0.251\textpm0.27 & 0.246\textpm0.28 & \textbf{0.339}\textpm0.17 & 0.334\textpm0.16 & 0.303\textpm0.18 \\
 & EU & -0.016\textpm0.27 & -0.002\textpm0.27 & 0.022\textpm0.29 & 0.01\textpm0.28 & 0.04\textpm0.26 & \textbf{0.062}\textpm0.23 \\
 & TU & 0.221\textpm0.28 & 0.245\textpm0.29 & 0.246\textpm0.28 & \textbf{0.336}\textpm0.17 & \textbf{0.336}\textpm0.16 & 0.305\textpm0.17 \\
\midrule
\multirow[t]{3}{*}{Obesity} & AU & 0.895\textpm0.07 & 0.896\textpm0.08 & 0.898\textpm0.08 & 0.899\textpm0.08 & 0.899\textpm0.08 & \textbf{0.903}\textpm0.08 \\
 & EU & 0.896\textpm0.08 & 0.896\textpm0.08 & 0.898\textpm0.08 & 0.9\textpm0.08 & 0.9\textpm0.08 & \textbf{0.903}\textpm0.08 \\
 & TU & 0.895\textpm0.08 & 0.896\textpm0.08 & 0.898\textpm0.08 & 0.9\textpm0.08 & 0.899\textpm0.08 & \textbf{0.903}\textpm0.08 \\
    \end{longtable}

    \normalsize
  \section{Additional Experiments with Ensembles of Gradient Boosted Trees (GBTs)}
  \label{asec:add_exp_gbts}

For the sake of completeness, we also include experimental results for error and OOD detection for the two other popular gradient boosting tree libraries, namely XGBoost \citep{DBLP:conf/kdd/ChenG16} and CatBoost \citep{DBLP:conf/nips/ProkhorenkovaGV18}.
We also create an ensemble of 10 XGBoost and CatBoost GBTs, respectively with subsample rate set to 0.5 to induce stochasticity in the training process. As our primary focus is on uncertainty quantification and not on predictive performance, we leave the parameters with the default values and do not perform any hyperparameter tuning.

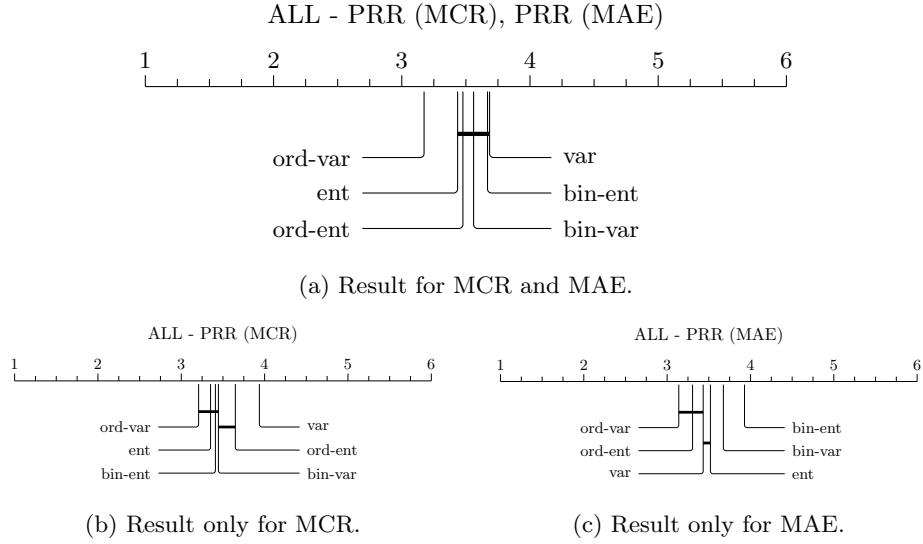
\begin{figure}[!htbp]
  \centering 
  \caption{CD diagrams for all uncertainty types (AU, EU and TU) using an ensemble of GBTs (XGBoost \citep{DBLP:conf/kdd/ChenG16}).}
  \label{fig:cd_all_xgb}
  \begin{subfigure}[t]{\linewidth}
    \centering 
\begin{tikzpicture}[
  treatment line/.style={rounded corners=1.5pt, line cap=round, shorten >=1pt},
  treatment label/.style={font=\small},
  group line/.style={ultra thick},
]

\begin{axis}[
  clip={false},
  axis x line={center},
  axis y line={none},
  axis line style={-},
  xmin={1},
  ymax={0},
  scale only axis={true},
  width={\axisdefaultwidth},
  ticklabel style={anchor=south, yshift=1.3*\pgfkeysvalueof{/pgfplots/major tick length}, font=\small},
  every tick/.style={draw=black},
  major tick style={yshift=.5*\pgfkeysvalueof{/pgfplots/major tick length}},
  minor tick style={yshift=.5*\pgfkeysvalueof{/pgfplots/minor tick length}},
  title style={yshift=\baselineskip},
  xmax={6},
  ymin={-4.5},
  height={5\baselineskip},
  xtick={1,2,3,4,5,6},
  minor x tick num={3},
  title={ALL - PRR (MCR), PRR (MAE)},
]

\draw[treatment line] ([yshift=-2pt] axis cs:3.1739130434782608, 0) |- (axis cs:2.6739130434782608, -2.0)
  node[treatment label, anchor=east] {ord-var};
\draw[treatment line] ([yshift=-2pt] axis cs:3.4355072463768117, 0) |- (axis cs:2.6739130434782608, -3.0)
  node[treatment label, anchor=east] {ent};
\draw[treatment line] ([yshift=-2pt] axis cs:3.4768115942028985, 0) |- (axis cs:2.6739130434782608, -4.0)
  node[treatment label, anchor=east] {ord-ent};
\draw[treatment line] ([yshift=-2pt] axis cs:3.560144927536232, 0) |- (axis cs:4.184782608695652, -4.0)
  node[treatment label, anchor=west] {bin-var};
\draw[treatment line] ([yshift=-2pt] axis cs:3.668840579710145, 0) |- (axis cs:4.184782608695652, -3.0)
  node[treatment label, anchor=west] {bin-ent};
\draw[treatment line] ([yshift=-2pt] axis cs:3.6847826086956523, 0) |- (axis cs:4.184782608695652, -2.0)
  node[treatment label, anchor=west] {var};
\draw[group line] (axis cs:3.4355072463768117, -1.3333333333333333) -- (axis cs:3.6847826086956523, -1.3333333333333333);

\end{axis}
\end{tikzpicture}
\captionsetup{justification=centering}
\subcaption{Result for MCR and MAE.}
\label{fig:cd_all_mcr_mae_xgb}
\end{subfigure}
\begin{subfigure}[t]{0.48\linewidth}
  \centering
\begin{tikzpicture}[scale=0.65,
  treatment line/.style={rounded corners=1.5pt, line cap=round, shorten >=1pt},
  treatment label/.style={font=\small},
  group line/.style={ultra thick},
]

\begin{axis}[
  clip={false},
  axis x line={center},
  axis y line={none},
  axis line style={-},
  xmin={1},
  ymax={0},
  scale only axis={true},
  width={\axisdefaultwidth},
  ticklabel style={anchor=south, yshift=1.3*\pgfkeysvalueof{/pgfplots/major tick length}, font=\small},
  every tick/.style={draw=black},
  major tick style={yshift=.5*\pgfkeysvalueof{/pgfplots/major tick length}},
  minor tick style={yshift=.5*\pgfkeysvalueof{/pgfplots/minor tick length}},
  title style={yshift=\baselineskip},
  xmax={6},
  ymin={-4.5},
  height={5\baselineskip},
  xtick={1,2,3,4,5,6},
  minor x tick num={3},
  title={ALL - PRR (MCR)},
]

\draw[treatment line] ([yshift=-2pt] axis cs:3.207971014492754, 0) |- (axis cs:2.707971014492754, -2.0)
  node[treatment label, anchor=east] {ord-var};
\draw[treatment line] ([yshift=-2pt] axis cs:3.351449275362319, 0) |- (axis cs:2.707971014492754, -3.0)
  node[treatment label, anchor=east] {ent};
\draw[treatment line] ([yshift=-2pt] axis cs:3.4094202898550723, 0) |- (axis cs:2.707971014492754, -4.0)
  node[treatment label, anchor=east] {bin-ent};
\draw[treatment line] ([yshift=-2pt] axis cs:3.4456521739130435, 0) |- (axis cs:4.4369565217391305, -4.0)
  node[treatment label, anchor=west] {bin-var};
\draw[treatment line] ([yshift=-2pt] axis cs:3.648550724637681, 0) |- (axis cs:4.4369565217391305, -3.0)
  node[treatment label, anchor=west] {ord-ent};
\draw[treatment line] ([yshift=-2pt] axis cs:3.9369565217391305, 0) |- (axis cs:4.4369565217391305, -2.0)
  node[treatment label, anchor=west] {var};
\draw[group line] (axis cs:3.4456521739130435, -2.0) -- (axis cs:3.648550724637681, -2.0);
\draw[group line] (axis cs:3.207971014492754, -1.3333333333333333) -- (axis cs:3.4456521739130435, -1.3333333333333333);

\end{axis}
\end{tikzpicture}
\captionsetup{justification=centering}
\subcaption{Result only for MCR.}
\label{fig:cd_all_mcr_xgb}
\end{subfigure}
\begin{subfigure}[t]{0.48\linewidth}
  \centering
\begin{tikzpicture}[scale=0.65,
  treatment line/.style={rounded corners=1.5pt, line cap=round, shorten >=1pt},
  treatment label/.style={font=\small},
  group line/.style={ultra thick},
]

\begin{axis}[
  clip={false},
  axis x line={center},
  axis y line={none},
  axis line style={-},
  xmin={1},
  ymax={0},
  scale only axis={true},
  width={\axisdefaultwidth},
  ticklabel style={anchor=south, yshift=1.3*\pgfkeysvalueof{/pgfplots/major tick length}, font=\small},
  every tick/.style={draw=black},
  major tick style={yshift=.5*\pgfkeysvalueof{/pgfplots/major tick length}},
  minor tick style={yshift=.5*\pgfkeysvalueof{/pgfplots/minor tick length}},
  title style={yshift=\baselineskip},
  xmax={6},
  ymin={-4.5},
  height={5\baselineskip},
  xtick={1,2,3,4,5,6},
  minor x tick num={3},
  title={ALL - PRR (MAE)},
]

\draw[treatment line] ([yshift=-2pt] axis cs:3.139855072463768, 0) |- (axis cs:2.639855072463768, -2.0)
  node[treatment label, anchor=east] {ord-var};
\draw[treatment line] ([yshift=-2pt] axis cs:3.305072463768116, 0) |- (axis cs:2.639855072463768, -3.0)
  node[treatment label, anchor=east] {ord-ent};
\draw[treatment line] ([yshift=-2pt] axis cs:3.4326086956521737, 0) |- (axis cs:2.639855072463768, -4.0)
  node[treatment label, anchor=east] {var};
\draw[treatment line] ([yshift=-2pt] axis cs:3.5195652173913046, 0) |- (axis cs:4.428260869565218, -4.0)
  node[treatment label, anchor=west] {ent};
\draw[treatment line] ([yshift=-2pt] axis cs:3.67463768115942, 0) |- (axis cs:4.428260869565218, -3.0)
  node[treatment label, anchor=west] {bin-var};
\draw[treatment line] ([yshift=-2pt] axis cs:3.9282608695652175, 0) |- (axis cs:4.428260869565218, -2.0)
  node[treatment label, anchor=west] {bin-ent};
\draw[group line] (axis cs:3.4326086956521737, -2.6666666666666665) -- (axis cs:3.5195652173913046, -2.6666666666666665);
\draw[group line] (axis cs:3.139855072463768, -1.3333333333333333) -- (axis cs:3.4326086956521737, -1.3333333333333333);

\end{axis}
\end{tikzpicture}

\captionsetup{justification=centering}
\subcaption{Result only for MAE.}
\label{fig:cd_all_mae_xgb}
         \end{subfigure}              
\end{figure}

On the one hand, overall error detection results for all uncertainty types (AU, EU, and TU) resemble those obtained for LightGBM (cf.\ Figure \ref{fig:cd_all}), with ord-var significantly outperforming all other measures when considering MCR and MAE simultaneously (cf.\ Figures \ref{fig:cd_all_xgb} and \ref{fig:cd_all_cat}). On the other hand, nominal uncertainty measures like ent or bin-ent perform better for XGBoost and CatBoost, which leads to ord-ent and var falling behind overall. Or phrased differently, we can spot the weakness of var when it comes to MCR (cf.\ Figure \ref{fig:cd_all_mcr_xgb}) as well as the weaker performance of ord-ent compared to ord-var and var when it comes to MAE.
Here it becomes even clearer that, in particular, ord-var captures the inherent trade-off between exact hit rate and minimized error distance for ordinal classification best.

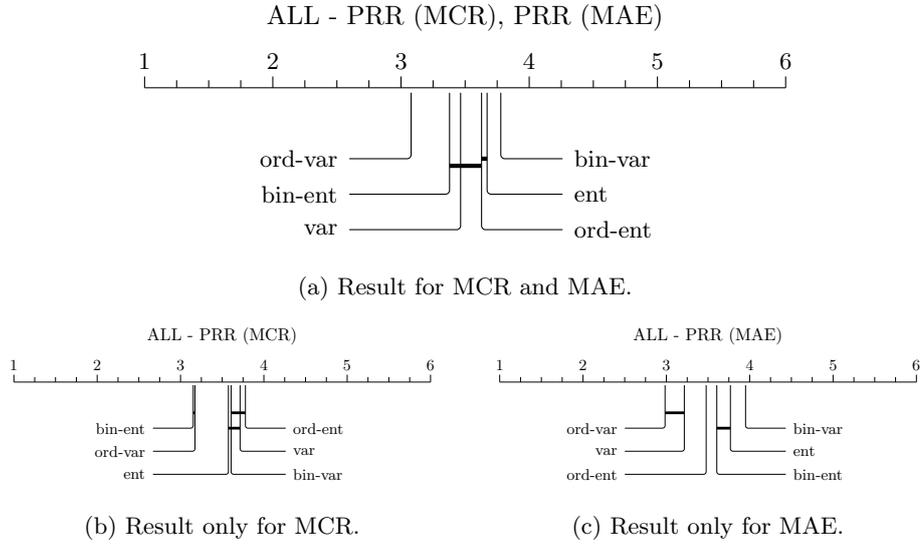
\begin{figure}[!htbp]
  \centering 
  \caption{CD diagram for OOD detection of the different uncertainty measures using an ensemble of GBTs (XGBoost \citep{DBLP:conf/kdd/ChenG16}).}
  \label{fig:cd_ood_xgb}

  \begin{tikzpicture}[
    treatment line/.style={rounded corners=1.5pt, line cap=round, shorten >=1pt},
    treatment label/.style={font=\small},
    group line/.style={ultra thick},
  ]
  
  \begin{axis}[
    clip={false},
    axis x line={center},
    axis y line={none},
    axis line style={-},
    xmin={1},
    ymax={0},
    scale only axis={true},
    width={\axisdefaultwidth},
    ticklabel style={anchor=south, yshift=1.3*\pgfkeysvalueof{/pgfplots/major tick length}, font=\small},
    every tick/.style={draw=black},
    major tick style={yshift=.5*\pgfkeysvalueof{/pgfplots/major tick length}},
    minor tick style={yshift=.5*\pgfkeysvalueof{/pgfplots/minor tick length}},
    title style={yshift=\baselineskip},
    xmax={6},
    ymin={-4.5},
    height={5\baselineskip},
    xtick={1,2,3,4,5,6},
    minor x tick num={3},
    title={EU - OOD},
  ]
  
  \draw[treatment line] ([yshift=-2pt] axis cs:2.667391304347826, 0) |- (axis cs:2.167391304347826, -2.0)
    node[treatment label, anchor=east] {ord-ent};
  \draw[treatment line] ([yshift=-2pt] axis cs:2.867391304347826, 0) |- (axis cs:2.167391304347826, -3.0)
    node[treatment label, anchor=east] {ent};
  \draw[treatment line] ([yshift=-2pt] axis cs:3.408695652173913, 0) |- (axis cs:2.167391304347826, -4.0)
    node[treatment label, anchor=east] {bin-ent};
  \draw[treatment line] ([yshift=-2pt] axis cs:3.7065217391304346, 0) |- (axis cs:4.969565217391304, -4.0)
    node[treatment label, anchor=west] {var};
  \draw[treatment line] ([yshift=-2pt] axis cs:3.880434782608696, 0) |- (axis cs:4.969565217391304, -3.0)
    node[treatment label, anchor=west] {ord-var};
  \draw[treatment line] ([yshift=-2pt] axis cs:4.469565217391304, 0) |- (axis cs:4.969565217391304, -2.0)
    node[treatment label, anchor=west] {bin-var};
  \draw[group line] (axis cs:3.408695652173913, -2.0) -- (axis cs:3.880434782608696, -2.0);
  \draw[group line] (axis cs:2.867391304347826, -2.2) -- (axis cs:3.7065217391304346, -2.2);
  
  \end{axis}
  \end{tikzpicture}

\end{figure}

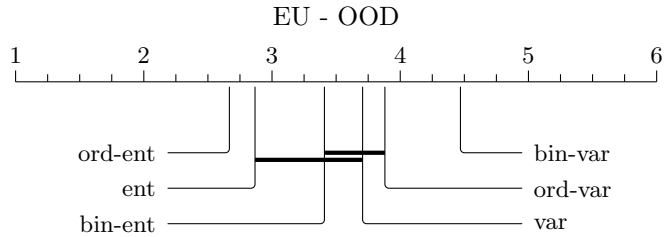
\begin{figure}[!htbp]
  \centering 
  \caption{CD diagram for OOD detection of the different uncertainty measures using an ensemble of GBTs (CatBoost \citep{DBLP:conf/nips/ProkhorenkovaGV18}).}
  \label{fig:cd_ood_cat}

  \begin{tikzpicture}[
    treatment line/.style={rounded corners=1.5pt, line cap=round, shorten >=1pt},
    treatment label/.style={font=\small},
    group line/.style={ultra thick},
  ]
  
  \begin{axis}[
    clip={false},
    axis x line={center},
    axis y line={none},
    axis line style={-},
    xmin={1},
    ymax={0},
    scale only axis={true},
    width={\axisdefaultwidth},
    ticklabel style={anchor=south, yshift=1.3*\pgfkeysvalueof{/pgfplots/major tick length}, font=\small},
    every tick/.style={draw=black},
    major tick style={yshift=.5*\pgfkeysvalueof{/pgfplots/major tick length}},
    minor tick style={yshift=.5*\pgfkeysvalueof{/pgfplots/minor tick length}},
    title style={yshift=\baselineskip},
    xmax={6},
    ymin={-4.5},
    height={5\baselineskip},
    xtick={1,2,3,4,5,6},
    minor x tick num={3},
    title={EU - OOD},
  ]
  
  \draw[treatment line] ([yshift=-2pt] axis cs:2.9217391304347826, 0) |- (axis cs:2.4217391304347826, -2.0)
    node[treatment label, anchor=east] {ord-ent};
  \draw[treatment line] ([yshift=-2pt] axis cs:3.017391304347826, 0) |- (axis cs:2.4217391304347826, -3.0)
    node[treatment label, anchor=east] {var};
  \draw[treatment line] ([yshift=-2pt] axis cs:3.126086956521739, 0) |- (axis cs:2.4217391304347826, -4.0)
    node[treatment label, anchor=east] {ent};
  \draw[treatment line] ([yshift=-2pt] axis cs:3.5369565217391306, 0) |- (axis cs:5.008695652173913, -4.0)
    node[treatment label, anchor=west] {ord-var};
  \draw[treatment line] ([yshift=-2pt] axis cs:3.889130434782609, 0) |- (axis cs:5.008695652173913, -3.0)
    node[treatment label, anchor=west] {bin-ent};
  \draw[treatment line] ([yshift=-2pt] axis cs:4.508695652173913, 0) |- (axis cs:5.008695652173913, -2.0)
    node[treatment label, anchor=west] {bin-var};
  \draw[group line] (axis cs:3.126086956521739, -2.6666666666666665) -- (axis cs:3.5369565217391306, -2.6666666666666665);
  \draw[group line] (axis cs:3.5369565217391306, -2.0) -- (axis cs:3.889130434782609, -2.0);
  \draw[group line] (axis cs:3.017391304347826, -2.0) -- (axis cs:3.126086956521739, -2.0);
  \draw[group line] (axis cs:2.9217391304347826, -1.3333333333333333) -- (axis cs:3.017391304347826, -1.3333333333333333);
  
  \end{axis}
  \end{tikzpicture}

\end{figure}

Just like for LightGBM (cf.\ Figure \ref{fig:ood_lgbm}), ord-ent performs best when it comes to OOD detection based on measured epistemic uncertainty (cf.\ Figures \ref{fig:cd_ood_xgb} and \ref{fig:cd_ood_cat}), and the OCS decomposition measures can be considered competitive to the other measures when it comes to OOD detection.


\begin{figure}[!htbp]
  \centering 
  \caption{CD diagrams for all uncertainty types (AU, EU and TU) using an ensemble of GBTs (CatBoost \citep{DBLP:conf/nips/ProkhorenkovaGV18}).}
  \label{fig:cd_all_cat}
  \begin{subfigure}[t]{\linewidth}
    \centering 
\begin{tikzpicture}[
  treatment line/.style={rounded corners=1.5pt, line cap=round, shorten >=1pt},
  treatment label/.style={font=\small},
  group line/.style={ultra thick},
]

\begin{axis}[
  clip={false},
  axis x line={center},
  axis y line={none},
  axis line style={-},
  xmin={1},
  ymax={0},
  scale only axis={true},
  width={\axisdefaultwidth},
  ticklabel style={anchor=south, yshift=1.3*\pgfkeysvalueof{/pgfplots/major tick length}, font=\small},
  every tick/.style={draw=black},
  major tick style={yshift=.5*\pgfkeysvalueof{/pgfplots/major tick length}},
  minor tick style={yshift=.5*\pgfkeysvalueof{/pgfplots/minor tick length}},
  title style={yshift=\baselineskip},
  xmax={6},
  ymin={-4.5},
  height={5\baselineskip},
  xtick={1,2,3,4,5,6},
  minor x tick num={3},
  title={ALL - PRR (MCR), PRR (MAE)},
]

\draw[treatment line] ([yshift=-2pt] axis cs:3.078623188405797, 0) |- (axis cs:2.578623188405797, -2.0)
  node[treatment label, anchor=east] {ord-var};
\draw[treatment line] ([yshift=-2pt] axis cs:3.378623188405797, 0) |- (axis cs:2.578623188405797, -3.0)
  node[treatment label, anchor=east] {bin-ent};
\draw[treatment line] ([yshift=-2pt] axis cs:3.4648550724637683, 0) |- (axis cs:2.578623188405797, -4.0)
  node[treatment label, anchor=east] {var};
\draw[treatment line] ([yshift=-2pt] axis cs:3.628260869565217, 0) |- (axis cs:4.278260869565218, -4.0)
  node[treatment label, anchor=west] {ord-ent};
\draw[treatment line] ([yshift=-2pt] axis cs:3.671376811594203, 0) |- (axis cs:4.278260869565218, -3.0)
  node[treatment label, anchor=west] {ent};
\draw[treatment line] ([yshift=-2pt] axis cs:3.7782608695652176, 0) |- (axis cs:4.278260869565218, -2.0)
  node[treatment label, anchor=west] {bin-var};
\draw[group line] (axis cs:3.628260869565217, -2.0) -- (axis cs:3.671376811594203, -2.0);
\draw[group line] (axis cs:3.378623188405797, -2.2) -- (axis cs:3.628260869565217, -2.2);

\end{axis}
\end{tikzpicture}
\captionsetup{justification=centering}
\subcaption{Result for MCR and MAE.}
\label{fig:cd_all_mcr_mae_cat}
\end{subfigure}
\begin{subfigure}[t]{0.48\linewidth}
  \centering
\begin{tikzpicture}[scale=0.65,
  treatment line/.style={rounded corners=1.5pt, line cap=round, shorten >=1pt},
  treatment label/.style={font=\small},
  group line/.style={ultra thick},
]

\begin{axis}[
  clip={false},
  axis x line={center},
  axis y line={none},
  axis line style={-},
  xmin={1},
  ymax={0},
  scale only axis={true},
  width={\axisdefaultwidth},
  ticklabel style={anchor=south, yshift=1.3*\pgfkeysvalueof{/pgfplots/major tick length}, font=\small},
  every tick/.style={draw=black},
  major tick style={yshift=.5*\pgfkeysvalueof{/pgfplots/major tick length}},
  minor tick style={yshift=.5*\pgfkeysvalueof{/pgfplots/minor tick length}},
  title style={yshift=\baselineskip},
  xmax={6},
  ymin={-4.5},
  height={5\baselineskip},
  xtick={1,2,3,4,5,6},
  minor x tick num={3},
  title={ALL - PRR (MCR)},
]

\draw[treatment line] ([yshift=-2pt] axis cs:3.152173913043478, 0) |- (axis cs:2.652173913043478, -2.0)
  node[treatment label, anchor=east] {bin-ent};
\draw[treatment line] ([yshift=-2pt] axis cs:3.1731884057971014, 0) |- (axis cs:2.652173913043478, -3.0)
  node[treatment label, anchor=east] {ord-var};
\draw[treatment line] ([yshift=-2pt] axis cs:3.5746376811594205, 0) |- (axis cs:2.652173913043478, -4.0)
  node[treatment label, anchor=east] {ent};
\draw[treatment line] ([yshift=-2pt] axis cs:3.606521739130435, 0) |- (axis cs:4.278985507246377, -4.0)
  node[treatment label, anchor=west] {bin-var};
\draw[treatment line] ([yshift=-2pt] axis cs:3.714492753623188, 0) |- (axis cs:4.278985507246377, -3.0)
  node[treatment label, anchor=west] {var};
\draw[treatment line] ([yshift=-2pt] axis cs:3.778985507246377, 0) |- (axis cs:4.278985507246377, -2.0)
  node[treatment label, anchor=west] {ord-ent};
\draw[group line] (axis cs:3.152173913043478, -1.3333333333333333) -- (axis cs:3.1731884057971014, -1.3333333333333333);
\draw[group line] (axis cs:3.5746376811594205, -2.0) -- (axis cs:3.714492753623188, -2.0);
\draw[group line] (axis cs:3.606521739130435, -1.3333333333333333) -- (axis cs:3.778985507246377, -1.3333333333333333);

\end{axis}
\end{tikzpicture}
\captionsetup{justification=centering}
\subcaption{Result only for MCR.}
\label{fig:cd_all_mcr_cat}
\end{subfigure}
\begin{subfigure}[t]{0.48\linewidth}
  \centering
\begin{tikzpicture}[scale=0.65,
  treatment line/.style={rounded corners=1.5pt, line cap=round, shorten >=1pt},
  treatment label/.style={font=\small},
  group line/.style={ultra thick},
]

\begin{axis}[
  clip={false},
  axis x line={center},
  axis y line={none},
  axis line style={-},
  xmin={1},
  ymax={0},
  scale only axis={true},
  width={\axisdefaultwidth},
  ticklabel style={anchor=south, yshift=1.3*\pgfkeysvalueof{/pgfplots/major tick length}, font=\small},
  every tick/.style={draw=black},
  major tick style={yshift=.5*\pgfkeysvalueof{/pgfplots/major tick length}},
  minor tick style={yshift=.5*\pgfkeysvalueof{/pgfplots/minor tick length}},
  title style={yshift=\baselineskip},
  xmax={6},
  ymin={-4.5},
  height={5\baselineskip},
  xtick={1,2,3,4,5,6},
  minor x tick num={3},
  title={ALL - PRR (MAE)},
]

\draw[treatment line] ([yshift=-2pt] axis cs:2.9840579710144928, 0) |- (axis cs:2.4840579710144928, -2.0)
  node[treatment label, anchor=east] {ord-var};
\draw[treatment line] ([yshift=-2pt] axis cs:3.215217391304348, 0) |- (axis cs:2.4840579710144928, -3.0)
  node[treatment label, anchor=east] {var};
\draw[treatment line] ([yshift=-2pt] axis cs:3.477536231884058, 0) |- (axis cs:2.4840579710144928, -4.0)
  node[treatment label, anchor=east] {ord-ent};
\draw[treatment line] ([yshift=-2pt] axis cs:3.6050724637681157, 0) |- (axis cs:4.45, -4.0)
  node[treatment label, anchor=west] {bin-ent};
\draw[treatment line] ([yshift=-2pt] axis cs:3.7681159420289854, 0) |- (axis cs:4.45, -3.0)
  node[treatment label, anchor=west] {ent};
\draw[treatment line] ([yshift=-2pt] axis cs:3.95, 0) |- (axis cs:4.45, -2.0)
  node[treatment label, anchor=west] {bin-var};
\draw[group line] (axis cs:2.9840579710144928, -1.3333333333333333) -- (axis cs:3.215217391304348, -1.3333333333333333);
\draw[group line] (axis cs:3.6050724637681157, -2.0) -- (axis cs:3.7681159420289854, -2.0);

\end{axis}
\end{tikzpicture}

\captionsetup{justification=centering}
\subcaption{Result only for MAE.}
\label{fig:cd_all_mae_cat}
         \end{subfigure}              
\end{figure}

\section{Comparison of Predictive Performance and Prediction Rejection Ratios (PRRs)}
\label{appendix:sec:comparison_prrs}

\begin{figure}[!htbp]
  \centering
   \begin{subfigure}[t]{0.48\linewidth}
           \centering
           \includegraphics[width=\linewidth]{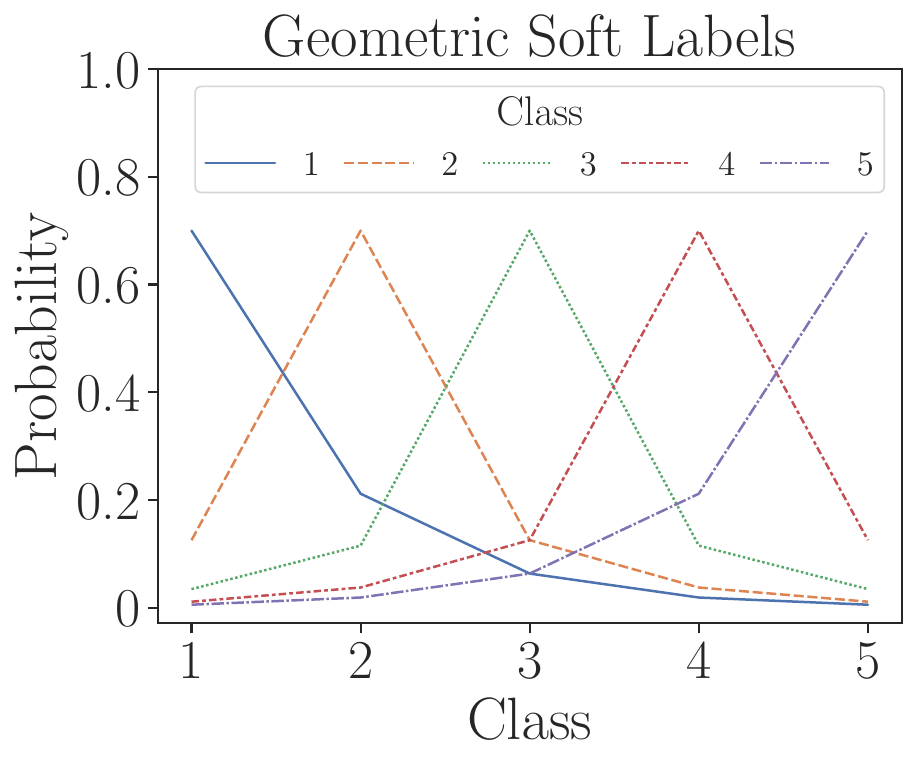}
           \subcaption{$\alpha=0.2$.}
       \end{subfigure} 
         \begin{subfigure}[t]{0.48\linewidth}
           \centering
           \includegraphics[width=\linewidth]{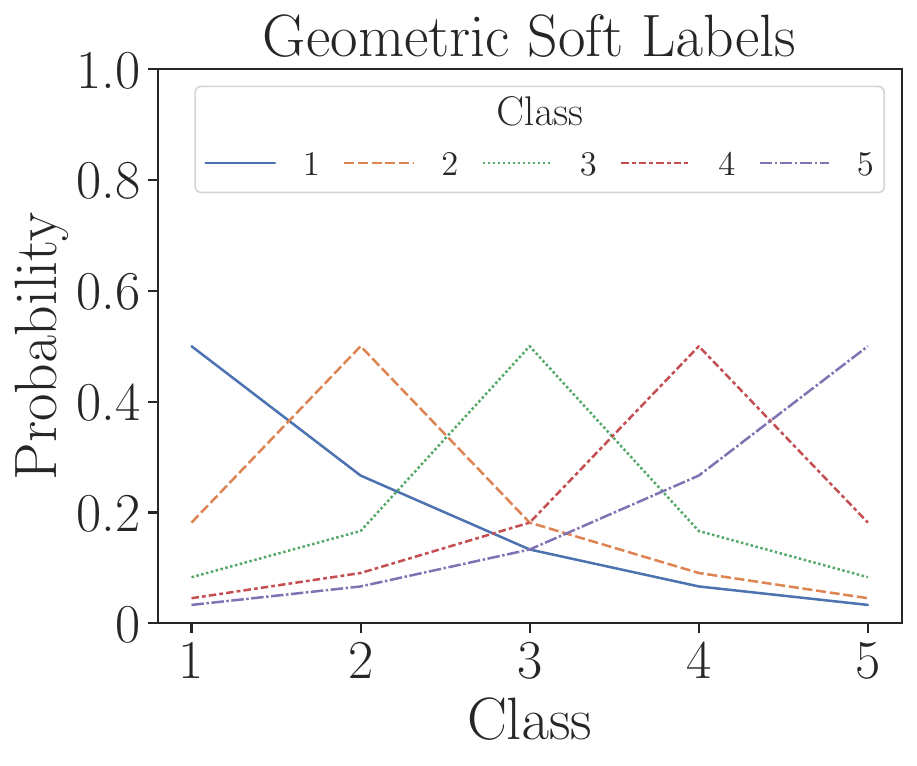}
           \subcaption{$\alpha=0.5$.}
       \end{subfigure}    
        \caption{Unimodal label smoothing for five classes based on the geometric distribution \citep{DBLP:conf/pkdd/HaasH23}.}
          \label{fig:geo}
\end{figure}

To complement the error and OOD detection experiments, Table \ref{tab:pred_perf} presents the predictive performance of the ensembles utilized, based on their base predictors: MLP with CE loss \citep{scikit-learn}, CatBoost \citep{DBLP:conf/nips/ProkhorenkovaGV18}, XGBoost \citep{DBLP:conf/kdd/ChenG16}, and LightGBM \citep{DBLP:conf/nips/KeMFWCMYL17}, averaged across all datasets. Additionally, we include two widely used ordinal losses as alternatives to the CE loss: the squared EMD loss \citep{DBLP:journals/corr/HouYS16} and the QWK loss \citep{DBLP:journals/prl/TorrePV18}. This comparison highlights the differences in performance between predictors trained with CE loss and those trained with specialized ordinal losses.
Moreover, we include a unimodal soft labeling (ULS) approach based on the geometric distribution, using LightGBM as the base learner. This method transforms deterministic one-hot (0/1) encoded labels into soft, unimodal probability distributions (cf.\ Figure \ref{fig:geo} for an illustration) \citep{DBLP:conf/pkdd/HaasH23}. The probability distribution is defined as follows:
\begin{equation*}
p^\text{GEO}(k) =
\begin{cases}
1-\alpha, & \text{if } k = c, \\
\frac{1}{G} \; \alpha^{|c-k|+1}(1-\alpha), & \text{if } k \neq c,
\end{cases}
\end{equation*}
where $\alpha$ is the smoothing factor, $k$ denotes the $k$-th class, and $c$ represents the index of the true label in the one-hot (0/1) encoded label vector $\vec{y}$, with $y_c = 1$ and $y_k = 0$ for all other classes.
$G$ serves as a normalizing constant, ensuring that $\sum_{k=1}^K p^\text{GEO}(k) = 1$. It is defined as:
\begin{equation*}
G = \sum_{k \neq c} \alpha^{|c-k|}(1-\alpha).
\end{equation*}
Unimodal soft labeling is a widely used method in ordinal classification that serves two purposes: First, it acts as a regularization technique akin to label smoothing, and second, it converts a standard predictor into an ordinal predictor by enforcing the assumption that adjacent classes are more likely than distant ones. In this approach, instead of subsampling per iteration, the 10 trees in the ensemble are smoothed using various smoothing factors, $\alpha = \{0.05, 0.1, 0.15, 0.2, 0.25, 0.3, 0.35, 0.4, 0.45, 0.5\}$.

To derive the final probabilistic prediction, we average the predicted probabilities of all ensemble members. The results are obtained using 10-fold cross-validation and are averaged across all datasets. As our primary focus is on uncertainty quantification rather than predictive performance, we refrain from performing hyperparameter tuning on the GBTs and adhere to the MLP architecture specified in Table \ref{appendix:tab:MLP_params}.

The metrics considered include accuracy (ACC), the accuracy of predictions allowing for errors in adjacent classes (1-OFF) \citep{berchez2025dlordinal}, the mean absolute error (MAE), the mean squared error (MSE), the quadratic weighted kappa (QWK), the negative log likelihood (NLL), the Brier score (BS) \citep{brier1950verification}, the ranked probability score (RPS) \citep{epstein1969scoring}, and the expected calibration error (ECE) \citep{DBLP:journals/ml/FilhoSPSKF23}.
To obtain final deterministic predictions, we make a decision that minimizes the expected loss over the predictive probability distributions. For instance, for MAE, we use the $l_1$ loss, and for MSE, we use the $l_2$ loss. For ACC and QWK, we choose the class with the maximal probability (arg max), and for 1-OFF, we select the two classes with the highest probabilities, respectively.

CatBoost is the best base predictor when it comes to ACC, 1-OFF, and MAE. 
Followed by the squared EMD loss and XGBoost. The squared EMD loss leads to the best calibration with the best overall results on RPS, NLL, BS, and ECE. It is also the best predictor on MSE and very competitive on all other metrics. 
In particular, the RPS is of interest in probabilistic ordinal classification as it is a proper scoring rule for ordinal outcomes \citep{epstein1969scoring,DBLP:conf/miccai/Galdran23}:
$$
\text{RPS} = \frac{1}{N} \sum_{i=1}^N \left( \sum_{k=1}^{K-1} \Bigl(  F_k(\vec{p}_i) - F_k(\vec{y}_i) \Bigr)^2 \right),
$$
where $\vec{p}$ is the predicted probability distribution, $\vec{y}$ is the true one-hot (0/1) encoded probability distribution, and $F_k(\vec{p})$ and  $F_k(\vec{y})$ are the respective cumulative probability distributions of class $k$. The RPS assigns lower scores to probabilistic forecasts that allocate high probabilities to classes that are close to the correct class. Since the EMD loss is equivalent to the RPS metric in measuring the squared distance of the cumulative probabilities between predicted and true outcomes, it is plausible that EMD outperforms other predictors on the RPS metric:
$$ 
  l_{\text{EMD}}(\vec{y},\vec{p}) = \sum_{k=1}^{K-1} \Bigl( F_k(\vec{p}) - F_k(\vec{y})  \Bigr)^2
$$
The same applies to the QWK loss, which leads to the best results on QWK. Remarkably, this is also true for XGBoost with CE loss, although XGBoost exhibits a higher standard deviation on QWK. From this evaluation, we can conclude that ordinal predictors indeed outperform predictors using CE loss on important ordinal metrics like QWK and RPS.
However, as noted by \cite{DBLP:conf/acl/KasaGGRPBM24}, there appears to be a trade-off between nominal and ordinal performance, as improvements in ordinal metrics are achieved at the expense of nominal metrics.

When comparing the ULS approach with the standard LightGBM ensemble trained using CE loss, we observe that applying unimodal label smoothing effectively transforms a nominal predictor into an ordinal predictor, leading to improvements across all performance metrics, including the RPS and NLL. However, for metrics such as the BS and ECE, the standard LightGBM ensemble demonstrates superior performance.

\begin{landscape}
  \vspace*{80px}
\begin{table}[!hbt]
  \caption{Predictive performance and calibration of the different ensembles by their base predictors averaged over all datasets.}
  \label{tab:pred_perf}
\begin{tabular}{l|lllllllll}
  \toprule
  Predictor & ACC ($\uparrow$) & 1-OFF ($\uparrow$) & MAE ($\downarrow$) & MSE ($\downarrow$) & QWK ($\uparrow$) & RPS ($\downarrow$) & NLL ($\downarrow$) & BS ($\downarrow$) & ECE ($\downarrow$) \\
  \midrule
  MLP  & 0.628\textpm0.199 & 0.891\textpm0.121 & 0.503\textpm0.349 & 0.782\textpm0.756 & 0.676\textpm0.262 & 0.387\textpm0.270 & 1.400\textpm1.028 & 0.541\textpm0.290 & 0.067\textpm0.065 \\
  CatBoost  & \textbf{0.639}\textpm0.193 & \textbf{0.900}\textpm0.113 & \textbf{0.470}\textpm0.328 & 0.698\textpm0.658 & 0.693\textpm0.239 & 0.345\textpm0.229 & 0.982\textpm0.543 & 0.490\textpm0.243 & 0.048\textpm0.040 \\
  XGBoost  & 0.637\textpm0.200 & 0.895\textpm0.124 & 0.477\textpm0.347 & 0.698\textpm0.667 & \textbf{0.696}\textpm0.242 & 0.353\textpm0.238 & 1.050\textpm0.579 & 0.503\textpm0.251 & 0.054\textpm0.042 \\
  LightGBM & 0.621\textpm0.223 & 0.886\textpm0.132 & 0.521\textpm0.435 & 0.812\textpm1.052 & 0.649\textpm0.299 & 0.368\textpm0.283 & 0.991\textpm0.533 & 0.488\textpm0.235 & 0.040\textpm0.026 \\
  \midrule
  ULS  & 0.625\textpm0.197 & 0.899\textpm0.115 & 0.483\textpm0.329 & 0.714\textpm0.667 & 0.688\textpm0.244 & 0.353\textpm0.214 & 0.964\textpm0.407 & 0.493\textpm0.193 & 0.045\textpm0.035 \\
  EMD & 0.635\textpm0.188 & 0.898\textpm0.116 & 0.471\textpm0.300 & \textbf{0.689}\textpm0.611 & 0.686\textpm0.245 & \textbf{0.335}\textpm0.210 & \textbf{0.892}\textpm0.449 &\textbf{0.467}\textpm0.217 & \textbf{0.017}\textpm0.010 \\
  QWK  & 0.595\textpm0.196 & 0.885\textpm0.119 & 0.537\textpm0.339 & 0.823\textpm0.725 & \textbf{0.696}\textpm0.216 & 0.409\textpm0.245 & 1.250\textpm0.634 & 0.568\textpm0.241 & 0.070\textpm0.057 \\
  \bottomrule
  \end{tabular}
\end{table}
\end{landscape}

\begin{figure}[!htbp]

  \centering    
    \begin{subfigure}[t]{0.3\linewidth}
      \centering
      \includegraphics[width=\linewidth]{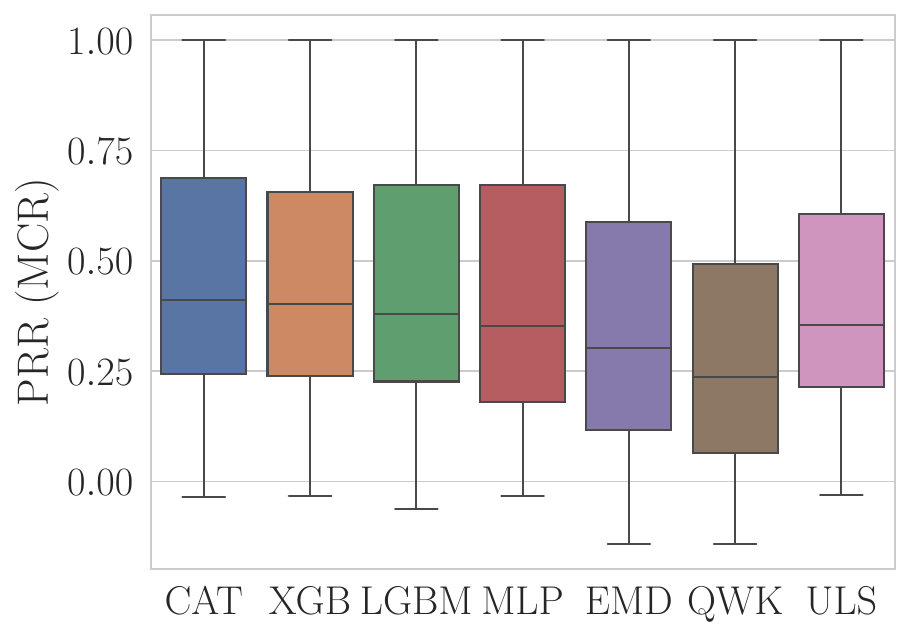}
      \caption{AU - PRR(MCR)}
  \end{subfigure} 
    \begin{subfigure}[t]{0.3\linewidth}
      \centering
      \includegraphics[width=\linewidth]{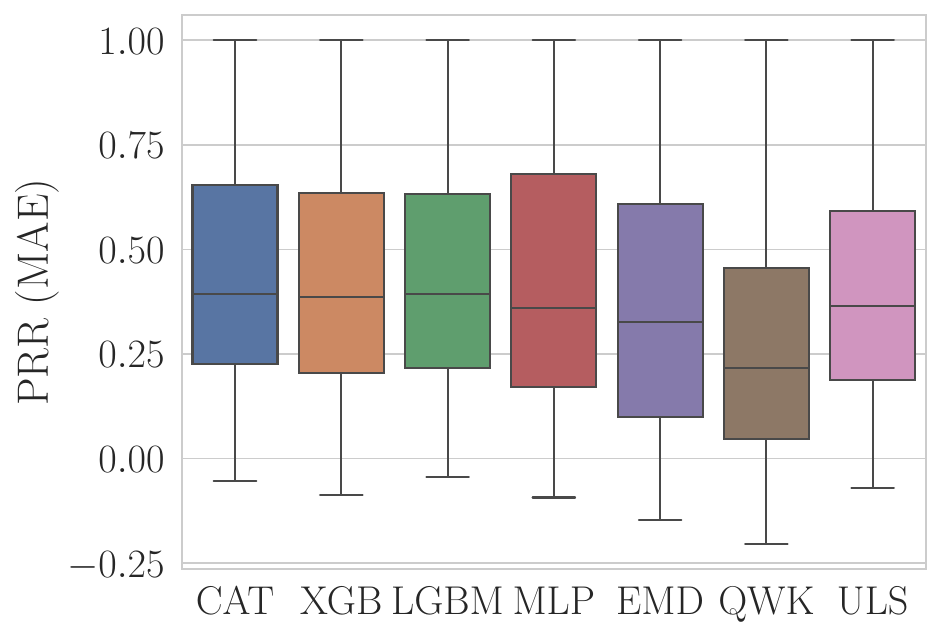}
      \caption{AU - PRR(MAE)}
  \end{subfigure} 
  \begin{subfigure}[t]{0.3\linewidth}
   \centering
   \includegraphics[width=\linewidth]{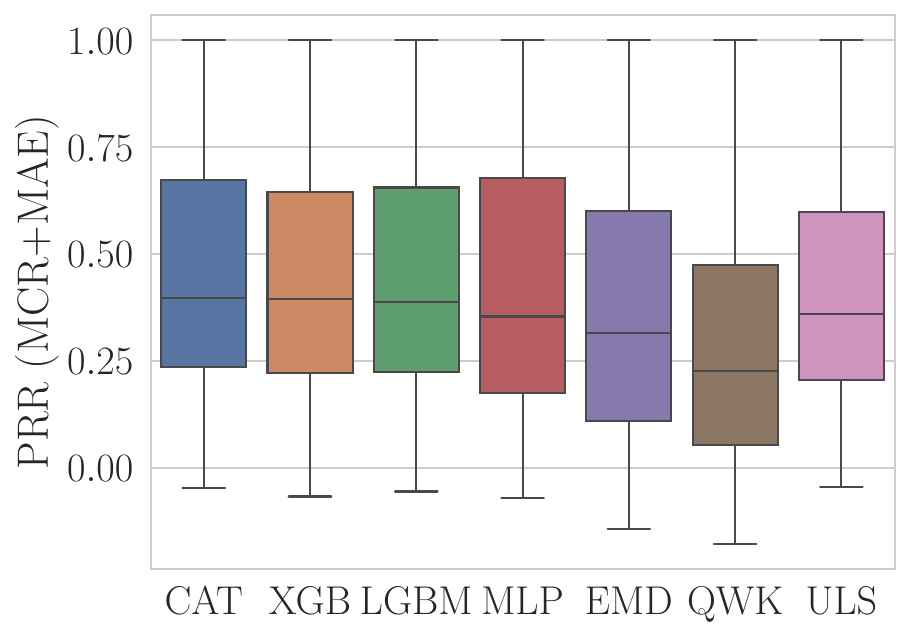}
   \caption{AU - PRR(MCR+MAE)}
\end{subfigure}   
    \begin{subfigure}[t]{0.3\linewidth}
      \centering
      \includegraphics[width=\linewidth]{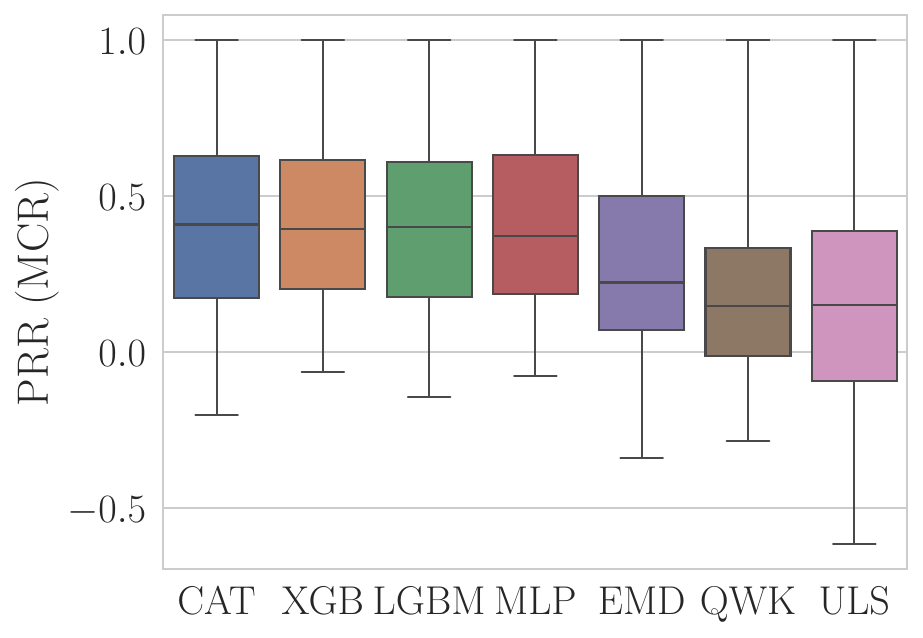}
      \caption{EU - PRR(MCR)}
  \end{subfigure} 
    \begin{subfigure}[t]{0.3\linewidth}
      \centering
      \includegraphics[width=\linewidth]{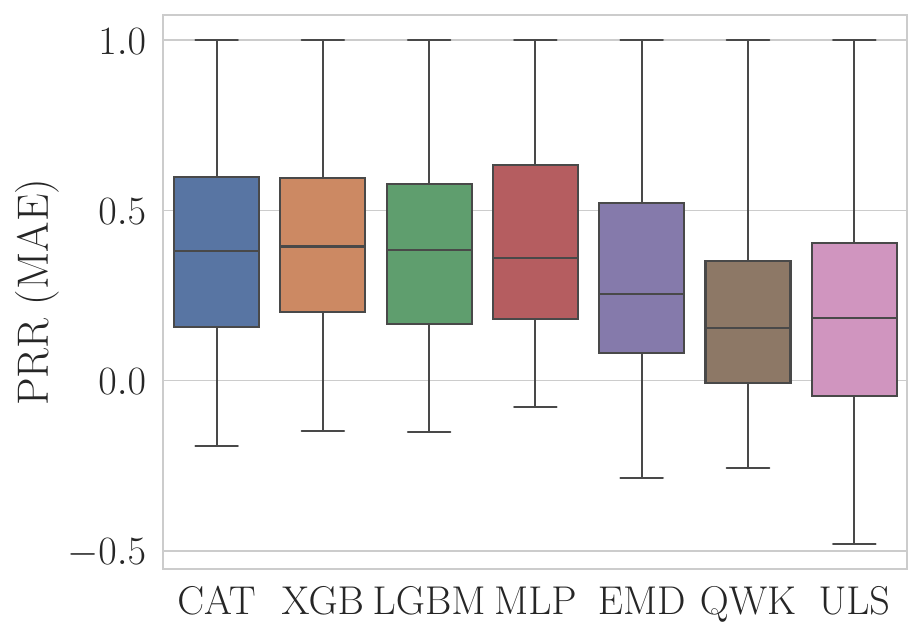}
      \caption{EU - PRR(MAE)}
  \end{subfigure} 
  \begin{subfigure}[t]{0.3\linewidth}
   \centering
   \includegraphics[width=\linewidth]{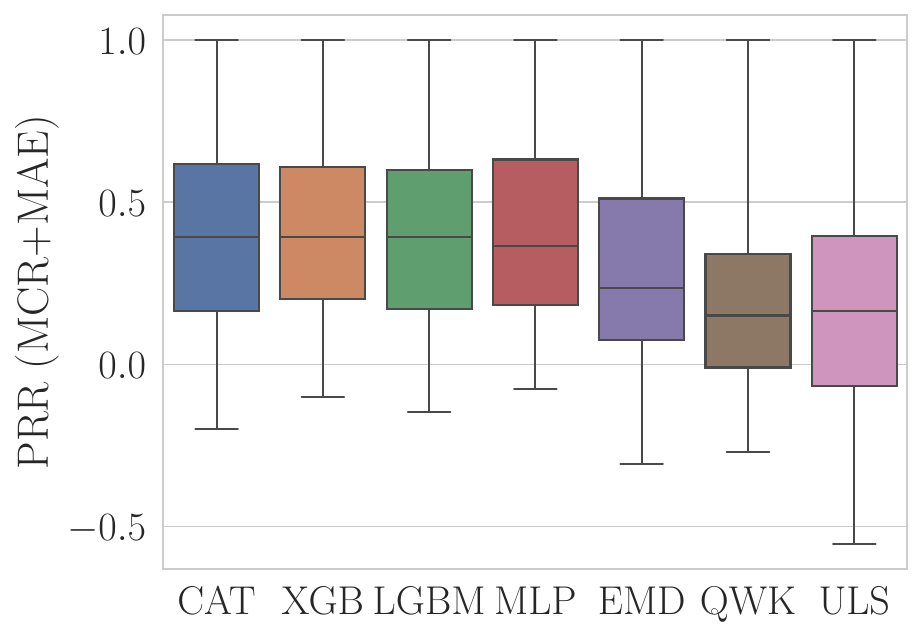}
   \caption{EU - PRR(MCR+MAE)}
\end{subfigure}  
    \begin{subfigure}[t]{0.3\linewidth}
      \centering
      \includegraphics[width=\linewidth]{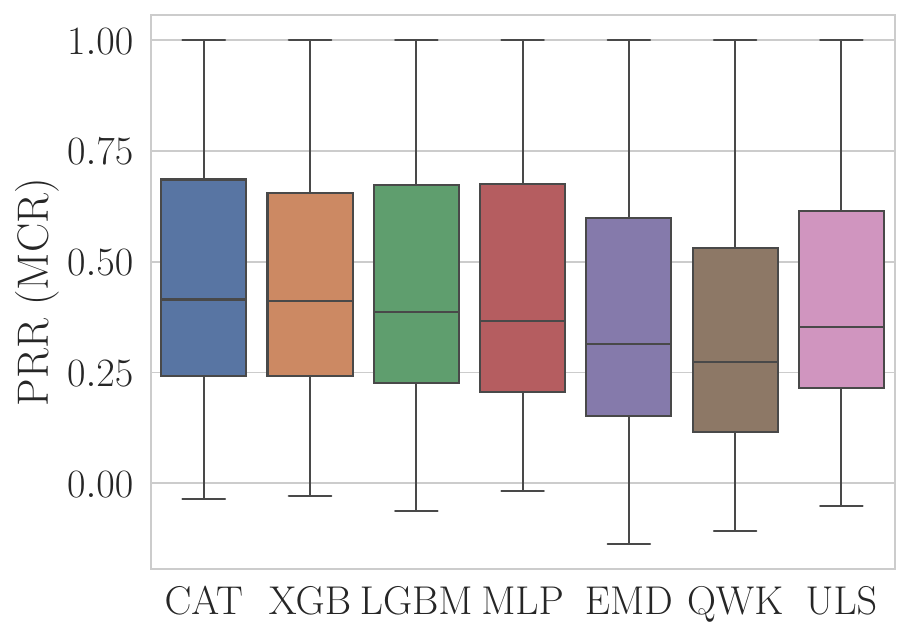}
      \caption{TU - PRR(MCR)}
  \end{subfigure} 
    \begin{subfigure}[t]{0.3\linewidth}
      \centering
      \includegraphics[width=\linewidth]{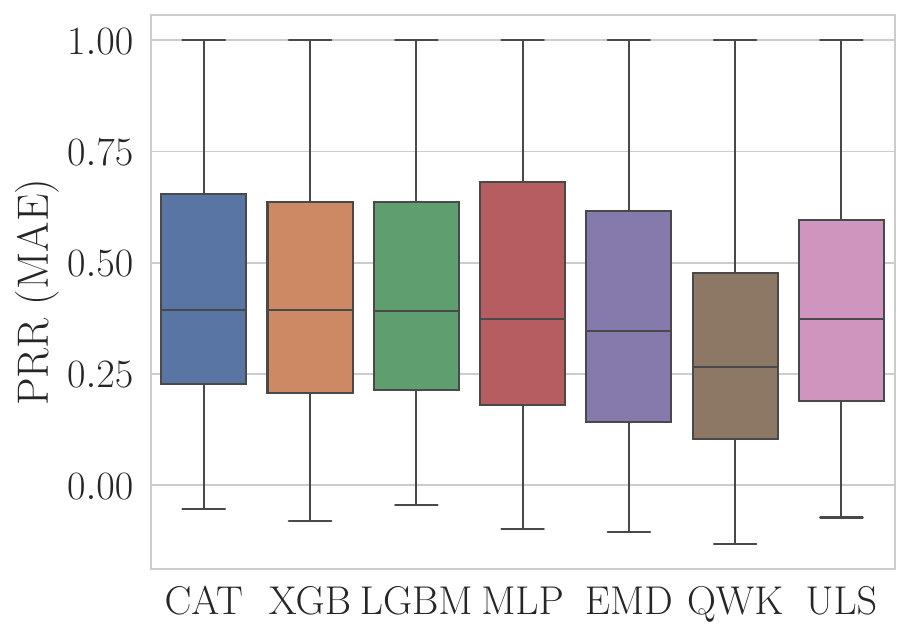}
      \caption{TU - PRR(MAE)}
  \end{subfigure} 
  \begin{subfigure}[t]{0.3\linewidth}
   \centering
   \includegraphics[width=\linewidth]{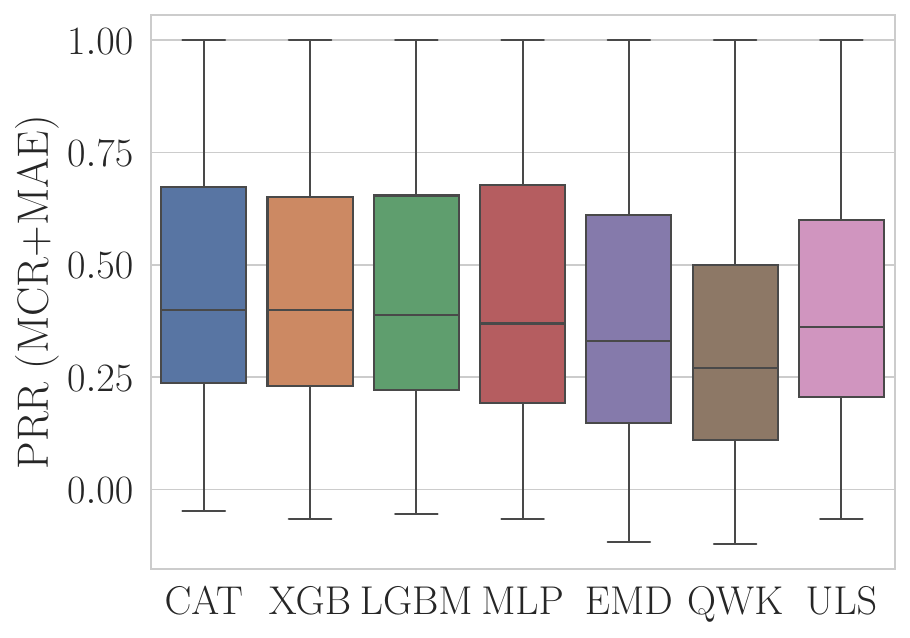}
   \caption{TU - PRR(MCR+MAE)}
  \end{subfigure}  
  \begin{subfigure}[t]{0.3\linewidth}
    \centering
    \includegraphics[width=\linewidth]{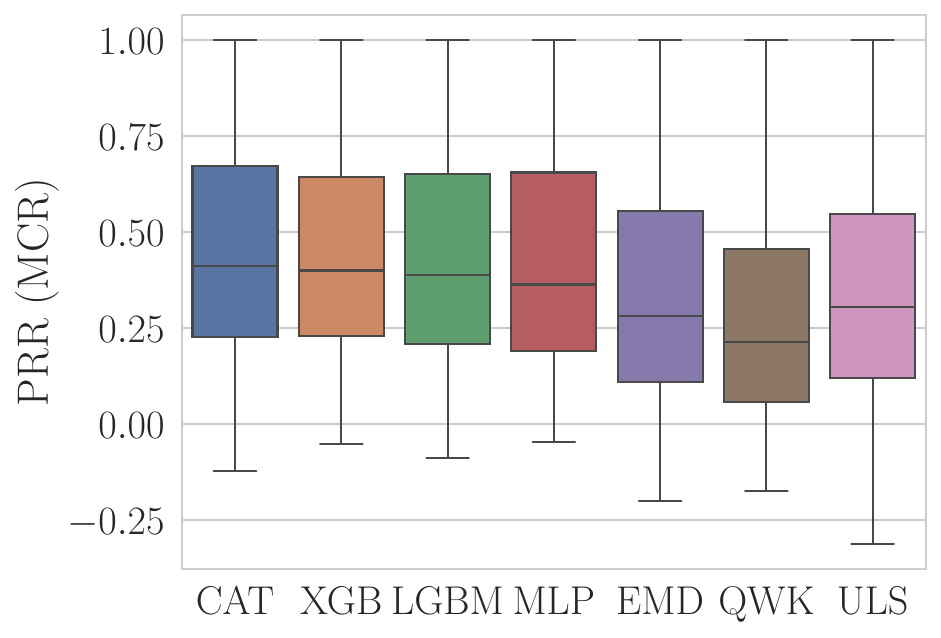}
    \caption{ALL - PRR(MCR)}
\end{subfigure} 
  \begin{subfigure}[t]{0.3\linewidth}
    \centering
    \includegraphics[width=\linewidth]{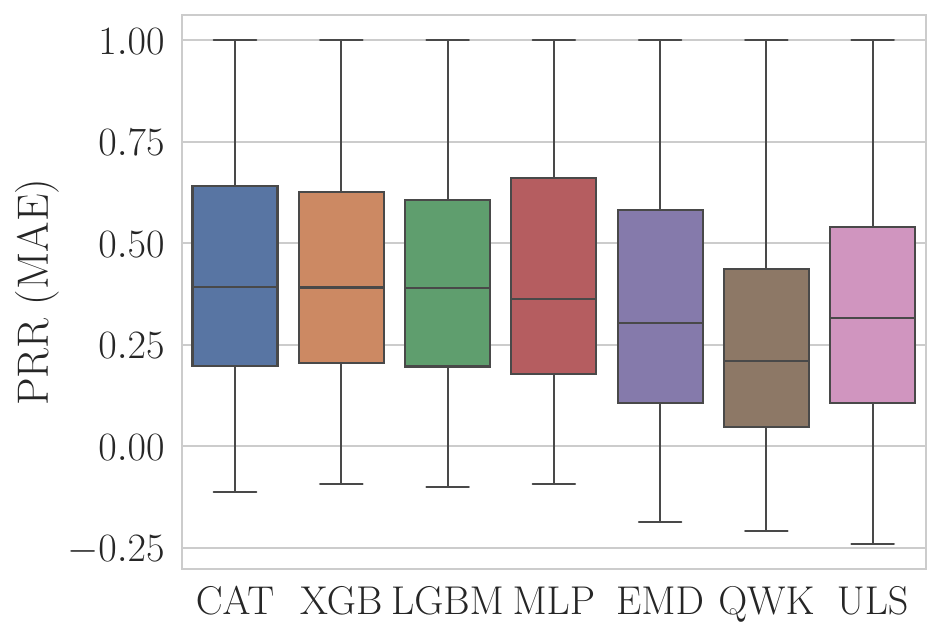}
    \caption{ALL - PRR(MAE)}
\end{subfigure} 
\begin{subfigure}[t]{0.3\linewidth}
 \centering
 \includegraphics[width=\linewidth]{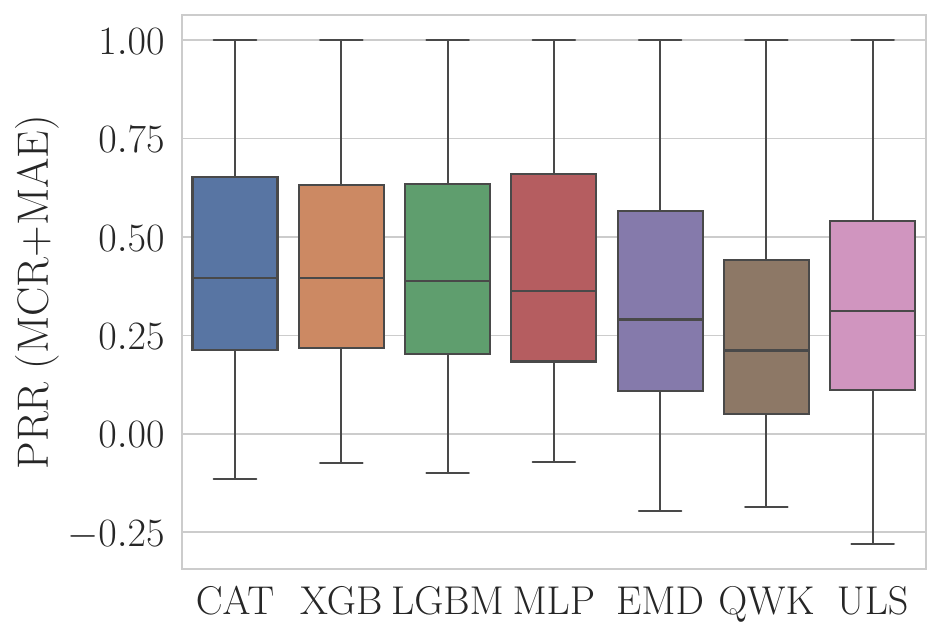}
 \caption{ALL - PRR(MCR+MAE)}
\end{subfigure}       
        \caption{Raw PRRs over all datasets and uncertainty measures by uncertainty type (AU, EU, TU, or All), performance metric (MCR, MAE), and base predictor (CatBoost, XGBoost, LightGBM, MLP, EMD, QWK, or ULS).}
          \label{fig:prr_comp}
\end{figure}

Figure \ref{fig:prr_comp} displays the attainable PRRs across all datasets and uncertainty measures, grouped by base predictor. It is important to note that PRRs are independent of predictive performance and solely measure the performance of uncertainty quantification. Specifically, they represent the area between the uncertainty measure-based rejection and random rejection, compared to the area between optimal and random rejection (cf.\ Figure \ref{fig:prr}). As one can see, regardless of the uncertainty measure, the base predictors using CE loss (CatBoost, XGBoost, LightGBM, and the MLP with CE loss) achieve higher PRRs compared to those using ordinal losses (EMD, QWK, and ULS).

\begin{figure}[!htbp]
  \centering 
  
 \begin{subfigure}[t]{0.48\linewidth}
  \centering
  \begin{tikzpicture}[scale=0.65,
    treatment line/.style={rounded corners=1.5pt, line cap=round, shorten >=1pt},
    treatment label/.style={font=\small},
    group line/.style={ultra thick},
  ]
  
  \begin{axis}[
    clip={false},
    axis x line={center},
    axis y line={none},
    axis line style={-},
    xmin={1},
    ymax={0},
    scale only axis={true},
    width={\axisdefaultwidth},
    ticklabel style={anchor=south, yshift=1.3*\pgfkeysvalueof{/pgfplots/major tick length}, font=\small},
    every tick/.style={draw=black},
    major tick style={yshift=.5*\pgfkeysvalueof{/pgfplots/major tick length}},
    minor tick style={yshift=.5*\pgfkeysvalueof{/pgfplots/minor tick length}},
    title style={yshift=\baselineskip},
    xmax={7},
    ymin={-4.5},
    height={5\baselineskip},
    xtick={1,2,3,4,5,6,7},
    minor x tick num={1},
    title={AU - PRR (MCR)},
  ]
  
  \draw[treatment line] ([yshift=-2pt] axis cs:3.590217391304348, 0) |- (axis cs:3.0068840579710145, -2.5)
    node[treatment label, anchor=east] {LGBM};
  \draw[treatment line] ([yshift=-2pt] axis cs:3.6365942028985505, 0) |- (axis cs:3.0068840579710145, -3.5)
    node[treatment label, anchor=east] {XGB};
  \draw[treatment line] ([yshift=-2pt] axis cs:3.6420289855072463, 0) |- (axis cs:3.0068840579710145, -4.5)
    node[treatment label, anchor=east] {CAT};
  \draw[treatment line] ([yshift=-2pt] axis cs:3.706159420289855, 0) |- (axis cs:5.4981884057971016, -5.0)
    node[treatment label, anchor=west] {MLP};
  \draw[treatment line] ([yshift=-2pt] axis cs:4.0793478260869565, 0) |- (axis cs:5.4981884057971016, -4.0)
    node[treatment label, anchor=west] {ULS};
  \draw[treatment line] ([yshift=-2pt] axis cs:4.430797101449276, 0) |- (axis cs:5.4981884057971016, -3.0)
    node[treatment label, anchor=west] {EMD};
  \draw[treatment line] ([yshift=-2pt] axis cs:4.9148550724637685, 0) |- (axis cs:5.4981884057971016, -2.0)
    node[treatment label, anchor=west] {QWK};
  \draw[group line] (axis cs:3.6365942028985505, -2.3333333333333335) -- (axis cs:3.706159420289855, -2.3333333333333335);
  \draw[group line] (axis cs:3.706159420289855, -2.6666666666666665) -- (axis cs:4.0793478260869565, -2.6666666666666665);
  \draw[group line] (axis cs:3.590217391304348, -1.6666666666666667) -- (axis cs:3.6420289855072463, -1.6666666666666667);
  
  \end{axis}
  \end{tikzpicture}
 \end{subfigure}
 \begin{subfigure}[t]{0.48\linewidth}
  \centering
  \begin{tikzpicture}[scale=0.65,
    treatment line/.style={rounded corners=1.5pt, line cap=round, shorten >=1pt},
    treatment label/.style={font=\small},
    group line/.style={ultra thick},
  ]
  
  \begin{axis}[
    clip={false},
    axis x line={center},
    axis y line={none},
    axis line style={-},
    xmin={1},
    ymax={0},
    scale only axis={true},
    width={\axisdefaultwidth},
    ticklabel style={anchor=south, yshift=1.3*\pgfkeysvalueof{/pgfplots/major tick length}, font=\small},
    every tick/.style={draw=black},
    major tick style={yshift=.5*\pgfkeysvalueof{/pgfplots/major tick length}},
    minor tick style={yshift=.5*\pgfkeysvalueof{/pgfplots/minor tick length}},
    title style={yshift=\baselineskip},
    xmax={7},
    ymin={-4.5},
    height={5\baselineskip},
    xtick={1,2,3,4,5,6,7},
    minor x tick num={1},
    title={AU - PRR (MAE)},
  ]
  
  \draw[treatment line] ([yshift=-2pt] axis cs:3.708695652173913, 0) |- (axis cs:3.1253623188405797, -2.5)
    node[treatment label, anchor=east] {LGBM};
  \draw[treatment line] ([yshift=-2pt] axis cs:3.722826086956522, 0) |- (axis cs:3.1253623188405797, -3.5)
    node[treatment label, anchor=east] {XGB};
  \draw[treatment line] ([yshift=-2pt] axis cs:3.7268115942028985, 0) |- (axis cs:3.1253623188405797, -4.5)
    node[treatment label, anchor=east] {CAT};
  \draw[treatment line] ([yshift=-2pt] axis cs:3.81231884057971, 0) |- (axis cs:5.446014492753623, -5.0)
    node[treatment label, anchor=west] {MLP};
  \draw[treatment line] ([yshift=-2pt] axis cs:3.9956521739130433, 0) |- (axis cs:5.446014492753623, -4.0)
    node[treatment label, anchor=west] {ULS};
  \draw[treatment line] ([yshift=-2pt] axis cs:4.171014492753623, 0) |- (axis cs:5.446014492753623, -3.0)
    node[treatment label, anchor=west] {EMD};
  \draw[treatment line] ([yshift=-2pt] axis cs:4.86268115942029, 0) |- (axis cs:5.446014492753623, -2.0)
    node[treatment label, anchor=west] {QWK};
  \draw[group line] (axis cs:3.708695652173913, -1.6666666666666667) -- (axis cs:3.9956521739130433, -1.6666666666666667);
  
  \end{axis}
  \end{tikzpicture}
 \end{subfigure}
 \begin{subfigure}[t]{0.48\linewidth}
  \centering
  \begin{tikzpicture}[scale=0.65,
    treatment line/.style={rounded corners=1.5pt, line cap=round, shorten >=1pt},
    treatment label/.style={font=\small},
    group line/.style={ultra thick},
  ]
  
  \begin{axis}[
    clip={false},
    axis x line={center},
    axis y line={none},
    axis line style={-},
    xmin={1},
    ymax={0},
    scale only axis={true},
    width={\axisdefaultwidth},
    ticklabel style={anchor=south, yshift=1.3*\pgfkeysvalueof{/pgfplots/major tick length}, font=\small},
    every tick/.style={draw=black},
    major tick style={yshift=.5*\pgfkeysvalueof{/pgfplots/major tick length}},
    minor tick style={yshift=.5*\pgfkeysvalueof{/pgfplots/minor tick length}},
    title style={yshift=\baselineskip},
    xmax={7},
    ymin={-4.5},
    height={5\baselineskip},
    xtick={1,2,3,4,5,6,7},
    minor x tick num={1},
    title={AU - PRR (MCR+MAE)},
  ]
  
  \draw[treatment line] ([yshift=-2pt] axis cs:3.6494565217391304, 0) |- (axis cs:3.066123188405797, -2.5)
    node[treatment label, anchor=east] {LGBM};
  \draw[treatment line] ([yshift=-2pt] axis cs:3.6797101449275362, 0) |- (axis cs:3.066123188405797, -3.5)
    node[treatment label, anchor=east] {XGB};
  \draw[treatment line] ([yshift=-2pt] axis cs:3.6844202898550726, 0) |- (axis cs:3.066123188405797, -4.5)
    node[treatment label, anchor=east] {CAT};
  \draw[treatment line] ([yshift=-2pt] axis cs:3.7592391304347825, 0) |- (axis cs:5.472101449275362, -5.0)
    node[treatment label, anchor=west] {MLP};
  \draw[treatment line] ([yshift=-2pt] axis cs:4.0375, 0) |- (axis cs:5.472101449275362, -4.0)
    node[treatment label, anchor=west] {ULS};
  \draw[treatment line] ([yshift=-2pt] axis cs:4.300905797101449, 0) |- (axis cs:5.472101449275362, -3.0)
    node[treatment label, anchor=west] {EMD};
  \draw[treatment line] ([yshift=-2pt] axis cs:4.888768115942029, 0) |- (axis cs:5.472101449275362, -2.0)
    node[treatment label, anchor=west] {QWK};
  \draw[group line] (axis cs:3.6797101449275362, -2.3333333333333335) -- (axis cs:3.7592391304347825, -2.3333333333333335);
  \draw[group line] (axis cs:3.7592391304347825, -2.6666666666666665) -- (axis cs:4.0375, -2.6666666666666665);
  \draw[group line] (axis cs:3.6494565217391304, -1.6666666666666667) -- (axis cs:3.6844202898550726, -1.6666666666666667);
  
  \end{axis}
  \end{tikzpicture}
 \end{subfigure}
 \begin{subfigure}[t]{0.48\linewidth}
  \centering
  \begin{tikzpicture}[scale=0.65,
    treatment line/.style={rounded corners=1.5pt, line cap=round, shorten >=1pt},
    treatment label/.style={font=\small},
    group line/.style={ultra thick},
  ]
  
  \begin{axis}[
    clip={false},
    axis x line={center},
    axis y line={none},
    axis line style={-},
    xmin={1},
    ymax={0},
    scale only axis={true},
    width={\axisdefaultwidth},
    ticklabel style={anchor=south, yshift=1.3*\pgfkeysvalueof{/pgfplots/major tick length}, font=\small},
    every tick/.style={draw=black},
    major tick style={yshift=.5*\pgfkeysvalueof{/pgfplots/major tick length}},
    minor tick style={yshift=.5*\pgfkeysvalueof{/pgfplots/minor tick length}},
    title style={yshift=\baselineskip},
    xmax={7},
    ymin={-4.5},
    height={5\baselineskip},
    xtick={1,2,3,4,5,6,7},
    minor x tick num={1},
    title={EU - PRR (MCR)},
  ]
  
  \draw[treatment line] ([yshift=-2pt] axis cs:3.2235507246376813, 0) |- (axis cs:2.640217391304348, -2.5)
    node[treatment label, anchor=east] {XGB};
  \draw[treatment line] ([yshift=-2pt] axis cs:3.3764492753623188, 0) |- (axis cs:2.640217391304348, -3.5)
    node[treatment label, anchor=east] {MLP};
  \draw[treatment line] ([yshift=-2pt] axis cs:3.527536231884058, 0) |- (axis cs:2.640217391304348, -4.5)
    node[treatment label, anchor=east] {CAT};
  \draw[treatment line] ([yshift=-2pt] axis cs:3.5608695652173914, 0) |- (axis cs:5.569202898550724, -5.0)
    node[treatment label, anchor=west] {LGBM};
  \draw[treatment line] ([yshift=-2pt] axis cs:4.413043478260869, 0) |- (axis cs:5.569202898550724, -4.0)
    node[treatment label, anchor=west] {EMD};
  \draw[treatment line] ([yshift=-2pt] axis cs:4.9126811594202895, 0) |- (axis cs:5.569202898550724, -3.0)
    node[treatment label, anchor=west] {QWK};
  \draw[treatment line] ([yshift=-2pt] axis cs:4.985869565217391, 0) |- (axis cs:5.569202898550724, -2.0)
    node[treatment label, anchor=west] {ULS};
  \draw[group line] (axis cs:3.2235507246376813, -1.6666666666666667) -- (axis cs:3.5608695652173914, -1.6666666666666667);
  \draw[group line] (axis cs:4.9126811594202895, -1.3333333333333333) -- (axis cs:4.985869565217391, -1.3333333333333333);
  
  \end{axis}
  \end{tikzpicture}
 \end{subfigure}
 \begin{subfigure}[t]{0.48\linewidth}
  \centering
  \begin{tikzpicture}[scale=0.65,
    treatment line/.style={rounded corners=1.5pt, line cap=round, shorten >=1pt},
    treatment label/.style={font=\small},
    group line/.style={ultra thick},
  ]
  
  \begin{axis}[
    clip={false},
    axis x line={center},
    axis y line={none},
    axis line style={-},
    xmin={1},
    ymax={0},
    scale only axis={true},
    width={\axisdefaultwidth},
    ticklabel style={anchor=south, yshift=1.3*\pgfkeysvalueof{/pgfplots/major tick length}, font=\small},
    every tick/.style={draw=black},
    major tick style={yshift=.5*\pgfkeysvalueof{/pgfplots/major tick length}},
    minor tick style={yshift=.5*\pgfkeysvalueof{/pgfplots/minor tick length}},
    title style={yshift=\baselineskip},
    xmax={7},
    ymin={-4.5},
    height={5\baselineskip},
    xtick={1,2,3,4,5,6,7},
    minor x tick num={1},
    title={EU - PRR (MAE)},
  ]
  
  \draw[treatment line] ([yshift=-2pt] axis cs:3.325, 0) |- (axis cs:2.7416666666666667, -2.5)
    node[treatment label, anchor=east] {XGB};
  \draw[treatment line] ([yshift=-2pt] axis cs:3.546376811594203, 0) |- (axis cs:2.7416666666666667, -3.5)
    node[treatment label, anchor=east] {MLP};
  \draw[treatment line] ([yshift=-2pt] axis cs:3.666304347826087, 0) |- (axis cs:2.7416666666666667, -4.5)
    node[treatment label, anchor=east] {LGBM};
  \draw[treatment line] ([yshift=-2pt] axis cs:3.7126811594202898, 0) |- (axis cs:5.453260869565217, -5.0)
    node[treatment label, anchor=west] {CAT};
  \draw[treatment line] ([yshift=-2pt] axis cs:4.1268115942028984, 0) |- (axis cs:5.453260869565217, -4.0)
    node[treatment label, anchor=west] {EMD};
  \draw[treatment line] ([yshift=-2pt] axis cs:4.7528985507246375, 0) |- (axis cs:5.453260869565217, -3.0)
    node[treatment label, anchor=west] {ULS};
  \draw[treatment line] ([yshift=-2pt] axis cs:4.869927536231884, 0) |- (axis cs:5.453260869565217, -2.0)
    node[treatment label, anchor=west] {QWK};
  \draw[group line] (axis cs:3.325, -1.6666666666666667) -- (axis cs:3.666304347826087, -1.6666666666666667);
  \draw[group line] (axis cs:3.546376811594203, -2.3333333333333335) -- (axis cs:3.7126811594202898, -2.3333333333333335);
  \draw[group line] (axis cs:4.7528985507246375, -1.3333333333333333) -- (axis cs:4.869927536231884, -1.3333333333333333);
  
  \end{axis}
  \end{tikzpicture}
 \end{subfigure}
 \begin{subfigure}[t]{0.48\linewidth}
  \centering
  \begin{tikzpicture}[scale=0.65,
    treatment line/.style={rounded corners=1.5pt, line cap=round, shorten >=1pt},
    treatment label/.style={font=\small},
    group line/.style={ultra thick},
  ]
  
  \begin{axis}[
    clip={false},
    axis x line={center},
    axis y line={none},
    axis line style={-},
    xmin={1},
    ymax={0},
    scale only axis={true},
    width={\axisdefaultwidth},
    ticklabel style={anchor=south, yshift=1.3*\pgfkeysvalueof{/pgfplots/major tick length}, font=\small},
    every tick/.style={draw=black},
    major tick style={yshift=.5*\pgfkeysvalueof{/pgfplots/major tick length}},
    minor tick style={yshift=.5*\pgfkeysvalueof{/pgfplots/minor tick length}},
    title style={yshift=\baselineskip},
    xmax={7},
    ymin={-4.5},
    height={5\baselineskip},
    xtick={1,2,3,4,5,6,7},
    minor x tick num={1},
    title={EU - PRR (MCR+MAE)},
  ]
  
  \draw[treatment line] ([yshift=-2pt] axis cs:3.2742753623188405, 0) |- (axis cs:2.690942028985507, -2.5)
    node[treatment label, anchor=east] {XGB};
  \draw[treatment line] ([yshift=-2pt] axis cs:3.461413043478261, 0) |- (axis cs:2.690942028985507, -3.5)
    node[treatment label, anchor=east] {MLP};
  \draw[treatment line] ([yshift=-2pt] axis cs:3.6135869565217393, 0) |- (axis cs:2.690942028985507, -4.5)
    node[treatment label, anchor=east] {LGBM};
  \draw[treatment line] ([yshift=-2pt] axis cs:3.6201086956521737, 0) |- (axis cs:5.47463768115942, -5.0)
    node[treatment label, anchor=west] {CAT};
  \draw[treatment line] ([yshift=-2pt] axis cs:4.269927536231884, 0) |- (axis cs:5.47463768115942, -4.0)
    node[treatment label, anchor=west] {EMD};
  \draw[treatment line] ([yshift=-2pt] axis cs:4.869384057971015, 0) |- (axis cs:5.47463768115942, -3.0)
    node[treatment label, anchor=west] {ULS};
  \draw[treatment line] ([yshift=-2pt] axis cs:4.891304347826087, 0) |- (axis cs:5.47463768115942, -2.0)
    node[treatment label, anchor=west] {QWK};
  \draw[group line] (axis cs:3.2742753623188405, -1.6666666666666667) -- (axis cs:3.6135869565217393, -1.6666666666666667);
  \draw[group line] (axis cs:3.461413043478261, -2.3333333333333335) -- (axis cs:3.6201086956521737, -2.3333333333333335);
  \draw[group line] (axis cs:4.869384057971015, -1.3333333333333333) -- (axis cs:4.891304347826087, -1.3333333333333333);
  
  \end{axis}
  \end{tikzpicture}
 \end{subfigure}
 \begin{subfigure}[t]{0.48\linewidth}
  \centering
  \begin{tikzpicture}[scale=0.65,
    treatment line/.style={rounded corners=1.5pt, line cap=round, shorten >=1pt},
    treatment label/.style={font=\small},
    group line/.style={ultra thick},
  ]
  
  \begin{axis}[
    clip={false},
    axis x line={center},
    axis y line={none},
    axis line style={-},
    xmin={1},
    ymax={0},
    scale only axis={true},
    width={\axisdefaultwidth},
    ticklabel style={anchor=south, yshift=1.3*\pgfkeysvalueof{/pgfplots/major tick length}, font=\small},
    every tick/.style={draw=black},
    major tick style={yshift=.5*\pgfkeysvalueof{/pgfplots/major tick length}},
    minor tick style={yshift=.5*\pgfkeysvalueof{/pgfplots/minor tick length}},
    title style={yshift=\baselineskip},
    xmax={7},
    ymin={-4.5},
    height={5\baselineskip},
    xtick={1,2,3,4,5,6,7},
    minor x tick num={1},
    title={TU - PRR (MCR)},
  ]
  
  \draw[treatment line] ([yshift=-2pt] axis cs:3.6478260869565218, 0) |- (axis cs:3.0644927536231883, -2.5)
    node[treatment label, anchor=east] {LGBM};
  \draw[treatment line] ([yshift=-2pt] axis cs:3.678985507246377, 0) |- (axis cs:3.0644927536231883, -3.5)
    node[treatment label, anchor=east] {XGB};
  \draw[treatment line] ([yshift=-2pt] axis cs:3.699275362318841, 0) |- (axis cs:3.0644927536231883, -4.5)
    node[treatment label, anchor=east] {CAT};
  \draw[treatment line] ([yshift=-2pt] axis cs:3.7115942028985507, 0) |- (axis cs:5.306521739130434, -5.0)
    node[treatment label, anchor=west] {MLP};
  \draw[treatment line] ([yshift=-2pt] axis cs:4.209420289855072, 0) |- (axis cs:5.306521739130434, -4.0)
    node[treatment label, anchor=west] {ULS};
  \draw[treatment line] ([yshift=-2pt] axis cs:4.329710144927536, 0) |- (axis cs:5.306521739130434, -3.0)
    node[treatment label, anchor=west] {EMD};
  \draw[treatment line] ([yshift=-2pt] axis cs:4.723188405797101, 0) |- (axis cs:5.306521739130434, -2.0)
    node[treatment label, anchor=west] {QWK};
  \draw[group line] (axis cs:3.6478260869565218, -1.6666666666666667) -- (axis cs:3.7115942028985507, -1.6666666666666667);
  
  \end{axis}
  \end{tikzpicture}
 \end{subfigure}
 \begin{subfigure}[t]{0.48\linewidth}
  \centering
  \begin{tikzpicture}[scale=0.65,
    treatment line/.style={rounded corners=1.5pt, line cap=round, shorten >=1pt},
    treatment label/.style={font=\small},
    group line/.style={ultra thick},
  ]
  
  \begin{axis}[
    clip={false},
    axis x line={center},
    axis y line={none},
    axis line style={-},
    xmin={1},
    ymax={0},
    scale only axis={true},
    width={\axisdefaultwidth},
    ticklabel style={anchor=south, yshift=1.3*\pgfkeysvalueof{/pgfplots/major tick length}, font=\small},
    every tick/.style={draw=black},
    major tick style={yshift=.5*\pgfkeysvalueof{/pgfplots/major tick length}},
    minor tick style={yshift=.5*\pgfkeysvalueof{/pgfplots/minor tick length}},
    title style={yshift=\baselineskip},
    xmax={7},
    ymin={-4.5},
    height={5\baselineskip},
    xtick={1,2,3,4,5,6,7},
    minor x tick num={1},
    title={TU - PRR (MAE)},
  ]
  
  \draw[treatment line] ([yshift=-2pt] axis cs:3.764130434782609, 0) |- (axis cs:3.1807971014492753, -2.5)
    node[treatment label, anchor=east] {LGBM};
  \draw[treatment line] ([yshift=-2pt] axis cs:3.794927536231884, 0) |- (axis cs:3.1807971014492753, -3.5)
    node[treatment label, anchor=east] {XGB};
  \draw[treatment line] ([yshift=-2pt] axis cs:3.7967391304347826, 0) |- (axis cs:3.1807971014492753, -4.5)
    node[treatment label, anchor=east] {CAT};
  \draw[treatment line] ([yshift=-2pt] axis cs:3.833695652173913, 0) |- (axis cs:5.337681159420289, -5.0)
    node[treatment label, anchor=west] {MLP};
  \draw[treatment line] ([yshift=-2pt] axis cs:3.973913043478261, 0) |- (axis cs:5.337681159420289, -4.0)
    node[treatment label, anchor=west] {EMD};
  \draw[treatment line] ([yshift=-2pt] axis cs:4.082246376811594, 0) |- (axis cs:5.337681159420289, -3.0)
    node[treatment label, anchor=west] {ULS};
  \draw[treatment line] ([yshift=-2pt] axis cs:4.754347826086956, 0) |- (axis cs:5.337681159420289, -2.0)
    node[treatment label, anchor=west] {QWK};
  \draw[group line] (axis cs:3.764130434782609, -1.6666666666666667) -- (axis cs:4.082246376811594, -1.6666666666666667);
  
  \end{axis}
  \end{tikzpicture}
 \end{subfigure}
 \begin{subfigure}[t]{0.48\linewidth}
  \centering
  \begin{tikzpicture}[scale=0.65,
    treatment line/.style={rounded corners=1.5pt, line cap=round, shorten >=1pt},
    treatment label/.style={font=\small},
    group line/.style={ultra thick},
  ]
  
  \begin{axis}[
    clip={false},
    axis x line={center},
    axis y line={none},
    axis line style={-},
    xmin={1},
    ymax={0},
    scale only axis={true},
    width={\axisdefaultwidth},
    ticklabel style={anchor=south, yshift=1.3*\pgfkeysvalueof{/pgfplots/major tick length}, font=\small},
    every tick/.style={draw=black},
    major tick style={yshift=.5*\pgfkeysvalueof{/pgfplots/major tick length}},
    minor tick style={yshift=.5*\pgfkeysvalueof{/pgfplots/minor tick length}},
    title style={yshift=\baselineskip},
    xmax={7},
    ymin={-4.5},
    height={5\baselineskip},
    xtick={1,2,3,4,5,6,7},
    minor x tick num={1},
    title={TU - PRR (MCR+MAE)},
  ]
  
  \draw[treatment line] ([yshift=-2pt] axis cs:3.705978260869565, 0) |- (axis cs:3.1226449275362316, -2.5)
    node[treatment label, anchor=east] {LGBM};
  \draw[treatment line] ([yshift=-2pt] axis cs:3.7369565217391303, 0) |- (axis cs:3.1226449275362316, -3.5)
    node[treatment label, anchor=east] {XGB};
  \draw[treatment line] ([yshift=-2pt] axis cs:3.7480072463768117, 0) |- (axis cs:3.1226449275362316, -4.5)
    node[treatment label, anchor=east] {CAT};
  \draw[treatment line] ([yshift=-2pt] axis cs:3.772644927536232, 0) |- (axis cs:5.322101449275362, -5.0)
    node[treatment label, anchor=west] {MLP};
  \draw[treatment line] ([yshift=-2pt] axis cs:4.145833333333333, 0) |- (axis cs:5.322101449275362, -4.0)
    node[treatment label, anchor=west] {ULS};
  \draw[treatment line] ([yshift=-2pt] axis cs:4.151811594202899, 0) |- (axis cs:5.322101449275362, -3.0)
    node[treatment label, anchor=west] {EMD};
  \draw[treatment line] ([yshift=-2pt] axis cs:4.738768115942029, 0) |- (axis cs:5.322101449275362, -2.0)
    node[treatment label, anchor=west] {QWK};
  \draw[group line] (axis cs:3.705978260869565, -1.6666666666666667) -- (axis cs:3.772644927536232, -1.6666666666666667);
  
  \end{axis}
  \end{tikzpicture}
 \end{subfigure}
 \begin{subfigure}[t]{0.48\linewidth}
  \centering
  \begin{tikzpicture}[scale=0.65,
    treatment line/.style={rounded corners=1.5pt, line cap=round, shorten >=1pt},
    treatment label/.style={font=\small},
    group line/.style={ultra thick},
  ]
  
  \begin{axis}[
    clip={false},
    axis x line={center},
    axis y line={none},
    axis line style={-},
    xmin={1},
    ymax={0},
    scale only axis={true},
    width={\axisdefaultwidth},
    ticklabel style={anchor=south, yshift=1.3*\pgfkeysvalueof{/pgfplots/major tick length}, font=\small},
    every tick/.style={draw=black},
    major tick style={yshift=.5*\pgfkeysvalueof{/pgfplots/major tick length}},
    minor tick style={yshift=.5*\pgfkeysvalueof{/pgfplots/minor tick length}},
    title style={yshift=\baselineskip},
    xmax={7},
    ymin={-4.5},
    height={5\baselineskip},
    xtick={1,2,3,4,5,6,7},
    minor x tick num={1},
    title={ALL - PRR (MCR)},
  ]
  
  \draw[treatment line] ([yshift=-2pt] axis cs:3.5130434782608697, 0) |- (axis cs:2.9297101449275362, -2.5)
    node[treatment label, anchor=east] {XGB};
  \draw[treatment line] ([yshift=-2pt] axis cs:3.5980676328502414, 0) |- (axis cs:2.9297101449275362, -3.5)
    node[treatment label, anchor=east] {MLP};
  \draw[treatment line] ([yshift=-2pt] axis cs:3.5996376811594204, 0) |- (axis cs:2.9297101449275362, -4.5)
    node[treatment label, anchor=east] {LGBM};
  \draw[treatment line] ([yshift=-2pt] axis cs:3.6229468599033816, 0) |- (axis cs:5.433574879227053, -5.0)
    node[treatment label, anchor=west] {CAT};
  \draw[treatment line] ([yshift=-2pt] axis cs:4.391183574879227, 0) |- (axis cs:5.433574879227053, -4.0)
    node[treatment label, anchor=west] {EMD};
  \draw[treatment line] ([yshift=-2pt] axis cs:4.4248792270531405, 0) |- (axis cs:5.433574879227053, -3.0)
    node[treatment label, anchor=west] {ULS};
  \draw[treatment line] ([yshift=-2pt] axis cs:4.85024154589372, 0) |- (axis cs:5.433574879227053, -2.0)
    node[treatment label, anchor=west] {QWK};
  \draw[group line] (axis cs:4.391183574879227, -2.0) -- (axis cs:4.4248792270531405, -2.0);
  \draw[group line] (axis cs:3.5130434782608697, -1.6666666666666667) -- (axis cs:3.5996376811594204, -1.6666666666666667);
  \draw[group line] (axis cs:3.5980676328502414, -2.3333333333333335) -- (axis cs:3.6229468599033816, -2.3333333333333335);
  
  \end{axis}
  \end{tikzpicture}
\end{subfigure}
\begin{subfigure}[t]{0.48\linewidth}
  \centering
  \begin{tikzpicture}[scale=0.65,
    treatment line/.style={rounded corners=1.5pt, line cap=round, shorten >=1pt},
    treatment label/.style={font=\small},
    group line/.style={ultra thick},
  ]
  
  \begin{axis}[
    clip={false},
    axis x line={center},
    axis y line={none},
    axis line style={-},
    xmin={1},
    ymax={0},
    scale only axis={true},
    width={\axisdefaultwidth},
    ticklabel style={anchor=south, yshift=1.3*\pgfkeysvalueof{/pgfplots/major tick length}, font=\small},
    every tick/.style={draw=black},
    major tick style={yshift=.5*\pgfkeysvalueof{/pgfplots/major tick length}},
    minor tick style={yshift=.5*\pgfkeysvalueof{/pgfplots/minor tick length}},
    title style={yshift=\baselineskip},
    xmax={7},
    ymin={-4.5},
    height={5\baselineskip},
    xtick={1,2,3,4,5,6,7},
    minor x tick num={1},
    title={ALL - PRR (MAE)},
  ]
  
  \draw[treatment line] ([yshift=-2pt] axis cs:3.6142512077294686, 0) |- (axis cs:3.030917874396135, -2.5)
    node[treatment label, anchor=east] {XGB};
  \draw[treatment line] ([yshift=-2pt] axis cs:3.7130434782608694, 0) |- (axis cs:3.030917874396135, -3.5)
    node[treatment label, anchor=east] {LGBM};
  \draw[treatment line] ([yshift=-2pt] axis cs:3.7307971014492756, 0) |- (axis cs:3.030917874396135, -4.5)
    node[treatment label, anchor=east] {MLP};
  \draw[treatment line] ([yshift=-2pt] axis cs:3.745410628019324, 0) |- (axis cs:5.41231884057971, -5.0)
    node[treatment label, anchor=west] {CAT};
  \draw[treatment line] ([yshift=-2pt] axis cs:4.090579710144928, 0) |- (axis cs:5.41231884057971, -4.0)
    node[treatment label, anchor=west] {EMD};
  \draw[treatment line] ([yshift=-2pt] axis cs:4.276932367149758, 0) |- (axis cs:5.41231884057971, -3.0)
    node[treatment label, anchor=west] {ULS};
  \draw[treatment line] ([yshift=-2pt] axis cs:4.828985507246377, 0) |- (axis cs:5.41231884057971, -2.0)
    node[treatment label, anchor=west] {QWK};
  \draw[group line] (axis cs:4.090579710144928, -2.0) -- (axis cs:4.276932367149758, -2.0);
  \draw[group line] (axis cs:3.6142512077294686, -1.6666666666666667) -- (axis cs:3.7307971014492756, -1.6666666666666667);
  \draw[group line] (axis cs:3.7130434782608694, -2.3333333333333335) -- (axis cs:3.745410628019324, -2.3333333333333335);
  
  \end{axis}
  \end{tikzpicture}
\end{subfigure}
\begin{subfigure}[t]{0.48\linewidth}
  \centering
  \begin{tikzpicture}[scale=0.65,
    treatment line/.style={rounded corners=1.5pt, line cap=round, shorten >=1pt},
    treatment label/.style={font=\small},
    group line/.style={ultra thick},
  ]
  
  \begin{axis}[
    clip={false},
    axis x line={center},
    axis y line={none},
    axis line style={-},
    xmin={1},
    ymax={0},
    scale only axis={true},
    width={\axisdefaultwidth},
    ticklabel style={anchor=south, yshift=1.3*\pgfkeysvalueof{/pgfplots/major tick length}, font=\small},
    every tick/.style={draw=black},
    major tick style={yshift=.5*\pgfkeysvalueof{/pgfplots/major tick length}},
    minor tick style={yshift=.5*\pgfkeysvalueof{/pgfplots/minor tick length}},
    title style={yshift=\baselineskip},
    xmax={7},
    ymin={-4.5},
    height={5\baselineskip},
    xtick={1,2,3,4,5,6,7},
    minor x tick num={1},
    title={ALL - PRR (MCR+MAE)},
  ]
  
  \draw[treatment line] ([yshift=-2pt] axis cs:3.563647342995169, 0) |- (axis cs:2.9803140096618357, -2.5)
    node[treatment label, anchor=east] {XGB};
  \draw[treatment line] ([yshift=-2pt] axis cs:3.656340579710145, 0) |- (axis cs:2.9803140096618357, -3.5)
    node[treatment label, anchor=east] {LGBM};
  \draw[treatment line] ([yshift=-2pt] axis cs:3.6644323671497583, 0) |- (axis cs:2.9803140096618357, -4.5)
    node[treatment label, anchor=east] {MLP};
  \draw[treatment line] ([yshift=-2pt] axis cs:3.6841787439613527, 0) |- (axis cs:5.422946859903381, -5.0)
    node[treatment label, anchor=west] {CAT};
  \draw[treatment line] ([yshift=-2pt] axis cs:4.240881642512077, 0) |- (axis cs:5.422946859903381, -4.0)
    node[treatment label, anchor=west] {EMD};
  \draw[treatment line] ([yshift=-2pt] axis cs:4.350905797101449, 0) |- (axis cs:5.422946859903381, -3.0)
    node[treatment label, anchor=west] {ULS};
  \draw[treatment line] ([yshift=-2pt] axis cs:4.839613526570048, 0) |- (axis cs:5.422946859903381, -2.0)
    node[treatment label, anchor=west] {QWK};
  \draw[group line] (axis cs:4.240881642512077, -2.0) -- (axis cs:4.350905797101449, -2.0);
  \draw[group line] (axis cs:3.563647342995169, -1.6666666666666667) -- (axis cs:3.6644323671497583, -1.6666666666666667);
  \draw[group line] (axis cs:3.656340579710145, -2.3333333333333335) -- (axis cs:3.6841787439613527, -2.3333333333333335);
  
  \end{axis}
  \end{tikzpicture}
\end{subfigure}
\caption{CD diagrams displaying the ranks of the different base predictors in terms of attainable PRRs per performance measure (MCR, MAE) and uncertainty type (AU, EU, TU, All). Base predictors which are not significantly different at $p=0.05$ are connected.}
\label{fig:prr_comp_cds}
\end{figure}
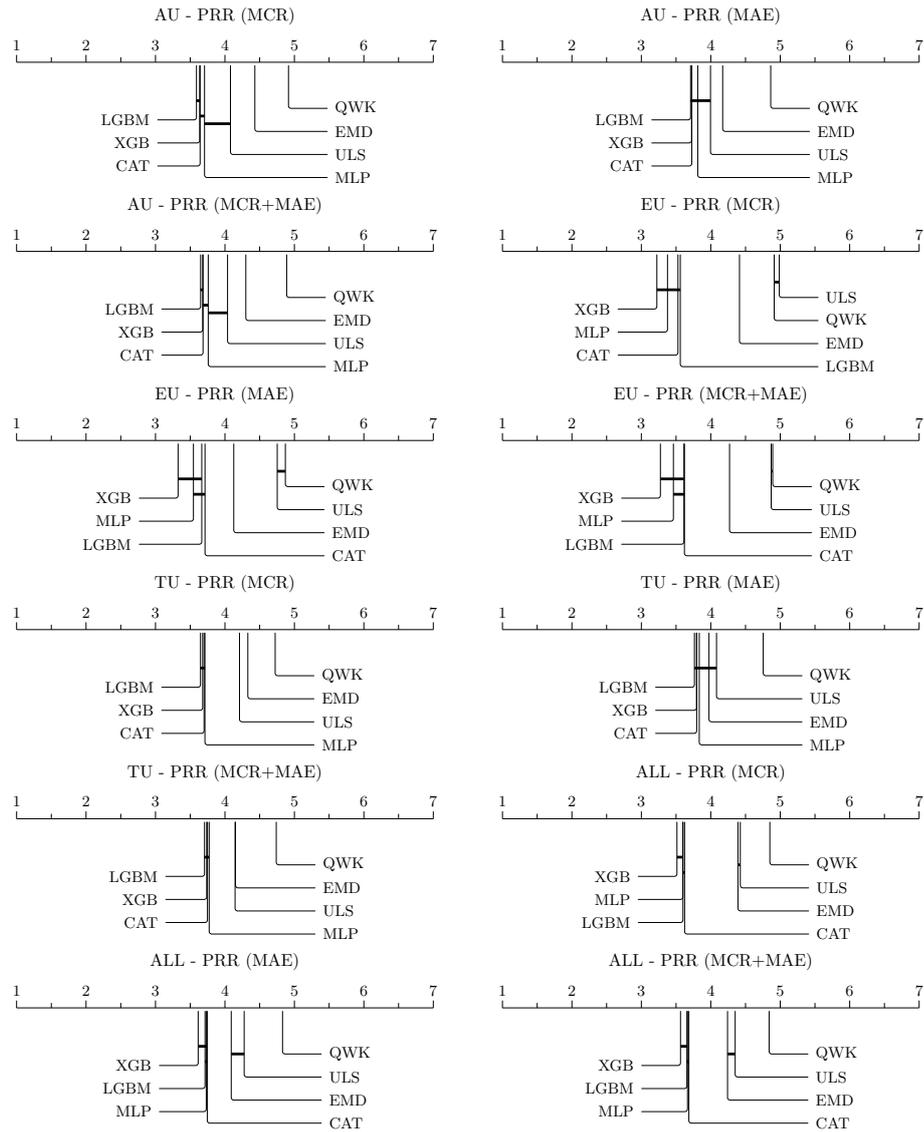

When examining the CD diagrams in Figure \ref{fig:prr_comp_cds}, which are based on Wilcoxon signed-rank tests, we observe that these differences are statistically significant in most cases at a significance level of $p = 0.05$ for various uncertainty types and combinations of the considered metrics, namely MCR and MAE.
Notably, QWK consistently performs the worst in terms of PRR. This is presumably due to the fact that it penalizes deviations from the true class quadratically, compared to EMD, which does so linearly, as noted by \cite{DBLP:conf/miccai/Galdran23}. Additionally, EMD does not impose a strong penalty in the tails, as CDFs are monotonic; hence, the difference between CDFs will be small in the tails \citep{DBLP:conf/acl/KasaGGRPBM24}. This is particularly evident in the PRRs for MAE, where ULS and EMD may not completely reassign distant class probability mass compared to QWK, thereby allowing for better uncertainty quantification.

\begin{figure}[!htbp]
  \centering
   \begin{subfigure}[t]{0.19\linewidth}
        \centering
          \includegraphics[width=\linewidth]{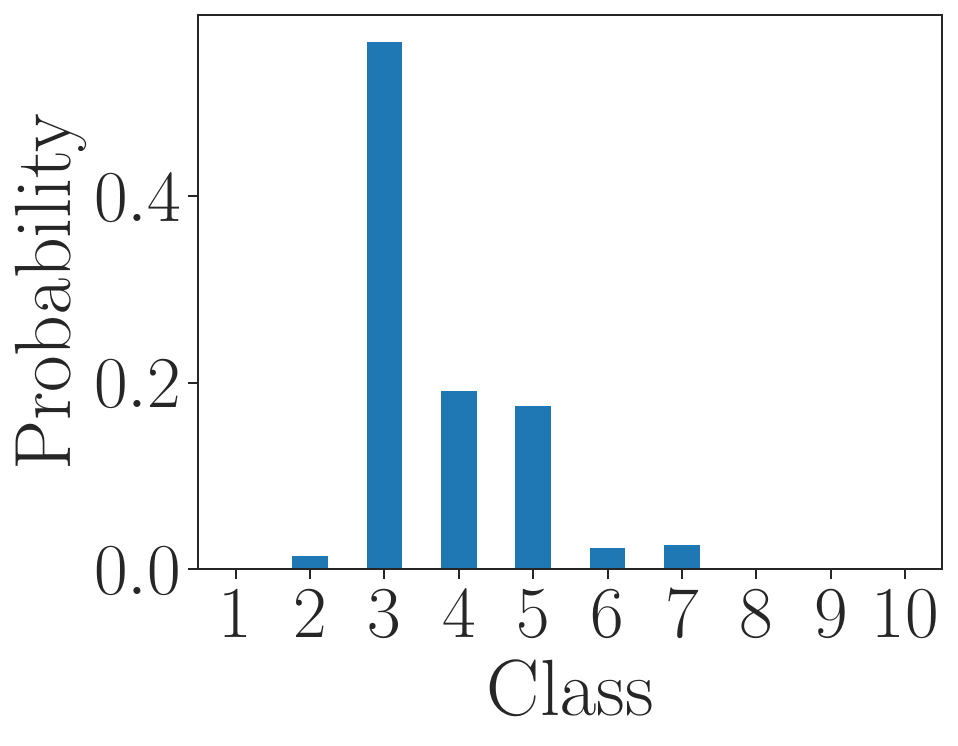}           
       \end{subfigure} 
         \begin{subfigure}[t]{0.19\linewidth}
           \centering
           \includegraphics[width=\linewidth]{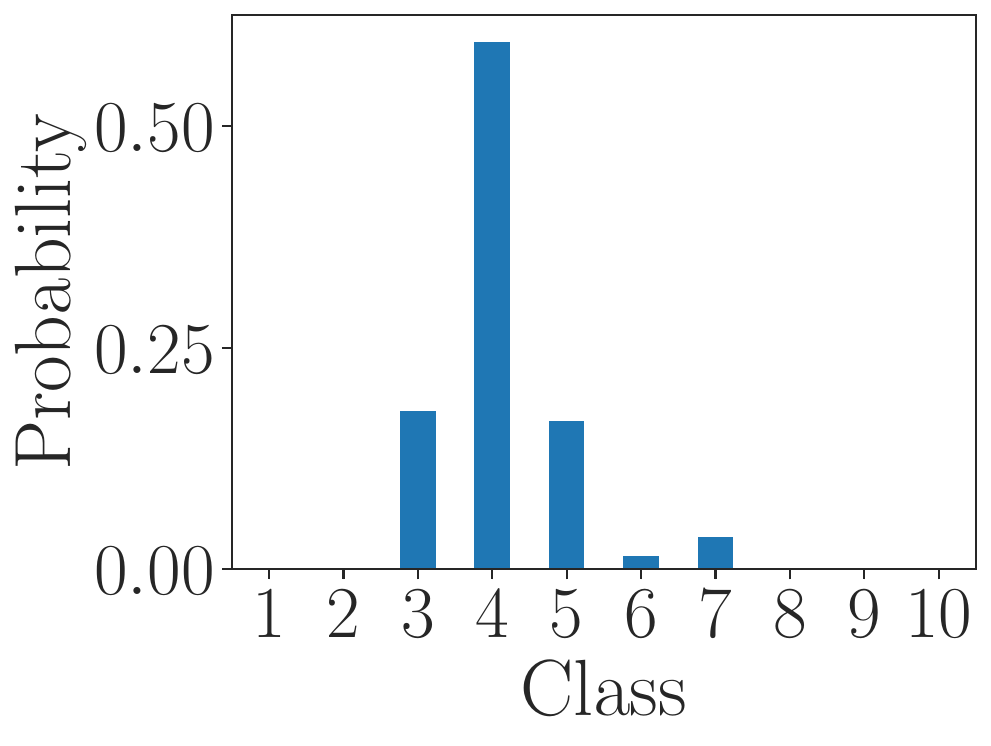}
       \end{subfigure}
         \begin{subfigure}[t]{0.19\linewidth}
           \centering
           \includegraphics[width=\linewidth]{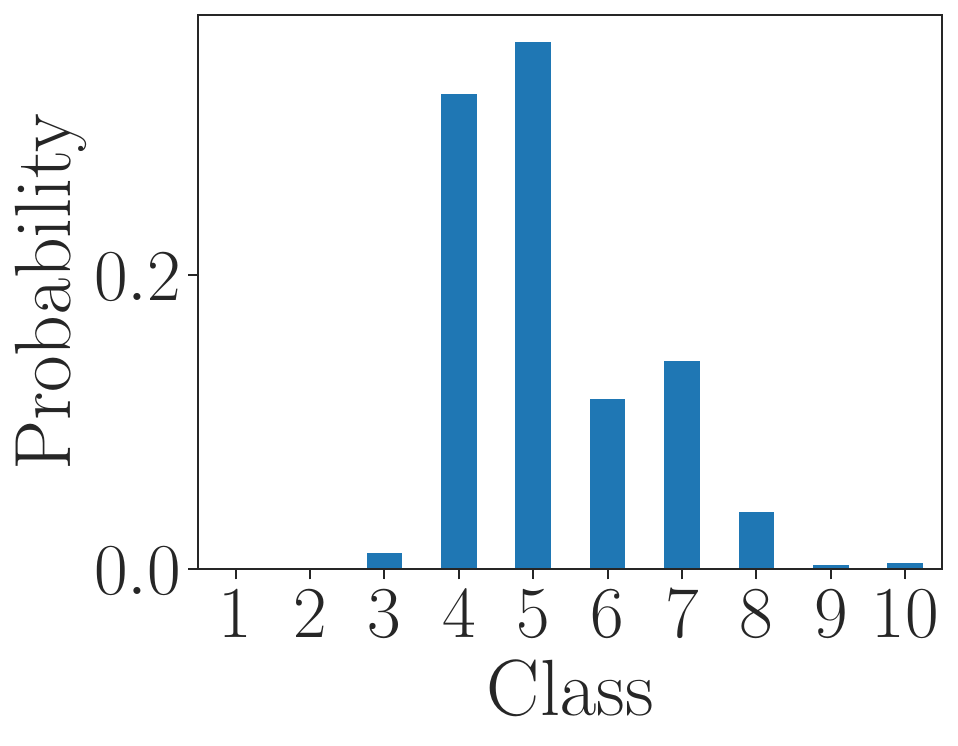}
  \end{subfigure}  
   \begin{subfigure}[t]{0.19\linewidth}
           \centering
           \includegraphics[width=\linewidth]{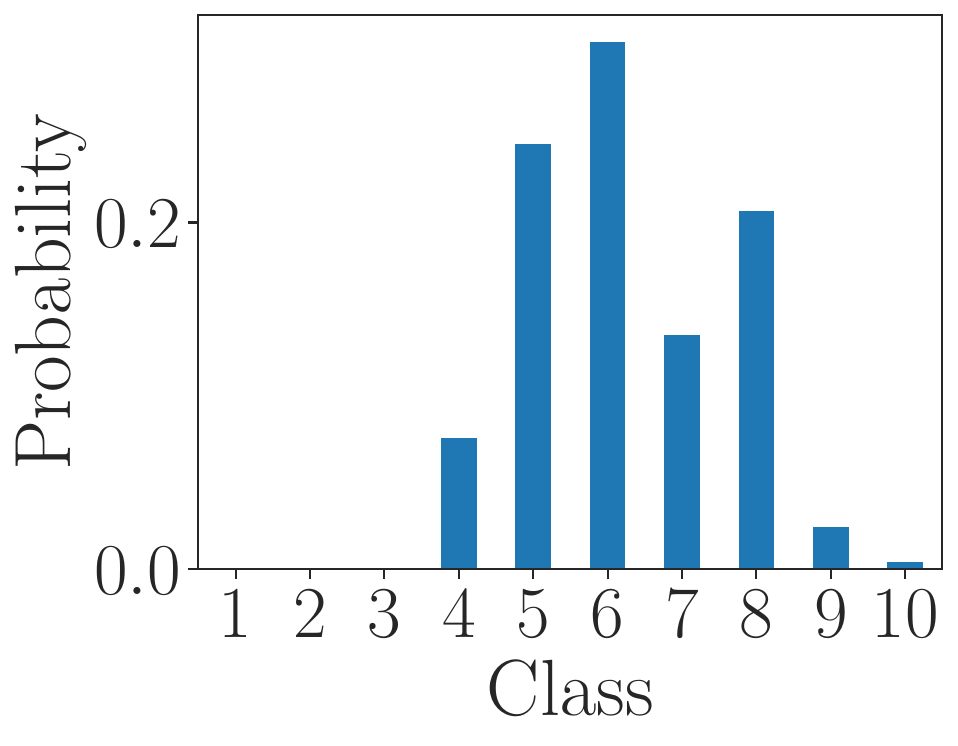}
       \end{subfigure}     
              \begin{subfigure}[t]{0.19\linewidth}
        \centering
          \includegraphics[width=\linewidth]{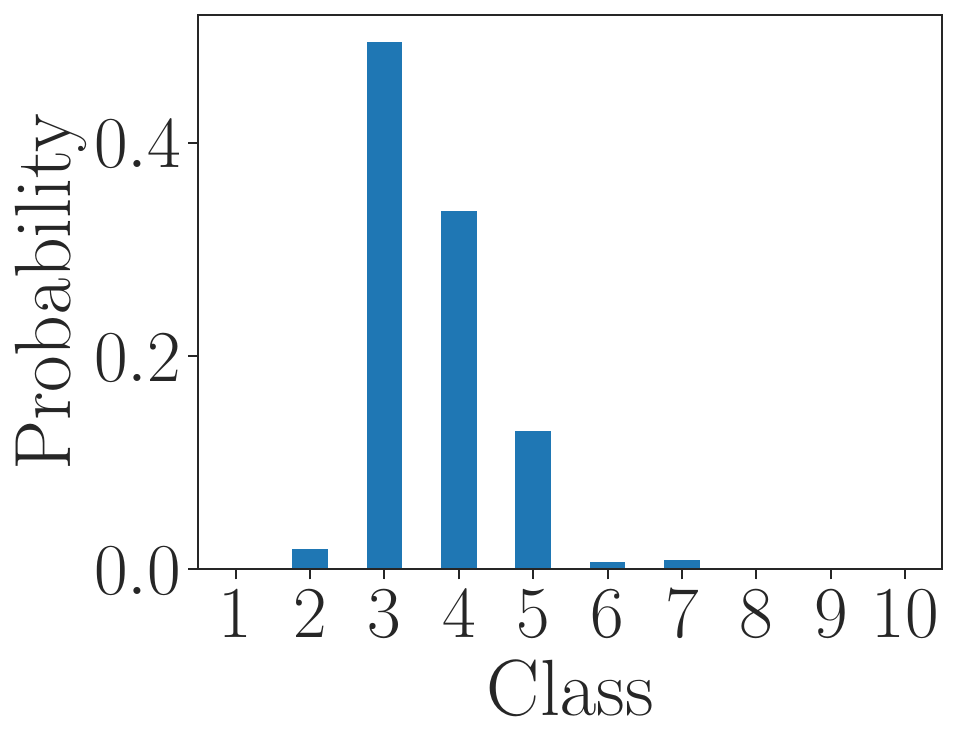}           
       \end{subfigure} 
       \begin{subfigure}[t]{0.19\linewidth}
        \centering
          \includegraphics[width=\linewidth]{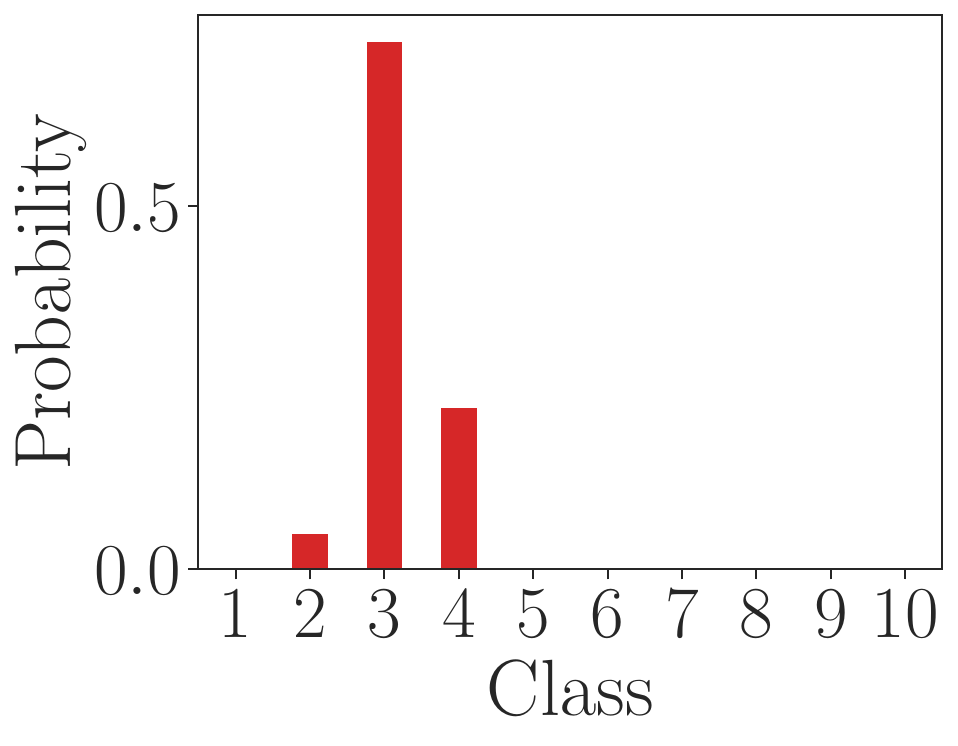}           
       \end{subfigure} 
         \begin{subfigure}[t]{0.19\linewidth}
           \centering
           \includegraphics[width=\linewidth]{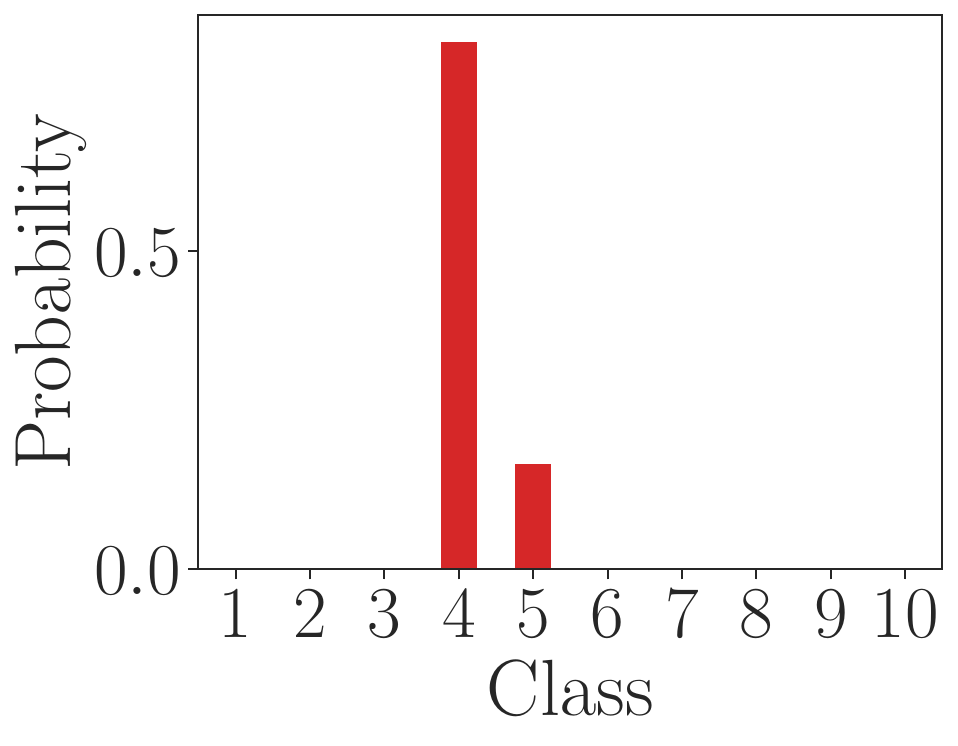}
       \end{subfigure}
         \begin{subfigure}[t]{0.19\linewidth}
           \centering
           \includegraphics[width=\linewidth]{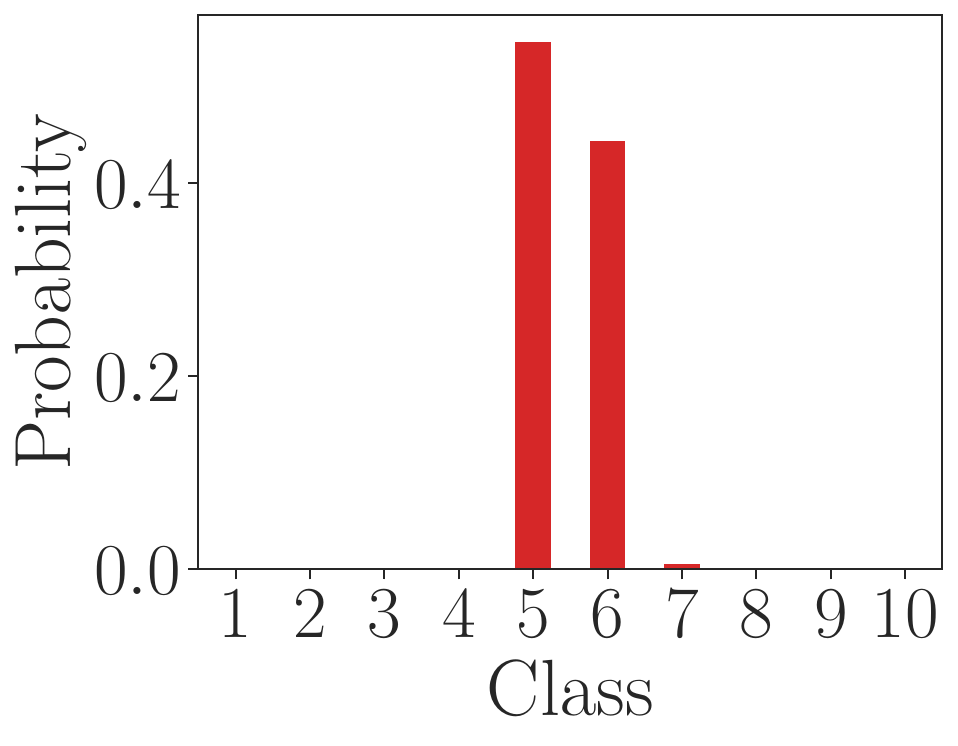}
  \end{subfigure}  
   \begin{subfigure}[t]{0.19\linewidth}
           \centering
           \includegraphics[width=\linewidth]{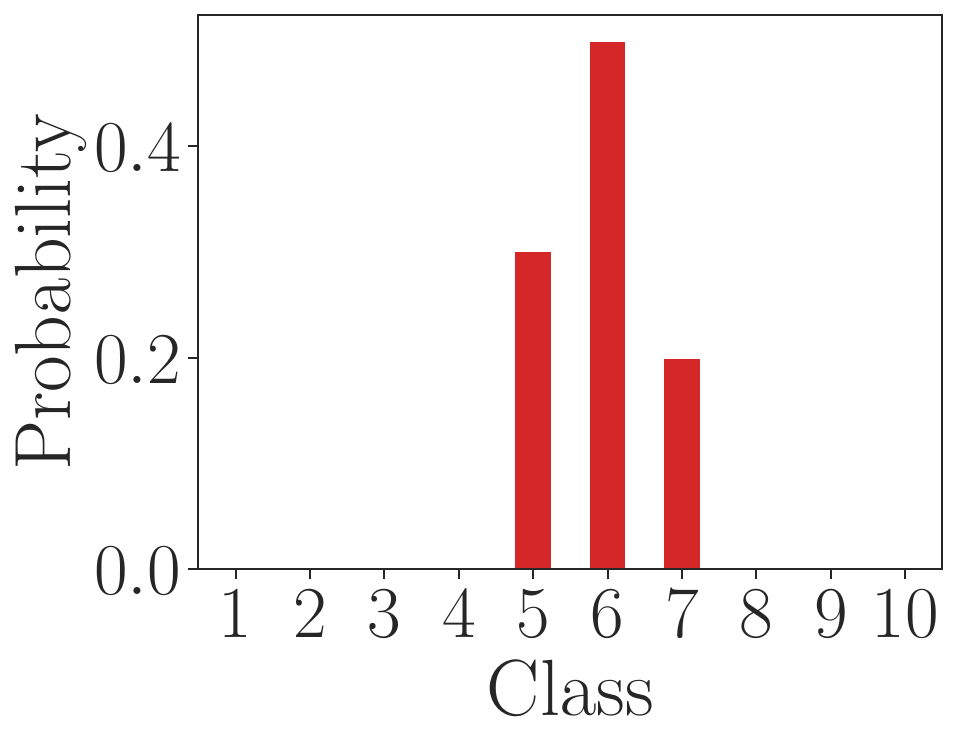}
       \end{subfigure}   
       \begin{subfigure}[t]{0.19\linewidth}
        \centering
          \includegraphics[width=\linewidth]{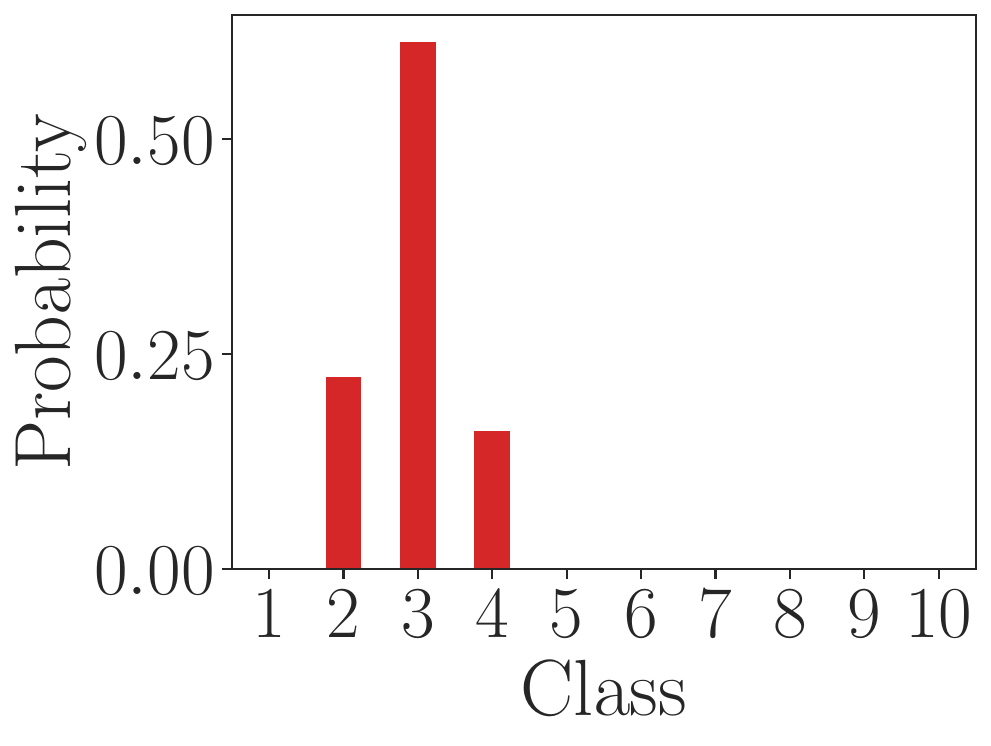}           
       \end{subfigure} 
         \begin{subfigure}[t]{0.19\linewidth}
           \centering
           \includegraphics[width=\linewidth]{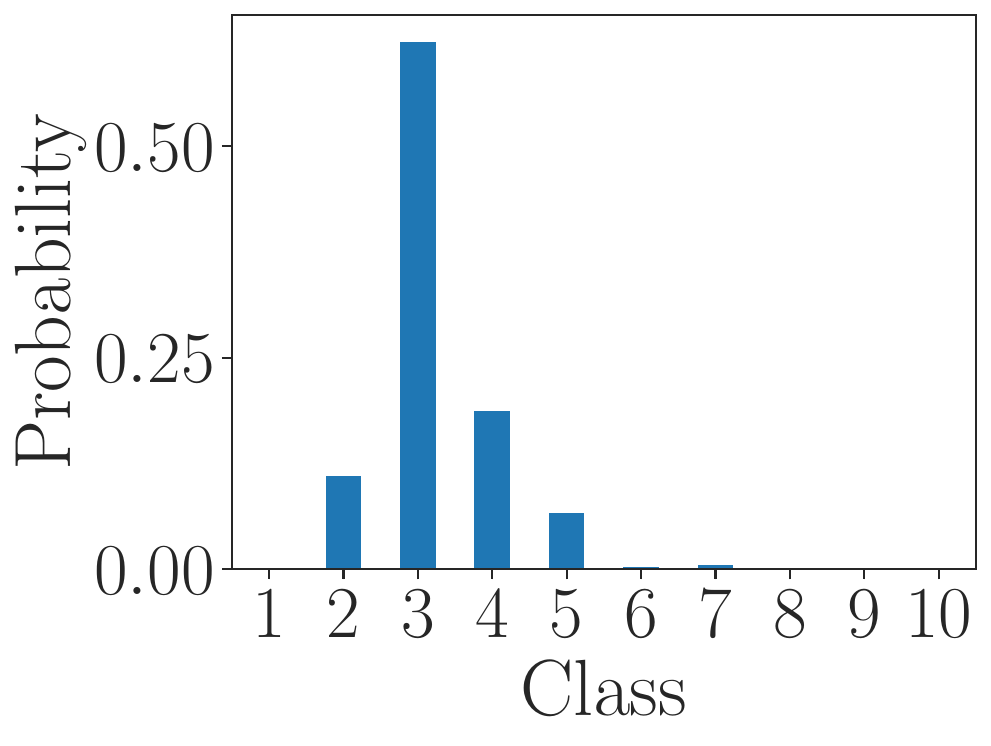}
       \end{subfigure}
         \begin{subfigure}[t]{0.19\linewidth}
           \centering
           \includegraphics[width=\linewidth]{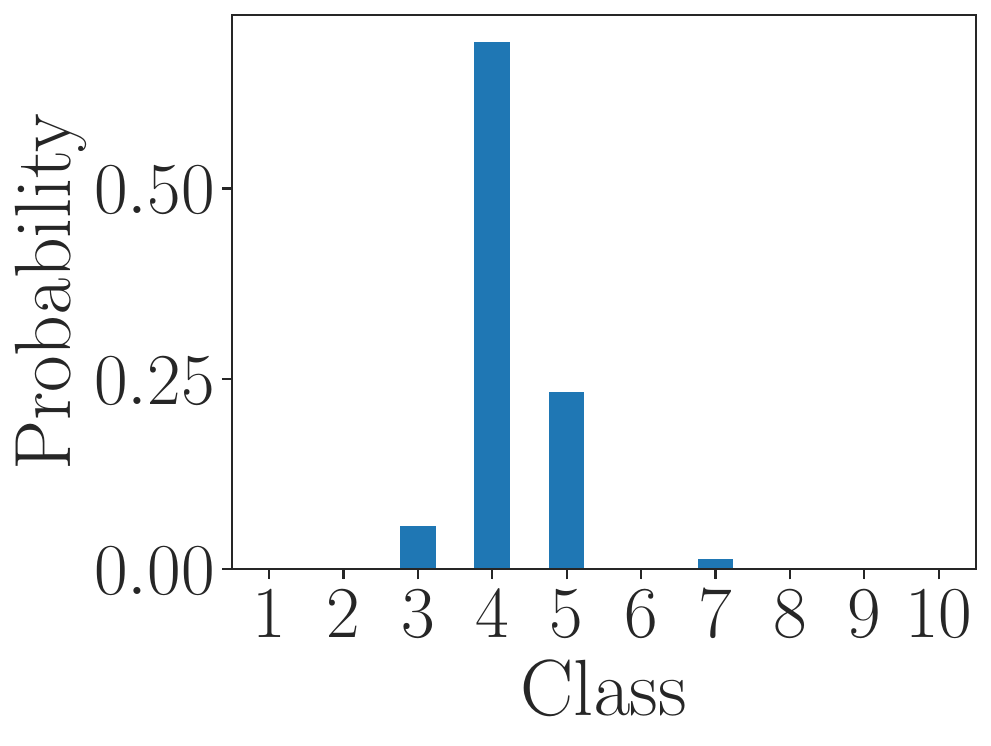}
  \end{subfigure}  
   \begin{subfigure}[t]{0.19\linewidth}
           \centering
           \includegraphics[width=\linewidth]{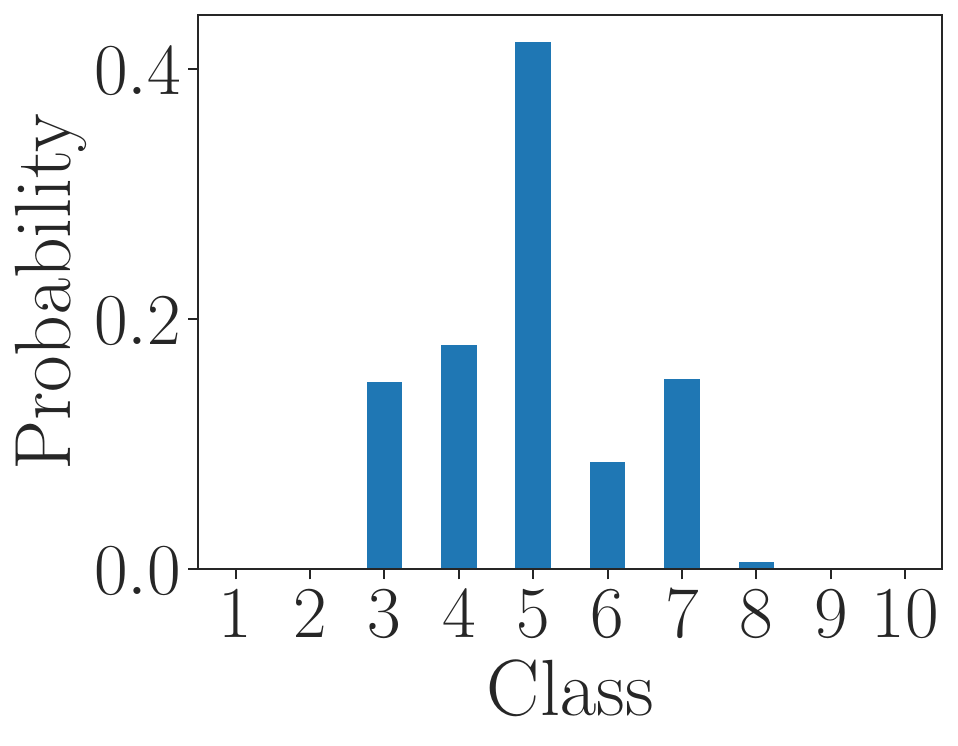}
       \end{subfigure}  
              \begin{subfigure}[t]{0.19\linewidth}
        \centering
          \includegraphics[width=\linewidth]{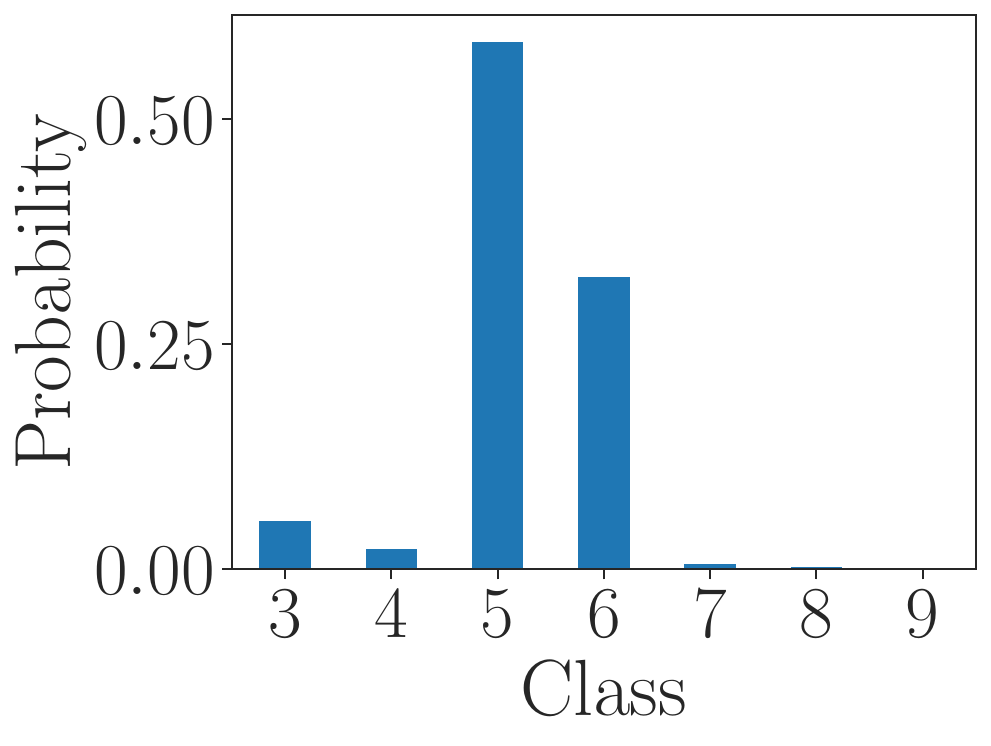}           
       \end{subfigure} 
         \begin{subfigure}[t]{0.19\linewidth}
           \centering
           \includegraphics[width=\linewidth]{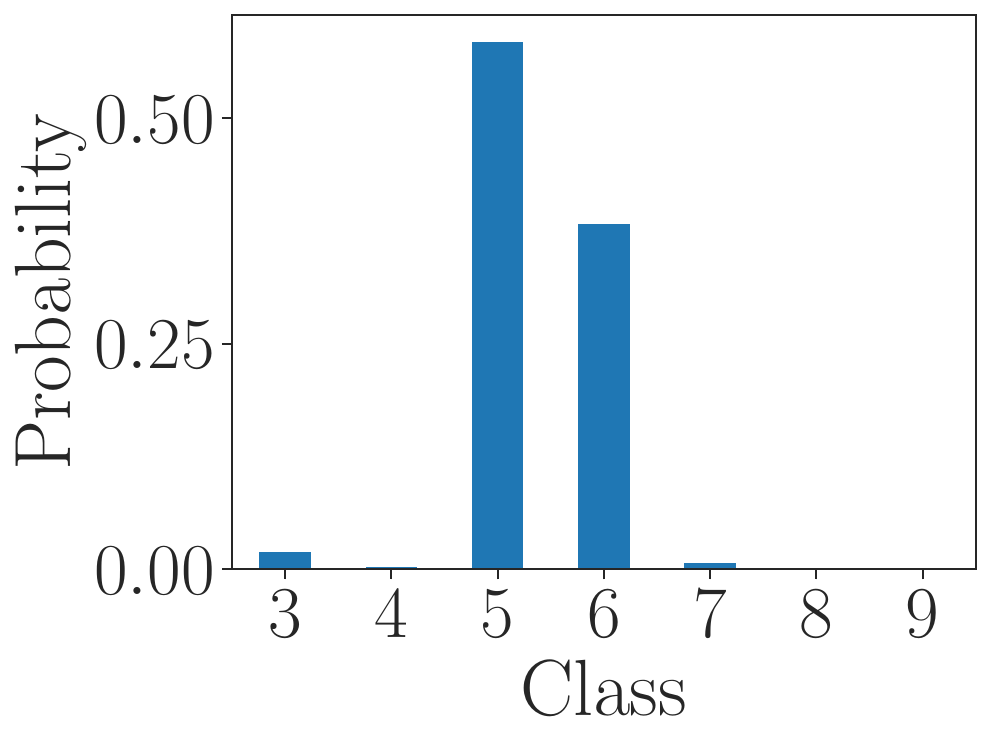}
       \end{subfigure}
         \begin{subfigure}[t]{0.19\linewidth}
           \centering
           \includegraphics[width=\linewidth]{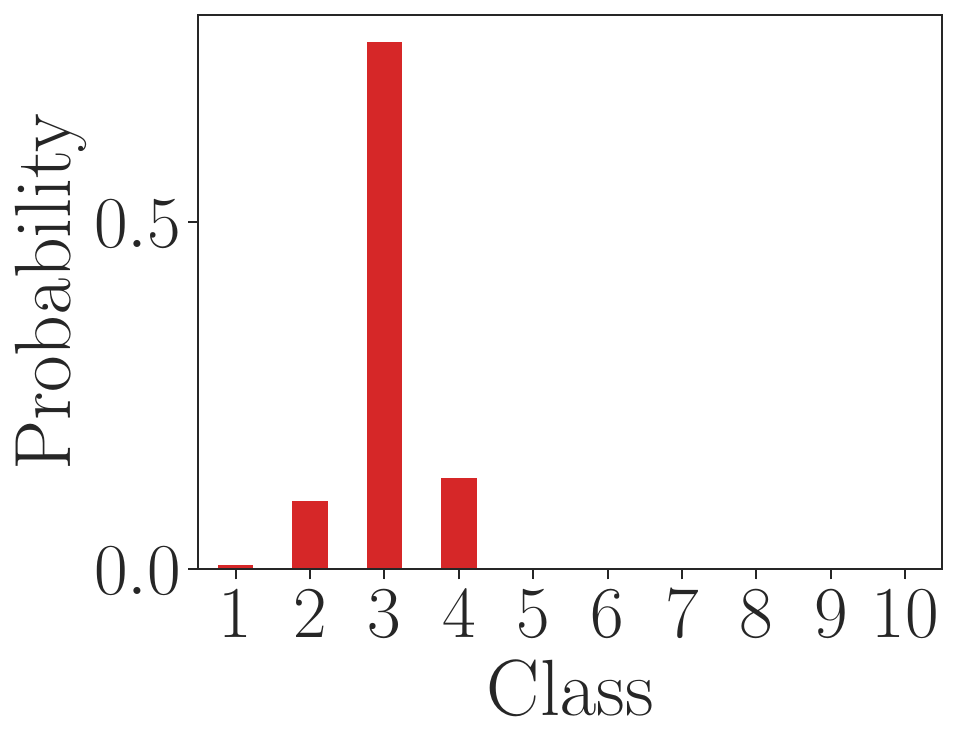}
       \end{subfigure}
         \begin{subfigure}[t]{0.19\linewidth}
           \centering
           \includegraphics[width=\linewidth]{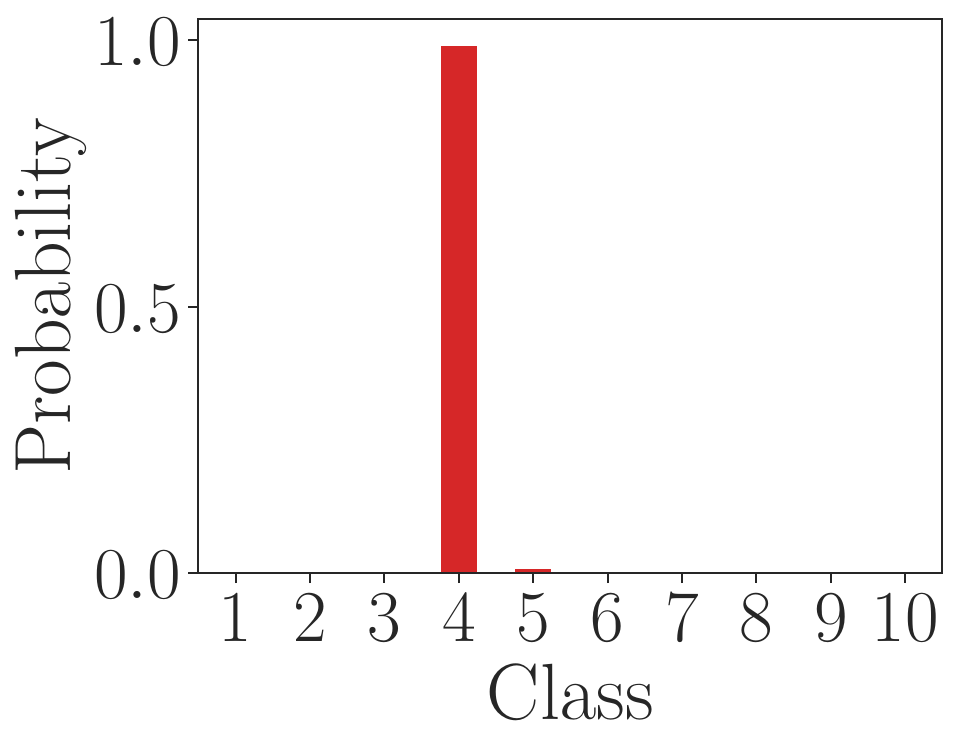}
  \end{subfigure}  
   \begin{subfigure}[t]{0.19\linewidth}
           \centering
           \includegraphics[width=\linewidth]{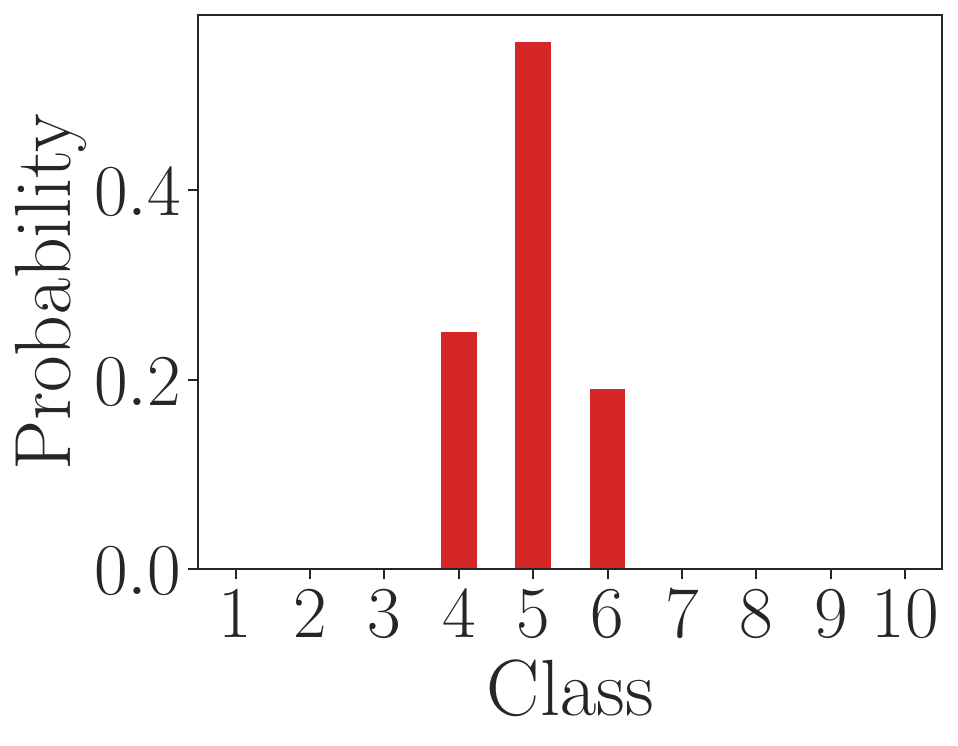}
       \end{subfigure}   
              \begin{subfigure}[t]{0.19\linewidth}
        \centering
          \includegraphics[width=\linewidth]{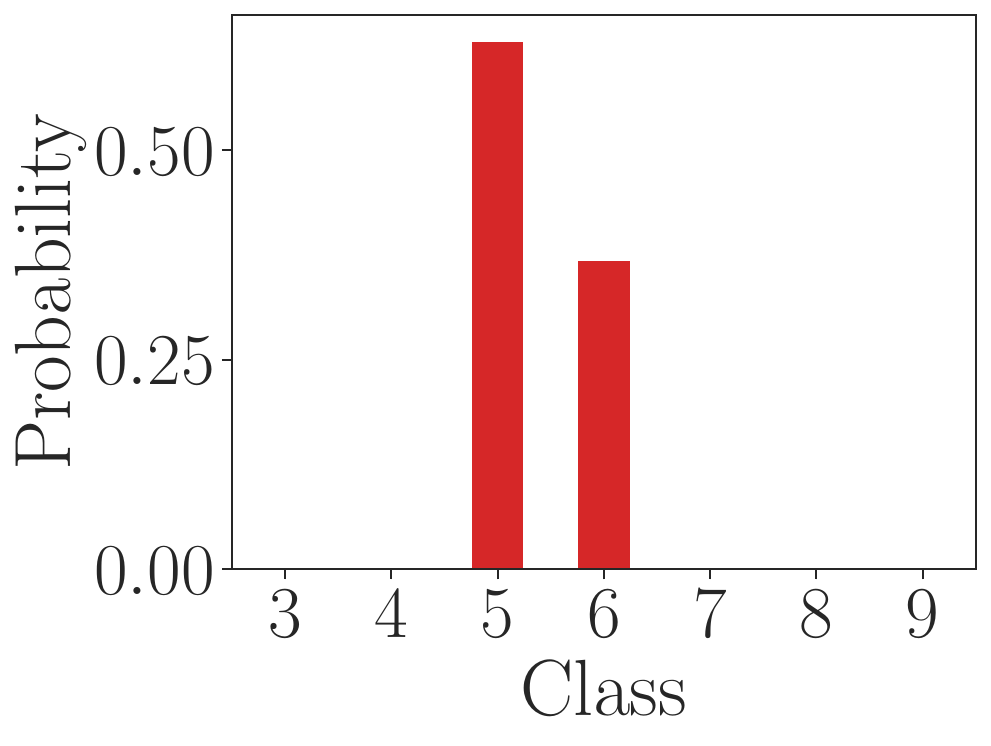}           
       \end{subfigure} 
         \begin{subfigure}[t]{0.19\linewidth}
           \centering
           \includegraphics[width=\linewidth]{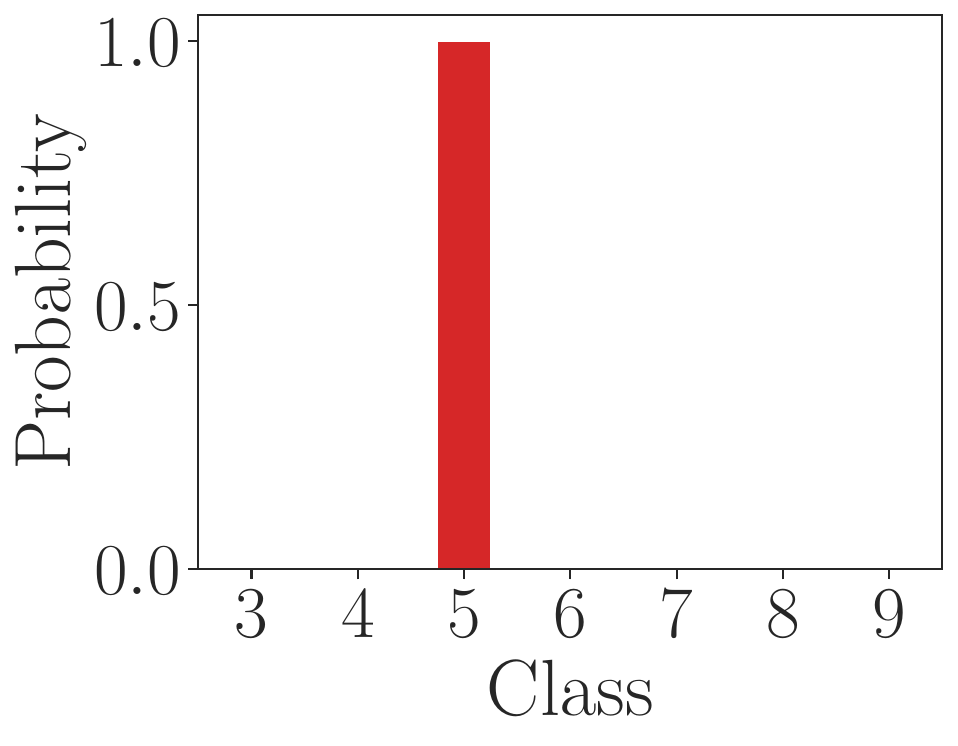}
          \end{subfigure}  
         \begin{subfigure}[t]{0.19\linewidth}
           \centering
           \includegraphics[width=\linewidth]{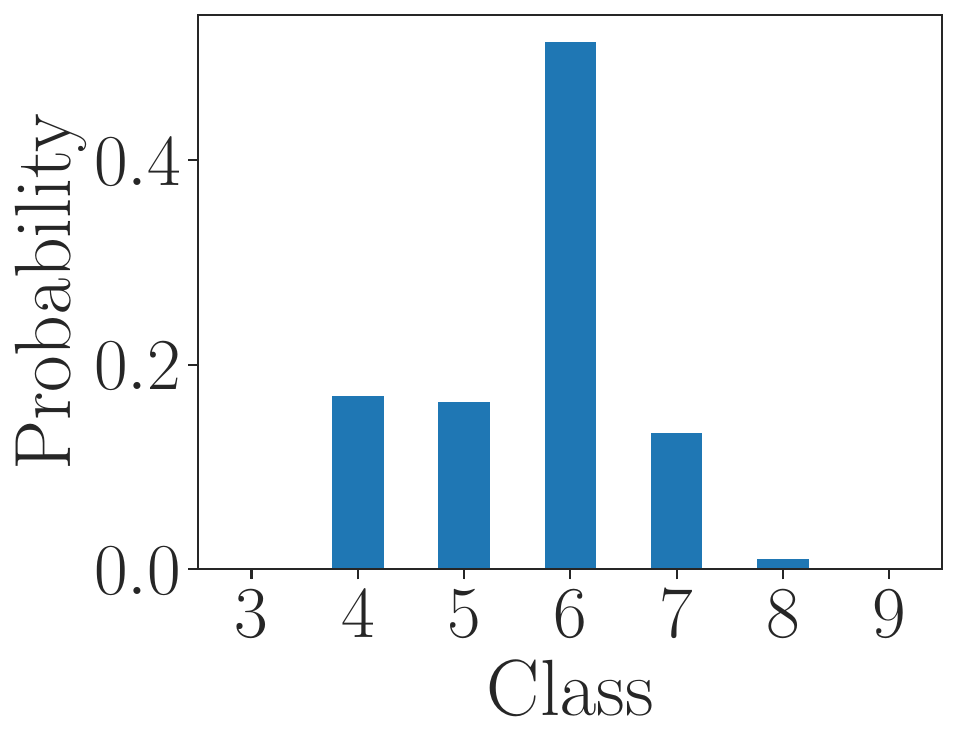}
  \end{subfigure}  
   \begin{subfigure}[t]{0.19\linewidth}
           \centering
           \includegraphics[width=\linewidth]{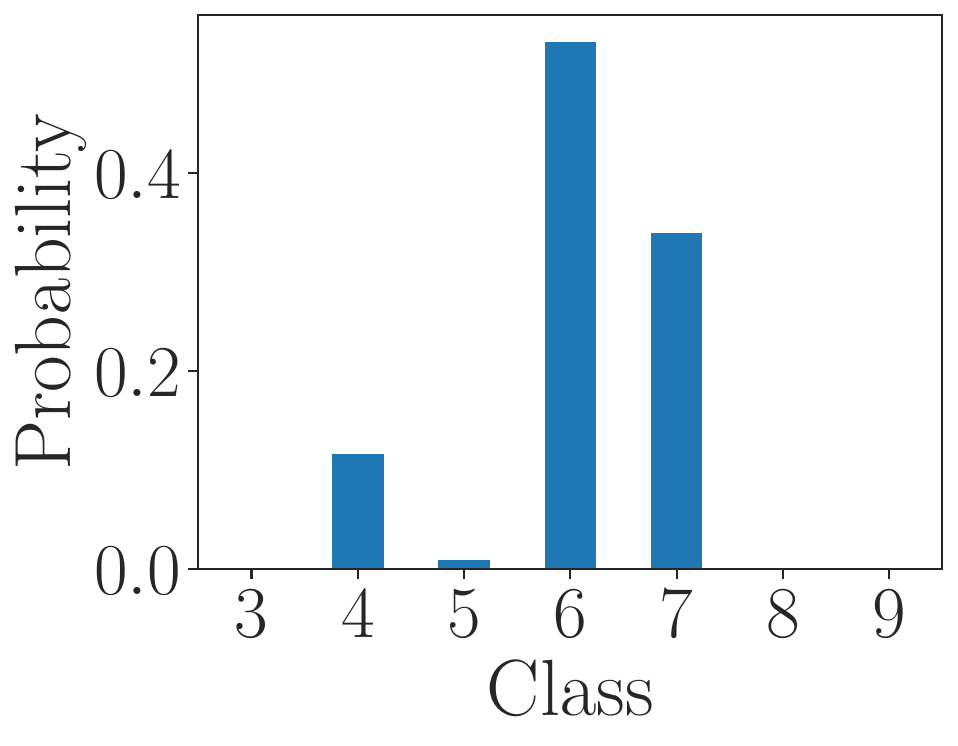}
       \end{subfigure}   
       \begin{subfigure}[t]{0.19\linewidth}
        \centering
          \includegraphics[width=\linewidth]{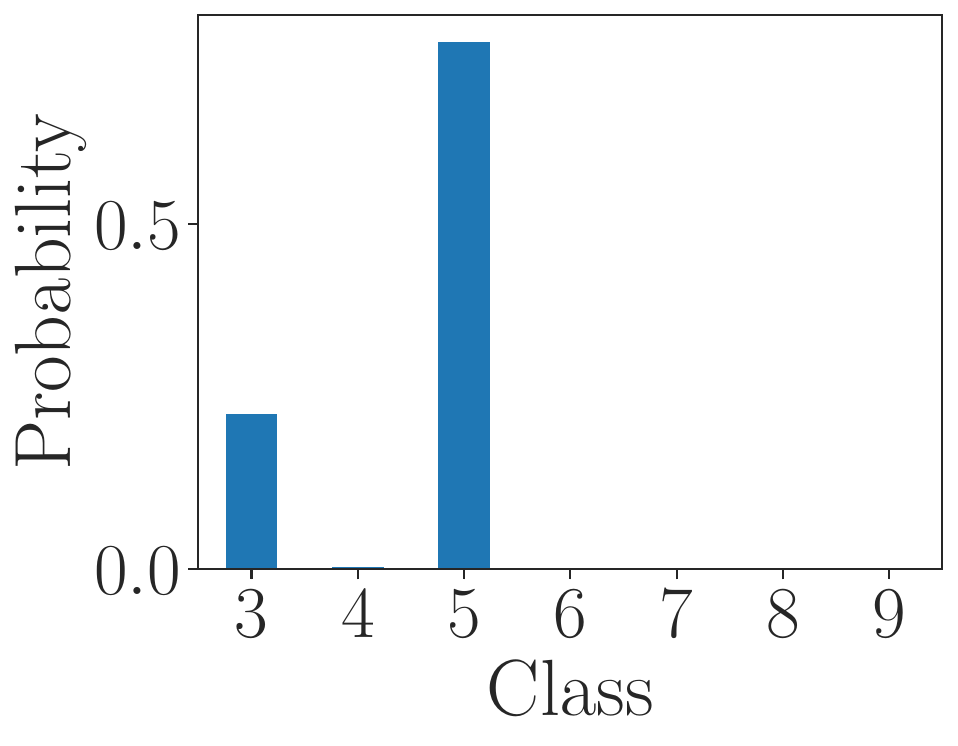}           
       \end{subfigure} 
         \begin{subfigure}[t]{0.19\linewidth}
           \centering
           \includegraphics[width=\linewidth]{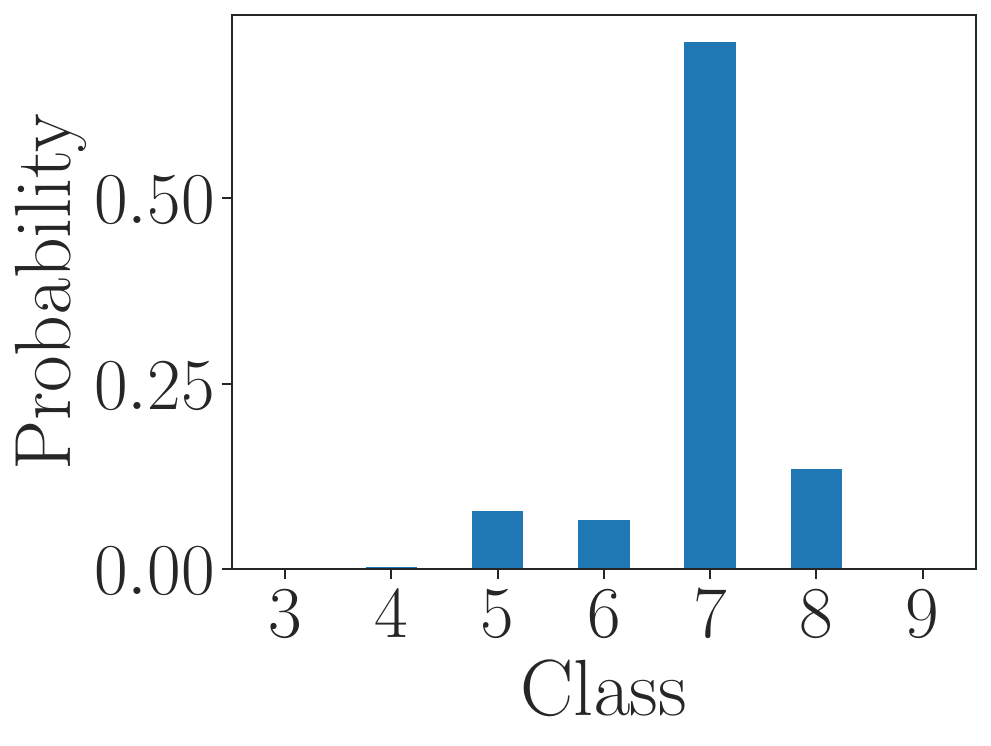}
       \end{subfigure}
         \begin{subfigure}[t]{0.19\linewidth}
           \centering
           \includegraphics[width=\linewidth]{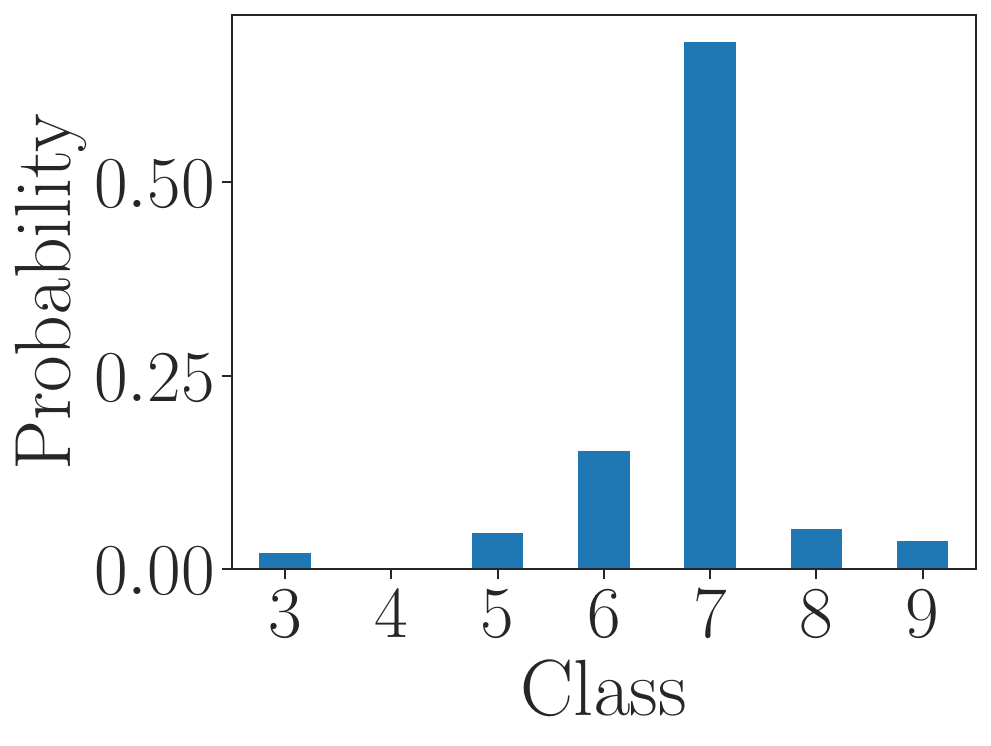}
  \end{subfigure}  
         \begin{subfigure}[t]{0.19\linewidth}
           \centering
           \includegraphics[width=\linewidth]{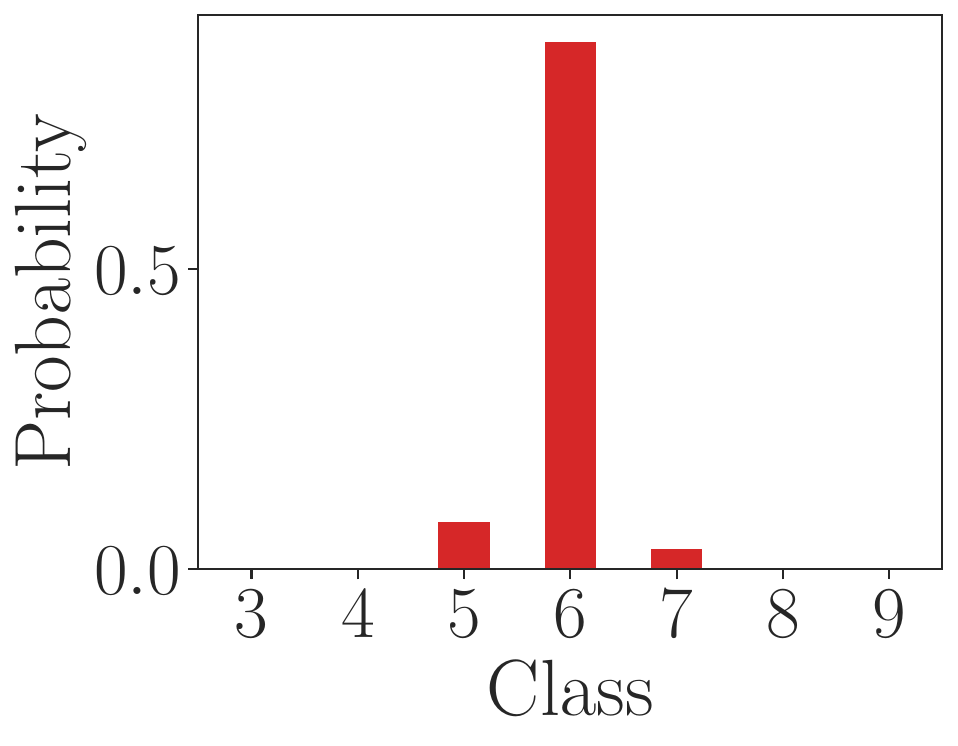}
  \end{subfigure}  
   \begin{subfigure}[t]{0.19\linewidth}
           \centering
           \includegraphics[width=\linewidth]{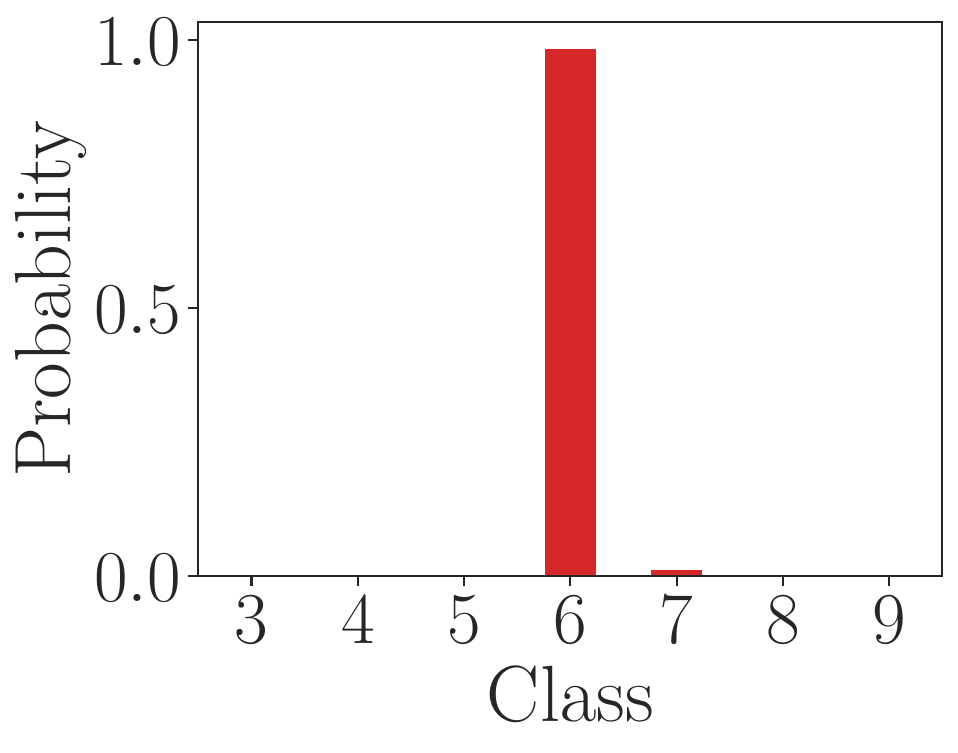}
       \end{subfigure}   
       \begin{subfigure}[t]{0.19\linewidth}
        \centering
          \includegraphics[width=\linewidth]{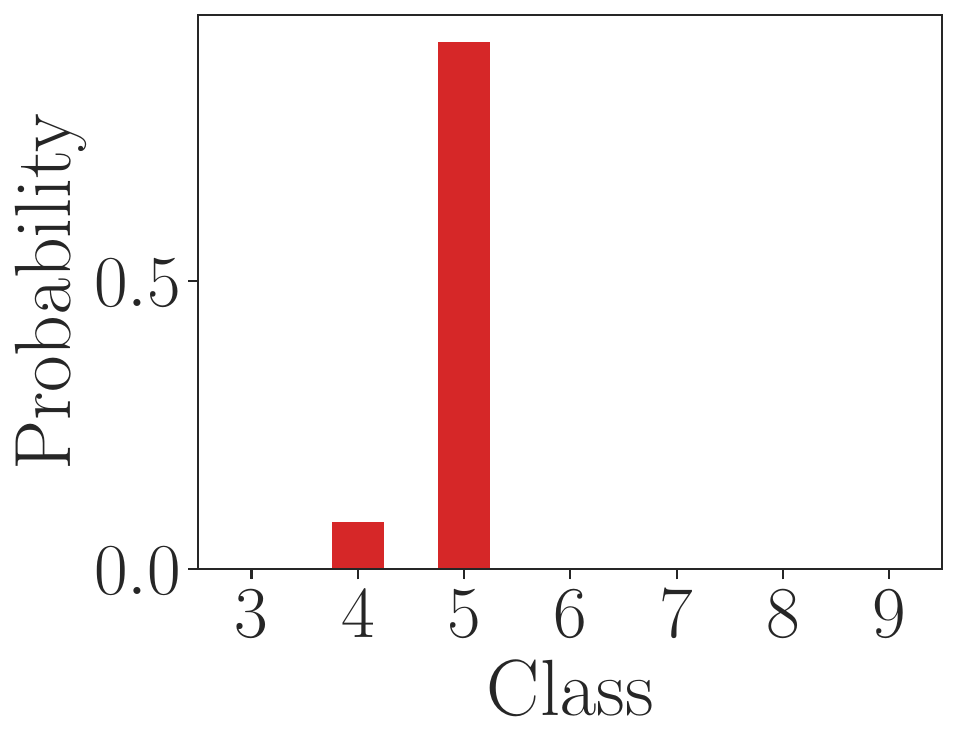}           
       \end{subfigure} 
         \begin{subfigure}[t]{0.19\linewidth}
           \centering
           \includegraphics[width=\linewidth]{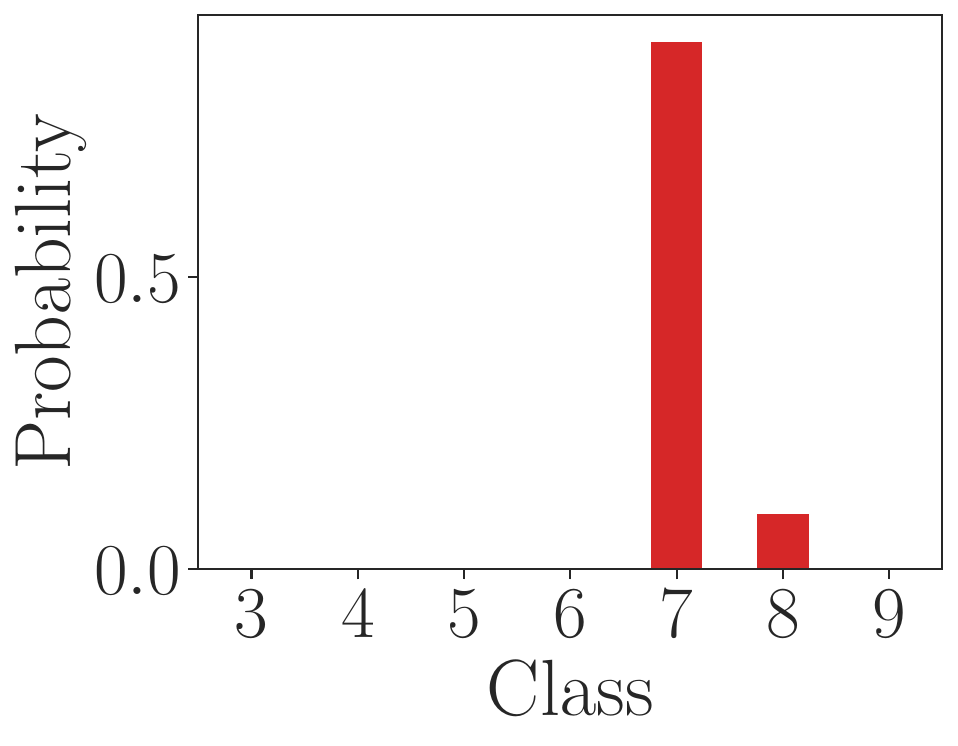}
       \end{subfigure}
         \begin{subfigure}[t]{0.19\linewidth}
           \centering
           \includegraphics[width=\linewidth]{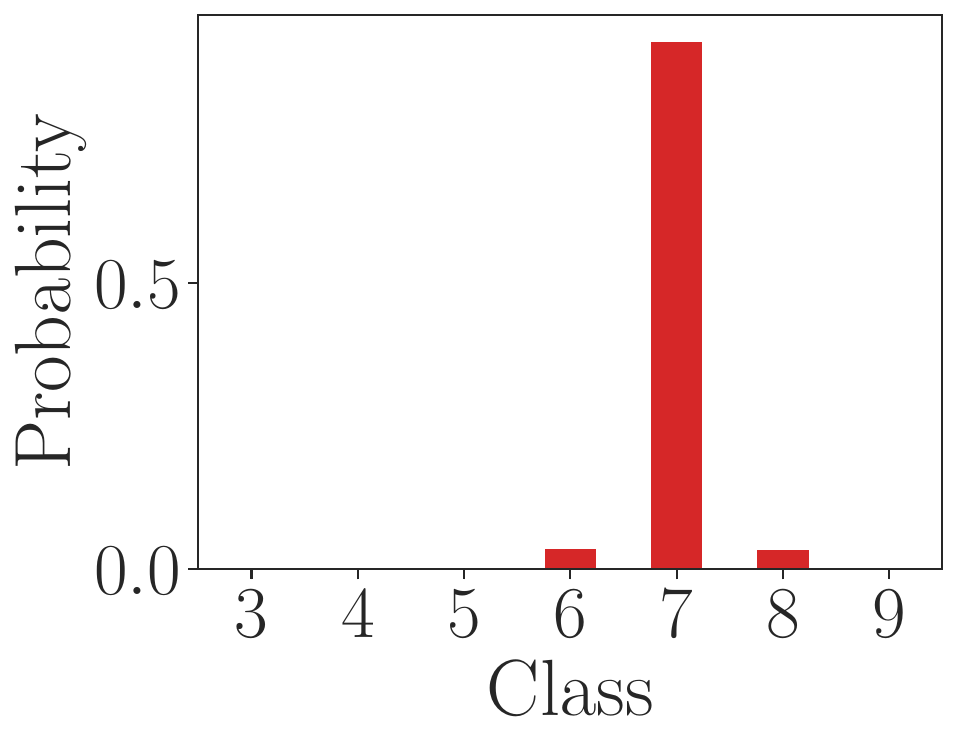}
          \end{subfigure}  
   \begin{subfigure}[t]{0.19\linewidth}
           \centering
           \includegraphics[width=\linewidth]{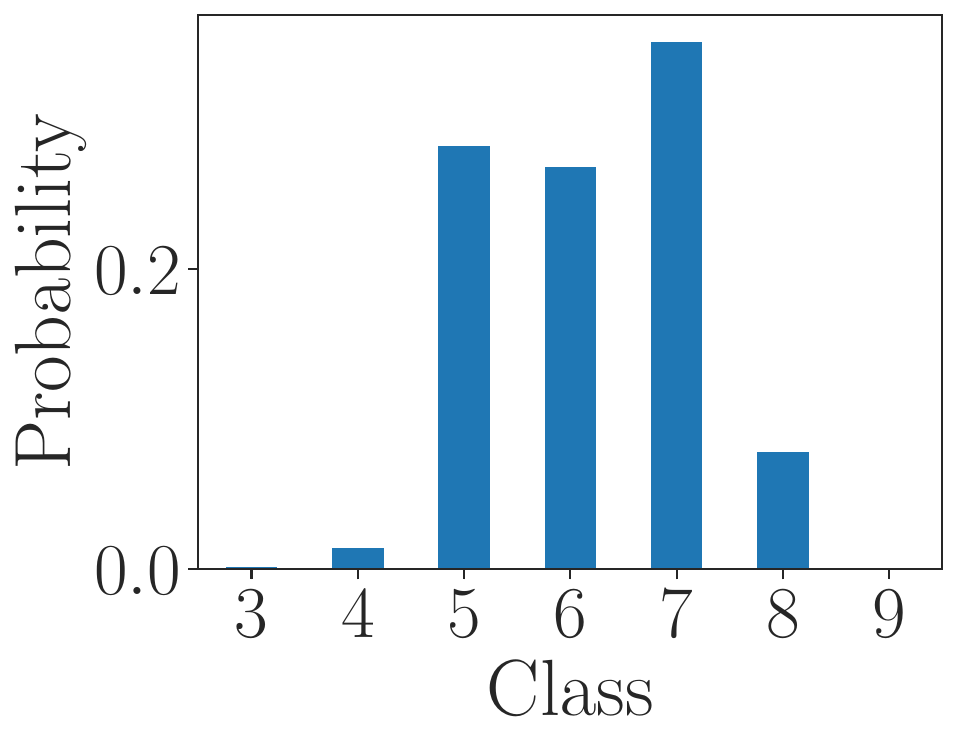}
       \end{subfigure}  
        \begin{subfigure}[t]{0.19\linewidth}
          \centering
            \includegraphics[width=\linewidth]{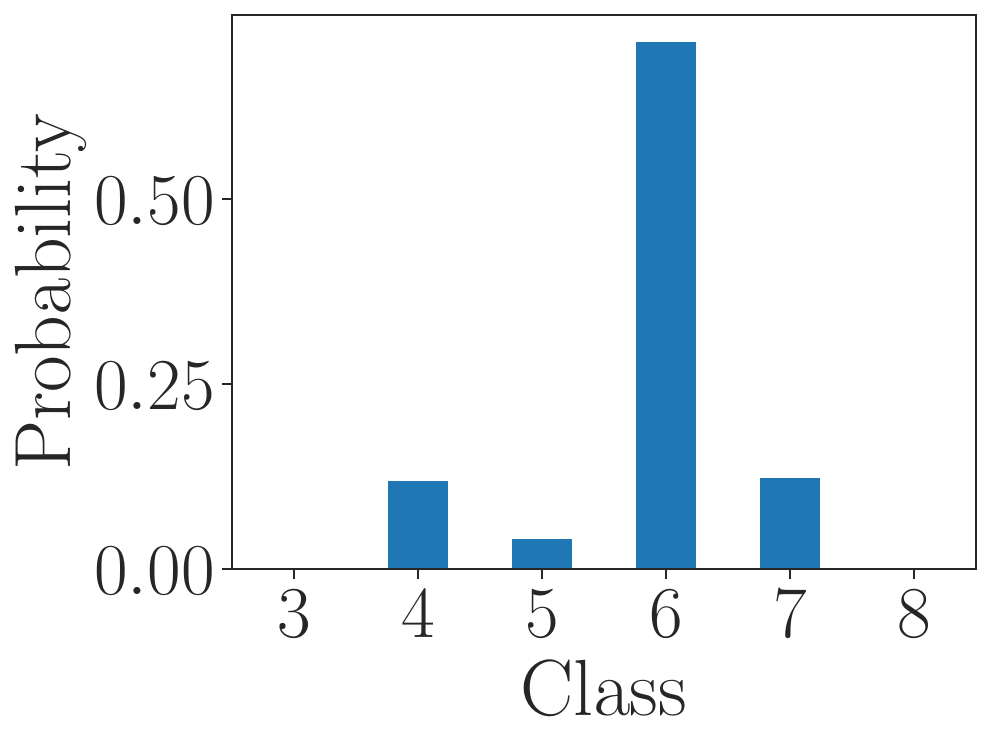}           
         \end{subfigure} 
           \begin{subfigure}[t]{0.19\linewidth}
             \centering
             \includegraphics[width=\linewidth]{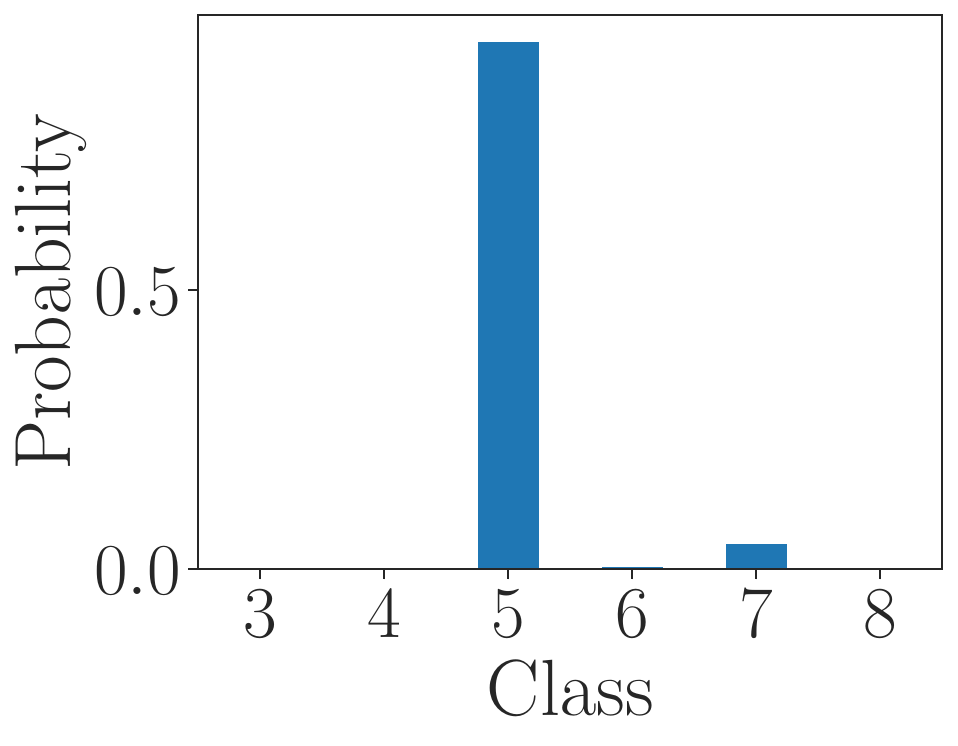}
         \end{subfigure}
           \begin{subfigure}[t]{0.19\linewidth}
             \centering
             \includegraphics[width=\linewidth]{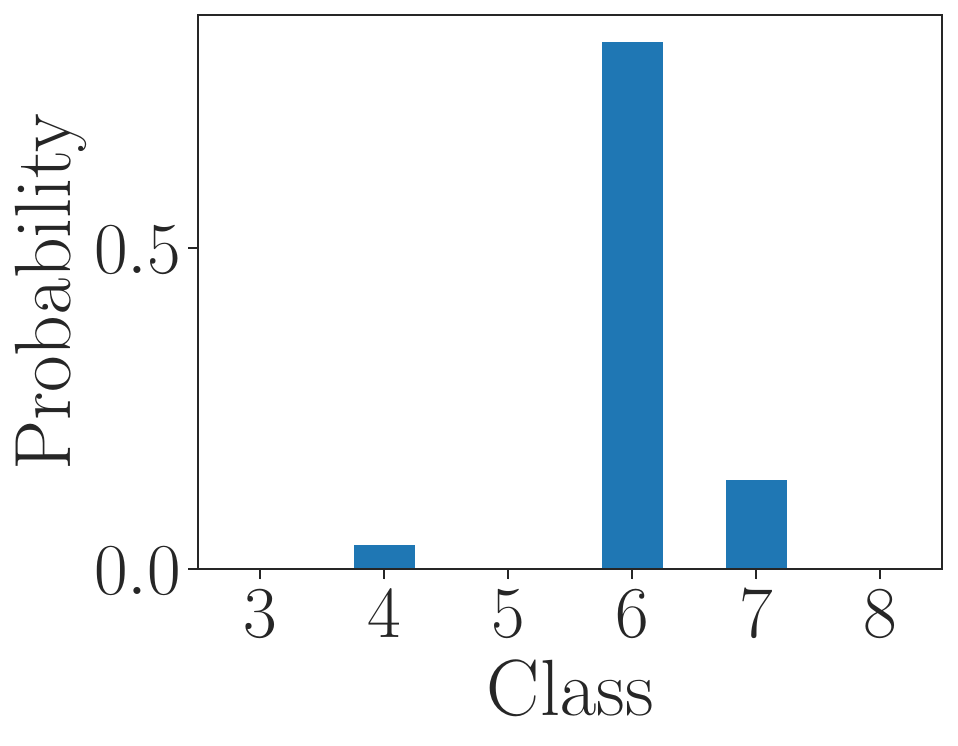}
    \end{subfigure}  
     \begin{subfigure}[t]{0.19\linewidth}
             \centering
             \includegraphics[width=\linewidth]{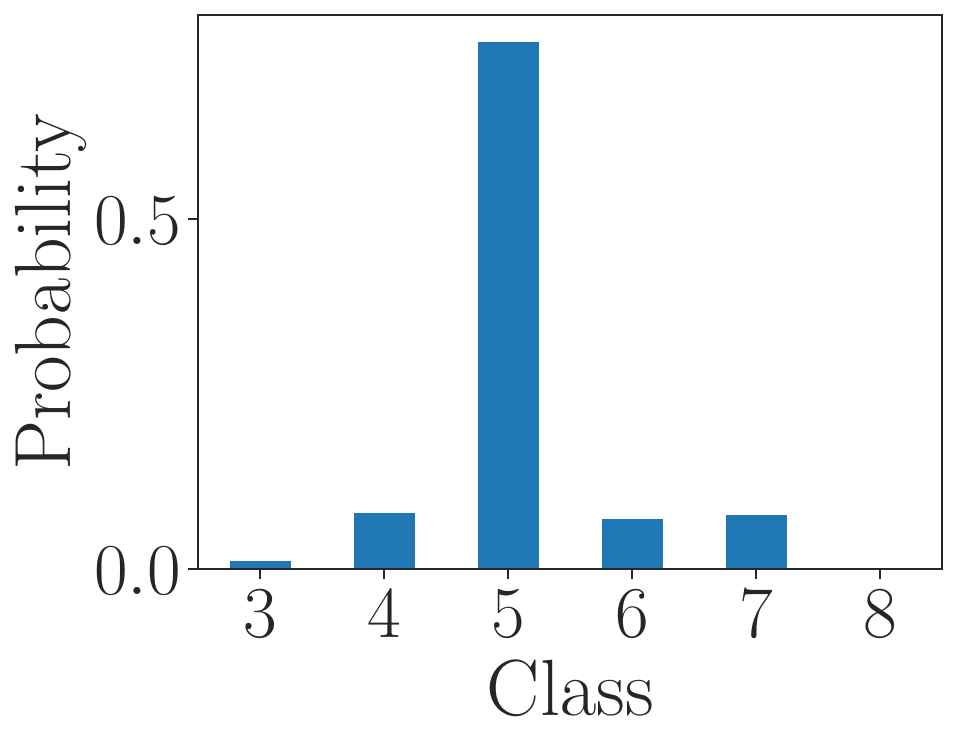}
         \end{subfigure}     
              \begin{subfigure}[t]{0.19\linewidth}
           \centering
           \includegraphics[width=\linewidth]{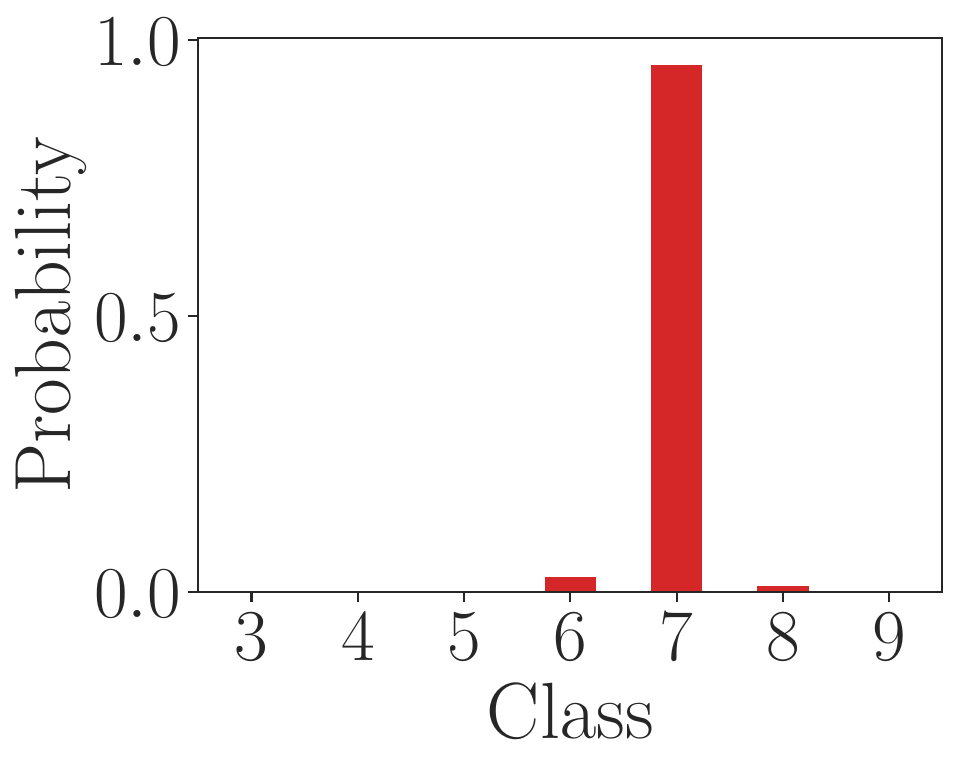}
       \end{subfigure} 
         \begin{subfigure}[t]{0.19\linewidth}
          \centering
            \includegraphics[width=\linewidth]{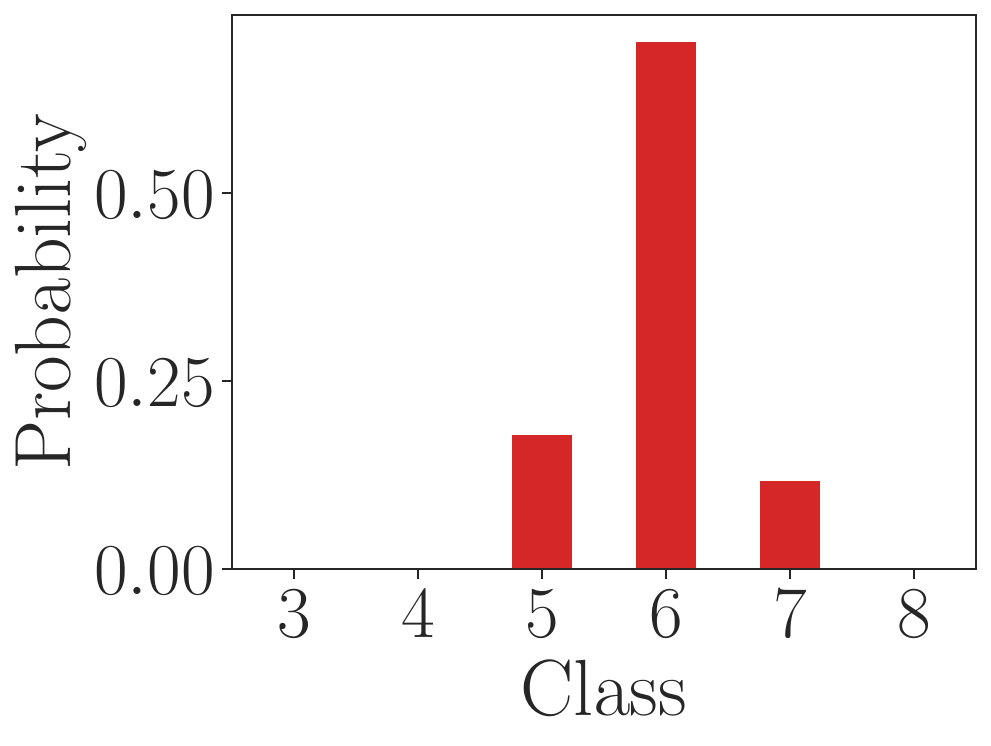}           
         \end{subfigure} 
           \begin{subfigure}[t]{0.19\linewidth}
             \centering
             \includegraphics[width=\linewidth]{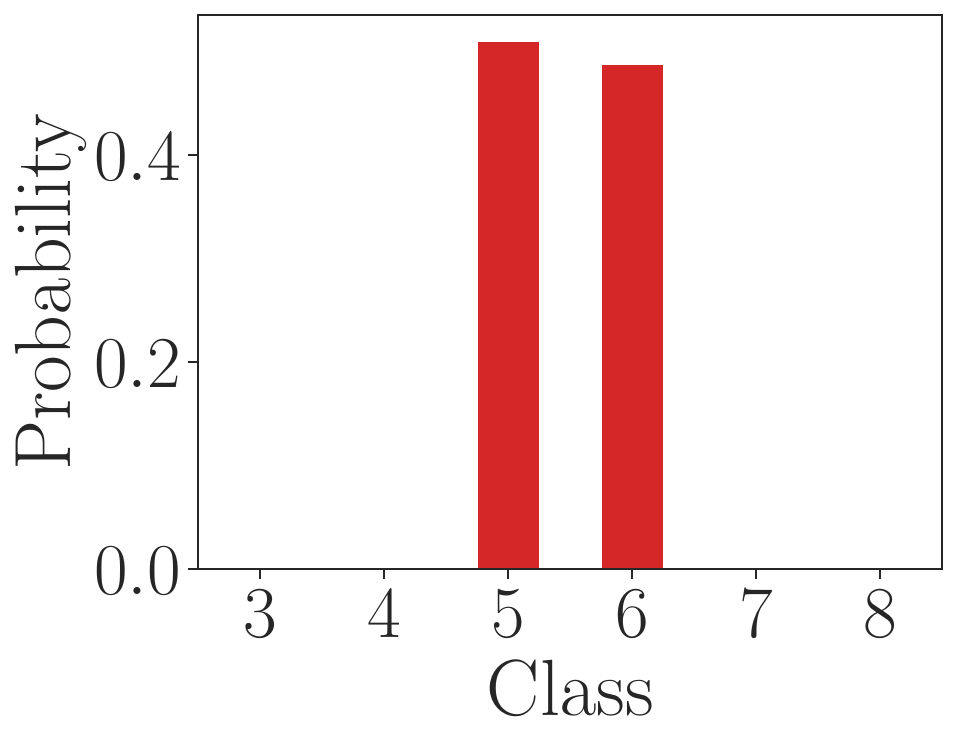}
         \end{subfigure}
           \begin{subfigure}[t]{0.19\linewidth}
             \centering
             \includegraphics[width=\linewidth]{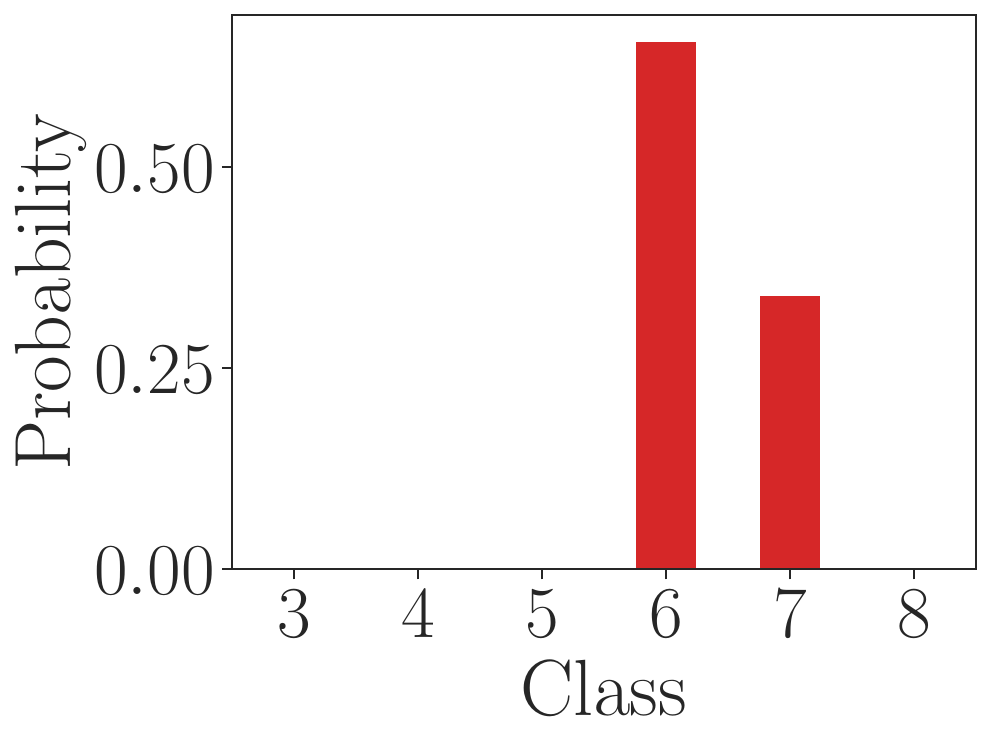}
    \end{subfigure}  
     \begin{subfigure}[t]{0.19\linewidth}
             \centering
             \includegraphics[width=\linewidth]{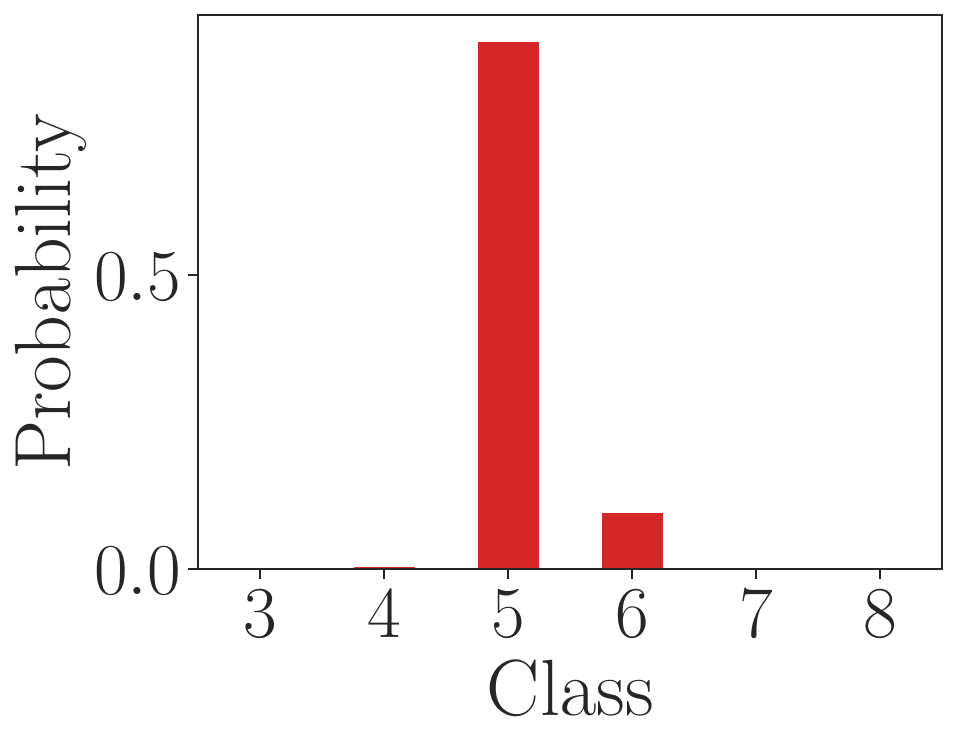}
         \end{subfigure}  
        \caption{Comparison of exemplary predictive probability distributions using the MLP with CE loss and the MLP with QWK loss on the Abalone, White Wine, and Red Wine datasets. Odd rows (blue) display the CE loss distributions, while even rows (red) show the corresponding QWK loss distributions.}
        \label{fig:probas}
\end{figure}

Overall, we can conclude that although ordinal losses deliver good ordinal predictions on ordinal metrics, which are even well-calibrated in the case of EMD, they bias the predictor to output unimodal, compressed probability distributions. See Figure \ref{fig:probas} for an illustration of this phenomenon, where we compare the predictive probability distributions of the CE loss and the QWK loss. This behavior, as also demonstrated by \cite{DBLP:journals/prl/TorrePV18} for the QWK loss, appears to negatively impact uncertainty quantification.
Although EMD loss may improve calibration in probabilistic ordinal classification on average, it introduces a bias in the predictive probabilities by enforcing a unimodality assumption that is not universally valid. This inductive bias can obscure information critical for reliable uncertainty quantification in probabilistic ordinal classification.
This observation is particularly notable in our experiments, as most datasets exhibit a unimodal prior distribution of class labels, $p(y)$, which might suggest that the predictive distributions, $p(y \,|\, \vec{x})$, are also unimodal in most cases, which aligns with standard assumptions in ordinal classification. However, when uncertainty quantification is a priority, CE loss emerges as a more suitable choice, as it provides unbiased predictive probability distributions in ordinal classification.

\section{Experiments with Ensemble of Multi-Layer Perceptron (MLP)}
  \label{asec:prrs_mlp}

  This section shows additional experimental results using an ensemble of Multi-Layer Perceptrons (MLPs) \citep{scikit-learn} instead of an ensemble of GBTs (cf.\ Section \ref{sec:experiments}) to approximate Bayesian inference.

  \subsection{Experimental Setup}

  For the experiments, the same datasets are used as in Section \ref{sec:experiments} (cf.\ Table \ref{tab:ord_benchmark}). Besides the preprocessing applied in Section \ref{sec:experiments}, numerical features are also standardized. The parameters of the MLPs are displayed in Table \ref{appendix:tab:MLP_params}. Please note that our focus is not on predictive performance, but on uncertainty quantification, so we deliberately do not perform extensive hyperparameter tuning. Nonetheless, we selected an architecture for the MLPs that leads to competitive performance compared to LightGBM, or GBTs in general.
  To create diversity among the MLPs in the ensemble, we provide different seeds, leading to different random number generations for weights and bias initialization.

  \begin{table}[!hbt]
    \begin{tabular}{ll}
    \toprule
    \textbf{Parameter}         & \textbf{Value} \\
    \midrule
    Hidden Layer Sizes             & [128,64,32] \\
    Activation Function        & ReLU \\
    Solver                     & Adam \\
    Maximum Epochs             & 200 \\
    Batch Size                 & 200 \\
    L2 Regularization (alpha)  & 1e-4 \\
    Learning Rate             & 1e-3 \\
    \bottomrule
    \end{tabular}
    \caption{MLP parameters \citep{scikit-learn}.}
    \label{appendix:tab:MLP_params}
    \end{table}

 \subsection{Rejection Curves}

\begin{figure}[!htbp]
  \centering
  \begin{subfigure}[t]{0.3\linewidth}
    \centering
    \includegraphics[width=\linewidth]{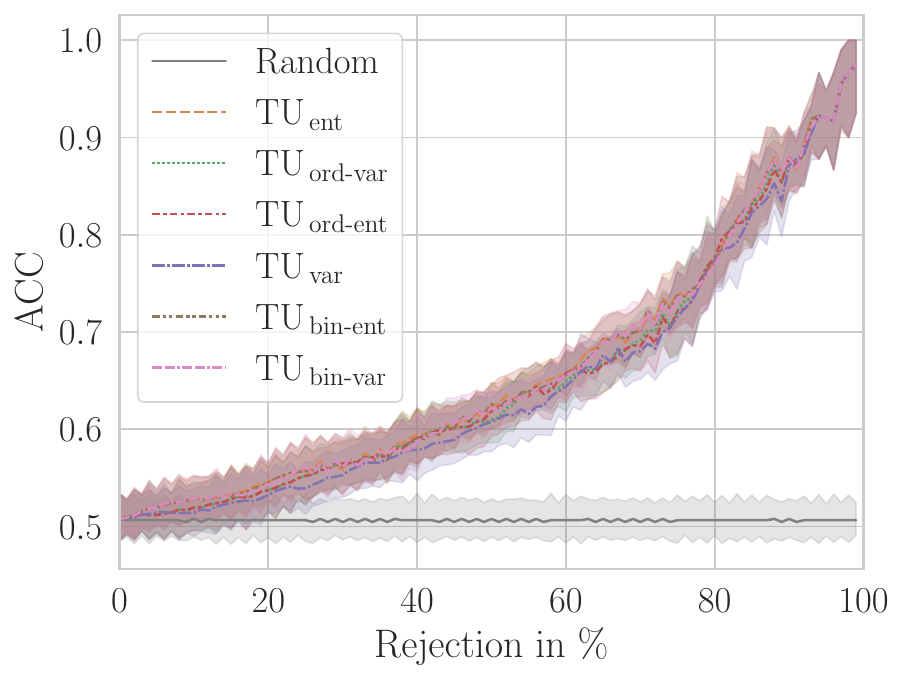}
    \subcaption{CMC TU}
\end{subfigure} 
  \begin{subfigure}[t]{0.3\linewidth}
    \centering
    \includegraphics[width=\linewidth]{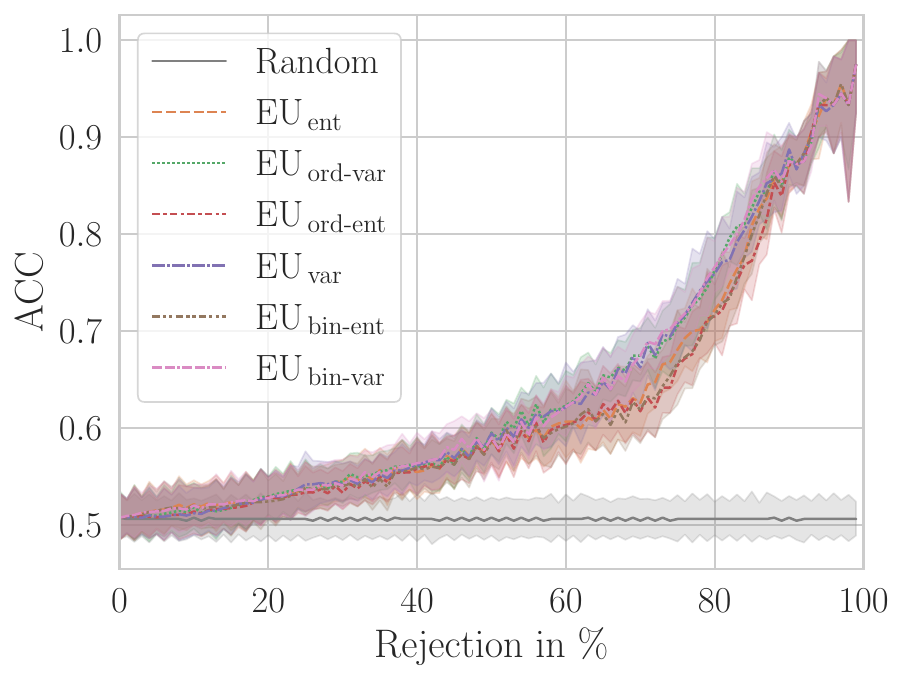}
    \subcaption{CMC EU}
\end{subfigure}  
\begin{subfigure}[t]{0.3\linewidth}
 \centering
 \includegraphics[width=\linewidth]{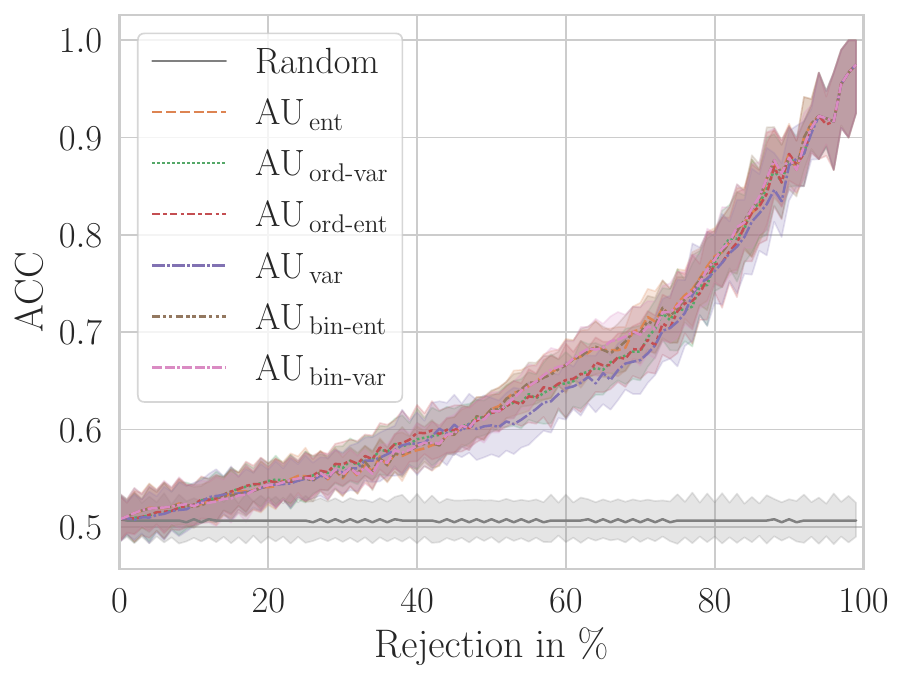}
 \subcaption{CMC AU}
\end{subfigure}   
    \begin{subfigure}[t]{0.3\linewidth}
      \centering
      \includegraphics[width=\linewidth]{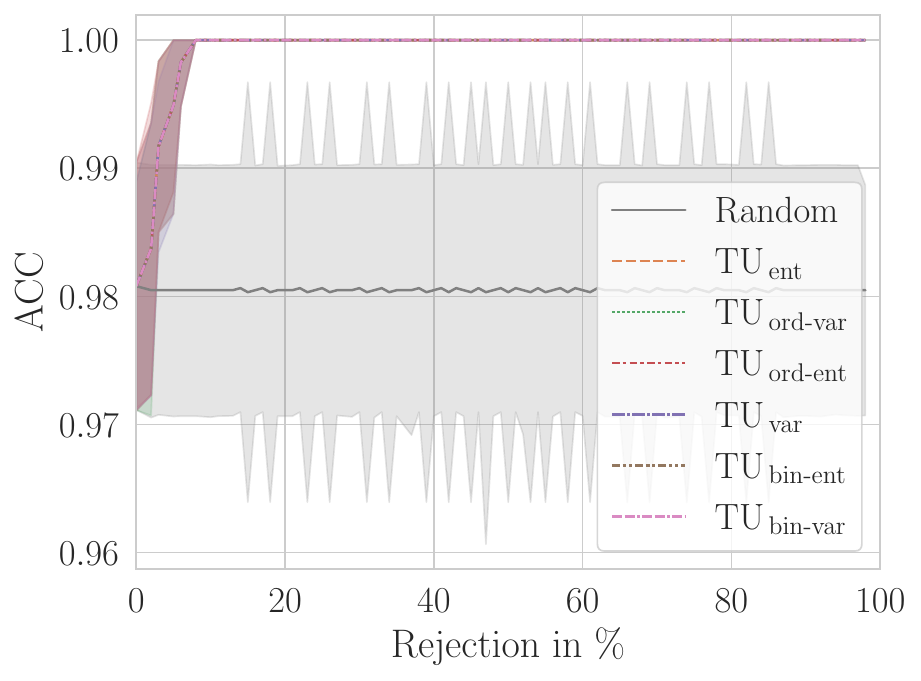}
      \subcaption{Balance Scale TU }
  \end{subfigure} 
    \begin{subfigure}[t]{0.3\linewidth}
      \centering
      \includegraphics[width=\linewidth]{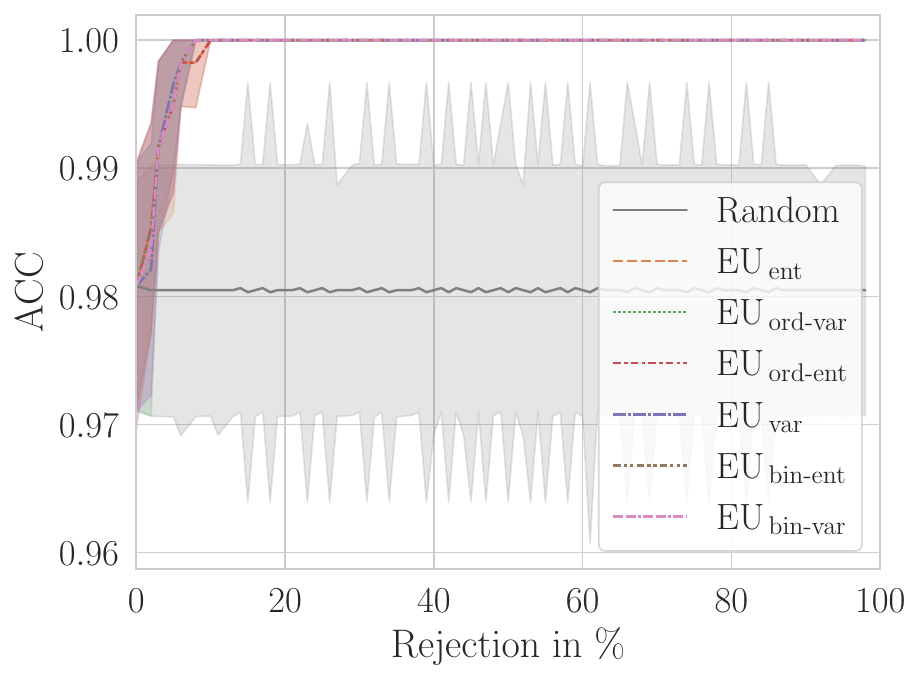}
      \subcaption{Balance Scale EU}
  \end{subfigure}  
  \begin{subfigure}[t]{0.3\linewidth}
   \centering
   \includegraphics[width=\linewidth]{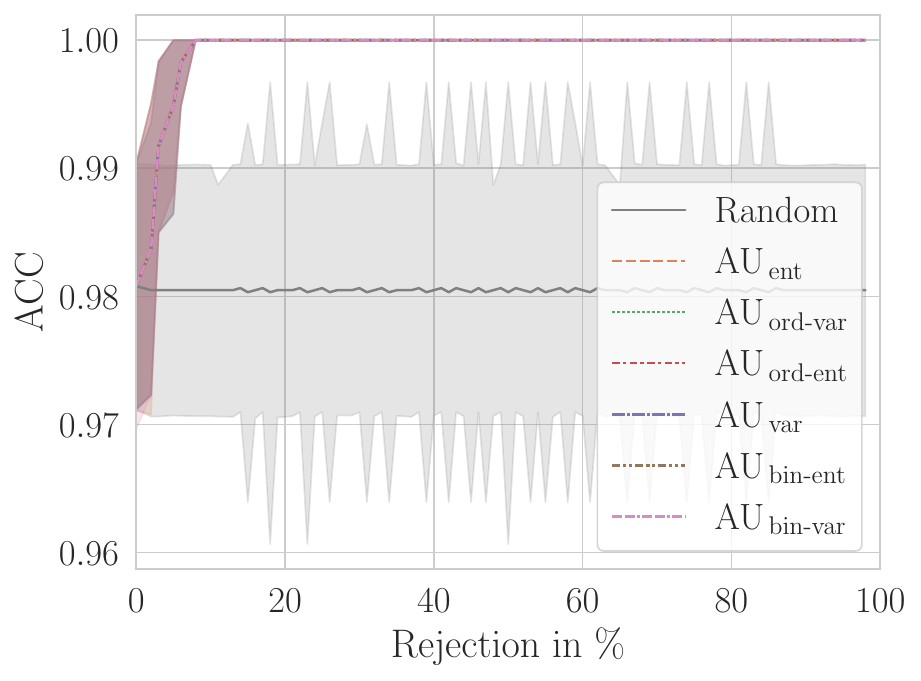}
   \subcaption{Balance Scale AU}
\end{subfigure}    
\begin{subfigure}[t]{0.3\linewidth}
  \centering
  \includegraphics[width=\linewidth]{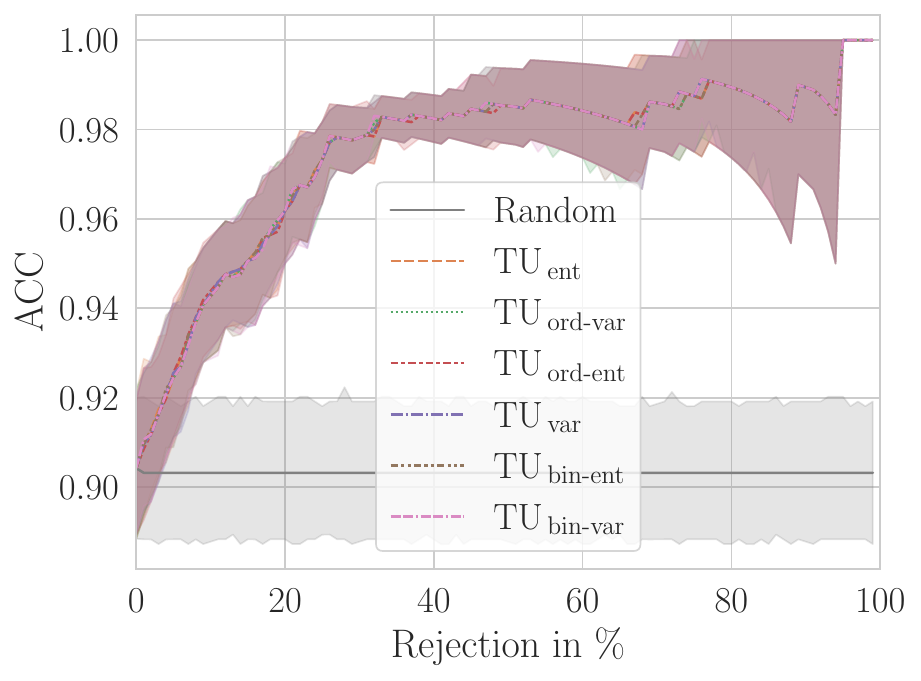}
  \subcaption{Stocks Domain TU }
\end{subfigure} 
\begin{subfigure}[t]{0.3\linewidth}
  \centering
  \includegraphics[width=\linewidth]{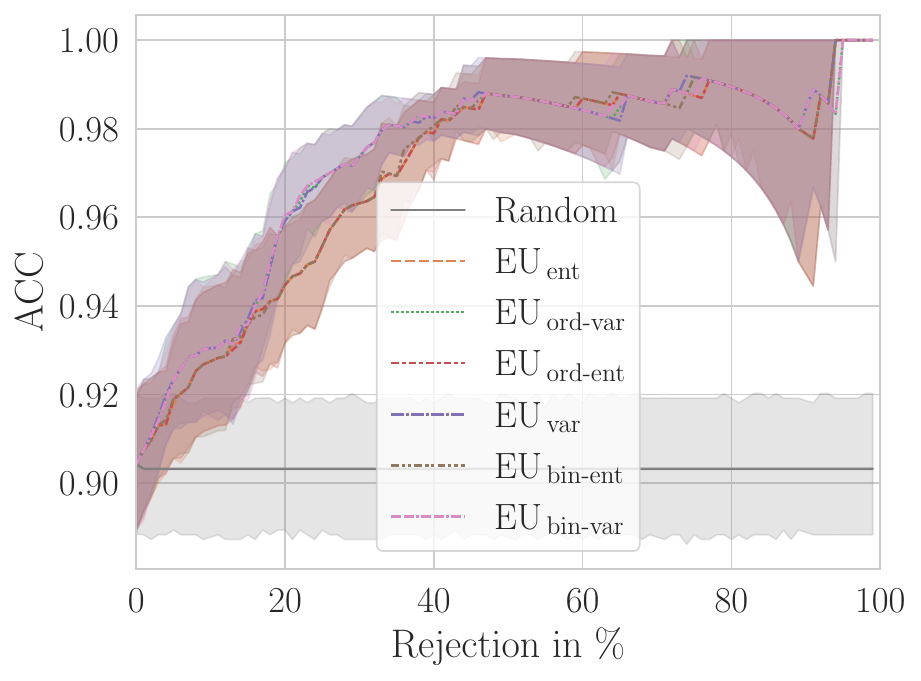}
  \subcaption{Stocks Domain EU}
\end{subfigure}  
\begin{subfigure}[t]{0.3\linewidth}
\centering
\includegraphics[width=\linewidth]{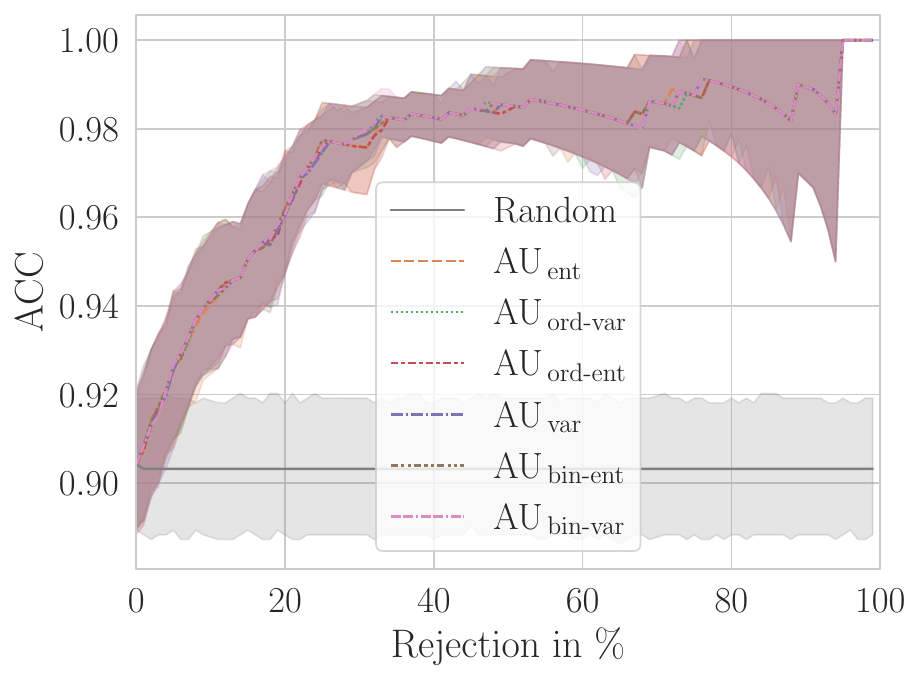}
\subcaption{Stocks Domain AU}
\end{subfigure}     
\begin{subfigure}[t]{0.3\linewidth}
  \centering
  \includegraphics[width=\linewidth]{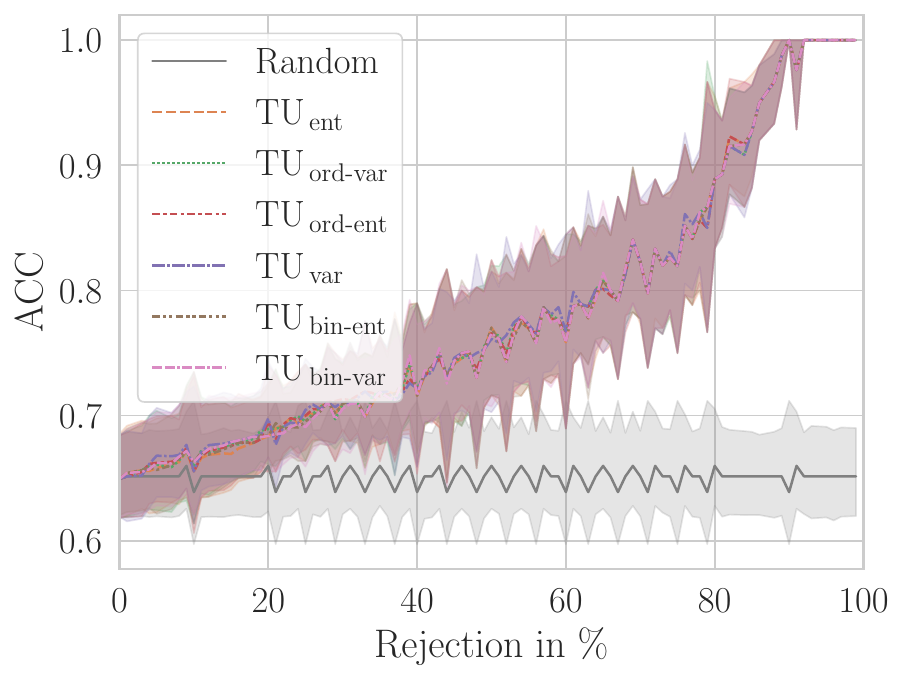}
  \subcaption{Eucalyptus TU}
\end{subfigure} 
\begin{subfigure}[t]{0.3\linewidth}
  \centering
  \includegraphics[width=\linewidth]{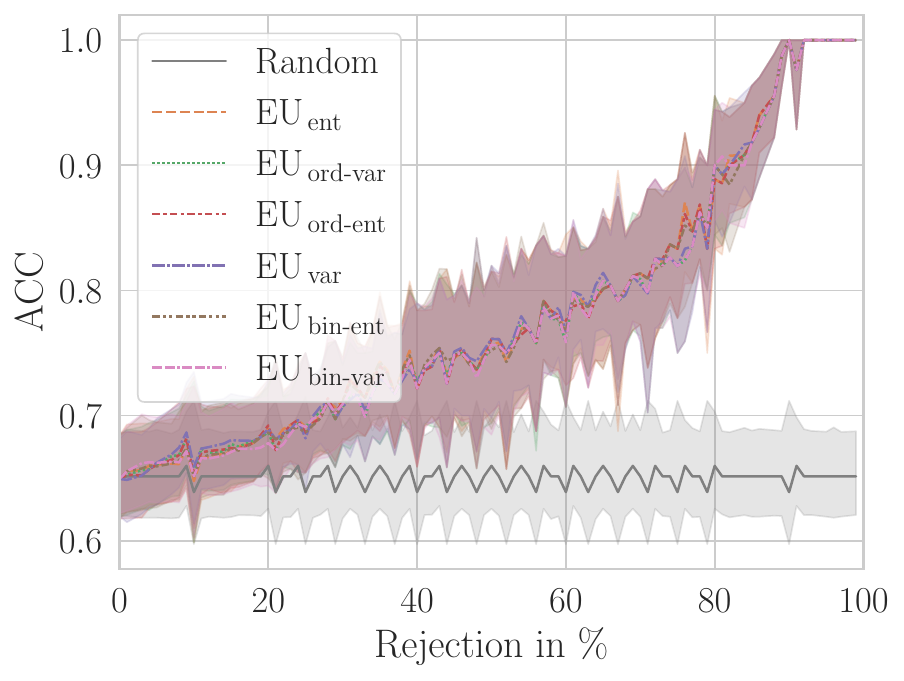}
  \subcaption{Eucalyptus EU}
\end{subfigure}  
\begin{subfigure}[t]{0.3\linewidth}
\centering
\includegraphics[width=\linewidth]{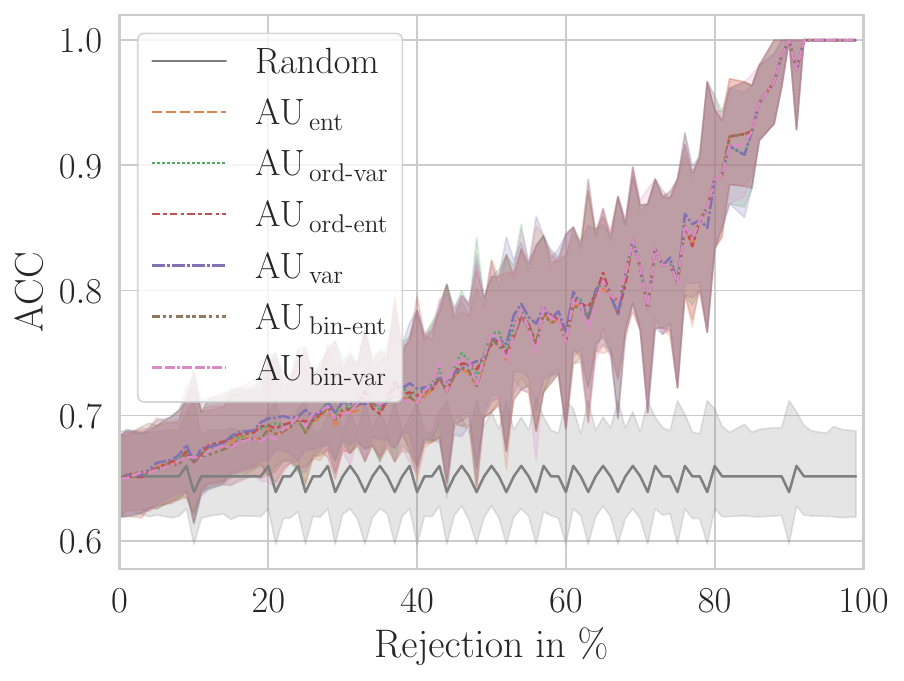}
\subcaption{Eucalyptus AU}
\end{subfigure}   
\begin{subfigure}[t]{0.3\linewidth}
  \centering
  \includegraphics[width=\linewidth]{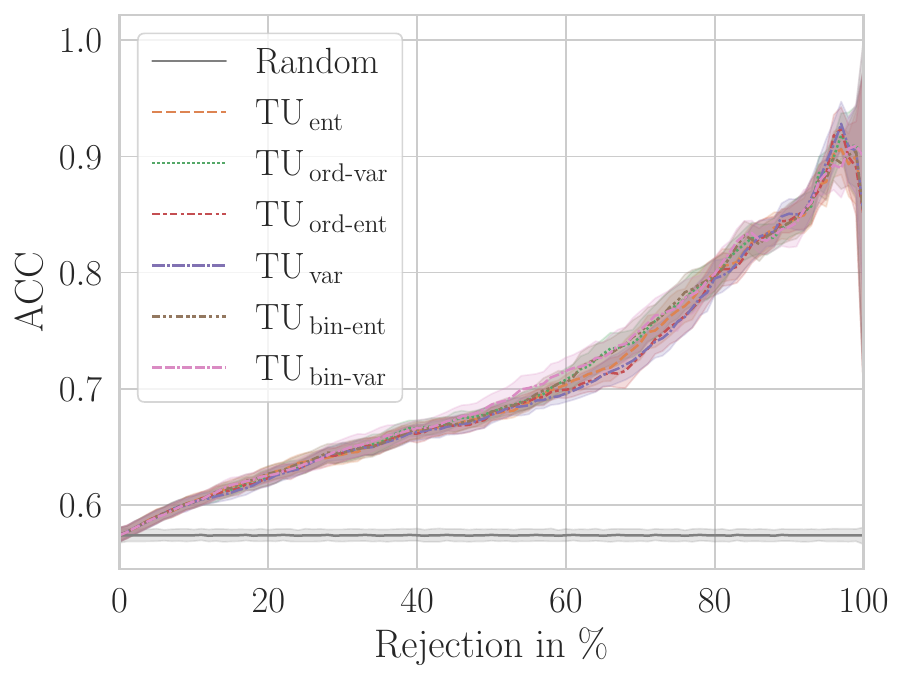}
  \subcaption{Abalone TU}
\end{subfigure} 
\begin{subfigure}[t]{0.3\linewidth}
  \centering
  \includegraphics[width=\linewidth]{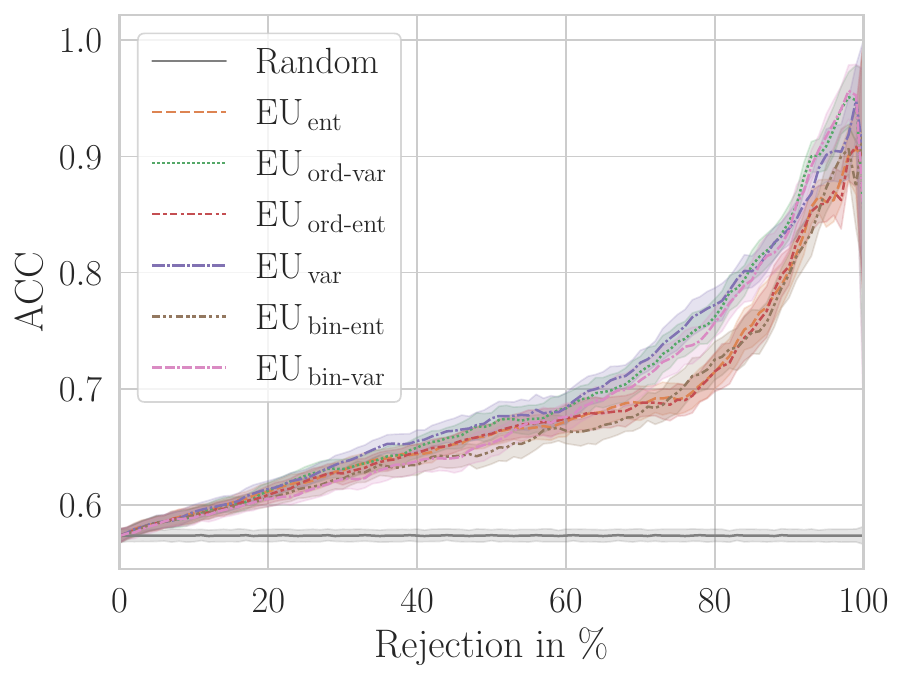}
  \subcaption{Abalone EU}
\end{subfigure}  
\begin{subfigure}[t]{0.3\linewidth}
\centering
\includegraphics[width=\linewidth]{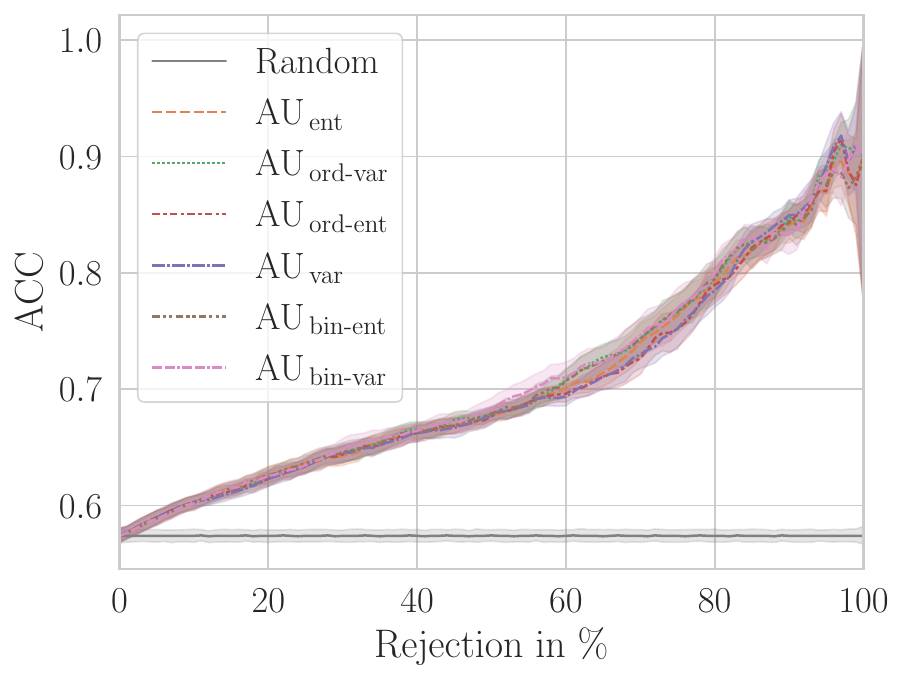}
\subcaption{Abalone AU}
\end{subfigure}
        \caption{Accuracy rejection curves for different datasets, uncertainty types (TU, EU, AU), and measures using an ensemble of MLPs for approximate Bayesian infernce.}
        \label{fig:acc_reject_mlp}
\end{figure}

\begin{figure}[!htbp]
  \centering
    \begin{subfigure}[t]{0.3\linewidth}
      \centering
      \includegraphics[width=\linewidth]{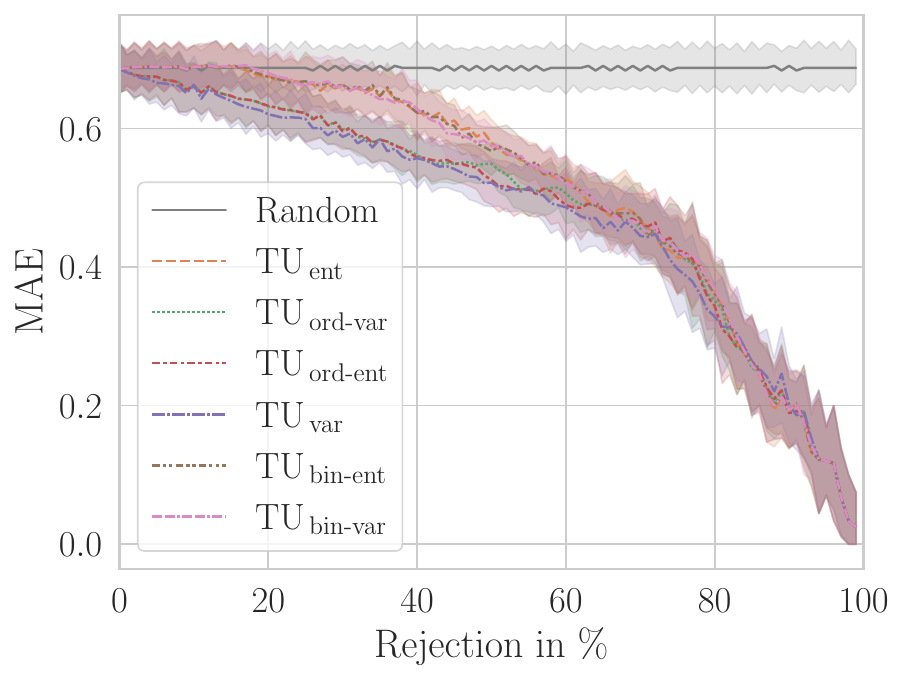}
      \subcaption{CMC TU }
  \end{subfigure} 
    \begin{subfigure}[t]{0.3\linewidth}
      \centering
      \includegraphics[width=\linewidth]{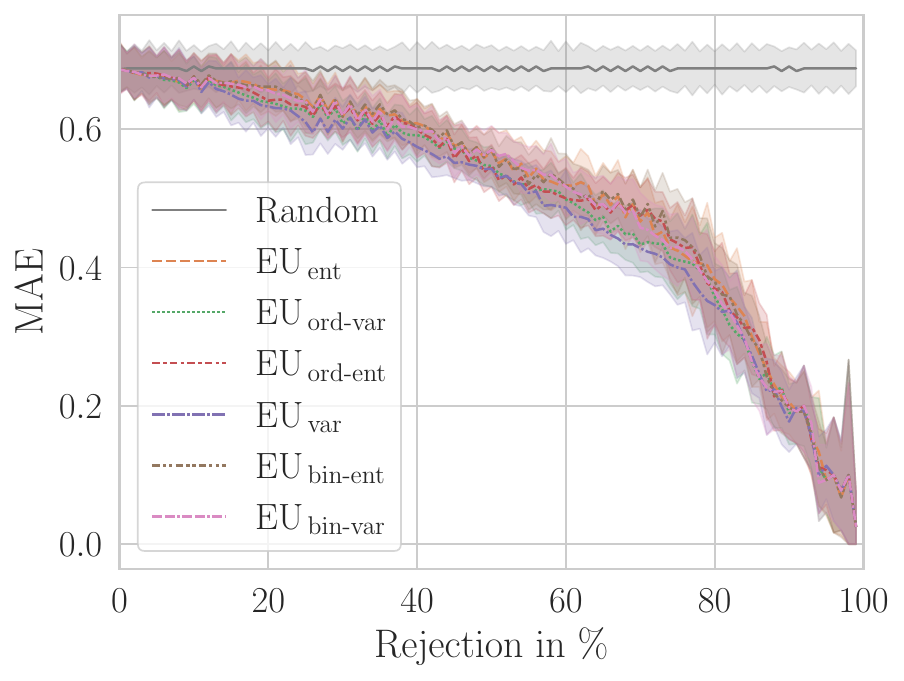}
      \subcaption{CMC EU}
  \end{subfigure}  
  \begin{subfigure}[t]{0.3\linewidth}
  \centering
  \includegraphics[width=\linewidth]{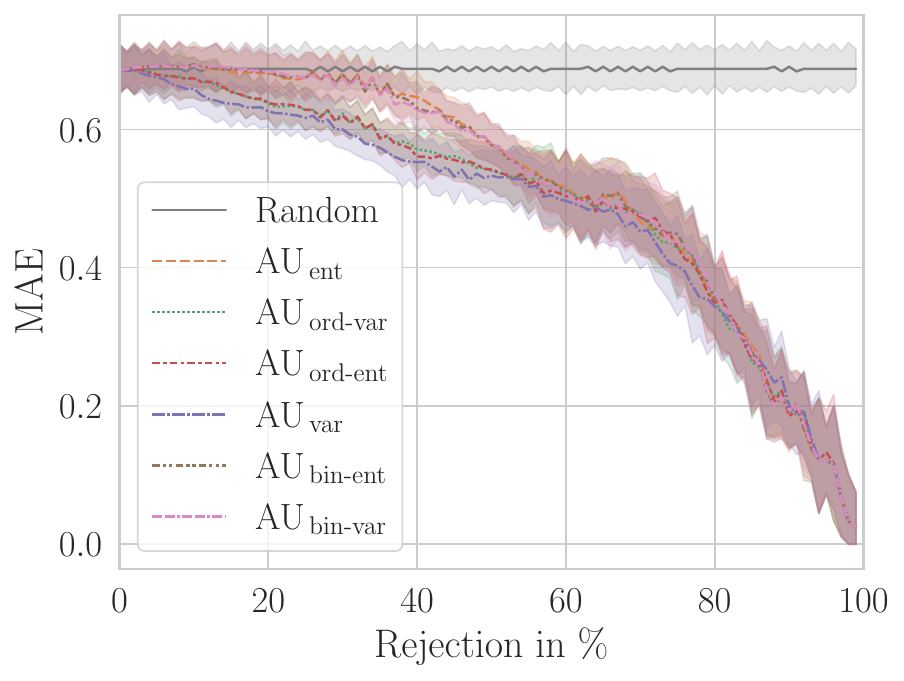}
  \subcaption{CMC AU}
  \end{subfigure}  
    \begin{subfigure}[t]{0.3\linewidth}
      \centering
      \includegraphics[width=\linewidth]{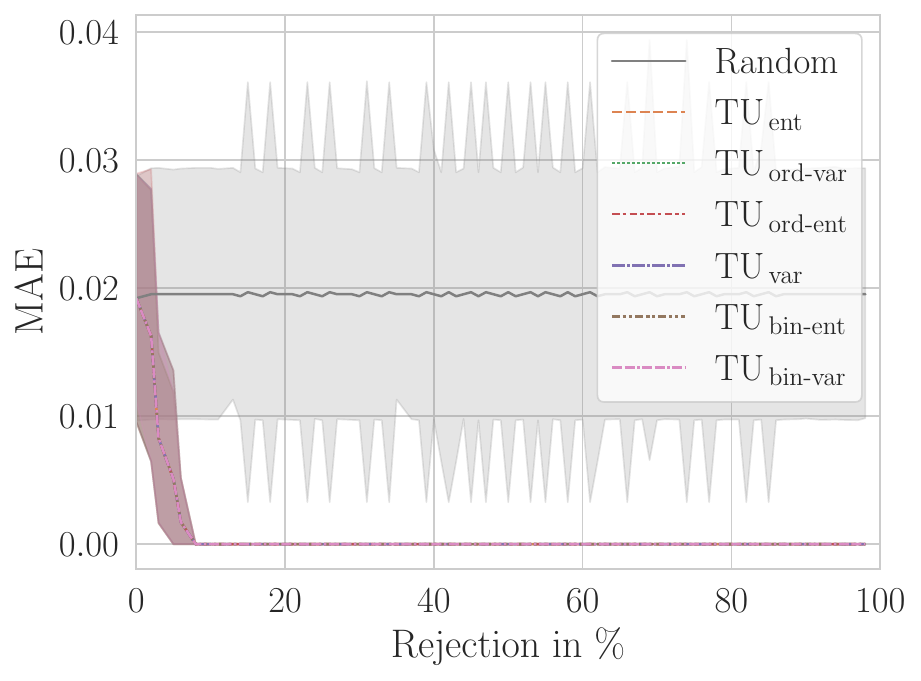}
      \subcaption{Balance Scale TU }
  \end{subfigure} 
    \begin{subfigure}[t]{0.3\linewidth}
      \centering
      \includegraphics[width=\linewidth]{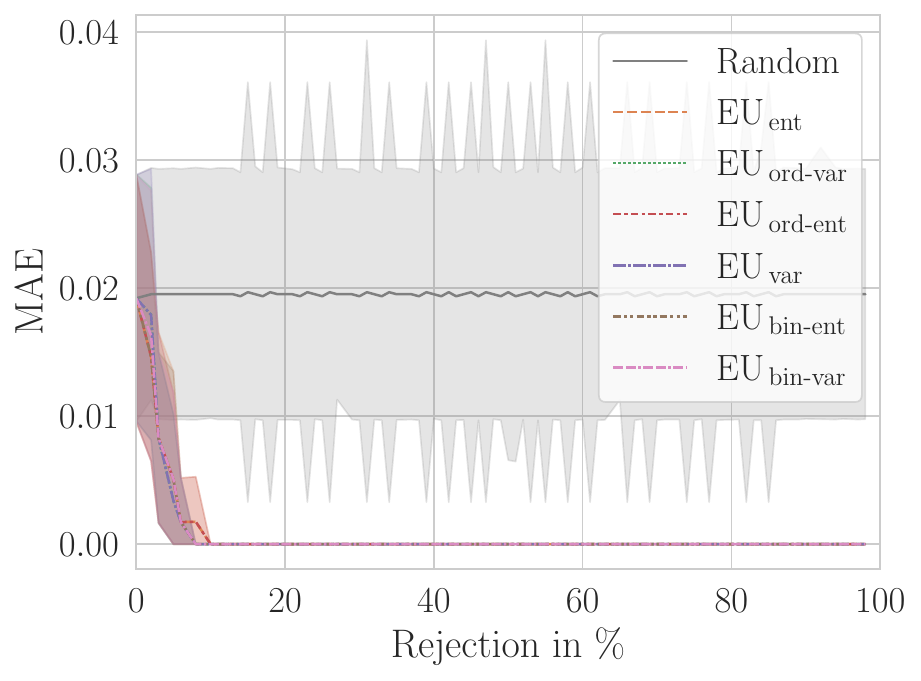}
      \subcaption{Balance Scale EU}
  \end{subfigure}  
  \begin{subfigure}[t]{0.3\linewidth}
   \centering
   \includegraphics[width=\linewidth]{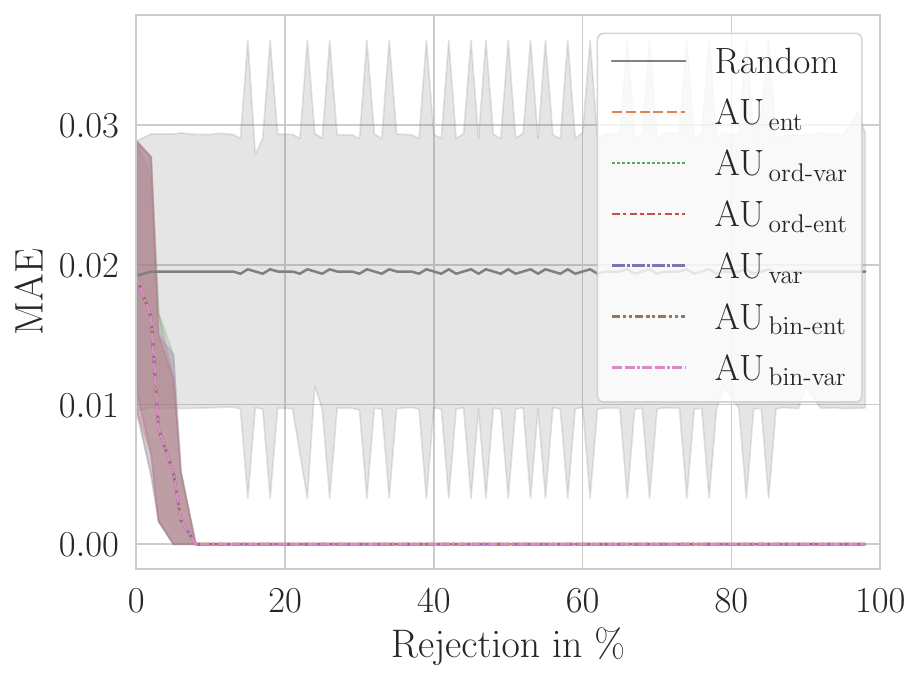}
   \subcaption{Balance Scale AU}
\end{subfigure}    
\begin{subfigure}[t]{0.3\linewidth}
  \centering
  \includegraphics[width=\linewidth]{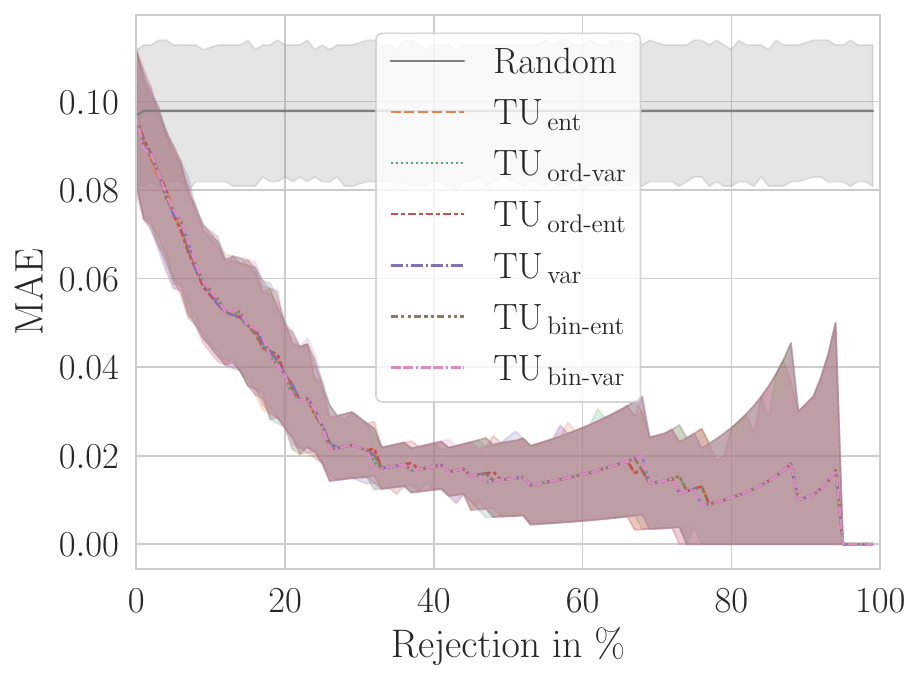}
  \subcaption{Stocks Domain TU }
\end{subfigure} 
\begin{subfigure}[t]{0.3\linewidth}
  \centering
  \includegraphics[width=\linewidth]{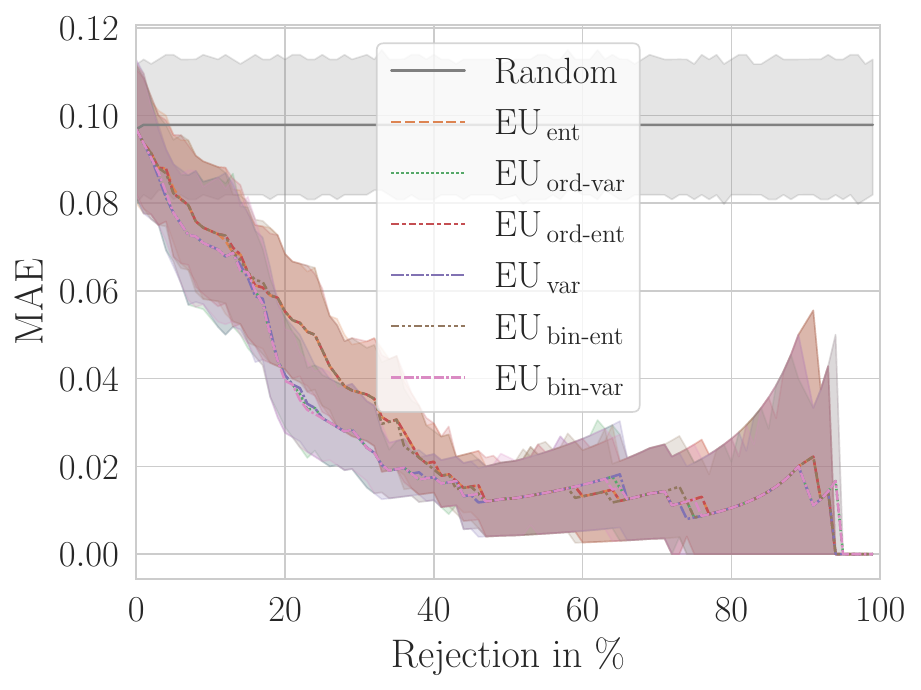}
  \subcaption{Stocks Domain EU}
\end{subfigure}  
\begin{subfigure}[t]{0.3\linewidth}
\centering
\includegraphics[width=\linewidth]{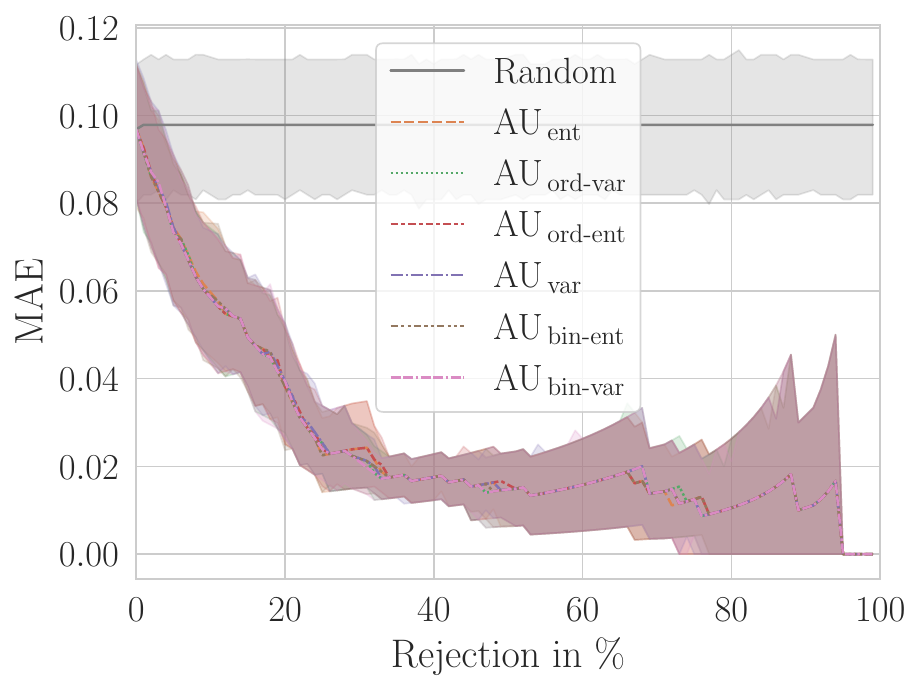}
\subcaption{Stocks Domain AU}
\end{subfigure}     
\begin{subfigure}[t]{0.3\linewidth}
  \centering
  \includegraphics[width=\linewidth]{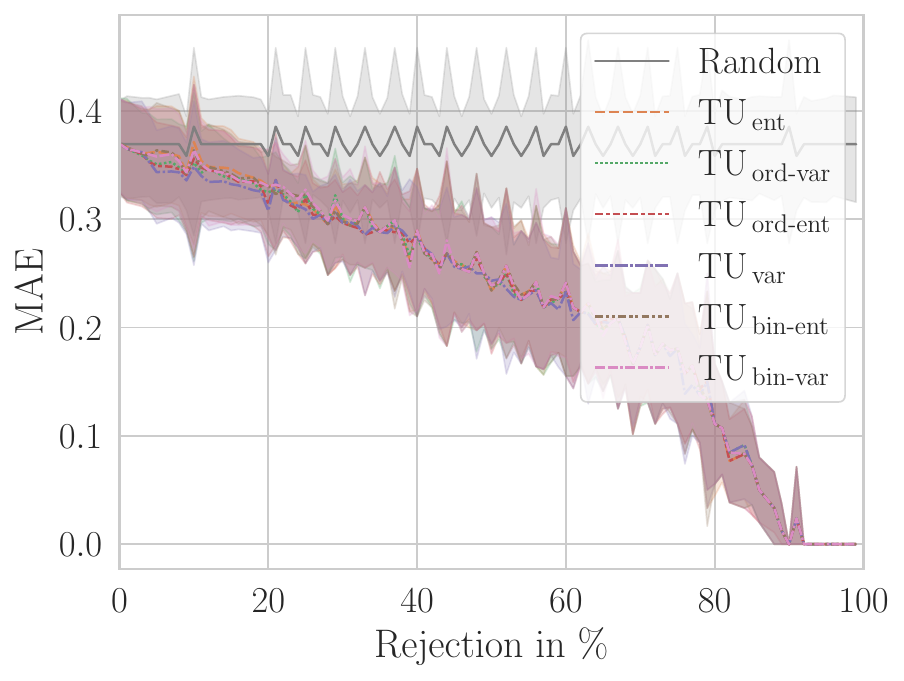}
  \subcaption{Eucalyptus TU }
\end{subfigure} 
\begin{subfigure}[t]{0.3\linewidth}
  \centering
  \includegraphics[width=\linewidth]{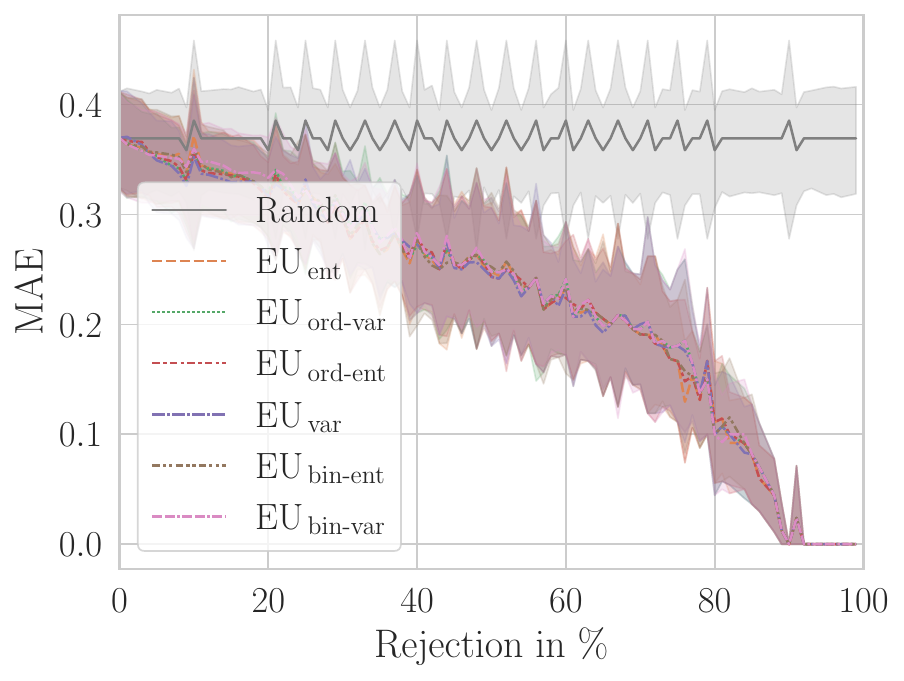}
  \subcaption{Eucalyptus EU}
\end{subfigure}  
\begin{subfigure}[t]{0.3\linewidth}
\centering
\includegraphics[width=\linewidth]{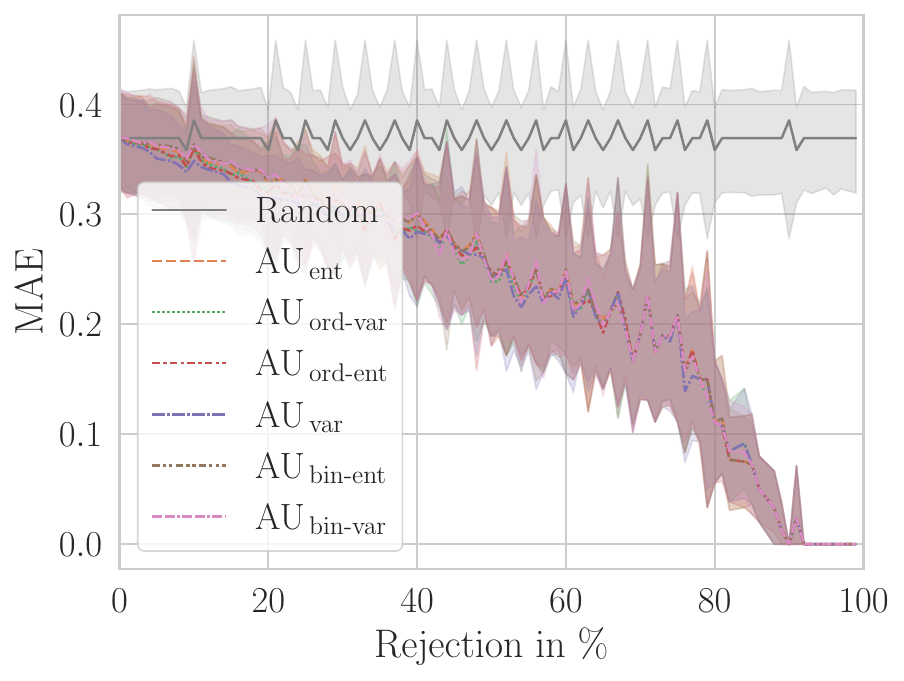}
\subcaption{Eucalyptus AU}
\end{subfigure}   
\begin{subfigure}[t]{0.3\linewidth}
  \centering
  \includegraphics[width=\linewidth]{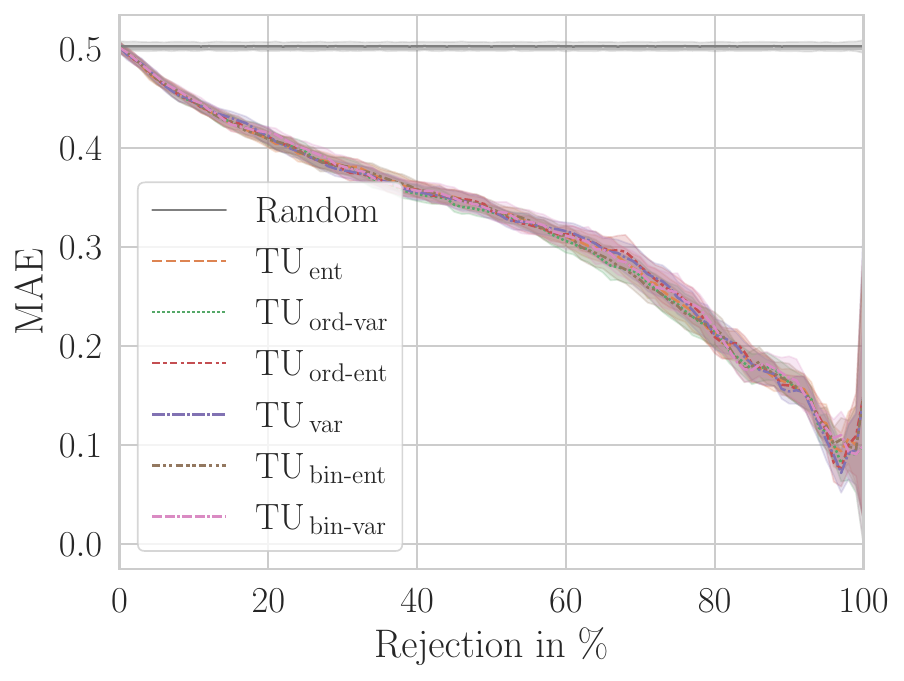}
  \subcaption{Abalone TU }
\end{subfigure} 
\begin{subfigure}[t]{0.3\linewidth}
  \centering
  \includegraphics[width=\linewidth]{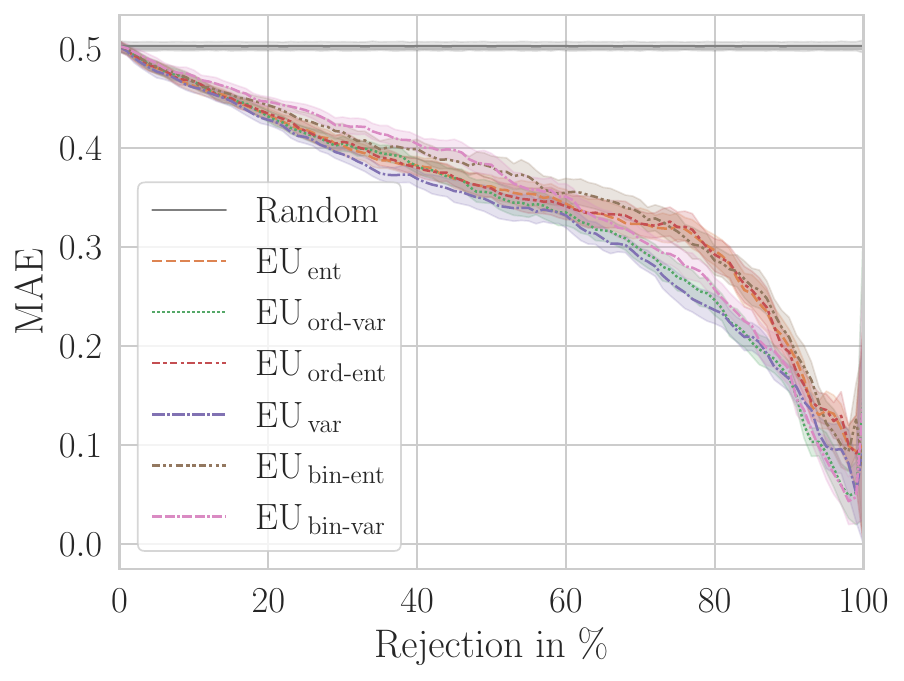}
  \subcaption{Abalone EU}
\end{subfigure}  
\begin{subfigure}[t]{0.3\linewidth}
\centering
\includegraphics[width=\linewidth]{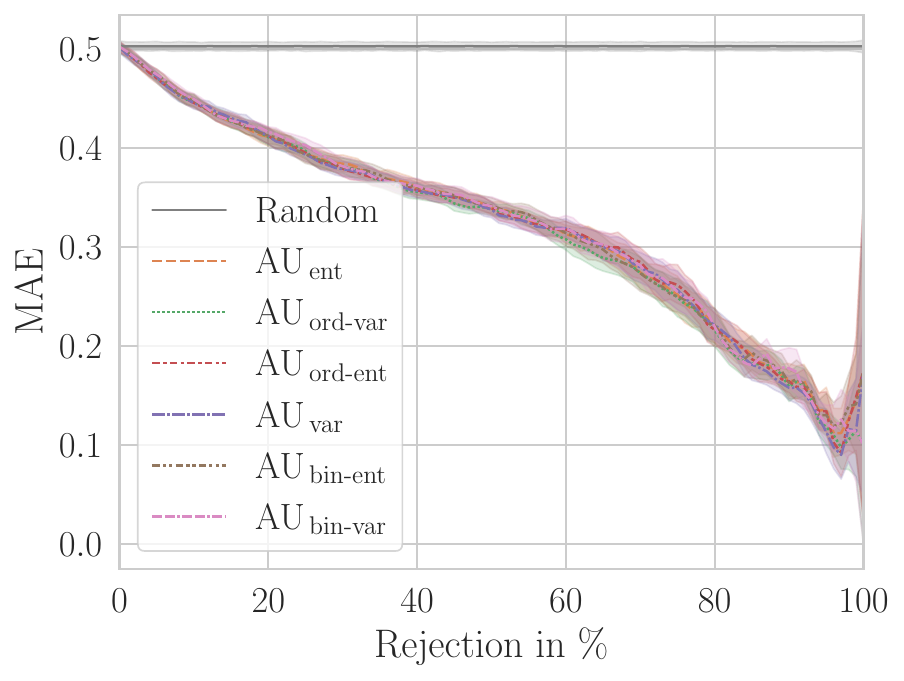}
\subcaption{Abalone AU}
\end{subfigure}  
        \caption{Mean absolute error rejection curves for different datasets, uncertainty types (TU, EU, AU), and measures using an ensemble of MLPs for approximate Bayesian inference.}
        \label{fig:mae_reject_mlp}
\end{figure}

 Again, we first display accuracy rejection curves to visually validate the quality of uncertainty quantification for some datasets (cf.\ Subsection \ref{subsec:rejection_curves}). Just like for the ensemble of GBTs, the investigated uncertainty methods all appear to be able to quantify epistemic and aleatoric uncertainty, as rejection curves monotonically increase in the case of ACC or decrease in the case of MAE, respectively (cf.\ Figures \ref{fig:acc_reject_mlp} and \ref{fig:mae_reject_mlp}).

 \subsection{Prediction-Rejection-Ratios (PRRs)}

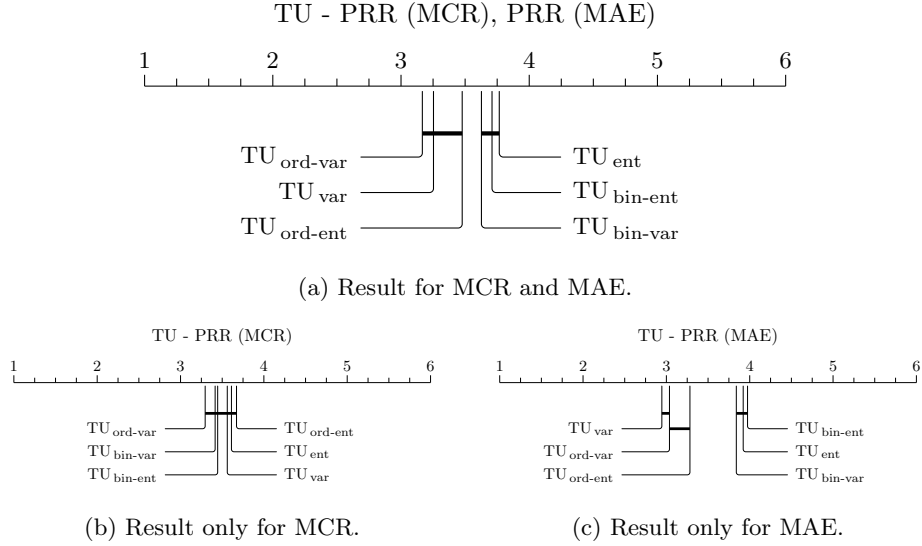
\begin{figure}[!htbp]
  \centering 
  \caption{CD diagrams for Epistemic Uncertainty (EU) using an ensemble of MLPs.}
  \label{fig:cd_epistemic_mlp}
  \begin{subfigure}[t]{\linewidth}
    \centering 
\begin{tikzpicture}[
  treatment line/.style={rounded corners=1.5pt, line cap=round, shorten >=1pt},
  treatment label/.style={font=\small},
  group line/.style={ultra thick},
]

\begin{axis}[
  clip={false},
  axis x line={center},
  axis y line={none},
  axis line style={-},
  xmin={1},
  ymax={0},
  scale only axis={true},
  width={\axisdefaultwidth},
  ticklabel style={anchor=south, yshift=1.3*\pgfkeysvalueof{/pgfplots/major tick length}, font=\small},
  every tick/.style={draw=black},
  major tick style={yshift=.5*\pgfkeysvalueof{/pgfplots/major tick length}},
  minor tick style={yshift=.5*\pgfkeysvalueof{/pgfplots/minor tick length}},
  title style={yshift=\baselineskip},
  xmax={6},
  ymin={-4.5},
  height={5\baselineskip},
  xtick={1,2,3,4,5,6},
  minor x tick num={3},
  title={EU - PRR (MCR), PRR (MAE)},
]

\draw[treatment line] ([yshift=-2pt] axis cs:2.9141304347826087, 0) |- (axis cs:2.4141304347826087, -2.0)
  node[treatment label, anchor=east] {$\text{EU}_{\,\text{var}}$};
\draw[treatment line] ([yshift=-2pt] axis cs:3.1021739130434782, 0) |- (axis cs:2.4141304347826087, -3.0)
  node[treatment label, anchor=east] {$\text{EU}_{\,\text{ord-var}}$};
\draw[treatment line] ([yshift=-2pt] axis cs:3.517391304347826, 0) |- (axis cs:2.4141304347826087, -4.0)
  node[treatment label, anchor=east] {$\text{EU}_{\,\text{ord-ent}}$};
\draw[treatment line] ([yshift=-2pt] axis cs:3.622826086956522, 0) |- (axis cs:4.5097826086956525, -4.0)
  node[treatment label, anchor=west] {$\text{EU}_{\,\text{ent}}$};
\draw[treatment line] ([yshift=-2pt] axis cs:3.833695652173913, 0) |- (axis cs:4.5097826086956525, -3.0)
  node[treatment label, anchor=west] {$\text{EU}_{\,\text{bin-ent}}$};
\draw[treatment line] ([yshift=-2pt] axis cs:4.0097826086956525, 0) |- (axis cs:4.5097826086956525, -2.0)
  node[treatment label, anchor=west] {$\text{EU}_{\,\text{bin-var}}$};
\draw[group line] (axis cs:3.517391304347826, -2.0) -- (axis cs:3.833695652173913, -2.0);

\end{axis}
\end{tikzpicture}
\captionsetup{justification=centering}
\subcaption{Result for MCR and MAE.}
\label{fig:cd_epistemic_mcr_mae_mlp}
\end{subfigure}
\begin{subfigure}[t]{0.48\linewidth}
  \centering
\begin{tikzpicture}[scale=0.65,
  treatment line/.style={rounded corners=1.5pt, line cap=round, shorten >=1pt},
  treatment label/.style={font=\small},
  group line/.style={ultra thick},
]

\begin{axis}[
  clip={false},
  axis x line={center},
  axis y line={none},
  axis line style={-},
  xmin={1},
  ymax={0},
  scale only axis={true},
  width={\axisdefaultwidth},
  ticklabel style={anchor=south, yshift=1.3*\pgfkeysvalueof{/pgfplots/major tick length}, font=\small},
  every tick/.style={draw=black},
  major tick style={yshift=.5*\pgfkeysvalueof{/pgfplots/major tick length}},
  minor tick style={yshift=.5*\pgfkeysvalueof{/pgfplots/minor tick length}},
  title style={yshift=\baselineskip},
  xmax={6},
  ymin={-4.5},
  height={5\baselineskip},
  xtick={1,2,3,4,5,6},
  minor x tick num={3},
  title={EU - PRR (MCR)},
]

\draw[treatment line] ([yshift=-2pt] axis cs:3.017391304347826, 0) |- (axis cs:2.517391304347826, -2.0)
  node[treatment label, anchor=east] {$\text{EU}_{\,\text{var}}$};
\draw[treatment line] ([yshift=-2pt] axis cs:3.108695652173913, 0) |- (axis cs:2.517391304347826, -3.0)
  node[treatment label, anchor=east] {$\text{EU}_{\,\text{ord-var}}$};
\draw[treatment line] ([yshift=-2pt] axis cs:3.5478260869565217, 0) |- (axis cs:2.517391304347826, -4.0)
  node[treatment label, anchor=east] {$\text{EU}_{\,\text{ent}}$};
\draw[treatment line] ([yshift=-2pt] axis cs:3.658695652173913, 0) |- (axis cs:4.495652173913044, -4.0)
  node[treatment label, anchor=west] {$\text{EU}_{\,\text{ord-ent}}$};
\draw[treatment line] ([yshift=-2pt] axis cs:3.6717391304347826, 0) |- (axis cs:4.495652173913044, -3.0)
  node[treatment label, anchor=west] {$\text{EU}_{\,\text{bin-ent}}$};
\draw[treatment line] ([yshift=-2pt] axis cs:3.9956521739130433, 0) |- (axis cs:4.495652173913044, -2.0)
  node[treatment label, anchor=west] {$\text{EU}_{\,\text{bin-var}}$};
\draw[group line] (axis cs:3.017391304347826, -1.3333333333333333) -- (axis cs:3.108695652173913, -1.3333333333333333);
\draw[group line] (axis cs:3.658695652173913, -1.3333333333333333) -- (axis cs:3.9956521739130433, -1.3333333333333333);
\draw[group line] (axis cs:3.5478260869565217, -2.0) -- (axis cs:3.6717391304347826, -2.0);

\end{axis}
\end{tikzpicture}
\captionsetup{justification=centering}
\subcaption{Result only for MCR.}
\label{fig:cd_epistemic_mcr_mlp}
\end{subfigure}
\begin{subfigure}[t]{0.48\linewidth}
  \centering
\begin{tikzpicture}[scale=0.65,
  treatment line/.style={rounded corners=1.5pt, line cap=round, shorten >=1pt},
  treatment label/.style={font=\small},
  group line/.style={ultra thick},
]

\begin{axis}[
  clip={false},
  axis x line={center},
  axis y line={none},
  axis line style={-},
  xmin={1},
  ymax={0},
  scale only axis={true},
  width={\axisdefaultwidth},
  ticklabel style={anchor=south, yshift=1.3*\pgfkeysvalueof{/pgfplots/major tick length}, font=\small},
  every tick/.style={draw=black},
  major tick style={yshift=.5*\pgfkeysvalueof{/pgfplots/major tick length}},
  minor tick style={yshift=.5*\pgfkeysvalueof{/pgfplots/minor tick length}},
  title style={yshift=\baselineskip},
  xmax={6},
  ymin={-4.5},
  height={5\baselineskip},
  xtick={1,2,3,4,5,6},
  minor x tick num={3},
  title={EU - PRR (MAE)},
]

\draw[treatment line] ([yshift=-2pt] axis cs:2.8108695652173914, 0) |- (axis cs:2.3108695652173914, -2.0)
  node[treatment label, anchor=east] {$\text{EU}_{\,\text{var}}$};
\draw[treatment line] ([yshift=-2pt] axis cs:3.0956521739130434, 0) |- (axis cs:2.3108695652173914, -3.0)
  node[treatment label, anchor=east] {$\text{EU}_{\,\text{ord-var}}$};
\draw[treatment line] ([yshift=-2pt] axis cs:3.376086956521739, 0) |- (axis cs:2.3108695652173914, -4.0)
  node[treatment label, anchor=east] {$\text{EU}_{\,\text{ord-ent}}$};
\draw[treatment line] ([yshift=-2pt] axis cs:3.6978260869565216, 0) |- (axis cs:4.523913043478261, -4.0)
  node[treatment label, anchor=west] {$\text{EU}_{\,\text{ent}}$};
\draw[treatment line] ([yshift=-2pt] axis cs:3.9956521739130433, 0) |- (axis cs:4.523913043478261, -3.0)
  node[treatment label, anchor=west] {$\text{EU}_{\,\text{bin-ent}}$};
\draw[treatment line] ([yshift=-2pt] axis cs:4.023913043478261, 0) |- (axis cs:4.523913043478261, -2.0)
  node[treatment label, anchor=west] {$\text{EU}_{\,\text{bin-var}}$};
\draw[group line] (axis cs:3.9956521739130433, -1.3333333333333333) -- (axis cs:4.023913043478261, -1.3333333333333333);
\draw[group line] (axis cs:3.6978260869565216, -2.0) -- (axis cs:3.9956521739130433, -2.0);
\draw[group line] (axis cs:3.0956521739130434, -2.0) -- (axis cs:3.376086956521739, -2.0);

\end{axis}
\end{tikzpicture}
\captionsetup{justification=centering}
\subcaption{Result only for MAE.}
\label{fig:cd_epistemic_mae_mlp}
         \end{subfigure}       
\end{figure}

 The following CD diagrams (cf.\ Figures \ref{fig:cd_total_mlp}, \ref{fig:cd_epistemic_mlp}, \ref{fig:cd_aleatoric_mlp}, and \ref{fig:cd_all_mlp}) show the ranks of the different uncertainty measures according to the obtained PRRs grouped by uncertainty type. The rankings resemble those for GBTs (cf.\ Subsection \ref{subsec:prrs}), with measures taking distance into account outperforming nominal measures. The results are even more significant than those for GBTs and underpin the superiority of the OCS decomposition method as well as variance in uncertainty quantification for ordinal classification and the disentanglement of aleatoric and epistemic uncertainty in this context.


\begin{figure}[!htbp]
  \centering 
  \caption{Critical difference (CD) diagrams\protect\footnotemark  for the evaluated Total Uncertainty (TU) measures and performance metrics based on a Friedman test followed by a post-hoc Holm-adjusted Wilcoxon signed-rank test \citep{demsar2006statistical,benavoli2016should} using an ensemble of MLPs for approximate Bayesian inference. Groups of uncertainty measures that are
  not significantly different (at p = 0.05) are connected.}
  \label{fig:cd_total_mlp}
  \begin{subfigure}[t]{\linewidth}
    \centering 
\begin{tikzpicture}[
  treatment line/.style={rounded corners=1.5pt, line cap=round, shorten >=1pt},
  treatment label/.style={font=\small},
  group line/.style={ultra thick},
]

\begin{axis}[
  clip={false},
  axis x line={center},
  axis y line={none},
  axis line style={-},
  xmin={1},
  ymax={0},
  scale only axis={true},
  width={\axisdefaultwidth},
  ticklabel style={anchor=south, yshift=1.3*\pgfkeysvalueof{/pgfplots/major tick length}, font=\small},
  every tick/.style={draw=black},
  major tick style={yshift=.5*\pgfkeysvalueof{/pgfplots/major tick length}},
  minor tick style={yshift=.5*\pgfkeysvalueof{/pgfplots/minor tick length}},
  title style={yshift=\baselineskip},
  xmax={6},
  ymin={-4.5},
  height={5\baselineskip},
  xtick={1,2,3,4,5,6},
  minor x tick num={3},
  title={TU - PRR (MCR), PRR (MAE)},
]

\draw[treatment line] ([yshift=-2pt] axis cs:3.166304347826087, 0) |- (axis cs:2.666304347826087, -2.0)
  node[treatment label, anchor=east] {$\text{TU}_{\,\text{ord-var}}$};
\draw[treatment line] ([yshift=-2pt] axis cs:3.253260869565217, 0) |- (axis cs:2.666304347826087, -3.0)
  node[treatment label, anchor=east] {$\text{TU}_{\,\text{var}}$};
\draw[treatment line] ([yshift=-2pt] axis cs:3.4771739130434782, 0) |- (axis cs:2.666304347826087, -4.0)
  node[treatment label, anchor=east] {$\text{TU}_{\,\text{ord-ent}}$};
\draw[treatment line] ([yshift=-2pt] axis cs:3.627173913043478, 0) |- (axis cs:4.266304347826087, -4.0)
  node[treatment label, anchor=west] {$\text{TU}_{\,\text{bin-var}}$};
\draw[treatment line] ([yshift=-2pt] axis cs:3.7097826086956522, 0) |- (axis cs:4.266304347826087, -3.0)
  node[treatment label, anchor=west] {$\text{TU}_{\,\text{bin-ent}}$};
\draw[treatment line] ([yshift=-2pt] axis cs:3.766304347826087, 0) |- (axis cs:4.266304347826087, -2.0)
  node[treatment label, anchor=west] {$\text{TU}_{\,\text{ent}}$};
\draw[group line] (axis cs:3.627173913043478, -1.3333333333333333) -- (axis cs:3.766304347826087, -1.3333333333333333);
\draw[group line] (axis cs:3.166304347826087, -1.3333333333333333) -- (axis cs:3.4771739130434782, -1.3333333333333333);

\end{axis}
\end{tikzpicture}
\captionsetup{justification=centering}
\subcaption{Result for MCR and MAE.}
\label{fig:cd_total_mcr_mae_mlp}
\end{subfigure}
\begin{subfigure}[t]{0.48\linewidth}
  \centering
\begin{tikzpicture}[scale=0.65,
  treatment line/.style={rounded corners=1.5pt, line cap=round, shorten >=1pt},
  treatment label/.style={font=\small},
  group line/.style={ultra thick},
]

\begin{axis}[
  clip={false},
  axis x line={center},
  axis y line={none},
  axis line style={-},
  xmin={1},
  ymax={0},
  scale only axis={true},
  width={\axisdefaultwidth},
  ticklabel style={anchor=south, yshift=1.3*\pgfkeysvalueof{/pgfplots/major tick length}, font=\small},
  every tick/.style={draw=black},
  major tick style={yshift=.5*\pgfkeysvalueof{/pgfplots/major tick length}},
  minor tick style={yshift=.5*\pgfkeysvalueof{/pgfplots/minor tick length}},
  title style={yshift=\baselineskip},
  xmax={6},
  ymin={-4.5},
  height={5\baselineskip},
  xtick={1,2,3,4,5,6},
  minor x tick num={3},
  title={TU - PRR (MCR)},
]

\draw[treatment line] ([yshift=-2pt] axis cs:3.2956521739130435, 0) |- (axis cs:2.7956521739130435, -2.0)
  node[treatment label, anchor=east] {$\text{TU}_{\,\text{ord-var}}$};
\draw[treatment line] ([yshift=-2pt] axis cs:3.4152173913043478, 0) |- (axis cs:2.7956521739130435, -3.0)
  node[treatment label, anchor=east] {$\text{TU}_{\,\text{bin-var}}$};
\draw[treatment line] ([yshift=-2pt] axis cs:3.4456521739130435, 0) |- (axis cs:2.7956521739130435, -4.0)
  node[treatment label, anchor=east] {$\text{TU}_{\,\text{bin-ent}}$};
\draw[treatment line] ([yshift=-2pt] axis cs:3.5608695652173914, 0) |- (axis cs:4.171739130434783, -4.0)
  node[treatment label, anchor=west] {$\text{TU}_{\,\text{var}}$};
\draw[treatment line] ([yshift=-2pt] axis cs:3.610869565217391, 0) |- (axis cs:4.171739130434783, -3.0)
  node[treatment label, anchor=west] {$\text{TU}_{\,\text{ent}}$};
\draw[treatment line] ([yshift=-2pt] axis cs:3.6717391304347826, 0) |- (axis cs:4.171739130434783, -2.0)
  node[treatment label, anchor=west] {$\text{TU}_{\,\text{ord-ent}}$};
\draw[group line] (axis cs:3.2956521739130435, -1.3333333333333333) -- (axis cs:3.6717391304347826, -1.3333333333333333);

\end{axis}
\end{tikzpicture}
\captionsetup{justification=centering}
\subcaption{Result only for MCR.}
\label{fig:cd_total_mcr_mlp}
\end{subfigure}
\begin{subfigure}[t]{0.48\linewidth}
\centering
\begin{tikzpicture}[scale=0.65,
  treatment line/.style={rounded corners=1.5pt, line cap=round, shorten >=1pt},
  treatment label/.style={font=\small},
  group line/.style={ultra thick},
]

\begin{axis}[
  clip={false},
  axis x line={center},
  axis y line={none},
  axis line style={-},
  xmin={1},
  ymax={0},
  scale only axis={true},
  width={\axisdefaultwidth},
  ticklabel style={anchor=south, yshift=1.3*\pgfkeysvalueof{/pgfplots/major tick length}, font=\small},
  every tick/.style={draw=black},
  major tick style={yshift=.5*\pgfkeysvalueof{/pgfplots/major tick length}},
  minor tick style={yshift=.5*\pgfkeysvalueof{/pgfplots/minor tick length}},
  title style={yshift=\baselineskip},
  xmax={6},
  ymin={-4.5},
  height={5\baselineskip},
  xtick={1,2,3,4,5,6},
  minor x tick num={3},
  title={TU - PRR (MAE)},
]

\draw[treatment line] ([yshift=-2pt] axis cs:2.9456521739130435, 0) |- (axis cs:2.4456521739130435, -2.0)
  node[treatment label, anchor=east] {$\text{TU}_{\,\text{var}}$};
\draw[treatment line] ([yshift=-2pt] axis cs:3.0369565217391306, 0) |- (axis cs:2.4456521739130435, -3.0)
  node[treatment label, anchor=east] {$\text{TU}_{\,\text{ord-var}}$};
\draw[treatment line] ([yshift=-2pt] axis cs:3.282608695652174, 0) |- (axis cs:2.4456521739130435, -4.0)
  node[treatment label, anchor=east] {$\text{TU}_{\,\text{ord-ent}}$};
\draw[treatment line] ([yshift=-2pt] axis cs:3.8391304347826085, 0) |- (axis cs:4.473913043478261, -4.0)
  node[treatment label, anchor=west] {$\text{TU}_{\,\text{bin-var}}$};
\draw[treatment line] ([yshift=-2pt] axis cs:3.9217391304347826, 0) |- (axis cs:4.473913043478261, -3.0)
  node[treatment label, anchor=west] {$\text{TU}_{\,\text{ent}}$};
\draw[treatment line] ([yshift=-2pt] axis cs:3.973913043478261, 0) |- (axis cs:4.473913043478261, -2.0)
  node[treatment label, anchor=west] {$\text{TU}_{\,\text{bin-ent}}$};
\draw[group line] (axis cs:3.8391304347826085, -1.3333333333333333) -- (axis cs:3.973913043478261, -1.3333333333333333);
\draw[group line] (axis cs:2.9456521739130435, -1.3333333333333333) -- (axis cs:3.0369565217391306, -1.3333333333333333);
\draw[group line] (axis cs:3.0369565217391306, -2.0) -- (axis cs:3.282608695652174, -2.0);

\end{axis}
\end{tikzpicture}
\captionsetup{justification=centering}
             \subcaption{Result only for MAE.}
             \label{fig:cd_total_mae_mlp}
         \end{subfigure}     

\end{figure} 

\footnotetext{\url{https://github.com/mirkobunse/critdd}}


\begin{figure}[!htbp]
  \centering 
  \caption{CD diagrams for Aleatoric Uncertainty (AU) using an ensemble of MLPs.}
  \label{fig:cd_aleatoric_mlp}
  \begin{subfigure}[t]{\linewidth}
    \centering 
\begin{tikzpicture}[
  treatment line/.style={rounded corners=1.5pt, line cap=round, shorten >=1pt},
  treatment label/.style={font=\small},
  group line/.style={ultra thick},
]

\begin{axis}[
  clip={false},
  axis x line={center},
  axis y line={none},
  axis line style={-},
  xmin={1},
  ymax={0},
  scale only axis={true},
  width={\axisdefaultwidth},
  ticklabel style={anchor=south, yshift=1.3*\pgfkeysvalueof{/pgfplots/major tick length}, font=\small},
  every tick/.style={draw=black},
  major tick style={yshift=.5*\pgfkeysvalueof{/pgfplots/major tick length}},
  minor tick style={yshift=.5*\pgfkeysvalueof{/pgfplots/minor tick length}},
  title style={yshift=\baselineskip},
  xmax={6},
  ymin={-4.5},
  height={5\baselineskip},
  xtick={1,2,3,4,5,6},
  minor x tick num={3},
  title={AU - PRR (MCR), PRR (MAE)},
]

\draw[treatment line] ([yshift=-2pt] axis cs:2.878260869565217, 0) |- (axis cs:2.378260869565217, -2.0)
  node[treatment label, anchor=east] {$\text{AU}_{\,\text{ord-var}}$};
\draw[treatment line] ([yshift=-2pt] axis cs:3.0445652173913045, 0) |- (axis cs:2.378260869565217, -3.0)
  node[treatment label, anchor=east] {$\text{AU}_{\,\text{var}}$};
\draw[treatment line] ([yshift=-2pt] axis cs:3.3554347826086954, 0) |- (axis cs:2.378260869565217, -4.0)
  node[treatment label, anchor=east] {$\text{AU}_{\,\text{ord-ent}}$};
\draw[treatment line] ([yshift=-2pt] axis cs:3.726086956521739, 0) |- (axis cs:4.5097826086956525, -4.0)
  node[treatment label, anchor=west] {$\text{AU}_{\,\text{bin-ent}}$};
\draw[treatment line] ([yshift=-2pt] axis cs:3.985869565217391, 0) |- (axis cs:4.5097826086956525, -3.0)
  node[treatment label, anchor=west] {$\text{AU}_{\,\text{bin-var}}$};
\draw[treatment line] ([yshift=-2pt] axis cs:4.0097826086956525, 0) |- (axis cs:4.5097826086956525, -2.0)
  node[treatment label, anchor=west] {$\text{AU}_{\,\text{ent}}$};
\draw[group line] (axis cs:3.726086956521739, -1.3333333333333333) -- (axis cs:4.0097826086956525, -1.3333333333333333);
\draw[group line] (axis cs:2.878260869565217, -1.3333333333333333) -- (axis cs:3.0445652173913045, -1.3333333333333333);

\end{axis}
\end{tikzpicture}
\captionsetup{justification=centering}
\subcaption{Result for MCR and MAE.}
\label{fig:cd_aleatoric_mcr_mae_mlp}
\end{subfigure}
\begin{subfigure}[t]{0.48\linewidth}
  \centering
\begin{tikzpicture}[scale=0.65,
  treatment line/.style={rounded corners=1.5pt, line cap=round, shorten >=1pt},
  treatment label/.style={font=\small},
  group line/.style={ultra thick},
]

\begin{axis}[
  clip={false},
  axis x line={center},
  axis y line={none},
  axis line style={-},
  xmin={1},
  ymax={0},
  scale only axis={true},
  width={\axisdefaultwidth},
  ticklabel style={anchor=south, yshift=1.3*\pgfkeysvalueof{/pgfplots/major tick length}, font=\small},
  every tick/.style={draw=black},
  major tick style={yshift=.5*\pgfkeysvalueof{/pgfplots/major tick length}},
  minor tick style={yshift=.5*\pgfkeysvalueof{/pgfplots/minor tick length}},
  title style={yshift=\baselineskip},
  xmax={6},
  ymin={-4.5},
  height={5\baselineskip},
  xtick={1,2,3,4,5,6},
  minor x tick num={3},
  title={AU - PRR (MCR)},
]

\draw[treatment line] ([yshift=-2pt] axis cs:2.965217391304348, 0) |- (axis cs:2.465217391304348, -2.0)
  node[treatment label, anchor=east] {$\text{AU}_{\,\text{ord-var}}$};
\draw[treatment line] ([yshift=-2pt] axis cs:3.3391304347826085, 0) |- (axis cs:2.465217391304348, -3.0)
  node[treatment label, anchor=east] {$\text{AU}_{\,\text{var}}$};
\draw[treatment line] ([yshift=-2pt] axis cs:3.467391304347826, 0) |- (axis cs:2.465217391304348, -4.0)
  node[treatment label, anchor=east] {$\text{AU}_{\,\text{ord-ent}}$};
\draw[treatment line] ([yshift=-2pt] axis cs:3.5, 0) |- (axis cs:4.404347826086957, -4.0)
  node[treatment label, anchor=west] {$\text{AU}_{\,\text{bin-ent}}$};
\draw[treatment line] ([yshift=-2pt] axis cs:3.8239130434782607, 0) |- (axis cs:4.404347826086957, -3.0)
  node[treatment label, anchor=west] {$\text{AU}_{\,\text{bin-var}}$};
\draw[treatment line] ([yshift=-2pt] axis cs:3.9043478260869566, 0) |- (axis cs:4.404347826086957, -2.0)
  node[treatment label, anchor=west] {$\text{AU}_{\,\text{ent}}$};
\draw[group line] (axis cs:2.965217391304348, -1.3333333333333333) -- (axis cs:3.3391304347826085, -1.3333333333333333);
\draw[group line] (axis cs:3.3391304347826085, -1.5333333333333332) -- (axis cs:3.9043478260869566, -1.5333333333333332);

\end{axis}
\end{tikzpicture}
\captionsetup{justification=centering}
\subcaption{Result only for MCR.}
\label{fig:cd_aleatoric_mcr_mlp}
\end{subfigure}
\begin{subfigure}[t]{0.48\linewidth}
  \centering
\begin{tikzpicture}[scale=0.65,
  treatment line/.style={rounded corners=1.5pt, line cap=round, shorten >=1pt},
  treatment label/.style={font=\small},
  group line/.style={ultra thick},
]

\begin{axis}[
  clip={false},
  axis x line={center},
  axis y line={none},
  axis line style={-},
  xmin={1},
  ymax={0},
  scale only axis={true},
  width={\axisdefaultwidth},
  ticklabel style={anchor=south, yshift=1.3*\pgfkeysvalueof{/pgfplots/major tick length}, font=\small},
  every tick/.style={draw=black},
  major tick style={yshift=.5*\pgfkeysvalueof{/pgfplots/major tick length}},
  minor tick style={yshift=.5*\pgfkeysvalueof{/pgfplots/minor tick length}},
  title style={yshift=\baselineskip},
  xmax={6},
  ymin={-4.5},
  height={5\baselineskip},
  xtick={1,2,3,4,5,6},
  minor x tick num={3},
  title={AU - PRR (MAE)},
]

\draw[treatment line] ([yshift=-2pt] axis cs:2.75, 0) |- (axis cs:2.25, -2.0)
  node[treatment label, anchor=east] {$\text{AU}_{\,\text{var}}$};
\draw[treatment line] ([yshift=-2pt] axis cs:2.791304347826087, 0) |- (axis cs:2.25, -3.0)
  node[treatment label, anchor=east] {$\text{AU}_{\,\text{ord-var}}$};
\draw[treatment line] ([yshift=-2pt] axis cs:3.243478260869565, 0) |- (axis cs:2.25, -4.0)
  node[treatment label, anchor=east] {$\text{AU}_{\,\text{ord-ent}}$};
\draw[treatment line] ([yshift=-2pt] axis cs:3.9521739130434783, 0) |- (axis cs:4.647826086956521, -4.0)
  node[treatment label, anchor=west] {$\text{AU}_{\,\text{bin-ent}}$};
\draw[treatment line] ([yshift=-2pt] axis cs:4.1152173913043475, 0) |- (axis cs:4.647826086956521, -3.0)
  node[treatment label, anchor=west] {$\text{AU}_{\,\text{ent}}$};
\draw[treatment line] ([yshift=-2pt] axis cs:4.147826086956521, 0) |- (axis cs:4.647826086956521, -2.0)
  node[treatment label, anchor=west] {$\text{AU}_{\,\text{bin-var}}$};
\draw[group line] (axis cs:3.9521739130434783, -1.3333333333333333) -- (axis cs:4.147826086956521, -1.3333333333333333);

\end{axis}
\end{tikzpicture}

\captionsetup{justification=centering}
\subcaption{Result only for MAE.}
\label{fig:cd_aleatoric_mae_mlp}
         \end{subfigure}                  
\end{figure}


\begin{figure}[!htbp]
  \centering 
  \caption{CD diagrams for all uncertainty types (AU, EU and TU) using an ensemble of MLPs.}
  \label{fig:cd_all_mlp}
  \begin{subfigure}[t]{\linewidth}
    \centering 
\begin{tikzpicture}[
  treatment line/.style={rounded corners=1.5pt, line cap=round, shorten >=1pt},
  treatment label/.style={font=\small},
  group line/.style={ultra thick},
]

\begin{axis}[
  clip={false},
  axis x line={center},
  axis y line={none},
  axis line style={-},
  xmin={1},
  ymax={0},
  scale only axis={true},
  width={\axisdefaultwidth},
  ticklabel style={anchor=south, yshift=1.3*\pgfkeysvalueof{/pgfplots/major tick length}, font=\small},
  every tick/.style={draw=black},
  major tick style={yshift=.5*\pgfkeysvalueof{/pgfplots/major tick length}},
  minor tick style={yshift=.5*\pgfkeysvalueof{/pgfplots/minor tick length}},
  title style={yshift=\baselineskip},
  xmax={6},
  ymin={-4.5},
  height={5\baselineskip},
  xtick={1,2,3,4,5,6},
  minor x tick num={3},
  title={ALL - PRR (MCR), PRR (MAE)},
]

\draw[treatment line] ([yshift=-2pt] axis cs:3.0489130434782608, 0) |- (axis cs:2.5489130434782608, -2.0)
  node[treatment label, anchor=east] {ord-var};
\draw[treatment line] ([yshift=-2pt] axis cs:3.0706521739130435, 0) |- (axis cs:2.5489130434782608, -3.0)
  node[treatment label, anchor=east] {var};
\draw[treatment line] ([yshift=-2pt] axis cs:3.45, 0) |- (axis cs:2.5489130434782608, -4.0)
  node[treatment label, anchor=east] {ord-ent};
\draw[treatment line] ([yshift=-2pt] axis cs:3.756521739130435, 0) |- (axis cs:4.374275362318841, -4.0)
  node[treatment label, anchor=west] {bin-ent};
\draw[treatment line] ([yshift=-2pt] axis cs:3.79963768115942, 0) |- (axis cs:4.374275362318841, -3.0)
  node[treatment label, anchor=west] {ent};
\draw[treatment line] ([yshift=-2pt] axis cs:3.8742753623188406, 0) |- (axis cs:4.374275362318841, -2.0)
  node[treatment label, anchor=west] {bin-var};
\draw[group line] (axis cs:3.756521739130435, -1.3333333333333333) -- (axis cs:3.8742753623188406, -1.3333333333333333);

\end{axis}
\end{tikzpicture}
\captionsetup{justification=centering}
\subcaption{Result for MCR and MAE.}
\label{fig:cd_all_mcr_mae_mlp}
\end{subfigure}
\begin{subfigure}[t]{0.48\linewidth}
  \centering
\begin{tikzpicture}[scale=0.65,
  treatment line/.style={rounded corners=1.5pt, line cap=round, shorten >=1pt},
  treatment label/.style={font=\small},
  group line/.style={ultra thick},
]

\begin{axis}[
  clip={false},
  axis x line={center},
  axis y line={none},
  axis line style={-},
  xmin={1},
  ymax={0},
  scale only axis={true},
  width={\axisdefaultwidth},
  ticklabel style={anchor=south, yshift=1.3*\pgfkeysvalueof{/pgfplots/major tick length}, font=\small},
  every tick/.style={draw=black},
  major tick style={yshift=.5*\pgfkeysvalueof{/pgfplots/major tick length}},
  minor tick style={yshift=.5*\pgfkeysvalueof{/pgfplots/minor tick length}},
  title style={yshift=\baselineskip},
  xmax={6},
  ymin={-4.5},
  height={5\baselineskip},
  xtick={1,2,3,4,5,6},
  minor x tick num={3},
  title={ALL - PRR (MCR)},
]

\draw[treatment line] ([yshift=-2pt] axis cs:3.1231884057971016, 0) |- (axis cs:2.6231884057971016, -2.0)
  node[treatment label, anchor=east] {ord-var};
\draw[treatment line] ([yshift=-2pt] axis cs:3.3057971014492753, 0) |- (axis cs:2.6231884057971016, -3.0)
  node[treatment label, anchor=east] {var};
\draw[treatment line] ([yshift=-2pt] axis cs:3.5391304347826087, 0) |- (axis cs:2.6231884057971016, -4.0)
  node[treatment label, anchor=east] {bin-ent};
\draw[treatment line] ([yshift=-2pt] axis cs:3.5992753623188407, 0) |- (axis cs:4.244927536231884, -4.0)
  node[treatment label, anchor=west] {ord-ent};
\draw[treatment line] ([yshift=-2pt] axis cs:3.68768115942029, 0) |- (axis cs:4.244927536231884, -3.0)
  node[treatment label, anchor=west] {ent};
\draw[treatment line] ([yshift=-2pt] axis cs:3.744927536231884, 0) |- (axis cs:4.244927536231884, -2.0)
  node[treatment label, anchor=west] {bin-var};
\draw[group line] (axis cs:3.1231884057971016, -1.3333333333333333) -- (axis cs:3.3057971014492753, -1.3333333333333333);
\draw[group line] (axis cs:3.5992753623188407, -1.3333333333333333) -- (axis cs:3.744927536231884, -1.3333333333333333);
\draw[group line] (axis cs:3.5391304347826087, -2.0) -- (axis cs:3.68768115942029, -2.0);
\draw[group line] (axis cs:3.3057971014492753, -2.2) -- (axis cs:3.5391304347826087, -2.2);

\end{axis}
\end{tikzpicture}
\captionsetup{justification=centering}
\subcaption{Result only for MCR.}
\label{fig:cd_all_mcr_mlp}
\end{subfigure}
\begin{subfigure}[t]{0.48\linewidth}
  \centering
\begin{tikzpicture}[scale=0.65,
  treatment line/.style={rounded corners=1.5pt, line cap=round, shorten >=1pt},
  treatment label/.style={font=\small},
  group line/.style={ultra thick},
]

\begin{axis}[
  clip={false},
  axis x line={center},
  axis y line={none},
  axis line style={-},
  xmin={1},
  ymax={0},
  scale only axis={true},
  width={\axisdefaultwidth},
  ticklabel style={anchor=south, yshift=1.3*\pgfkeysvalueof{/pgfplots/major tick length}, font=\small},
  every tick/.style={draw=black},
  major tick style={yshift=.5*\pgfkeysvalueof{/pgfplots/major tick length}},
  minor tick style={yshift=.5*\pgfkeysvalueof{/pgfplots/minor tick length}},
  title style={yshift=\baselineskip},
  xmax={6},
  ymin={-4.5},
  height={5\baselineskip},
  xtick={1,2,3,4,5,6},
  minor x tick num={3},
  title={ALL - PRR (MAE)},
]

\draw[treatment line] ([yshift=-2pt] axis cs:2.8355072463768116, 0) |- (axis cs:2.3355072463768116, -2.0)
  node[treatment label, anchor=east] {var};
\draw[treatment line] ([yshift=-2pt] axis cs:2.9746376811594204, 0) |- (axis cs:2.3355072463768116, -3.0)
  node[treatment label, anchor=east] {ord-var};
\draw[treatment line] ([yshift=-2pt] axis cs:3.300724637681159, 0) |- (axis cs:2.3355072463768116, -4.0)
  node[treatment label, anchor=east] {ord-ent};
\draw[treatment line] ([yshift=-2pt] axis cs:3.911594202898551, 0) |- (axis cs:4.503623188405797, -4.0)
  node[treatment label, anchor=west] {ent};
\draw[treatment line] ([yshift=-2pt] axis cs:3.973913043478261, 0) |- (axis cs:4.503623188405797, -3.0)
  node[treatment label, anchor=west] {bin-ent};
\draw[treatment line] ([yshift=-2pt] axis cs:4.003623188405797, 0) |- (axis cs:4.503623188405797, -2.0)
  node[treatment label, anchor=west] {bin-var};
\draw[group line] (axis cs:3.973913043478261, -1.3333333333333333) -- (axis cs:4.003623188405797, -1.3333333333333333);
\draw[group line] (axis cs:3.911594202898551, -2.0) -- (axis cs:3.973913043478261, -2.0);

\end{axis}
\end{tikzpicture}

\captionsetup{justification=centering}
\subcaption{Result only for MAE.}
\label{fig:cd_all_mae_mlp}
         \end{subfigure}              
\end{figure}

\subsection{Prediction Rejection Ratios (PRR) - Detailed Results}

In this subsection, we display the detailed prediction rejection ratio results for the different tabular ordinal benchmark datasets, uncertainty types, and measures using an ensemble of MLPs for approximate Bayesian inference. Table \ref{tab:prr_acc_mlp} shows results for misclassification rate (MCR) and Table \ref{tab:prr_mae_mlp} for mean absolute error (MAE).

\tiny 

\begin{longtable}{llrrrrrr}
  \caption{PRRs (MCR) using an ensemble of MLPs.} \label{tab:prr_acc_mlp} \\
  \toprule
   &  & \multicolumn{6}{c}{\textbf{PRR (MCR)} ($\uparrow$) } \\
   \midrule
   & \textbf{Measure} & bin-ent ($\uparrow$) & bin-var ($\uparrow$) & ent ($\uparrow$) & ord-ent ($\uparrow$) & ord-var ($\uparrow$) & var ($\uparrow$) \\
   \midrule
   \textbf{Dataset} & \textbf{Type} & & &  &  &  &  \\
  \endfirsthead
  \toprule
  &  & \multicolumn{6}{c}{\textbf{PRR (MCR)} ($\uparrow$)} \\
  \midrule
  & \textbf{Measure} & bin-ent ($\uparrow$) & bin-var ($\uparrow$) & ent ($\uparrow$) & ord-ent ($\uparrow$) & ord-var ($\uparrow$) & var ($\uparrow$) \\
  \midrule
  \textbf{Dataset} & \textbf{Type} &  &  &  & & & \\
  \midrule
  \endhead
  \midrule
  \multicolumn{6}{r}{Continued on next page} \\
  \midrule
  \endfoot
  \bottomrule
  \endlastfoot
  \midrule
  \multirow[t]{3}{*}{Abalone} & AU & \textbf{0.34}\textpm0.06 & 0.334\textpm0.05 & 0.328\textpm0.05 & 0.323\textpm0.05 & 0.333\textpm0.05 & 0.321\textpm0.05 \\
  & EU & 0.25\textpm0.07 & 0.228\textpm0.07 & 0.251\textpm0.07 & 0.247\textpm0.07 & 0.28\textpm0.07 & \textbf{0.293}\textpm0.07 \\
  & TU & \textbf{0.344}\textpm0.06 & 0.338\textpm0.06 & 0.331\textpm0.05 & 0.325\textpm0.05 & 0.337\textpm0.05 & 0.322\textpm0.05 \\
 \midrule
 \multirow[t]{3}{*}{Auto MPG} & AU & 0.305\textpm0.17 & 0.307\textpm0.18 & 0.31\textpm0.18 & 0.335\textpm0.18 & 0.343\textpm0.17 & \textbf{0.352}\textpm0.17 \\
  & EU & 0.323\textpm0.13 & 0.284\textpm0.14 & 0.3\textpm0.16 & 0.313\textpm0.17 & 0.33\textpm0.14 & \textbf{0.335}\textpm0.16 \\
  & TU & 0.316\textpm0.18 & 0.319\textpm0.18 & 0.318\textpm0.19 & 0.344\textpm0.17 & 0.351\textpm0.17 & \textbf{0.355}\textpm0.17 \\
 \midrule
 \multirow[t]{3}{*}{Automobile} & AU & 0.664\textpm0.29 & 0.651\textpm0.31 & 0.656\textpm0.3 & 0.666\textpm0.28 & 0.682\textpm0.27 & \textbf{0.686}\textpm0.27 \\
  & EU & 0.694\textpm0.27 & 0.718\textpm0.25 & 0.721\textpm0.25 & 0.721\textpm0.25 & 0.716\textpm0.27 & \textbf{0.729}\textpm0.26 \\
  & TU & 0.681\textpm0.29 & 0.676\textpm0.29 & 0.675\textpm0.3 & 0.696\textpm0.28 & 0.697\textpm0.29 & \textbf{0.712}\textpm0.28 \\
 \midrule
 \multirow[t]{3}{*}{Balance Scale} & AU & \textbf{0.68}\textpm0.47 & \textbf{0.68}\textpm0.47 & \textbf{0.68}\textpm0.47 & \textbf{0.68}\textpm0.47 & \textbf{0.68}\textpm0.47 & \textbf{0.68}\textpm0.47 \\
  & EU & \textbf{0.68}\textpm0.47 & 0.679\textpm0.47 & 0.678\textpm0.47 & 0.678\textpm0.47 & 0.678\textpm0.47 & 0.678\textpm0.47 \\
  & TU & \textbf{0.68}\textpm0.47 & \textbf{0.68}\textpm0.47 & \textbf{0.68}\textpm0.47 & \textbf{0.68}\textpm0.47 & \textbf{0.68}\textpm0.47 & \textbf{0.68}\textpm0.47 \\
 \midrule
 \multirow[c]{3}{*}{\shortstack{Wisconsin \\Breast Cancer}} & AU & 0.213\textpm0.2 & 0.219\textpm0.19 & \textbf{0.221}\textpm0.2 & 0.201\textpm0.25 & 0.186\textpm0.26 & 0.206\textpm0.25 \\
  & EU & 0.2\textpm0.31 & 0.211\textpm0.31 & 0.222\textpm0.29 & \textbf{0.223}\textpm0.32 & 0.211\textpm0.32 & 0.207\textpm0.33 \\
  & TU & 0.228\textpm0.26 & 0.219\textpm0.25 & \textbf{0.229}\textpm0.24 & 0.206\textpm0.28 & 0.196\textpm0.31 & 0.208\textpm0.29 \\
 \midrule
 \multirow[t]{3}{*}{ERA} & AU & \textbf{0.161}\textpm0.06 & 0.157\textpm0.07 & 0.152\textpm0.07 & 0.088\textpm0.1 & 0.102\textpm0.1 & 0.027\textpm0.09 \\
  & EU & -0.014\textpm0.13 & -0.006\textpm0.1 & \textbf{0.006}\textpm0.1 & -0.016\textpm0.1 & -0.009\textpm0.11 & -0.04\textpm0.11 \\
  & TU & \textbf{0.161}\textpm0.06 & 0.154\textpm0.07 & 0.152\textpm0.07 & 0.088\textpm0.1 & 0.101\textpm0.1 & 0.026\textpm0.09 \\
 \midrule
 \multirow[t]{3}{*}{ESL} & AU & \textbf{0.238}\textpm0.23 & 0.228\textpm0.22 & 0.218\textpm0.21 & 0.206\textpm0.21 & 0.236\textpm0.22 & 0.205\textpm0.21 \\
  & EU & 0.274\textpm0.23 & 0.285\textpm0.22 & 0.272\textpm0.22 & 0.282\textpm0.22 & 0.286\textpm0.23 & \textbf{0.298}\textpm0.21 \\
  & TU & \textbf{0.242}\textpm0.23 & 0.238\textpm0.22 & 0.223\textpm0.21 & 0.214\textpm0.21 & 0.241\textpm0.22 & 0.209\textpm0.21 \\
 \midrule
 \multirow[t]{3}{*}{Eucalyptus} & AU & 0.325\textpm0.09 & 0.322\textpm0.09 & 0.322\textpm0.09 & 0.33\textpm0.09 & 0.334\textpm0.09 & \textbf{0.342}\textpm0.09 \\
  & EU & 0.329\textpm0.09 & 0.337\textpm0.09 & 0.34\textpm0.09 & 0.34\textpm0.09 & 0.334\textpm0.09 & \textbf{0.341}\textpm0.08 \\
  & TU & 0.337\textpm0.09 & 0.335\textpm0.09 & 0.333\textpm0.09 & 0.34\textpm0.08 & 0.339\textpm0.08 & \textbf{0.345}\textpm0.08 \\
 \midrule
 \multirow[t]{3}{*}{Heart (CLE)} & AU & 0.592\textpm0.15 & 0.595\textpm0.14 & 0.595\textpm0.14 & 0.603\textpm0.13 & 0.596\textpm0.13 & \textbf{0.604}\textpm0.12 \\
  & EU & 0.467\textpm0.22 & 0.453\textpm0.22 & 0.476\textpm0.21 & 0.477\textpm0.23 & 0.483\textpm0.24 & \textbf{0.512}\textpm0.22 \\
  & TU & 0.56\textpm0.18 & 0.56\textpm0.17 & 0.562\textpm0.16 & 0.57\textpm0.15 & \textbf{0.572}\textpm0.16 & 0.57\textpm0.16 \\
 \midrule
 \multirow[t]{3}{*}{Boston Housing} & AU & 0.418\textpm0.24 & 0.412\textpm0.24 & 0.414\textpm0.23 & 0.423\textpm0.24 & 0.431\textpm0.24 & \textbf{0.437}\textpm0.25 \\
  & EU & 0.433\textpm0.19 & 0.453\textpm0.18 & 0.455\textpm0.18 & \textbf{0.467}\textpm0.18 & 0.448\textpm0.2 & 0.458\textpm0.2 \\
  & TU & 0.424\textpm0.22 & 0.425\textpm0.22 & 0.431\textpm0.22 & 0.439\textpm0.23 & 0.434\textpm0.23 & \textbf{0.442}\textpm0.24 \\
 \midrule
 \multirow[t]{3}{*}{LEV} & AU & 0.177\textpm0.08 & 0.177\textpm0.1 & 0.175\textpm0.11 & 0.171\textpm0.11 & 0.182\textpm0.1 & \textbf{0.185}\textpm0.11 \\
  & EU & 0.21\textpm0.06 & 0.193\textpm0.09 & 0.183\textpm0.09 & 0.181\textpm0.09 & \textbf{0.211}\textpm0.05 & 0.198\textpm0.06 \\
  & TU & 0.179\textpm0.08 & 0.177\textpm0.1 & 0.174\textpm0.11 & 0.172\textpm0.11 & 0.183\textpm0.1 & \textbf{0.185}\textpm0.11 \\
 \midrule
 \multirow[t]{3}{*}{Machine CPU} & AU & 0.609\textpm0.2 & 0.632\textpm0.19 & 0.649\textpm0.18 & 0.692\textpm0.15 & 0.68\textpm0.14 & \textbf{0.704}\textpm0.15 \\
  & EU & 0.591\textpm0.13 & 0.6\textpm0.13 & \textbf{0.619}\textpm0.12 & 0.613\textpm0.12 & 0.61\textpm0.13 & 0.611\textpm0.13 \\
  & TU & 0.596\textpm0.19 & 0.632\textpm0.17 & 0.647\textpm0.16 & 0.67\textpm0.14 & 0.656\textpm0.14 & \textbf{0.7}\textpm0.15 \\
 \midrule
 \multirow[t]{3}{*}{New Thyroid} & AU & \textbf{0.479}\textpm0.51 & 0.474\textpm0.5 & \textbf{0.479}\textpm0.51 & \textbf{0.479}\textpm0.51 & \textbf{0.479}\textpm0.51 & \textbf{0.479}\textpm0.51 \\
  & EU & \textbf{0.484}\textpm0.51 & \textbf{0.484}\textpm0.51 & \textbf{0.484}\textpm0.51 & 0.479\textpm0.51 & 0.479\textpm0.51 & 0.474\textpm0.5 \\
  & TU & 0.479\textpm0.51 & 0.479\textpm0.51 & \textbf{0.484}\textpm0.51 & 0.474\textpm0.5 & 0.474\textpm0.5 & 0.474\textpm0.5 \\
 \midrule
 \multirow[t]{3}{*}{Pyrimidines} & AU & 0.181\textpm0.29 & \textbf{0.189}\textpm0.29 & 0.167\textpm0.29 & 0.149\textpm0.36 & 0.136\textpm0.38 & 0.05\textpm0.42 \\
  & EU & 0.029\textpm0.49 & 0.015\textpm0.47 & 0.026\textpm0.34 & -0.044\textpm0.48 & \textbf{0.039}\textpm0.49 & 0.03\textpm0.54 \\
  & TU & 0.156\textpm0.3 & \textbf{0.189}\textpm0.24 & 0.175\textpm0.22 & 0.089\textpm0.39 & 0.128\textpm0.4 & 0.081\textpm0.49 \\
 \midrule
 \multirow[t]{3}{*}{Red Wine} & AU & 0.316\textpm0.13 & 0.305\textpm0.12 & 0.303\textpm0.12 & 0.311\textpm0.12 & 0.325\textpm0.12 & \textbf{0.327}\textpm0.11 \\
  & EU & 0.423\textpm0.12 & 0.434\textpm0.12 & \textbf{0.44}\textpm0.12 & 0.439\textpm0.13 & 0.429\textpm0.12 & 0.423\textpm0.13 \\
  & TU & \textbf{0.385}\textpm0.13 & 0.384\textpm0.13 & 0.381\textpm0.13 & 0.375\textpm0.13 & 0.381\textpm0.13 & 0.375\textpm0.13 \\
 \midrule
 \multirow[t]{3}{*}{SWD} & AU & 0.184\textpm0.11 & 0.18\textpm0.11 & 0.181\textpm0.1 & 0.19\textpm0.11 & 0.188\textpm0.11 & \textbf{0.192}\textpm0.11 \\
  & EU & 0.129\textpm0.11 & 0.101\textpm0.08 & 0.119\textpm0.08 & 0.117\textpm0.07 & 0.149\textpm0.11 & \textbf{0.171}\textpm0.11 \\
  & TU & 0.184\textpm0.11 & 0.18\textpm0.11 & 0.181\textpm0.1 & 0.19\textpm0.11 & 0.19\textpm0.11 & \textbf{0.192}\textpm0.11 \\
 \midrule
 \multirow[t]{3}{*}{Stocks Domain} & AU & 0.716\textpm0.07 & \textbf{0.717}\textpm0.07 & 0.715\textpm0.07 & 0.714\textpm0.08 & 0.716\textpm0.07 & 0.714\textpm0.07 \\
  & EU & \textbf{0.676}\textpm0.07 & 0.625\textpm0.08 & 0.622\textpm0.08 & 0.623\textpm0.08 & \textbf{0.676}\textpm0.07 & \textbf{0.676}\textpm0.07 \\
  & TU & 0.716\textpm0.07 & 0.716\textpm0.07 & \textbf{0.717}\textpm0.07 & 0.716\textpm0.07 & 0.716\textpm0.07 & 0.716\textpm0.07 \\
 \midrule
 \multirow[t]{3}{*}{TAE} & AU & 0.208\textpm0.38 & 0.224\textpm0.39 & 0.231\textpm0.39 & 0.217\textpm0.45 & \textbf{0.237}\textpm0.44 & 0.214\textpm0.45 \\
  & EU & \textbf{0.36}\textpm0.3 & 0.27\textpm0.28 & 0.295\textpm0.29 & 0.266\textpm0.28 & 0.332\textpm0.32 & 0.313\textpm0.39 \\
  & TU & 0.252\textpm0.36 & 0.267\textpm0.39 & \textbf{0.283}\textpm0.4 & 0.249\textpm0.46 & 0.256\textpm0.46 & 0.226\textpm0.48 \\
 \midrule
 \multirow[t]{3}{*}{Triazines} & AU & 0.194\textpm0.27 & 0.193\textpm0.27 & 0.194\textpm0.26 & 0.226\textpm0.29 & 0.221\textpm0.28 & \textbf{0.24}\textpm0.29 \\
  & EU & 0.273\textpm0.26 & 0.294\textpm0.28 & \textbf{0.295}\textpm0.27 & 0.286\textpm0.28 & 0.274\textpm0.25 & 0.266\textpm0.26 \\
  & TU & 0.214\textpm0.25 & 0.221\textpm0.25 & 0.214\textpm0.24 & 0.236\textpm0.26 & 0.229\textpm0.26 & \textbf{0.254}\textpm0.27 \\
 \midrule
 \multirow[t]{3}{*}{White Wine} & AU & 0.25\textpm0.06 & 0.246\textpm0.06 & 0.246\textpm0.06 & 0.254\textpm0.06 & \textbf{0.263}\textpm0.06 & \textbf{0.263}\textpm0.06 \\
  & EU & 0.363\textpm0.03 & 0.353\textpm0.04 & 0.356\textpm0.04 & 0.356\textpm0.05 & \textbf{0.37}\textpm0.04 & 0.361\textpm0.05 \\
  & TU & \textbf{0.341}\textpm0.05 & 0.331\textpm0.05 & 0.323\textpm0.05 & 0.315\textpm0.05 & 0.33\textpm0.05 & 0.311\textpm0.05 \\
 \midrule
 \multirow[t]{3}{*}{CMC} & AU & 0.316\textpm0.06 & 0.316\textpm0.06 & \textbf{0.317}\textpm0.06 & 0.313\textpm0.07 & 0.313\textpm0.07 & 0.293\textpm0.07 \\
  & EU & \textbf{0.255}\textpm0.07 & 0.218\textpm0.07 & 0.229\textpm0.07 & 0.212\textpm0.07 & 0.252\textpm0.07 & 0.246\textpm0.06 \\
  & TU & 0.328\textpm0.06 & 0.33\textpm0.05 & \textbf{0.333}\textpm0.06 & 0.308\textpm0.07 & 0.309\textpm0.06 & 0.284\textpm0.07 \\
 \midrule
 \multirow[t]{3}{*}{Grub Damage} & AU & 0.152\textpm0.28 & 0.159\textpm0.27 & 0.161\textpm0.26 & 0.182\textpm0.27 & 0.157\textpm0.27 & \textbf{0.2}\textpm0.26 \\
  & EU & 0.152\textpm0.27 & 0.14\textpm0.29 & 0.141\textpm0.31 & 0.167\textpm0.33 & \textbf{0.181}\textpm0.31 & 0.162\textpm0.31 \\
  & TU & 0.201\textpm0.26 & 0.212\textpm0.26 & \textbf{0.219}\textpm0.26 & 0.209\textpm0.28 & 0.195\textpm0.29 & 0.21\textpm0.28 \\
 \midrule
 \multirow[t]{3}{*}{Obesity} & AU & 0.858\textpm0.06 & 0.853\textpm0.06 & 0.854\textpm0.06 & 0.862\textpm0.06 & 0.864\textpm0.06 & \textbf{0.87}\textpm0.05 \\
  & EU & 0.887\textpm0.05 & 0.891\textpm0.05 & 0.892\textpm0.05 & \textbf{0.894}\textpm0.04 & 0.89\textpm0.05 & 0.891\textpm0.05 \\
  & TU & 0.881\textpm0.06 & 0.882\textpm0.06 & 0.882\textpm0.06 & 0.886\textpm0.06 & 0.884\textpm0.06 & \textbf{0.887}\textpm0.05 \\
 \end{longtable}

\begin{longtable}{llrrrrrr}
  \caption{PRRs (MAE) using an ensemble of MLPs.} \label{tab:prr_mae_mlp} \\
  \toprule
   &  & \multicolumn{6}{c}{\textbf{PRR (MAE)} ($\uparrow$) } \\
   \midrule
   & \textbf{Measure} & bin-ent ($\uparrow$) & bin-var ($\uparrow$) & ent ($\uparrow$) & ord-ent ($\uparrow$) & ord-var ($\uparrow$) & var ($\uparrow$) \\
   \midrule
   \textbf{Dataset} & \textbf{Type} & & &  &  &  &  \\
  \endfirsthead
  \toprule
  &  & \multicolumn{6}{c}{\textbf{PRR (MAE)} ($\uparrow$)} \\
  \midrule
  & \textbf{Measure} & bin-ent ($\uparrow$) & bin-var ($\uparrow$) & ent ($\uparrow$) & ord-ent ($\uparrow$) & ord-var ($\uparrow$) & var ($\uparrow$) \\
  \midrule
  \textbf{Dataset} & \textbf{Type} &  &  &  & & & \\
  \midrule
  \endhead
  \midrule
  \multicolumn{6}{r}{Continued on next page} \\
  \midrule
  \endfoot
  \bottomrule
  \endlastfoot
  \midrule
  \multirow[t]{3}{*}{Abalone} & AU & 0.382\textpm0.05 & 0.387\textpm0.04 & 0.387\textpm0.04 & 0.389\textpm0.04 & \textbf{0.394}\textpm0.04 & 0.39\textpm0.04 \\
  & EU & 0.288\textpm0.05 & 0.288\textpm0.05 & 0.317\textpm0.05 & 0.315\textpm0.05 & 0.333\textpm0.05 & \textbf{0.354}\textpm0.04 \\
  & TU & 0.389\textpm0.05 & 0.395\textpm0.05 & 0.393\textpm0.04 & 0.393\textpm0.04 & \textbf{0.398}\textpm0.04 & 0.393\textpm0.04 \\
 \midrule
 \multirow[t]{3}{*}{Auto MPG} & AU & 0.285\textpm0.14 & 0.294\textpm0.15 & 0.302\textpm0.15 & 0.342\textpm0.15 & 0.353\textpm0.15 & \textbf{0.365}\textpm0.16 \\
  & EU & 0.304\textpm0.16 & 0.281\textpm0.16 & 0.298\textpm0.18 & 0.313\textpm0.18 & 0.316\textpm0.16 & \textbf{0.325}\textpm0.16 \\
  & TU & 0.307\textpm0.16 & 0.314\textpm0.16 & 0.317\textpm0.17 & 0.359\textpm0.17 & 0.367\textpm0.16 & \textbf{0.373}\textpm0.17 \\
 \midrule
 \multirow[t]{3}{*}{Automobile} & AU & 0.657\textpm0.24 & 0.652\textpm0.25 & 0.654\textpm0.24 & 0.688\textpm0.23 & 0.693\textpm0.22 & \textbf{0.715}\textpm0.23 \\
  & EU & 0.692\textpm0.22 & 0.715\textpm0.2 & 0.72\textpm0.19 & 0.729\textpm0.21 & 0.72\textpm0.21 & \textbf{0.742}\textpm0.21 \\
  & TU & 0.672\textpm0.23 & 0.667\textpm0.23 & 0.669\textpm0.24 & 0.711\textpm0.22 & 0.711\textpm0.23 & \textbf{0.737}\textpm0.23 \\
 \midrule
 \multirow[t]{3}{*}{Balance Scale} & AU & \textbf{0.68}\textpm0.47 & \textbf{0.68}\textpm0.47 & \textbf{0.68}\textpm0.47 & \textbf{0.68}\textpm0.47 & \textbf{0.68}\textpm0.47 & \textbf{0.68}\textpm0.47 \\
  & EU & \textbf{0.68}\textpm0.47 & 0.679\textpm0.47 & 0.678\textpm0.47 & 0.678\textpm0.47 & 0.678\textpm0.47 & 0.678\textpm0.47 \\
  & TU & \textbf{0.68}\textpm0.47 & \textbf{0.68}\textpm0.47 & \textbf{0.68}\textpm0.47 & \textbf{0.68}\textpm0.47 & \textbf{0.68}\textpm0.47 & \textbf{0.68}\textpm0.47 \\
 \midrule
 \multirow[c]{3}{*}{\shortstack{Wisconsin \\Breast Cancer}} & AU & 0.119\textpm0.17 & 0.139\textpm0.16 & 0.141\textpm0.16 & 0.182\textpm0.17 & 0.164\textpm0.17 & \textbf{0.189}\textpm0.17 \\
  & EU & 0.069\textpm0.23 & 0.08\textpm0.24 & 0.067\textpm0.21 & 0.108\textpm0.23 & 0.112\textpm0.24 & \textbf{0.131}\textpm0.23 \\
  & TU & 0.064\textpm0.18 & 0.07\textpm0.19 & 0.086\textpm0.17 & 0.135\textpm0.19 & 0.115\textpm0.22 & \textbf{0.154}\textpm0.2 \\
 \midrule
 \multirow[t]{3}{*}{ERA} & AU & -0.002\textpm0.09 & -0.018\textpm0.09 & -0.027\textpm0.08 & -0.018\textpm0.13 & \textbf{0.002}\textpm0.13 & -0.022\textpm0.14 \\
  & EU & 0.014\textpm0.09 & 0.017\textpm0.12 & \textbf{0.021}\textpm0.13 & 0.007\textpm0.12 & 0.013\textpm0.08 & -0.013\textpm0.07 \\
  & TU & -0.003\textpm0.09 & -0.019\textpm0.08 & -0.025\textpm0.08 & -0.017\textpm0.13 & \textbf{0.002}\textpm0.13 & -0.022\textpm0.14 \\
 \midrule
 \multirow[t]{3}{*}{ESL} & AU & \textbf{0.229}\textpm0.25 & 0.218\textpm0.24 & 0.208\textpm0.23 & 0.195\textpm0.23 & 0.225\textpm0.24 & 0.192\textpm0.23 \\
  & EU & 0.266\textpm0.24 & \textbf{0.286}\textpm0.22 & 0.273\textpm0.23 & 0.28\textpm0.23 & 0.275\textpm0.25 & \textbf{0.286}\textpm0.23 \\
  & TU & \textbf{0.231}\textpm0.25 & 0.229\textpm0.24 & 0.214\textpm0.24 & 0.204\textpm0.23 & 0.23\textpm0.24 & 0.196\textpm0.23 \\
 \midrule
 \multirow[t]{3}{*}{Eucalyptus} & AU & 0.321\textpm0.1 & 0.317\textpm0.1 & 0.318\textpm0.1 & 0.334\textpm0.1 & 0.338\textpm0.11 & \textbf{0.351}\textpm0.1 \\
  & EU & 0.339\textpm0.11 & 0.351\textpm0.11 & 0.354\textpm0.11 & 0.356\textpm0.11 & 0.349\textpm0.11 & \textbf{0.357}\textpm0.1 \\
  & TU & 0.339\textpm0.11 & 0.339\textpm0.11 & 0.338\textpm0.1 & 0.351\textpm0.1 & 0.348\textpm0.1 & \textbf{0.36}\textpm0.1 \\
 \midrule
 \multirow[t]{3}{*}{Heart (CLE)} & AU & 0.536\textpm0.16 & 0.538\textpm0.15 & 0.538\textpm0.16 & 0.56\textpm0.12 & 0.555\textpm0.12 & \textbf{0.581}\textpm0.09 \\
  & EU & 0.36\textpm0.2 & 0.353\textpm0.2 & 0.376\textpm0.19 & 0.411\textpm0.18 & 0.403\textpm0.18 & \textbf{0.443}\textpm0.15 \\
  & TU & 0.483\textpm0.18 & 0.491\textpm0.17 & 0.496\textpm0.16 & 0.529\textpm0.11 & 0.519\textpm0.13 & \textbf{0.544}\textpm0.1 \\
 \midrule
 \multirow[t]{3}{*}{Boston Housing} & AU & 0.388\textpm0.26 & 0.382\textpm0.26 & 0.385\textpm0.26 & 0.398\textpm0.26 & 0.406\textpm0.26 & \textbf{0.414}\textpm0.27 \\
  & EU & 0.411\textpm0.22 & 0.434\textpm0.21 & 0.436\textpm0.21 & \textbf{0.451}\textpm0.22 & 0.426\textpm0.23 & 0.437\textpm0.23 \\
  & TU & 0.397\textpm0.25 & 0.399\textpm0.24 & 0.405\textpm0.25 & 0.414\textpm0.25 & 0.409\textpm0.25 & \textbf{0.418}\textpm0.26 \\
 \midrule
 \multirow[t]{3}{*}{LEV} & AU & 0.183\textpm0.09 & 0.185\textpm0.11 & 0.185\textpm0.12 & 0.184\textpm0.12 & 0.189\textpm0.11 & \textbf{0.194}\textpm0.12 \\
  & EU & 0.206\textpm0.1 & 0.198\textpm0.11 & 0.197\textpm0.1 & 0.195\textpm0.1 & \textbf{0.214}\textpm0.08 & 0.207\textpm0.07 \\
  & TU & 0.185\textpm0.09 & 0.185\textpm0.11 & 0.185\textpm0.12 & 0.184\textpm0.12 & 0.191\textpm0.11 & \textbf{0.194}\textpm0.12 \\
 \midrule
 \multirow[t]{3}{*}{Machine CPU} & AU & 0.544\textpm0.3 & 0.586\textpm0.29 & 0.613\textpm0.28 & 0.687\textpm0.22 & 0.671\textpm0.22 & \textbf{0.708}\textpm0.21 \\
  & EU & 0.552\textpm0.2 & 0.597\textpm0.13 & \textbf{0.626}\textpm0.14 & 0.613\textpm0.14 & 0.595\textpm0.18 & 0.599\textpm0.19 \\
  & TU & 0.559\textpm0.29 & 0.608\textpm0.27 & 0.626\textpm0.25 & 0.673\textpm0.21 & 0.657\textpm0.21 & \textbf{0.702}\textpm0.21 \\
 \midrule
 \multirow[t]{3}{*}{New Thyroid} & AU & 0.476\textpm0.5 & 0.473\textpm0.5 & 0.48\textpm0.51 & \textbf{0.487}\textpm0.51 & \textbf{0.487}\textpm0.51 & \textbf{0.487}\textpm0.51 \\
  & EU & \textbf{0.48}\textpm0.51 & 0.476\textpm0.5 & 0.476\textpm0.5 & 0.473\textpm0.5 & 0.476\textpm0.5 & 0.473\textpm0.5 \\
  & TU & 0.476\textpm0.5 & 0.476\textpm0.5 & \textbf{0.483}\textpm0.51 & 0.48\textpm0.51 & 0.48\textpm0.51 & 0.48\textpm0.51 \\
 \midrule
 \multirow[t]{3}{*}{Pyrimidines} & AU & 0.319\textpm0.46 & 0.29\textpm0.47 & 0.26\textpm0.45 & \textbf{0.366}\textpm0.48 & 0.348\textpm0.48 & 0.313\textpm0.52 \\
  & EU & 0.277\textpm0.47 & 0.263\textpm0.45 & 0.257\textpm0.43 & 0.274\textpm0.51 & \textbf{0.326}\textpm0.49 & 0.282\textpm0.54 \\
  & TU & 0.335\textpm0.4 & 0.309\textpm0.4 & 0.29\textpm0.37 & 0.359\textpm0.44 & \textbf{0.367}\textpm0.45 & 0.35\textpm0.53 \\
 \midrule
 \multirow[t]{3}{*}{Red Wine} & AU & 0.306\textpm0.12 & 0.296\textpm0.11 & 0.294\textpm0.11 & 0.306\textpm0.12 & 0.319\textpm0.12 & \textbf{0.327}\textpm0.12 \\
  & EU & 0.407\textpm0.12 & 0.42\textpm0.12 & 0.428\textpm0.13 & \textbf{0.429}\textpm0.13 & 0.417\textpm0.12 & 0.416\textpm0.13 \\
  & TU & \textbf{0.376}\textpm0.13 & 0.375\textpm0.12 & 0.373\textpm0.13 & 0.371\textpm0.13 & \textbf{0.376}\textpm0.13 & 0.373\textpm0.13 \\
 \midrule
 \multirow[t]{3}{*}{SWD} & AU & 0.121\textpm0.1 & 0.119\textpm0.1 & 0.124\textpm0.09 & 0.135\textpm0.1 & 0.134\textpm0.1 & \textbf{0.144}\textpm0.1 \\
  & EU & 0.143\textpm0.13 & 0.123\textpm0.13 & 0.136\textpm0.13 & 0.125\textpm0.11 & 0.144\textpm0.12 & \textbf{0.146}\textpm0.11 \\
  & TU & 0.122\textpm0.1 & 0.12\textpm0.1 & 0.122\textpm0.09 & 0.136\textpm0.1 & 0.136\textpm0.1 & \textbf{0.145}\textpm0.1 \\
 \midrule
 \multirow[t]{3}{*}{Stocks Domain} & AU & \textbf{0.721}\textpm0.08 & \textbf{0.721}\textpm0.08 & 0.719\textpm0.08 & 0.718\textpm0.08 & \textbf{0.721}\textpm0.08 & 0.719\textpm0.08 \\
  & EU & \textbf{0.679}\textpm0.07 & 0.628\textpm0.08 & 0.625\textpm0.08 & 0.625\textpm0.08 & 0.678\textpm0.07 & 0.678\textpm0.07 \\
  & TU & 0.72\textpm0.07 & \textbf{0.721}\textpm0.07 & \textbf{0.721}\textpm0.07 & 0.72\textpm0.07 & 0.72\textpm0.07 & 0.72\textpm0.07 \\
 \midrule
 \multirow[t]{3}{*}{TAE} & AU & 0.234\textpm0.35 & 0.251\textpm0.36 & 0.251\textpm0.36 & 0.277\textpm0.4 & \textbf{0.287}\textpm0.39 & 0.281\textpm0.4 \\
  & EU & 0.241\textpm0.36 & 0.143\textpm0.34 & 0.191\textpm0.35 & 0.15\textpm0.36 & 0.233\textpm0.37 & \textbf{0.249}\textpm0.41 \\
  & TU & 0.253\textpm0.36 & 0.266\textpm0.38 & 0.28\textpm0.39 & 0.294\textpm0.43 & \textbf{0.298}\textpm0.43 & 0.271\textpm0.45 \\
 \midrule
 \multirow[t]{3}{*}{Triazines} & AU & 0.264\textpm0.26 & 0.269\textpm0.25 & 0.27\textpm0.25 & 0.308\textpm0.25 & 0.3\textpm0.25 & \textbf{0.325}\textpm0.24 \\
  & EU & 0.291\textpm0.24 & 0.313\textpm0.23 & 0.32\textpm0.23 & \textbf{0.341}\textpm0.23 & 0.328\textpm0.24 & 0.339\textpm0.23 \\
  & TU & 0.274\textpm0.23 & 0.273\textpm0.22 & 0.272\textpm0.21 & 0.31\textpm0.23 & 0.307\textpm0.24 & \textbf{0.333}\textpm0.22 \\
 \midrule
 \multirow[t]{3}{*}{White Wine} & AU & 0.235\textpm0.06 & 0.233\textpm0.06 & 0.236\textpm0.06 & 0.25\textpm0.06 & 0.257\textpm0.06 & \textbf{0.264}\textpm0.06 \\
  & EU & 0.354\textpm0.03 & 0.349\textpm0.03 & 0.353\textpm0.04 & 0.353\textpm0.05 & \textbf{0.363}\textpm0.04 & 0.356\textpm0.05 \\
  & TU & \textbf{0.326}\textpm0.05 & 0.32\textpm0.05 & 0.314\textpm0.05 & 0.311\textpm0.05 & 0.323\textpm0.05 & 0.31\textpm0.05 \\
 \midrule
 \multirow[t]{3}{*}{CMC} & AU & 0.172\textpm0.08 & 0.17\textpm0.08 & 0.169\textpm0.08 & 0.246\textpm0.08 & 0.245\textpm0.08 & \textbf{0.278}\textpm0.08 \\
  & EU & 0.222\textpm0.07 & 0.211\textpm0.07 & 0.213\textpm0.07 & 0.231\textpm0.08 & 0.248\textpm0.07 & \textbf{0.267}\textpm0.07 \\
  & TU & 0.181\textpm0.06 & 0.18\textpm0.06 & 0.181\textpm0.06 & 0.27\textpm0.06 & 0.266\textpm0.06 & \textbf{0.289}\textpm0.06 \\
 \midrule
 \multirow[t]{3}{*}{Grub Damage} & AU & 0.127\textpm0.28 & 0.142\textpm0.26 & 0.143\textpm0.24 & 0.175\textpm0.25 & 0.154\textpm0.27 & \textbf{0.202}\textpm0.25 \\
  & EU & 0.072\textpm0.31 & 0.059\textpm0.32 & 0.062\textpm0.34 & 0.087\textpm0.35 & 0.103\textpm0.34 & \textbf{0.12}\textpm0.33 \\
  & TU & 0.128\textpm0.29 & 0.134\textpm0.3 & 0.143\textpm0.3 & 0.185\textpm0.28 & 0.174\textpm0.3 & \textbf{0.205}\textpm0.26 \\
 \midrule
 \multirow[t]{3}{*}{Obesity} & AU & 0.848\textpm0.06 & 0.843\textpm0.07 & 0.844\textpm0.07 & 0.854\textpm0.06 & 0.857\textpm0.06 & \textbf{0.865}\textpm0.05 \\
  & EU & 0.879\textpm0.06 & 0.884\textpm0.05 & 0.885\textpm0.05 & \textbf{0.889}\textpm0.05 & 0.884\textpm0.05 & 0.886\textpm0.05 \\
  & TU & 0.87\textpm0.07 & 0.872\textpm0.07 & 0.873\textpm0.07 & 0.879\textpm0.06 & 0.877\textpm0.06 & \textbf{0.881}\textpm0.05 \\
 \end{longtable}

\normalsize

\section{Out-Of-Distribution (OOD) Detection - Detailed Results}
\label{appendix:ood}

In this section, we provide detailed results for the OOD detection experiment in Subsection \ref{subsec:ood_gbt}, as well as additional OOD detection results using an ensemble of MLPs instead of GBTs.

\subsection{Ensemble of GBTs - Detailed Results for OOD detection}

\begin{figure}[!hbtp]
  \centering
   \begin{subfigure}[t]{0.48\linewidth}
           \centering
           \includegraphics[width=\linewidth]{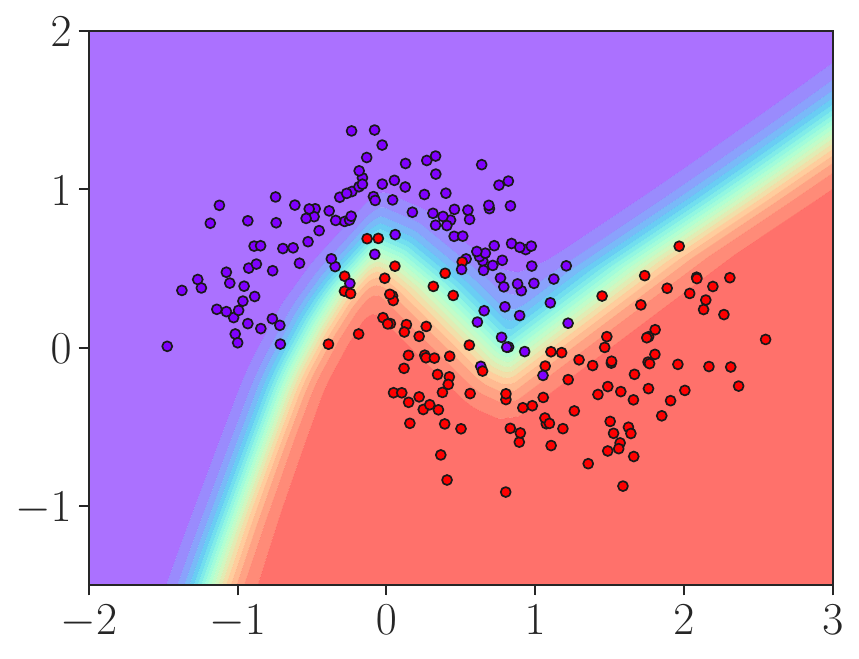}
           \subcaption{The MLP extrapolates smoothly, assigning high-confidence predictions even in OOD regions.}
       \end{subfigure}        
       \begin{subfigure}[t]{0.48\linewidth}
        \centering
        \includegraphics[width=\linewidth]{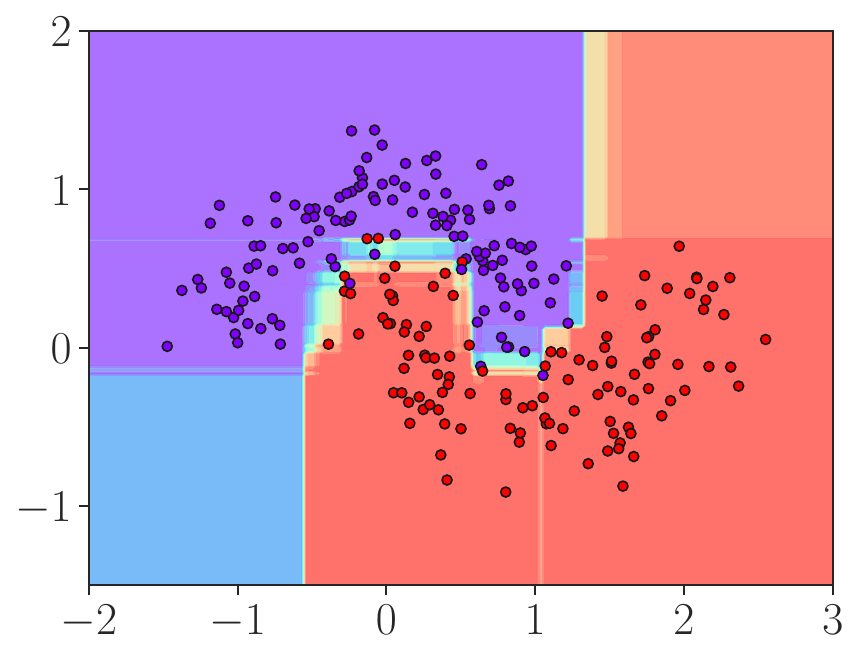}
        \subcaption{The GBT does not extrapolate outside the training data and assigns more uncertain probabilities instead (closer to 0.5 in the binary case) \citep{DBLP:conf/kdd/ChenG16}. }
    \end{subfigure}      
        \caption{Illustration of different behaviors of MLPs and GBTs when it comes to OOD data. In the case of an MLP, OOD data will be predicted confidently with low aleatoric uncertainty. In contrast, OOD data will lead to high aleatoric uncertainty with GBTs, as GBTs will not extrapolate. If an OOD sample falls outside the learned partitions, it is forced into the nearest known leaf. This results in high aleatoric uncertainty, as the OOD sample may be assigned to a leaf that contains a mix of different labels, leading to a less confident prediction.}
          \label{fig:extrapolation}
\end{figure}

Table \ref{tab:ood_gbt_detailed} displays detailed results for OOD detection inclduing all uncertainty types.
Notably, the best OOD performance is not in all cases achieved by measuring epistemic uncertainty. This is in stark contrast to the results obtained from the ensemble of MLPs, where, as expected, epistemic uncertainty very clearly outperforms TU as well as AU for OOD detection across all datasets (cf.\ Table \ref{tab:ood_mlp_detailed}).
Presumably, this is due to the fact that decision trees, unlike neural networks, do not extrapolate their decision function to OOD regions, leading to high-confidence predictions for OOD data with low aleatoric uncertainty. In contrast, in the case of GBTs, OOD data will end up in mixed leaves, also leading to comparably higher aleatoric uncertainty (cf.\ Figure \ref{fig:extrapolation} for an illustration).

\tiny

\begin{longtable}{llllllll}
  \caption{OOD using an ensemble of GBTs (LGBM \citep{DBLP:conf/nips/KeMFWCMYL17}).} \label{tab:ood_gbt_detailed} \\
  \toprule
  & & \multicolumn{6}{c}{\textbf{AUC-ROC} ($\uparrow$) } \\
  \midrule
   & \textbf{Measure} & bin-ent ($\uparrow$) & bin-var ($\uparrow$) & ent ($\uparrow$) & ord-ent ($\uparrow$) & ord-var ($\uparrow$) & var ($\uparrow$) \\
   \midrule
   \textbf{Dataset} &   \textbf{Type} &   &  &  &  &  &  \\
  \endfirsthead
  \toprule
  & & \multicolumn{6}{c}{\textbf{AUC-ROC} ($\uparrow$) } \\
  \midrule
   & \textbf{Measure} & bin-ent ($\uparrow$) & bin-var ($\uparrow$) & ent ($\uparrow$) & ord-ent ($\uparrow$) & ord-var ($\uparrow$) & var ($\uparrow$) \\
   \midrule
   \textbf{Dataset} &   \textbf{Type} &   &  &  &  &  &  \\
  \midrule
  \endhead
  \midrule
  \multicolumn{8}{r}{Continued on next page} \\
  \midrule
  \endfoot
  \bottomrule
  \endlastfoot
  \midrule
  \multirow[t]{3}{*}{Abalone} & AU & 0.678\textpm0.08 & 0.671\textpm0.11 & 0.675\textpm0.07 & 0.682\textpm0.05 & 0.685\textpm0.05 & 0.693\textpm0.07 \\
   & EU & 0.671\textpm0.14 & 0.631\textpm0.14 & 0.72\textpm0.13 & 0.795\textpm0.13 & 0.783\textpm0.14 & \textbf{0.818}\textpm0.13 \\
   & TU & 0.751\textpm0.12 & 0.737\textpm0.15 & 0.763\textpm0.11 & 0.808\textpm0.14 & 0.813\textpm0.13 & 0.813\textpm0.14 \\
  \midrule
  \multirow[t]{3}{*}{Auto MPG} & AU & 0.848\textpm0.19 & 0.803\textpm0.18 & 0.874\textpm0.18 & 0.95\textpm0.08 & 0.933\textpm0.09 & \textbf{0.967}\textpm0.04 \\
   & EU & 0.833\textpm0.11 & 0.731\textpm0.14 & 0.87\textpm0.13 & 0.932\textpm0.07 & 0.863\textpm0.11 & 0.917\textpm0.08 \\
   & TU & 0.85\textpm0.19 & 0.805\textpm0.18 & 0.879\textpm0.18 & 0.951\textpm0.07 & 0.93\textpm0.09 & 0.964\textpm0.04 \\
  \midrule
  \multirow[t]{3}{*}{Automobile} & AU & 0.911\textpm0.05 & 0.903\textpm0.05 & 0.916\textpm0.05 & 0.905\textpm0.06 & 0.9\textpm0.06 & 0.899\textpm0.06 \\
   & EU & 0.922\textpm0.05 & 0.912\textpm0.06 & \textbf{0.939}\textpm0.04 & 0.933\textpm0.05 & 0.921\textpm0.05 & 0.914\textpm0.05 \\
   & TU & 0.916\textpm0.04 & 0.91\textpm0.04 & 0.92\textpm0.05 & 0.908\textpm0.06 & 0.906\textpm0.06 & 0.901\textpm0.06 \\
  \midrule
  \multirow[t]{3}{*}{Balance Scale} & AU & 0.872\textpm0.06 & 0.862\textpm0.06 & \textbf{0.877}\textpm0.05 & 0.816\textpm0.07 & 0.807\textpm0.07 & 0.789\textpm0.07 \\
   & EU & 0.843\textpm0.09 & 0.852\textpm0.08 & 0.844\textpm0.08 & 0.833\textpm0.08 & 0.803\textpm0.07 & 0.758\textpm0.07 \\
   & TU & 0.87\textpm0.06 & 0.859\textpm0.06 & 0.876\textpm0.05 & 0.816\textpm0.07 & 0.807\textpm0.07 & 0.79\textpm0.07 \\
  \midrule
  \multirow[t]{3}{*}{Boston Housing} & AU & 0.507\textpm0.13 & 0.496\textpm0.12 & 0.528\textpm0.14 & 0.656\textpm0.13 & 0.618\textpm0.13 & 0.773\textpm0.12 \\
   & EU & 0.704\textpm0.12 & 0.585\textpm0.14 & 0.72\textpm0.12 & \textbf{0.786}\textpm0.11 & 0.651\textpm0.12 & 0.729\textpm0.12 \\
   & TU & 0.52\textpm0.13 & 0.501\textpm0.12 & 0.543\textpm0.14 & 0.67\textpm0.13 & 0.621\textpm0.13 & 0.769\textpm0.12 \\
  \midrule
  \multirow[t]{3}{*}{CMC} & AU & 0.363\textpm0.07 & 0.362\textpm0.07 & 0.365\textpm0.06 & 0.545\textpm0.07 & 0.54\textpm0.07 & 0.628\textpm0.06 \\
   & EU & 0.667\textpm0.09 & 0.582\textpm0.1 & 0.675\textpm0.08 & \textbf{0.702}\textpm0.1 & 0.647\textpm0.09 & 0.695\textpm0.09 \\
   & TU & 0.371\textpm0.07 & 0.368\textpm0.07 & 0.375\textpm0.06 & 0.555\textpm0.08 & 0.55\textpm0.08 & 0.634\textpm0.07 \\
  \midrule
  \multirow[t]{3}{*}{ERA} & AU & 0.218\textpm0.03 & 0.199\textpm0.02 & 0.222\textpm0.03 & 0.205\textpm0.03 & 0.2\textpm0.03 & 0.221\textpm0.03 \\
   & EU & \textbf{0.992}\textpm0.02 & 0.959\textpm0.03 & 0.989\textpm0.02 & \textbf{0.992}\textpm0.02 & 0.963\textpm0.03 & 0.925\textpm0.08 \\
   & TU & 0.258\textpm0.03 & 0.252\textpm0.02 & 0.258\textpm0.03 & 0.221\textpm0.03 & 0.216\textpm0.03 & 0.232\textpm0.03 \\
  \midrule
  \multirow[t]{3}{*}{ESL} & AU & 0.078\textpm0.06 & 0.084\textpm0.07 & 0.08\textpm0.06 & 0.1\textpm0.06 & 0.103\textpm0.06 & 0.118\textpm0.07 \\
   & EU & 0.596\textpm0.2 & 0.325\textpm0.21 & 0.58\textpm0.21 & \textbf{0.607}\textpm0.2 & 0.357\textpm0.22 & 0.377\textpm0.23 \\
   & TU & 0.088\textpm0.07 & 0.086\textpm0.07 & 0.087\textpm0.07 & 0.11\textpm0.07 & 0.107\textpm0.07 & 0.128\textpm0.08 \\
  \midrule
  \multirow[t]{3}{*}{Eucalyptus} & AU & 0.583\textpm0.05 & 0.568\textpm0.06 & 0.592\textpm0.05 & 0.646\textpm0.01 & 0.646\textpm0.01 & 0.646\textpm0.01 \\
   & EU & 0.581\textpm0.07 & 0.556\textpm0.08 & 0.601\textpm0.06 & 0.645\textpm0.03 & 0.637\textpm0.03 & \textbf{0.653}\textpm0.02 \\
   & TU & 0.59\textpm0.05 & 0.574\textpm0.06 & 0.599\textpm0.05 & 0.647\textpm0.01 & 0.646\textpm0.01 & 0.646\textpm0.01 \\
  \midrule
  \multirow[t]{3}{*}{Grub Damage} & AU & 0.522\textpm0.1 & 0.5\textpm0.09 & 0.533\textpm0.1 & 0.63\textpm0.07 & 0.616\textpm0.07 & 0.619\textpm0.07 \\
   & EU & 0.667\textpm0.08 & 0.62\textpm0.09 & 0.656\textpm0.09 & \textbf{0.712}\textpm0.06 & 0.662\textpm0.07 & 0.62\textpm0.06 \\
   & TU & 0.524\textpm0.11 & 0.506\textpm0.1 & 0.536\textpm0.1 & 0.632\textpm0.07 & 0.617\textpm0.07 & 0.619\textpm0.07 \\
  \midrule
  \multirow[t]{3}{*}{Heart (CLE)} & AU & 0.324\textpm0.06 & 0.316\textpm0.06 & 0.33\textpm0.05 & 0.342\textpm0.05 & 0.332\textpm0.05 & 0.341\textpm0.06 \\
   & EU & \textbf{0.405}\textpm0.15 & 0.368\textpm0.12 & 0.396\textpm0.13 & 0.39\textpm0.13 & 0.359\textpm0.11 & 0.339\textpm0.1 \\
   & TU & 0.325\textpm0.06 & 0.318\textpm0.06 & 0.325\textpm0.06 & 0.342\textpm0.05 & 0.334\textpm0.06 & 0.341\textpm0.06 \\
  \midrule
  \multirow[t]{3}{*}{LEV} & AU & 0.129\textpm0.03 & 0.151\textpm0.04 & 0.117\textpm0.03 & 0.12\textpm0.03 & 0.139\textpm0.04 & 0.133\textpm0.04 \\
   & EU & \textbf{0.776}\textpm0.13 & 0.65\textpm0.09 & 0.727\textpm0.14 & 0.727\textpm0.14 & 0.635\textpm0.09 & 0.629\textpm0.09 \\
   & TU & 0.145\textpm0.04 & 0.16\textpm0.05 & 0.127\textpm0.03 & 0.133\textpm0.04 & 0.147\textpm0.05 & 0.145\textpm0.05 \\
  \midrule
  \multirow[t]{3}{*}{Machine CPU} & AU & 0.309\textpm0.13 & 0.297\textpm0.12 & 0.33\textpm0.13 & 0.621\textpm0.14 & 0.553\textpm0.17 & \textbf{0.832}\textpm0.07 \\
   & EU & 0.432\textpm0.15 & 0.338\textpm0.13 & 0.459\textpm0.14 & 0.726\textpm0.13 & 0.471\textpm0.15 & 0.701\textpm0.14 \\
   & TU & 0.316\textpm0.13 & 0.297\textpm0.12 & 0.333\textpm0.12 & 0.626\textpm0.14 & 0.553\textpm0.17 & 0.829\textpm0.07 \\
  \midrule
  \multirow[t]{3}{*}{New Thyroid} & AU & 0.966\textpm0.03 & 0.957\textpm0.04 & \textbf{0.97}\textpm0.03 & 0.954\textpm0.03 & 0.95\textpm0.04 & 0.945\textpm0.04 \\
   & EU & 0.946\textpm0.05 & 0.946\textpm0.04 & 0.953\textpm0.04 & 0.937\textpm0.05 & 0.942\textpm0.05 & 0.94\textpm0.04 \\
   & TU & 0.964\textpm0.03 & 0.956\textpm0.04 & 0.966\textpm0.03 & 0.954\textpm0.03 & 0.949\textpm0.04 & 0.947\textpm0.04 \\
  \midrule
  \multirow[t]{3}{*}{Obesity} & AU & 0.614\textpm0.02 & 0.617\textpm0.02 & 0.614\textpm0.02 & 0.601\textpm0.02 & 0.607\textpm0.02 & 0.576\textpm0.02 \\
   & EU & \textbf{0.687}\textpm0.04 & 0.66\textpm0.03 & 0.682\textpm0.04 & 0.679\textpm0.04 & 0.661\textpm0.03 & 0.664\textpm0.03 \\
   & TU & 0.616\textpm0.02 & 0.617\textpm0.02 & 0.615\textpm0.02 & 0.603\textpm0.02 & 0.607\textpm0.02 & 0.576\textpm0.02 \\
  \midrule
  \multirow[t]{3}{*}{Pyrimidines} & AU & 0.627\textpm0.05 & 0.627\textpm0.05 & 0.627\textpm0.05 & 0.404\textpm0.14 & 0.443\textpm0.12 & 0.357\textpm0.17 \\
   & EU & 0.536\textpm0.26 & 0.478\textpm0.26 & 0.559\textpm0.27 & 0.635\textpm0.26 & \textbf{0.657}\textpm0.28 & 0.653\textpm0.25 \\
   & TU & 0.616\textpm0.05 & 0.616\textpm0.05 & 0.616\textpm0.05 & 0.404\textpm0.14 & 0.45\textpm0.12 & 0.377\textpm0.17 \\
  \midrule
  \multirow[t]{3}{*}{Red Wine} & AU & 0.755\textpm0.15 & 0.716\textpm0.13 & 0.793\textpm0.15 & 0.917\textpm0.09 & 0.885\textpm0.11 & 0.962\textpm0.03 \\
   & EU & 0.95\textpm0.03 & 0.878\textpm0.08 & 0.965\textpm0.03 & \textbf{0.975}\textpm0.02 & 0.929\textpm0.06 & 0.961\textpm0.03 \\
   & TU & 0.829\textpm0.13 & 0.777\textpm0.13 & 0.854\textpm0.13 & 0.936\textpm0.08 & 0.907\textpm0.1 & 0.967\textpm0.03 \\
  \midrule
  \multirow[t]{3}{*}{SWD} & AU & 0.181\textpm0.05 & 0.192\textpm0.06 & 0.171\textpm0.05 & 0.174\textpm0.05 & 0.184\textpm0.05 & 0.189\textpm0.04 \\
   & EU & \textbf{0.648}\textpm0.18 & 0.494\textpm0.11 & 0.59\textpm0.15 & 0.599\textpm0.15 & 0.492\textpm0.11 & 0.488\textpm0.1 \\
   & TU & 0.194\textpm0.06 & 0.207\textpm0.07 & 0.186\textpm0.05 & 0.177\textpm0.05 & 0.196\textpm0.05 & 0.194\textpm0.04 \\
  \midrule
  \multirow[t]{3}{*}{Stocks Domain} & AU & 0.747\textpm0.05 & 0.744\textpm0.05 & 0.752\textpm0.05 & 0.781\textpm0.05 & 0.772\textpm0.05 & 0.806\textpm0.05 \\
   & EU & 0.818\textpm0.1 & 0.78\textpm0.08 & 0.826\textpm0.09 & \textbf{0.836}\textpm0.09 & 0.79\textpm0.07 & 0.807\textpm0.08 \\
   & TU & 0.752\textpm0.06 & 0.747\textpm0.06 & 0.759\textpm0.06 & 0.784\textpm0.06 & 0.773\textpm0.06 & 0.807\textpm0.05 \\
  \midrule
  \multirow[t]{3}{*}{TAE} & AU & 0.832\textpm0.15 & 0.832\textpm0.15 & 0.834\textpm0.15 & 0.603\textpm0.13 & 0.601\textpm0.13 & 0.52\textpm0.06 \\
   & EU & 0.618\textpm0.26 & 0.641\textpm0.23 & 0.643\textpm0.25 & 0.613\textpm0.19 & 0.635\textpm0.17 & 0.629\textpm0.14 \\
   & TU & 0.848\textpm0.15 & 0.848\textpm0.15 & \textbf{0.85}\textpm0.15 & 0.611\textpm0.14 & 0.612\textpm0.13 & 0.52\textpm0.06 \\
  \midrule
  \multirow[t]{3}{*}{Triazines} & AU & 0.567\textpm0.11 & 0.521\textpm0.12 & 0.582\textpm0.12 & 0.697\textpm0.09 & 0.659\textpm0.09 & 0.69\textpm0.09 \\
   & EU & 0.847\textpm0.12 & 0.759\textpm0.2 & 0.873\textpm0.08 & \textbf{0.896}\textpm0.09 & 0.802\textpm0.19 & 0.808\textpm0.15 \\
   & TU & 0.574\textpm0.11 & 0.519\textpm0.12 & 0.595\textpm0.12 & 0.704\textpm0.09 & 0.678\textpm0.1 & 0.707\textpm0.1 \\
  \midrule
  \multirow[t]{3}{*}{White Wine} & AU & 0.644\textpm0.17 & 0.579\textpm0.16 & 0.705\textpm0.15 & 0.846\textpm0.08 & 0.793\textpm0.09 & 0.891\textpm0.06 \\
   & EU & 0.977\textpm0.04 & 0.937\textpm0.09 & \textbf{0.986}\textpm0.02 & 0.981\textpm0.03 & 0.938\textpm0.08 & 0.938\textpm0.08 \\
   & TU & 0.818\textpm0.09 & 0.763\textpm0.09 & 0.853\textpm0.09 & 0.899\textpm0.08 & 0.844\textpm0.09 & 0.91\textpm0.07 \\
  \midrule
  \multirow[c]{3}{*}{\shortstack{Wisconsin \\Breast Cancer}} & AU & 0.873\textpm0.06 & 0.857\textpm0.06 & \textbf{0.874}\textpm0.06 & 0.818\textpm0.08 & 0.811\textpm0.08 & 0.805\textpm0.08 \\
   & EU & 0.335\textpm0.2 & 0.346\textpm0.2 & 0.381\textpm0.19 & 0.381\textpm0.21 & 0.444\textpm0.23 & 0.524\textpm0.22 \\
   & TU & 0.852\textpm0.06 & 0.839\textpm0.07 & 0.854\textpm0.07 & 0.801\textpm0.08 & 0.797\textpm0.08 & 0.79\textpm0.1 \\
  \end{longtable}

\normalsize

\subsection{Ensemble of MLPs - Detailed Results for OOD detection}

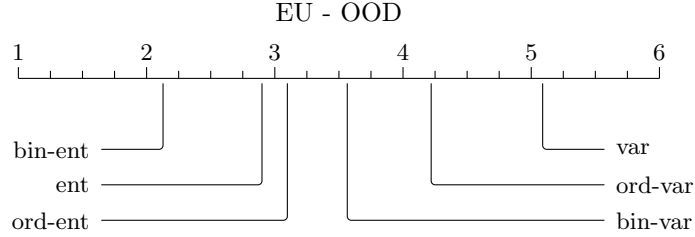
\begin{figure}[!htbp]
  \centering 
  \caption{CD diagram for OOD detection of the different uncertainty measures using an ensemble of MLPs.}
  \label{fig:ood_mlp}

  \begin{subfigure}[t]{\linewidth}
    \centering

\begin{tikzpicture}[
  treatment line/.style={rounded corners=1.5pt, line cap=round, shorten >=1pt},
  treatment label/.style={font=\small},
  group line/.style={ultra thick},
]

\begin{axis}[
  clip={false},
  axis x line={center},
  axis y line={none},
  axis line style={-},
  xmin={1},
  ymax={0},
  scale only axis={true},
  width={\axisdefaultwidth},
  ticklabel style={anchor=south, yshift=1.3*\pgfkeysvalueof{/pgfplots/major tick length}, font=\small},
  every tick/.style={draw=black},
  major tick style={yshift=.5*\pgfkeysvalueof{/pgfplots/major tick length}},
  minor tick style={yshift=.5*\pgfkeysvalueof{/pgfplots/minor tick length}},
  title style={yshift=\baselineskip},
  xmax={6},
  ymin={-4.5},
  height={5\baselineskip},
  xtick={1,2,3,4,5,6},
  minor x tick num={3},
  title={EU - OOD},
]

\draw[treatment line] ([yshift=-2pt] axis cs:2.128260869565217, 0) |- (axis cs:1.6282608695652172, -2.0)
  node[treatment label, anchor=east] {bin-ent};
\draw[treatment line] ([yshift=-2pt] axis cs:2.9, 0) |- (axis cs:1.6282608695652172, -3.0)
  node[treatment label, anchor=east] {ent};
\draw[treatment line] ([yshift=-2pt] axis cs:3.097826086956522, 0) |- (axis cs:1.6282608695652172, -4.0)
  node[treatment label, anchor=east] {ord-ent};
\draw[treatment line] ([yshift=-2pt] axis cs:3.5652173913043477, 0) |- (axis cs:5.589130434782609, -4.0)
  node[treatment label, anchor=west] {bin-var};
\draw[treatment line] ([yshift=-2pt] axis cs:4.219565217391304, 0) |- (axis cs:5.589130434782609, -3.0)
  node[treatment label, anchor=west] {ord-var};
\draw[treatment line] ([yshift=-2pt] axis cs:5.089130434782609, 0) |- (axis cs:5.589130434782609, -2.0)
  node[treatment label, anchor=west] {var};

\end{axis}
\end{tikzpicture}
\end{subfigure}
\end{figure}

In the case of an ensemble of MLPs, EU clearly outperforms TU and AU when it comes to OOD detection. The MLP will output overconfident predictions for OOD data with low to even no aleatoric uncertainty. Just like for GBTs (cf.\ Figure \ref{fig:ood_lgbm}), entropy-based measures outperform variance-based measures (cf.\ Figure \ref{fig:ood_mlp}), and the proposed OCS decomposition method can be considered competitive with existing standard and labelwise decompositions, though not excelling at OOD.

\tiny

\begin{longtable}{llllllll}
  \caption{OOD using an ensemble of MLPs.} \label{tab:ood_mlp_detailed} \\
  \toprule
  & & \multicolumn{6}{c}{\textbf{AUC-ROC} ($\uparrow$) } \\
  \midrule
   & \textbf{Measure} & bin-ent ($\uparrow$) & bin-var ($\uparrow$) & ent ($\uparrow$) & ord-ent ($\uparrow$) & ord-var ($\uparrow$) & var ($\uparrow$) \\
   \midrule
   \textbf{Dataset} &   \textbf{Type} &   &  &  &  &  &  \\
  \endfirsthead
  \toprule
  & & \multicolumn{6}{c}{\textbf{AUC-ROC} ($\uparrow$) } \\
  \midrule
   & \textbf{Measure} & bin-ent ($\uparrow$) & bin-var ($\uparrow$) & ent ($\uparrow$) & ord-ent ($\uparrow$) & ord-var ($\uparrow$) & var ($\uparrow$) \\
   \midrule
   \textbf{Dataset} &   \textbf{Type} &   &  &  &  &  &  \\
  \midrule
  \endhead
  \midrule
  \multicolumn{8}{r}{Continued on next page} \\
  \midrule
  \endfoot
  \bottomrule
  \endlastfoot
  \midrule
  \multirow[t]{3}{*}{Abalone} & AU & 0.0\textpm0.0 & 0.0\textpm0.0 & 0.0\textpm0.0 & 0.0\textpm0.0 & 0.0\textpm0.0 & 0.0\textpm0.0 \\
   & EU & \textbf{0.8}\textpm0.35 & 0.799\textpm0.35 & 0.799\textpm0.35 & 0.799\textpm0.35 & 0.799\textpm0.35 & 0.795\textpm0.35 \\
   & TU & 0.104\textpm0.06 & 0.11\textpm0.06 & 0.105\textpm0.07 & 0.186\textpm0.14 & 0.185\textpm0.14 & 0.238\textpm0.19 \\
  \midrule
  \multirow[t]{3}{*}{Auto MPG} & AU & 0.008\textpm0.01 & 0.015\textpm0.02 & 0.008\textpm0.01 & 0.021\textpm0.02 & 0.029\textpm0.02 & 0.06\textpm0.03 \\
   & EU & \textbf{1.0}\textpm0.0 & \textbf{1.0}\textpm0.0 & \textbf{1.0}\textpm0.0 & \textbf{1.0}\textpm0.0 & \textbf{1.0}\textpm0.0 & \textbf{1.0}\textpm0.0 \\
   & TU & 0.647\textpm0.06 & 0.682\textpm0.08 & 0.627\textpm0.06 & 0.87\textpm0.05 & 0.897\textpm0.05 & 0.913\textpm0.05 \\
  \midrule
  \multirow[t]{3}{*}{Automobile} & AU & 0.086\textpm0.07 & 0.094\textpm0.08 & 0.085\textpm0.07 & 0.085\textpm0.07 & 0.093\textpm0.08 & 0.09\textpm0.08 \\
   & EU & 0.569\textpm0.29 & 0.562\textpm0.29 & 0.564\textpm0.29 & \textbf{0.57}\textpm0.29 & 0.568\textpm0.29 & 0.562\textpm0.29 \\
   & TU & 0.481\textpm0.25 & 0.484\textpm0.25 & 0.478\textpm0.24 & 0.5\textpm0.25 & 0.501\textpm0.25 & 0.504\textpm0.25 \\
  \midrule
  \multirow[t]{3}{*}{Balance Scale} & AU & 0.331\textpm0.15 & 0.334\textpm0.15 & 0.331\textpm0.15 & 0.331\textpm0.15 & 0.334\textpm0.15 & 0.334\textpm0.15 \\
   & EU & \textbf{0.573}\textpm0.33 & 0.569\textpm0.33 & \textbf{0.573}\textpm0.33 & \textbf{0.573}\textpm0.33 & 0.569\textpm0.33 & 0.569\textpm0.33 \\
   & TU & 0.556\textpm0.33 & 0.557\textpm0.33 & 0.555\textpm0.33 & 0.555\textpm0.33 & 0.557\textpm0.33 & 0.556\textpm0.33 \\
  \midrule
  \multirow[t]{3}{*}{Boston Housing} & AU & 0.003\textpm0.01 & 0.003\textpm0.01 & 0.003\textpm0.01 & 0.002\textpm0.01 & 0.003\textpm0.01 & 0.003\textpm0.01 \\
   & EU & \textbf{0.731}\textpm0.09 & 0.728\textpm0.1 & 0.729\textpm0.1 & 0.728\textpm0.09 & 0.726\textpm0.09 & 0.721\textpm0.08 \\
   & TU & 0.589\textpm0.12 & 0.598\textpm0.13 & 0.58\textpm0.12 & 0.577\textpm0.12 & 0.591\textpm0.12 & 0.582\textpm0.12 \\
  \midrule
  \multirow[t]{3}{*}{CMC} & AU & 0.001\textpm0.0 & 0.001\textpm0.0 & 0.001\textpm0.0 & 0.001\textpm0.0 & 0.001\textpm0.0 & 0.001\textpm0.0 \\
   & EU & \textbf{0.924}\textpm0.2 & 0.915\textpm0.2 & 0.921\textpm0.2 & 0.921\textpm0.2 & 0.91\textpm0.2 & 0.898\textpm0.2 \\
   & TU & 0.553\textpm0.15 & 0.548\textpm0.15 & 0.558\textpm0.15 & 0.552\textpm0.17 & 0.544\textpm0.16 & 0.58\textpm0.18 \\
  \midrule
  \multirow[t]{3}{*}{ERA} & AU & 0.243\textpm0.02 & 0.236\textpm0.02 & 0.244\textpm0.02 & 0.251\textpm0.03 & 0.251\textpm0.03 & 0.264\textpm0.03 \\
   & EU & \textbf{0.926}\textpm0.03 & 0.844\textpm0.07 & 0.903\textpm0.04 & 0.909\textpm0.04 & 0.813\textpm0.08 & 0.752\textpm0.09 \\
   & TU & 0.259\textpm0.03 & 0.258\textpm0.03 & 0.258\textpm0.03 & 0.26\textpm0.03 & 0.258\textpm0.03 & 0.271\textpm0.03 \\
  \midrule
  \multirow[t]{3}{*}{ESL} & AU & 0.009\textpm0.01 & 0.012\textpm0.01 & 0.009\textpm0.01 & 0.013\textpm0.01 & 0.017\textpm0.01 & 0.025\textpm0.01 \\
   & EU & 0.397\textpm0.28 & 0.389\textpm0.28 & 0.396\textpm0.28 & \textbf{0.398}\textpm0.28 & 0.389\textpm0.28 & 0.388\textpm0.28 \\
   & TU & 0.121\textpm0.11 & 0.128\textpm0.11 & 0.114\textpm0.1 & 0.193\textpm0.18 & 0.19\textpm0.17 & 0.25\textpm0.24 \\
  \midrule
  \multirow[t]{3}{*}{Eucalyptus} & AU & 0.028\textpm0.01 & 0.03\textpm0.01 & 0.028\textpm0.01 & 0.029\textpm0.01 & 0.03\textpm0.01 & 0.03\textpm0.01 \\
   & EU & \textbf{1.0}\textpm0.0 & \textbf{1.0}\textpm0.0 & 0.999\textpm0.0 & 0.994\textpm0.01 & 0.991\textpm0.01 & 0.978\textpm0.03 \\
   & TU & 0.972\textpm0.03 & 0.97\textpm0.03 & 0.971\textpm0.03 & 0.95\textpm0.04 & 0.952\textpm0.04 & 0.94\textpm0.04 \\
  \midrule
  \multirow[t]{3}{*}{Grub Damage} & AU & 0.1\textpm0.04 & 0.109\textpm0.05 & 0.099\textpm0.05 & 0.099\textpm0.04 & 0.108\textpm0.04 & 0.115\textpm0.04 \\
   & EU & \textbf{0.934}\textpm0.13 & 0.915\textpm0.14 & 0.929\textpm0.13 & 0.917\textpm0.15 & 0.899\textpm0.16 & 0.868\textpm0.15 \\
   & TU & 0.745\textpm0.16 & 0.74\textpm0.16 & 0.743\textpm0.16 & 0.723\textpm0.15 & 0.729\textpm0.15 & 0.709\textpm0.14 \\
  \midrule
  \multirow[t]{3}{*}{Heart (CLE)} & AU & 0.016\textpm0.01 & 0.017\textpm0.01 & 0.015\textpm0.01 & 0.016\textpm0.01 & 0.017\textpm0.01 & 0.018\textpm0.01 \\
   & EU & \textbf{0.777}\textpm0.21 & 0.75\textpm0.21 & 0.754\textpm0.2 & 0.738\textpm0.21 & 0.72\textpm0.23 & 0.687\textpm0.22 \\
   & TU & 0.478\textpm0.14 & 0.503\textpm0.14 & 0.469\textpm0.13 & 0.476\textpm0.18 & 0.492\textpm0.18 & 0.481\textpm0.18 \\
  \midrule
  \multirow[t]{3}{*}{LEV} & AU & 0.012\textpm0.0 & 0.013\textpm0.0 & 0.011\textpm0.0 & 0.01\textpm0.0 & 0.012\textpm0.0 & 0.011\textpm0.0 \\
   & EU & \textbf{0.474}\textpm0.23 & 0.447\textpm0.26 & 0.469\textpm0.23 & 0.469\textpm0.23 & 0.446\textpm0.26 & 0.445\textpm0.26 \\
   & TU & 0.035\textpm0.03 & 0.048\textpm0.03 & 0.026\textpm0.02 & 0.024\textpm0.02 & 0.042\textpm0.03 & 0.03\textpm0.03 \\
  \midrule
  \multirow[t]{3}{*}{Machine CPU} & AU & 0.076\textpm0.05 & 0.083\textpm0.05 & 0.076\textpm0.05 & 0.073\textpm0.05 & 0.081\textpm0.05 & 0.074\textpm0.04 \\
   & EU & \textbf{0.237}\textpm0.09 & 0.177\textpm0.09 & 0.227\textpm0.08 & 0.226\textpm0.08 & 0.175\textpm0.09 & 0.171\textpm0.09 \\
   & TU & 0.092\textpm0.06 & 0.094\textpm0.06 & 0.091\textpm0.06 & 0.087\textpm0.05 & 0.09\textpm0.06 & 0.083\textpm0.05 \\
  \midrule
  \multirow[t]{3}{*}{New Thyroid} & AU & 0.146\textpm0.07 & 0.15\textpm0.08 & 0.146\textpm0.07 & 0.141\textpm0.07 & 0.145\textpm0.07 & 0.137\textpm0.07 \\
   & EU & \textbf{0.175}\textpm0.08 & 0.17\textpm0.08 & 0.174\textpm0.08 & 0.173\textpm0.08 & 0.168\textpm0.08 & 0.164\textpm0.08 \\
   & TU & 0.152\textpm0.08 & 0.152\textpm0.08 & 0.152\textpm0.08 & 0.146\textpm0.07 & 0.147\textpm0.07 & 0.14\textpm0.07 \\
  \midrule
  \multirow[t]{3}{*}{Obesity} & AU & 0.004\textpm0.01 & 0.004\textpm0.01 & 0.004\textpm0.01 & 0.004\textpm0.01 & 0.004\textpm0.01 & 0.004\textpm0.01 \\
   & EU & \textbf{0.92}\textpm0.19 & 0.911\textpm0.19 & 0.917\textpm0.19 & 0.913\textpm0.19 & 0.909\textpm0.19 & 0.906\textpm0.19 \\
   & TU & 0.859\textpm0.17 & 0.857\textpm0.17 & 0.859\textpm0.17 & 0.869\textpm0.17 & 0.868\textpm0.17 & 0.872\textpm0.18 \\
  \midrule
  \multirow[t]{3}{*}{Pyrimidines} & AU & 0.0\textpm0.0 & 0.0\textpm0.0 & 0.0\textpm0.0 & 0.0\textpm0.0 & 0.0\textpm0.0 & 0.0\textpm0.0 \\
   & EU & \textbf{0.898}\textpm0.22 & 0.888\textpm0.22 & 0.888\textpm0.22 & 0.846\textpm0.2 & 0.846\textpm0.2 & 0.814\textpm0.2 \\
   & TU & 0.692\textpm0.23 & 0.66\textpm0.21 & 0.692\textpm0.23 & 0.715\textpm0.24 & 0.715\textpm0.25 & 0.691\textpm0.26 \\
  \midrule
  \multirow[t]{3}{*}{Red Wine} & AU & 0.0\textpm0.0 & 0.0\textpm0.0 & 0.0\textpm0.0 & 0.0\textpm0.0 & 0.0\textpm0.0 & 0.0\textpm0.0 \\
   & EU & \textbf{0.998}\textpm0.01 & 0.99\textpm0.03 & 0.995\textpm0.01 & \textbf{0.998}\textpm0.0 & 0.997\textpm0.0 & 0.984\textpm0.01 \\
   & TU & 0.777\textpm0.1 & 0.802\textpm0.1 & 0.751\textpm0.1 & 0.763\textpm0.07 & 0.799\textpm0.08 & 0.787\textpm0.08 \\
  \midrule
  \multirow[t]{3}{*}{SWD} & AU & 0.004\textpm0.0 & 0.005\textpm0.0 & 0.004\textpm0.0 & 0.004\textpm0.0 & 0.005\textpm0.0 & 0.005\textpm0.0 \\
   & EU & \textbf{0.363}\textpm0.3 & 0.352\textpm0.3 & 0.361\textpm0.3 & 0.361\textpm0.3 & 0.352\textpm0.3 & 0.351\textpm0.3 \\
   & TU & 0.044\textpm0.04 & 0.047\textpm0.04 & 0.041\textpm0.03 & 0.04\textpm0.03 & 0.043\textpm0.04 & 0.042\textpm0.04 \\
  \midrule
  \multirow[t]{3}{*}{Stocks Domain} & AU & 0.339\textpm0.08 & 0.383\textpm0.09 & 0.339\textpm0.08 & 0.375\textpm0.08 & 0.417\textpm0.09 & 0.46\textpm0.09 \\
   & EU & \textbf{0.977}\textpm0.05 & 0.969\textpm0.06 & \textbf{0.977}\textpm0.05 & \textbf{0.977}\textpm0.05 & 0.969\textpm0.06 & 0.97\textpm0.06 \\
   & TU & 0.81\textpm0.07 & 0.804\textpm0.07 & 0.815\textpm0.08 & 0.889\textpm0.08 & 0.875\textpm0.07 & 0.902\textpm0.08 \\
  \midrule
  \multirow[t]{3}{*}{TAE} & AU & 0.018\textpm0.02 & 0.022\textpm0.02 & 0.018\textpm0.02 & 0.017\textpm0.02 & 0.019\textpm0.02 & 0.017\textpm0.02 \\
   & EU & \textbf{0.906}\textpm0.17 & 0.892\textpm0.18 & \textbf{0.906}\textpm0.17 & \textbf{0.906}\textpm0.17 & 0.888\textpm0.18 & 0.873\textpm0.17 \\
   & TU & 0.47\textpm0.2 & 0.475\textpm0.21 & 0.461\textpm0.2 & 0.445\textpm0.2 & 0.455\textpm0.21 & 0.445\textpm0.21 \\
  \midrule
  \multirow[t]{3}{*}{Triazines} & AU & 0.0\textpm0.0 & 0.0\textpm0.0 & 0.0\textpm0.0 & 0.0\textpm0.0 & 0.0\textpm0.0 & 0.0\textpm0.0 \\
   & EU & \textbf{0.609}\textpm0.31 & 0.594\textpm0.3 & 0.604\textpm0.31 & 0.604\textpm0.32 & 0.581\textpm0.3 & 0.568\textpm0.3 \\
   & TU & 0.505\textpm0.27 & 0.504\textpm0.27 & 0.509\textpm0.28 & 0.48\textpm0.26 & 0.475\textpm0.25 & 0.47\textpm0.25 \\
  \midrule
  \multirow[t]{3}{*}{White Wine} & AU & 0.0\textpm0.0 & 0.0\textpm0.0 & 0.0\textpm0.0 & 0.0\textpm0.0 & 0.0\textpm0.0 & 0.0\textpm0.0 \\
   & EU & \textbf{0.989}\textpm0.01 & 0.972\textpm0.04 & 0.977\textpm0.03 & 0.982\textpm0.01 & 0.975\textpm0.02 & 0.951\textpm0.03 \\
   & TU & 0.489\textpm0.14 & 0.509\textpm0.13 & 0.478\textpm0.15 & 0.605\textpm0.14 & 0.617\textpm0.13 & 0.648\textpm0.12 \\
  \midrule
  \multirow[c]{3}{*}{\shortstack{Wisconsin \\Breast Cancer}} & AU & 0.0\textpm0.0 & 0.0\textpm0.0 & 0.0\textpm0.0 & 0.0\textpm0.0 & 0.0\textpm0.0 & 0.0\textpm0.0 \\
   & EU & \textbf{0.907}\textpm0.19 & 0.878\textpm0.2 & 0.878\textpm0.2 & 0.862\textpm0.21 & 0.839\textpm0.22 & 0.807\textpm0.24 \\
   & TU & 0.614\textpm0.18 & 0.617\textpm0.17 & 0.601\textpm0.18 & 0.585\textpm0.22 & 0.598\textpm0.23 & 0.577\textpm0.23 \\
  \end{longtable}

\normalsize

\newpage




\end{appendices}


\bibliography{sn-bibliography}

\end{document}